\documentclass[10pt,twocolumn,letterpaper]{article}
\usepackage[pagenumbers]{cvpr} 

\graphicspath{{./}}

\definecolor{linkblue}{rgb}{0.21,0.49,0.74}
\usepackage[pagebackref,breaklinks,colorlinks,allcolors=linkblue]{hyperref}
\hypersetup{pdftitle={Intuition Vectors from Pretrained Representations Steer Diffusion-Based Reasoning},pdfauthor={Shahar Haim and Daniel C. McNamee}}

\title{Intuition Vectors from Pretrained Representations \\ Steer Diffusion-Based Reasoning}

\author{Shahar Haim and Daniel C. McNamee\\
{\normalsize Champalimaud Research, Centre for the Unknown}\\
{\normalsize Lisbon, Portugal}\\
{\small\texttt{\{shahar.haim,daniel.mcnamee\}@research.fchampalimaud.org}}
}

\makeatletter
\def\ps@preprint{%
  \let\@oddhead\@empty
  \let\@evenhead\@empty
  \def\@oddfoot{\rlap{\small Preprint}\hfil\thepage\hfil}%
  \let\@evenfoot\@oddfoot
}
\makeatother

\begin{document}
\maketitle
\thispagestyle{preprint}
\begin{abstract}

Abstract visual reasoning requires identifying relationships between few examples and applying them to novel inputs. Recent advances have been driven by large language and vision-language models or specialized domain-specific architectures. It remains unclear whether abstract reasoning can be supported directly by the relational geometry of pretrained visual representations, without relying on language-based inference or task-specific representation learning. We investigate this possibility using frozen self-supervised vision encoders. With latent vector arithmetic, we confirmed that the representational geometry of two self-supervised encoders support abstract concept discrimination on Bongard problems and encode systematic relationships between input-output pairs in ARC tasks. Simple vector differences between image embeddings, which we call \emph{intuition vectors}, capture transformations that generalize across instances of the same task. These vectors distinguish among 794 abstract reasoning tasks with up to 87.2\% accuracy, preserve relational structure across task-invariant visual transformations, and guide output retrieval for unseen inputs. Given this, we tested the hypothesis that this relational structure can steer generative reasoning by conditioning a diffusion transformer operating in pretrained visual latent space. On held-out ARC tasks, intuition-vector conditioning outperforms conditioning on complete demonstrations while requiring nearly six times less test-time compute, and supports zero-shot inference. Taken together, our findings suggest that pretrained visual representations provide a transferable substrate for abstract reasoning, enabling a modular approach in which the relational structure inferred from demonstrations steer generative reasoning.

\end{abstract}

\section{Introduction}
\begin{figure*}[!t]
    \centering
    \captionsetup[subfigure]{skip=1pt}
    \begin{subfigure}[t]{0.195\textwidth}
        \centering
        \includegraphics[width=\linewidth]{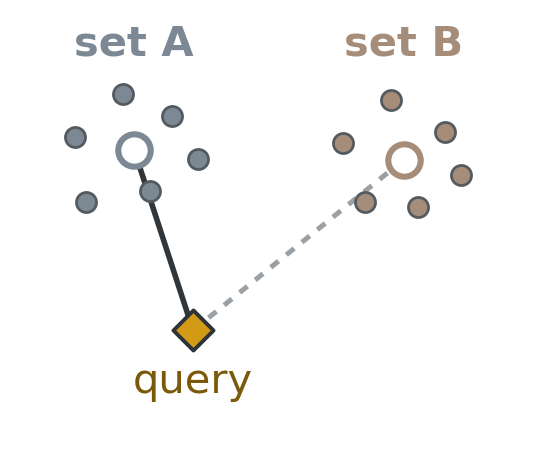}
        \caption{Concepts (\S\ref{sec:concepts})}
        \label{fig:concept-centroid}
    \end{subfigure}
    \hfill
    \begin{subfigure}[t]{0.195\textwidth}
        \centering
        \includegraphics[width=\linewidth]{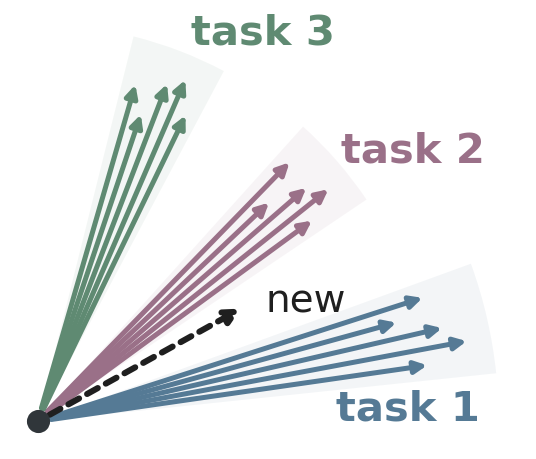}
        \caption{Task identity (\S\ref{sec:task-identity})}
        \label{fig:concept-population}
    \end{subfigure}
    \hfill
    \begin{subfigure}[t]{0.195\textwidth}
        \centering
        \includegraphics[width=\linewidth]{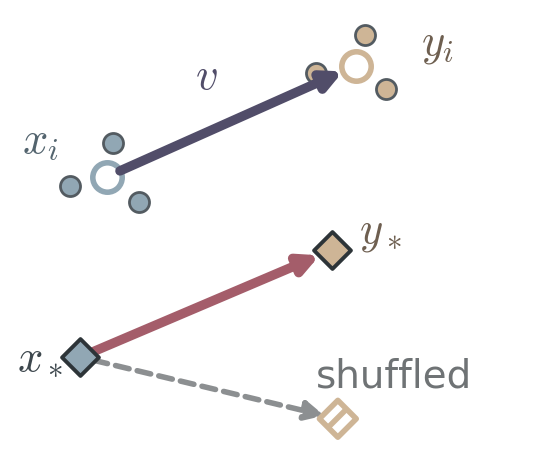}
        \caption{Alignment (\S\ref{sec:alignment})}
        \label{fig:concept-alignment}
    \end{subfigure}
    \hfill
    \begin{subfigure}[t]{0.195\textwidth}
        \centering
        \includegraphics[width=\linewidth]{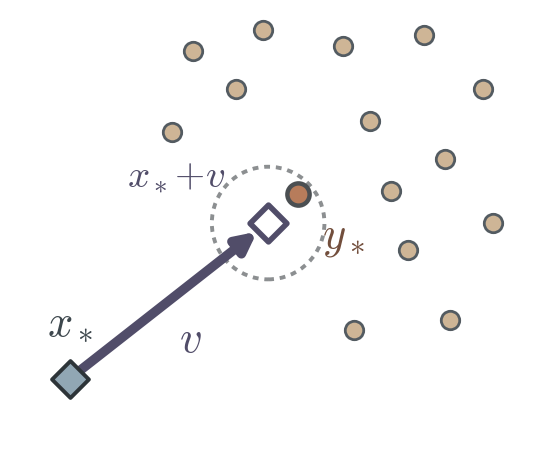}
        \caption{Retrieval (\S\ref{sec:alignment})}
        \label{fig:concept-transport}
    \end{subfigure}
    \hfill
    \begin{subfigure}[t]{0.195\textwidth}
        \centering
        \includegraphics[width=\linewidth]{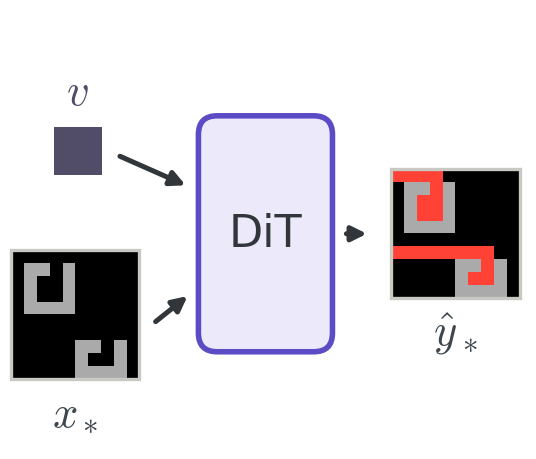}
        \caption{Generation (\S\ref{sec:rae-results})}
        \label{fig:concept-generation}
    \end{subfigure}
    \caption{
        From visual concept discrimination to transformation-conditioned generation.
        \textbf{(a)} In Bongard, points represent individual images: a query is assigned to the
        nearer of two support centroids, testing concept discrimination.
        \textbf{(b--e)} In ARC, input--output embedding differences represent transformations.
        \textbf{(b)} Single-pair difference vectors $E(y)-E(x)$ identify the task across
        different instances.
        \textbf{(c)} The intuition vector $v$, averaged over support pairs, discriminates
        the held-out transformation vector and a cell-shuffled control.
        \textbf{(d)} Adding $v$ to the query embedding guides retrieval of candidate outputs.
        \textbf{(e)} Conditioning a trained diffusion transformer on $v$ and the query steers generation of the answer.
    }
    \label{fig:concept}
\end{figure*}

Pretrained visual representations can support the discrimination of novel concepts from few examples without adapting the underlying encoder \citep{sorcher2022geometry, bitton2022vasrvisualanalogiessituation, vompa2026linearseparabilityceilingaligning}. Moving from concept discrimination to transformation reasoning shifts the focus from identifying concepts shared across images to inferring rules that relate inputs and outputs across examples. Such rule inference is central to the ARC-AGI-1 and 2 benchmarks \citep{chollet2019measureintelligence, chollet2026arcagi2}, where some methods construct an explicit intermediate task representation, such as a symbolic program \citep{li2024combininginductiontransductionabstract} or latent program code \citep{macfarlane2025searchinglatentprogramspaces}, before applying it to a test input. Others predict outputs directly through task-specific training or adaptation \citep{hu2025arcvisionproblem, akyurek2025surprisingeffectivenesstesttimetraining}, without first constructing a program.

General-purpose visual pretraining offers a potential source of reusable structure for either form of inference. Self-supervised encoders support a range of perceptual tasks \citep{simeoni2025dinov3, he2021maskedautoencodersscalablevision}, and their features have also been used as visual latent spaces for learned predictive and generative models \citep{zhou2025dinowmworldmodelspretrained, zheng2025rae, singh2026raev2}. Our question concerns an additional role for pretrained representations: can their relational structure supply a task-level transformation signal, extracted from demonstrations without adapting the encoder, that can then steer exact-output generation in the pretrained latent space?

We address this question through a progression from representation analysis to generative execution (Fig.~\ref{fig:concept}). Using frozen DINOv3 and MAE encoders, we first test concept discrimination on Bongard problems. On ARC, we test task discrimination using single-pair input-output embedding differences, then average these differences across demonstrations to form \emph{intuition vectors} for alignment and output retrieval. Pixel projections and untrained encoders provide controls. Finally, we test whether this extracted task-level signal can guide answer generation with the Reasoning Representation Autoencoder (RRAE). A diffusion transformer is trained to generate answer latents from the query and intuition vector in the frozen DINOv3 representation space; conditioning on complete input-output examples and learned task tokens provides comparisons.

Our contributions are as follows: (i) We show that frozen self-supervised visual embeddings support abstract concept discrimination in Bongard, while input-output differences identify ARC transformation tasks across instances and robust to visual augmentation. (ii) We show that intuition vectors extracted directly from frozen pretrained embeddings provide a transferable relational signal that guides output retrieval for novel inputs through latent vector arithmetic. (iii) We introduce RRAE, adapting RAEv2 to execute transformations conditioned on intuition vectors, and demonstrate zero-shot exact-output generation on held-out ARC1 tasks. With comparable training compute, intuition-vector conditioning achieves higher cell accuracy than full-demonstration conditioning on both ARC1\&2 after test-time training, while requiring roughly one-sixth of the test-time compute.

\section{Related work}

\subsection{Bongard and ARC: concepts and transformations}
Bongard problems test abstract concept discrimination from positive and negative examples \citep{bongard1970pattern}. Bongard-LOGO uses geometric drawings \citep{nie2020bongardlogo}, while Bongard-OpenWorld uses natural images and free-form concepts \citep{wu2024bongardopen}. Common approaches include dedicated few-shot models that classify queries by distances to support examples or class prototypes \citep{nie2020bongardlogo,raghuraman2024supportsetcontextmattersbongard}. Recent work probes the linear separability of abstract concepts in pretrained VLM visual embeddings \citep{vompa2026linearseparabilityceilingaligning}. We focus on abstract concept discrimination in frozen self-supervised visual encoders, without language-mediated inference or reasoning-specific representation learning. ARC-AGI-1 and 2 require inferring transformations from demonstration input-output grids and generating output grids for novel queries \citep{chollet2019measureintelligence,chollet2026arcagi2}.
A test query is solved if either of two proposed output grids exactly matches the target (pass@2).
Inductive approaches infer a symbolic or latent program before applying it to a query \citep{li2024combininginductiontransductionabstract,macfarlane2025searchinglatentprogramspaces}, whereas transductive models predict outputs directly, often with test-time adaptation \citep{hu2025arcvisionproblem,akyurek2025surprisingeffectivenesstesttimetraining}. Their complementarity motivates hybrid approaches \citep{li2024combininginductiontransductionabstract}. Limited data availability and the need to study conceptual generalization have motivated additional datasets and generators: ARC-GEN generates further instances of existing ARC rules \citep{moffitt2025arcgenmimeticproceduralbenchmark}, BARC supplies synthetic ARC-style tasks \citep{li2024combininginductiontransductionabstract}, and ConceptARC organizes problems around core spatial and semantic concepts \citep{moskvichev2023conceptarcbenchmarkevaluatingunderstanding}.

\subsection{Visual analogies through vector arithmetic}
Visual analogy methods represent relationships between images through operations in latent space. \citet{reed2015deep} learn embeddings and a generator for analogy completion using image-pair differences; \citet{thoms2023solving} apply vector arithmetic in a VAE trained on ARC, with limited success on complex tasks. \citet{bitton2022vasrvisualanalogiessituation} introduce VASR and evaluate zero-shot vector arithmetic in pretrained visual features on natural-image analogies. These studies establish precedents for both analogy-specific embedding learning and arithmetic in general-purpose pretrained spaces. We examine the latter in frozen self-supervised encoders on abstract reasoning, analyzing task identity, robustness, and relational sensitivity, then testing whether the extracted transformation signal can condition exact-output generation.

\subsection{Latent visual reasoning}
Recent work has extended multimodal language models with intermediate latent visual representations that support reasoning beyond text-based computation \citep{li2025latentvisual,li2026latentimplicit}. Complementary work has investigated abstract visual reasoning without language-mediated inference: \citet{hu2025arcvisionproblem} formulated ARC as an image-to-image translation problem, training a vision transformer from scratch on ARC tasks and adapting it to unseen tasks through test-time training. \citet{deliege2026implicit} further study task embeddings optimized at test time, separating task-code fitting from backbone adaptation. While these approaches learn or adapt models to perform visual reasoning, we investigate whether relational structure already present in pretrained visual representations can be exploited for abstract reasoning without task-specific adaptation of the visual encoder.

\subsection{Visual prompting and generative conditioning}
A complementary line of work exploits pretrained image-generation models for visual inference through prompting and conditional generation. \citet{bar2022visual} demonstrated visual in-context learning through image inpainting, while \citet{gabeur2026image} showed that pretrained image generators can be adapted for general-purpose visual perception. Subsequent studies demonstrated zero-shot perceptual capabilities without task-specific fine-tuning \citep{liu2026opensource,di2026image}. Representation autoencoders use frozen pretrained visual features as latent spaces for generation \citep{zheng2025rae,singh2026raev2}. We investigate an additional role for frozen self-supervised representations: extracting transformation representations from demonstration embeddings through vector arithmetic and using them to condition a learned generator on unseen tasks.

\section{Methods}

\subsection{Self-supervised encoders and controls}
We analyzed frozen DINOv3-Large/16 \citep{simeoni2025dinov3} and ViT-MAE-Large/16 \citep{he2021maskedautoencodersscalablevision} encoders at 224 and 480 px, recording the CLS representation after each of their 24 transformer blocks; MAE was run without masking. We focused on the Large configurations since they provide substantial representational capacity while remaining tractable for layer-wise analyses.  Three control representations isolate low-level similarity and architectural effects: pixels mapped to the encoders' 1024 dimensions by Gaussian random projection, a randomly initialized ViT with the DINOv3-Large architecture, and its untrained patch-embedding layer with mean-pooled patch tokens. Pixel and patchifier results average ten seeds. ARC grids use the standard ten-color palette and distinct padding; at 480 px, each cell occupies one patch. No parameters were fitted for these geometric analyses. Rendering details and further descriptions of the controls are in App.~\ref{sec:supp-representation-methods}.

\subsection{Bongard concept discrimination}
We evaluated the standard Bongard-LOGO test subsets and all 200 Bongard-OpenWorld problems. For each problem, one query per class was held out, centroids were formed from the remaining six images, and the two nearest-centroid decisions were averaged to obtain problem-level accuracy (Fig.~\ref{fig:concept-centroid}). For each encoder, resolution, and benchmark, the layer was selected on training problems and held fixed for testing. LOGO accuracy is the equal-weight mean over its four test subsets. Comparisons use paired permutation tests with Bonferroni correction (App.~\ref{sec:supp-representation-methods}).

\subsection{ARC intuition vectors}\label{sec:methods-intuition}
For a task with $K$ support input-output pairs $\{(x_i,y_i)\}_{i=1\ldots K}$ and frozen representation $E_l$ at layer $l$, we define the \emph{intuition vector} (Fig.~\ref{fig:concept})
\begin{equation}
v_l=\frac{1}{K}\sum_{i=1}^{K}\left[E_l(y_i)-E_l(x_i)\right]~~. \label{eqn:intuition_vector}
\end{equation}
We hypothesized that this vector implicitly encodes the transformation rule. To test this, we evaluated ARC-AGI-1 (400 training and 400 evaluation tasks), ARC-AGI-2 (1,000 training and 120 evaluation tasks), and generated instances of 794 ARC training tasks from ARC-GEN~\citep{moffitt2025arcgenmimeticproceduralbenchmark}.

\paragraph{Task classification.}
We formed single-pair intuition vector \emph{instances} ($K=1$) for 397,000 ARC-GEN instances (500 distinct pairs per task) and classified each by Euclidean distance to task centroids, under leave-one-out evaluation (Fig.~\ref{fig:concept-population}). At 224 px, we tested robustness to color and orientation by classifying recolored or rotated/reflected instances against reference sets of instances with varying augmentation diversity, including configurations in which the classified instance's color permutation or orientation was absent from its task-specific reference instances (App.~\ref{sec:supp-representation-methods}).

\paragraph{Alignment.}
We measured the angle between $v_l$ and the held-out test vector $E_l(y_*)-E_l(x_*)$ (Fig.~\ref{fig:concept-alignment}). Cross-task vectors and cell-shuffled answers, which preserve grid dimensions and color counts, served as control baselines to measure sensitivity to task and visual spatial organization.

\paragraph{Output retrieval.}
We transported the query to $\hat z=E_l(x_*)+v_l$ and ranked all candidate outputs in the corresponding dataset by Euclidean distance to $\hat z$, reporting top-1 accuracy (Fig.~\ref{fig:concept-transport}; 1,782 candidates for 419 ARC-1 evaluation queries and 526 for 167 ARC-2 evaluation queries). Ranking from $E_l(x_*)$ alone provides the baseline. Paired transport gains were compared between representations using task-level permutation tests, Bonferroni-adjusted over 12 comparisons per resolution (App.~\ref{sec:supp-representation-methods}).

On ARC-GEN, vectors from the original ARC training pairs transported 79,400 unique generated queries (100 per task), held out from the original pairs, against a shared pool of 368,442 unique outputs. We also analyzed gain across eight equal-count task bins by the fraction of canvas cells that change. Sampling and bin construction are detailed in App.~\ref{sec:supp-representation-methods}; near-miss retrieval against corrupted copies of the answer is described in App.~\ref{app:noisy_arc}.

\paragraph{Layer selection and intervals.}
\label{sec:methods-layers}
For each encoder and resolution, we selected a single transformer layer maximizing mean transported top-1 retrieval accuracy on the ARC-1 and ARC-2 evaluation sets. The selected layers were then fixed for all task classification, alignment, and retrieval analyses where they were evaluated on ARC-GEN generated instances without further layer selection.
Selected layers are listed in App.~\ref{sec:supp-representation-methods}. Unless stated otherwise, ARC intervals are $95\%$ task-bootstrap intervals; gains and angle differences retain task pairing.

\subsection{The Reasoning Representation Autoencoder}
\label{sec:methods-rae}
A Representation Autoencoder (RAE)~\citep{zheng2025rae} pairs a frozen pretrained vision encoder with a trained image decoder. A diffusion transformer (DiT) learns to generate the encoder's patch-token latents using flow matching, and the decoder maps them back to pixels. We adapt RAEv2~\citep{singh2026raev2} with DINOv3-L to few-shot ARC reasoning as the Reasoning Representation Autoencoder (RRAE) (Fig.~\ref{fig:rae2-method}). We use up to three demonstration pairs and compare three task-conditioning formats, each provided alongside the query representation:
(1) the intuition vector $\bar v$, $p_\theta(y_*\mid x_*,\bar v)$;
(2) the full demonstration grids (CLS and patch tokens), $p_\theta(y_*\mid x_*,\{(x_i,y_i)\}_{i=1}^{K})$; and
(3) a learned task token $\tau_k$~\citep{hu2026arc}, $p_\theta(y_*\mid x_*,\tau_k)$.
Here $\bar v=\frac{1}{|L|}\sum_{l\in L}v_l$ averages the intuition vectors at 224 px over $L=\{11,13,\dots,23\}$ (zero-indexed). Both CLS representations and patch-token latents average these seven layers (App.~\ref{sec:supp-rae}).

Intuition vectors identify tasks across rotations and reflections (Sec.~\ref{sec:task-identity}; Supp.~Fig.~\ref{fig:supp-orientation}). To specify the orientation of the presented grids, both single-token models receive an additional learned orientation token shared across tasks.

\paragraph{Data and training.}
We train the RRAE on ARC1 and ARC2 training data, ARC-GEN~\citep{moffitt2025arcgenmimeticproceduralbenchmark}, BARC~\citep{li2024combininginductiontransductionabstract}, and ConceptARC~\citep{moskvichev2023conceptarcbenchmarkevaluatingunderstanding} (166,503 tasks), and evaluate on 120 held-out ARC1 evaluation tasks and the 120 ARC2 evaluation tasks.

We retain RAEv2's ViT-XL decoder (416M parameters) and world-model DiT$^{\mathrm{DH}}$-XL configuration (623--624M parameters), with a further 171M parameters in the task-token model's learned task table (Table~\ref{tab:rae-params}). The decoder is first trained to reconstruct 480-px grids from frozen DINOv3 patch tokens, one per cell, using L1 pixel and palette losses, and then frozen. Its reconstruction data include grids from held-out tasks; these tasks are excluded from DiT training (App.~\ref{sec:supp-rae}). The DiT is trained from scratch to recover an answer's patch tokens from their noised version, conditioned on the query and task conditioning format. Following RAEv2~\citep{singh2026raev2}, we use representation alignment by supervising an intermediate patch-latent prediction alongside the final prediction with the same flow-matching objective in velocity space, with equal coefficients and twice the cell weight on non-dominant colors. We additionally predict the target's layer-averaged CLS representation using a cosine loss weighted by 0.1. Each training example draws one query--target pair from all pairs of a training task, including its original test pairs, and uses up to three remaining pairs as demonstrations. A shared random rotation/reflection and color permutation augment every grid. Architecture, objectives, data splits, and optimization are detailed in App.~\ref{sec:supp-rae}.

\paragraph{Inference.}
We generate answers from Gaussian noise in two diffusion steps under 32 views (8 rotations/reflections $\times$ 4 color permutations), invert the augmentations, and aggregate the answers. The two attempts are the cell-wise majority vote and the most frequent grid excluding the first attempt; either must exactly match the target. We also report cell accuracy of the first attempt. Sampling-step and view-count analyses appear in Supp.~Figs.~\ref{fig:arc-rae-sampler-calls} and~\ref{fig:arc-rae-view-voting}.

\paragraph{Test-time training (TTT).}
Following~\citep{akyurek2025surprisingeffectivenesstesttimetraining,hu2026arc}, we fine-tune all DiT weights on the task's demonstrations and reset them afterwards. Each fitting example holds out one demonstration as a pseudo-query, uses up to three remaining demonstrations as support, and applies one of the 32 augmentations. Each Adam step uses two examples. The intuition vector is recomputed from their augmented supports; the task-token model jointly fits its zero-initialized token. The test answer is never used for fitting. TTT results average three seeds.

\paragraph{Model comparisons.}
We compare conditioning variants at approximately matched GPU training time (epochs 192, 48, and 192 for intuition vectors, full grids, and task tokens); the single-token models use about $24\%$ fewer training FLOPs (App.~\ref{sec:supp-rae}). Two-sided sign-flip tests compare per-task exact-match scores averaged over test queries and TTT seeds (20,000 permutations).

\section{Results}

\subsection{Embedding-space neighborhoods discriminate abstract concepts}
\label{sec:concepts}
\begin{figure}[t]
    \centering
    \captionsetup[subfigure]{skip=1pt}
    \begin{subfigure}[t]{\linewidth}
        \centering
        \includegraphics[width=\linewidth]{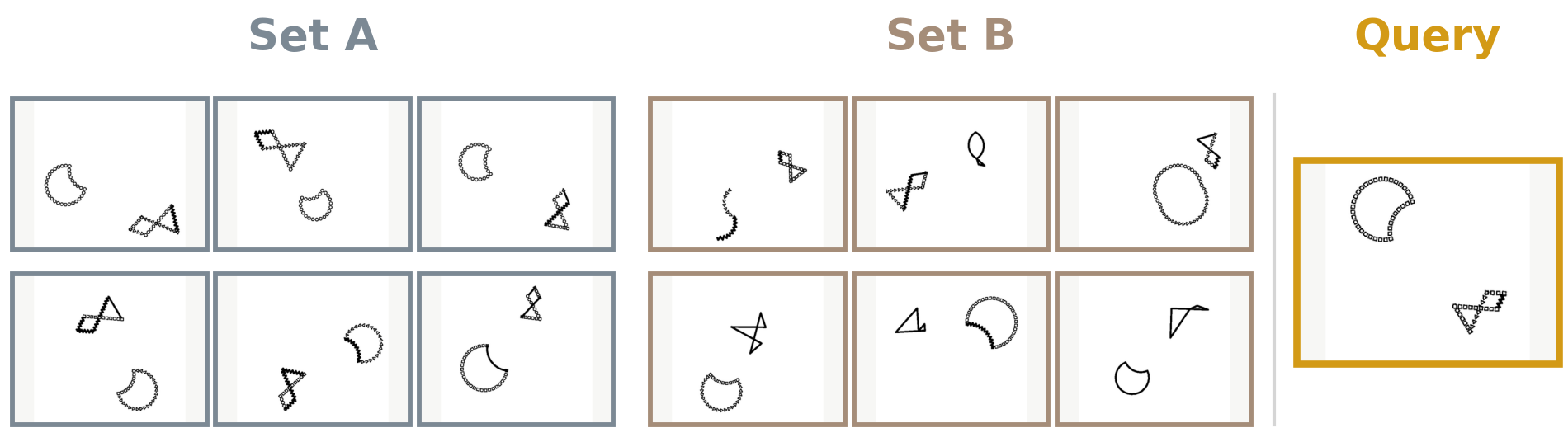}
        \caption{Bongard-LOGO problem.}
        \label{fig:bongard2-logo-example}
    \end{subfigure}

    \vspace{0.2em}
    \begin{subfigure}[t]{\linewidth}
        \centering
        \includegraphics[width=\linewidth]{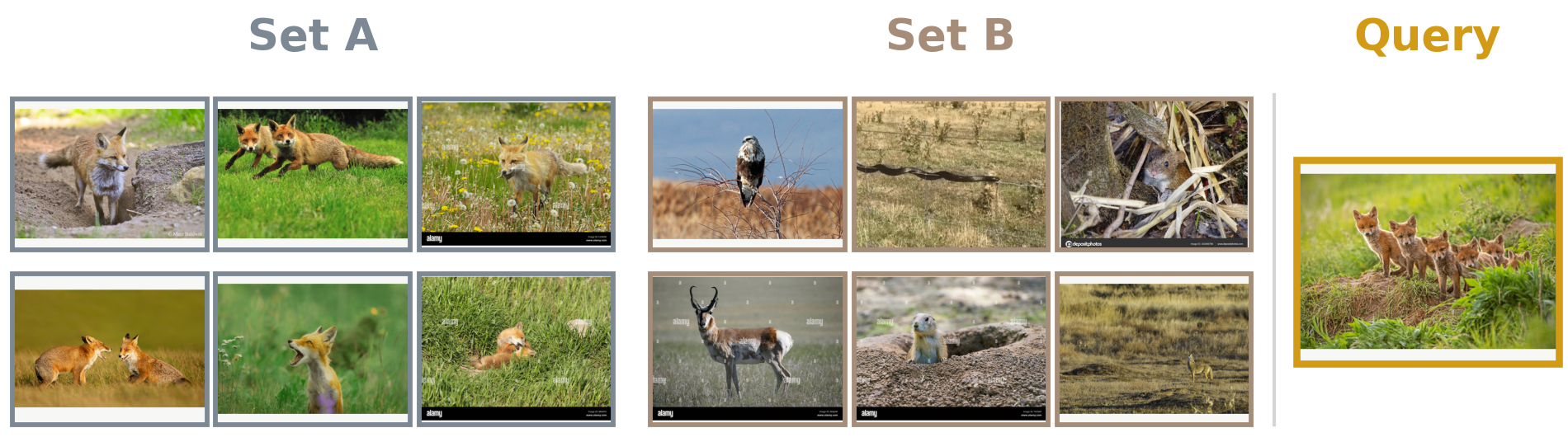}
        \caption{Bongard-OpenWorld problem.}
        \label{fig:bongard2-ow-example}
    \end{subfigure}

    \vspace{0.2em}
    \begin{subfigure}[t]{\linewidth}
        \centering
        \includegraphics[width=\linewidth]{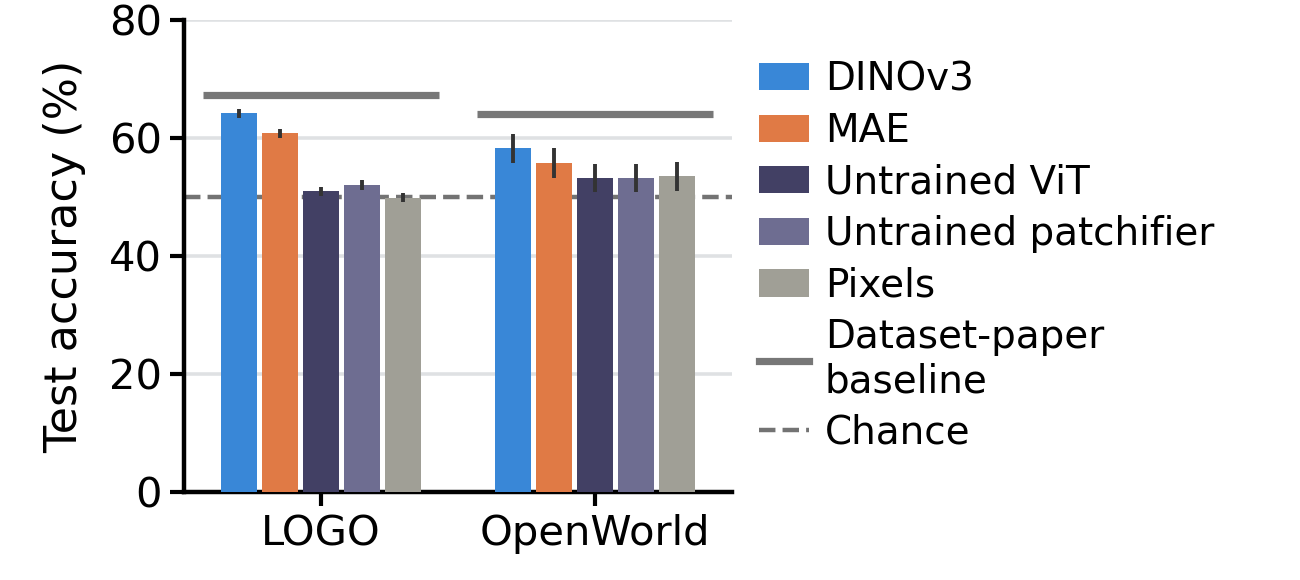}
        \caption{Nearest-centroid test accuracy, 224 px.}
        \label{fig:bongard2-bars}
    \end{subfigure}
    \caption{
        Concept discrimination in frozen representations.
        \textbf{(a,b)} Each problem gives six images per set; a held-out query is assigned to the
        nearer set centroid (Fig.~\ref{fig:concept-centroid}).
        \textbf{(c)} Test accuracy at the layer selected on training problems. LOGO: equal-weight
        mean of its four test subsets; OpenWorld: 200 problems. Gray segments: dataset-paper
        baselines; dashed: chance. Error bars: SEM across problems (pixels and patchifier: mean
        SEM over ten seeds).
    }
    \label{fig:bongard2}
\end{figure}
Bongard problems provide a direct test of whether relative position in a frozen embedding space supports abstract classification. For each problem, we form positive and negative support centroids and assign each held-out panel to the nearer centroid (Fig.~\ref{fig:concept-centroid}). At 224 px, with layers selected on training problems, DINOv3 gives the strongest results: Bongard--LOGO reaches $64.2\%$ accuracy and Bongard-OpenWorld $58.2\%$; MAE reaches $60.8\%$ and $55.8\%$, respectively (Fig.~\ref{fig:bongard2-bars}). Pixels, the untrained ViT, and the untrained patchifier remain close to chance. On LOGO, every trained encoder exceeds every control (paired permutation tests, Bonferroni-adjusted $p<.001$). On OpenWorld at 224 px, DINOv3 and MAE exceed chance ($p=.001$ and $p=.03$) while the controls do not ($p\geq.14$). Results at 480 px are similar; resolution comparisons, detailed control scores, and layer profiles are reported in App.~\ref{sec:supp-geometric-results}.

Together, these results show that pretrained representation geometry exposes abstract concept information beyond pixel similarity or architectural effects on synthetic drawings, and above chance on natural images.

\begin{figure*}[!t]
    \centering
    \captionsetup[subfigure]{skip=1pt}
    \includegraphics[width=0.8\textwidth]{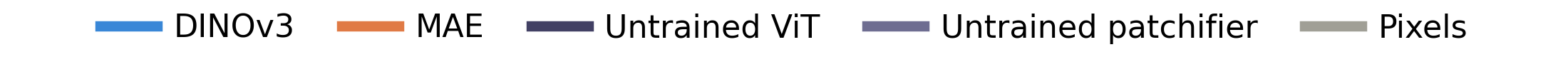}

    \vspace{0.1em}
    \begin{subfigure}[t]{0.29\textwidth}
        \centering
        \includegraphics[width=\linewidth]{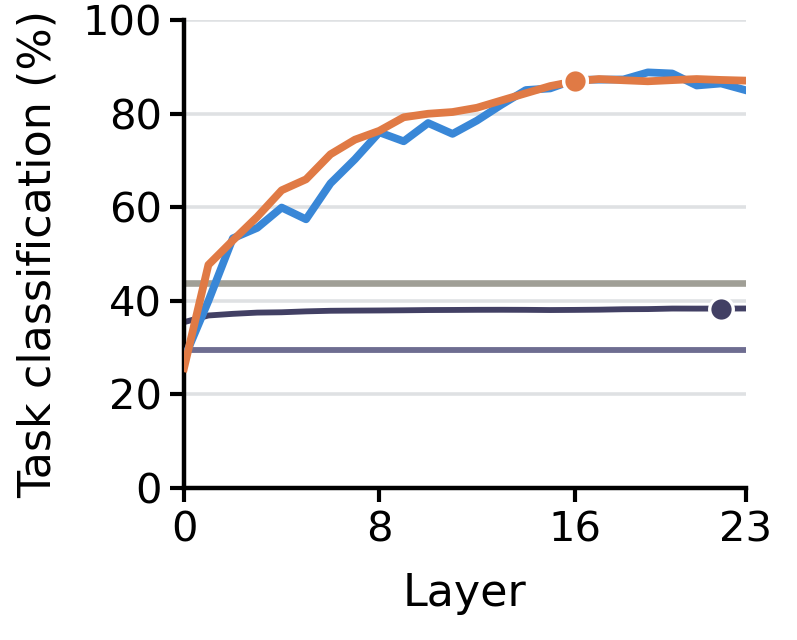}
        \caption{Task classification by layer.}
        \label{fig:task-id-layer}
    \end{subfigure}
    \hfill
    \begin{subfigure}[t]{0.29\textwidth}
        \centering
        \includegraphics[width=\linewidth]{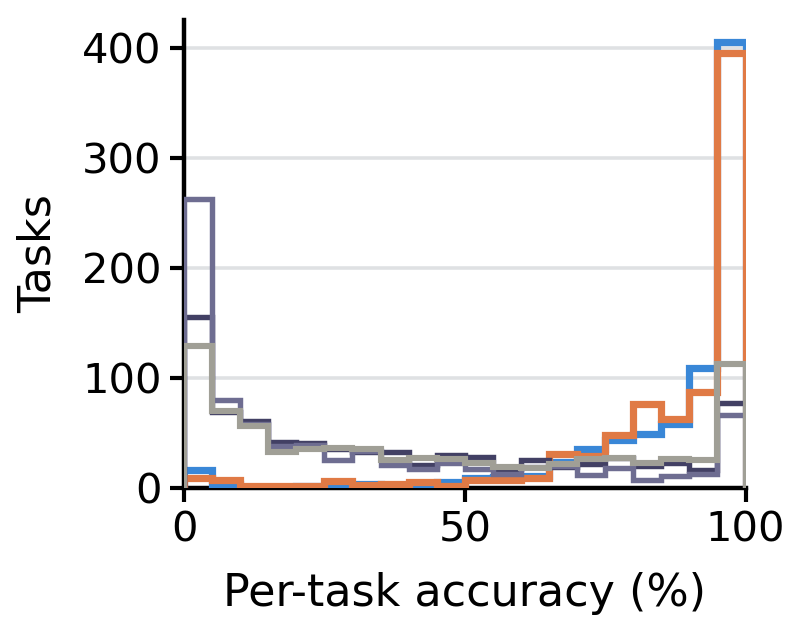}
        \caption{Per-task accuracy.}
        \label{fig:task-id-histogram}
    \end{subfigure}
    \hfill
    \begin{subfigure}[t]{0.40\textwidth}
        \centering
        \includegraphics[width=\linewidth]{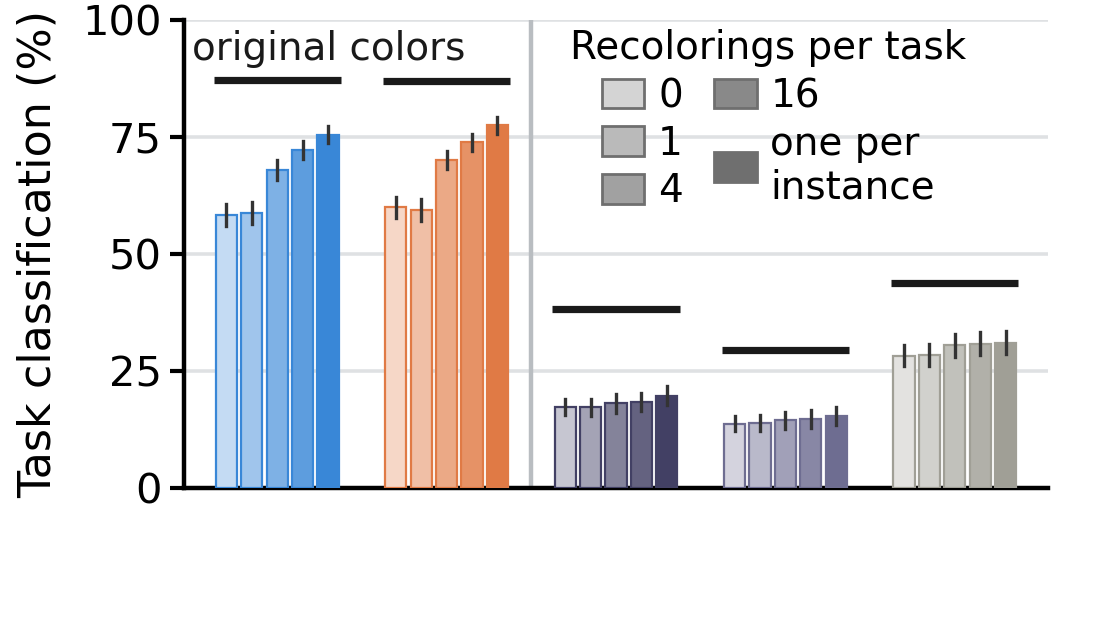}
        \caption{Recolored queries and references sets.}
        \label{fig:task-id-color}
    \end{subfigure}
    \caption{
        Single-pair intuition vectors identify their task (ARC-GEN, 794 tasks $\times$ 500
        instances, 224 px; chance $0.13\%$).
        \textbf{(a)} Leave-one-out nearest-centroid accuracy across layers; markers: the layers
        used throughout (Sec.~\ref{sec:methods-layers}).
        \textbf{(b)} Per-task accuracy at those layers.
        \textbf{(c)} Each classified instance is recolored by a permutation of the nine non-black
        colors, and classified against each task reference set recolored with 0--16 permutations, or one independently sampled per reference instance, so that color carries no task information. Black lines: original colors.
        Rotations and reflections: Supp.~Fig.~\ref{fig:supp-orientation}.
        Pixel and patchifier values average ten seeds; whiskers: $95\%$ task-bootstrap intervals.
    }

    \label{fig:task-id}
\end{figure*}

\subsection{Intuition vectors identify tasks beyond color and orientation}
\label{sec:task-identity}

We next ask whether relational geometry can encode transformations rather than only static concepts. We represent the transformation in a single ARC input-output pair by its output$ - $input embedding difference i.e. its intuition vector instance (Eqn.~\ref{eqn:intuition_vector}, $K=1$). We first ask an across-task question: whether these single-pair vectors preserve task identity across many instances and hundreds of competing tasks. We tested this using leave-one-out nearest-centroid classification of 397,000 intuition vector instances from 794 ARC-GEN tasks \citep{moffitt2025arcgenmimeticproceduralbenchmark}, whose success requires within-task consistency relative to between-task separation (Fig.~\ref{fig:concept-population}). At the selected layers (Sec.~\ref{sec:methods-layers}) and 224 px, accuracy reached $87.2\%$ and $86.9\%$ for DINOv3 and MAE, compared with $38.3\%$ for the untrained ViT, $29.4\%$ for the untrained patchifier, and $43.9\%$ for pixels (chance $0.13\%$; Fig.~\ref{fig:task-id-histogram}). At 480 px, pretrained accuracy reached $73.4\%$--$77.3\%$, versus $21.7\%$--$43.9\%$ for the controls (Supp.~Fig.~\ref{fig:supp-arcgen-classification-480}).

The pretrained advantage emerges with depth: both encoders start at or below the controls ($25.6\%$--$27.6\%$ at the earliest layer at 224 px) and rise through the network, whereas the untrained ViT stays between $35.4\%$ and $38.4\%$ at every layer (Fig.~\ref{fig:task-id-layer}).

Because many ARC tasks involve specific colors, this signal could partly reflect color identity. We therefore recolored every instance with a permutation of the nine non-black colors (which is usually background), applied to input and output alike (Fig.~\ref{fig:task-id-color}; Supp.~Fig.~\ref{fig:color-permutation}). Recoloring the classified instance lowered accuracy in every representation, to $58.3\%$--$60.0\%$ for the pretrained encoders and $13.8\%$--$28.3\%$ for the controls. We then recolored the reference instances as well, using independently sampled permutations for each instance so that color carries no task information. This restored $17.2$--$17.5$ points for the pretrained encoders, bringing accuracy to $75.5\%$--$77.5\%$, but only $1.6$--$2.8$ points for the controls. The pretrained encoders thus remained above every control even when the controls retained their original colors (Fig.~\ref{fig:task-id-color}).

Rotating or reflecting classified and reference instances gives the same separation. With every reference instance in one of the seven orientations other than the classified instance's, the pretrained encoders reach $83.8\%$--$84.3\%$, against $24.5\%$--$32.9\%$ for the controls (Supp.~Fig.~\ref{fig:supp-orientation}). Together, the depth and augmentation results indicate that pretrained intuition vector instances encode transformation features shared across task instances, beyond consistent color patterns and orientation.

\begin{figure*}[!t]
    \centering
    \captionsetup[subfigure]{skip=1pt}
    \includegraphics[width=0.8\textwidth]{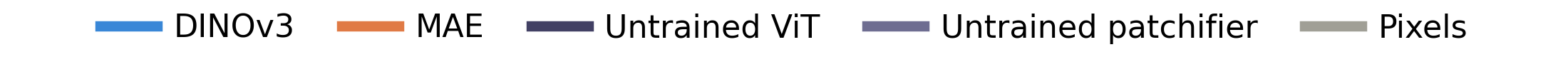}

    \vspace{0.1em}
    \begin{subfigure}[t]{0.27\textwidth}
        \centering
        \includegraphics[width=\linewidth]{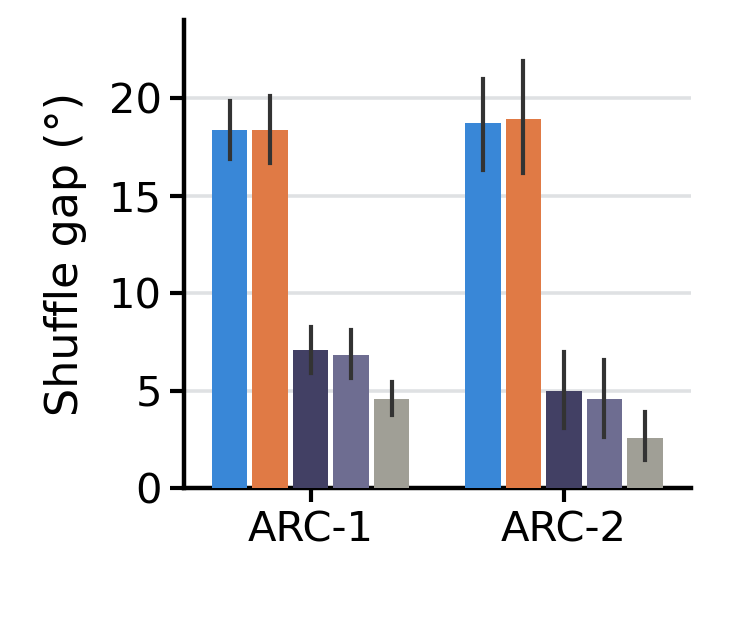}
        \caption{Sensitivity to spatial layout.}
        \label{fig:transport-gap}
    \end{subfigure}
    \hfill
    \begin{subfigure}[t]{0.38\textwidth}
        \centering
        \includegraphics[width=\linewidth]{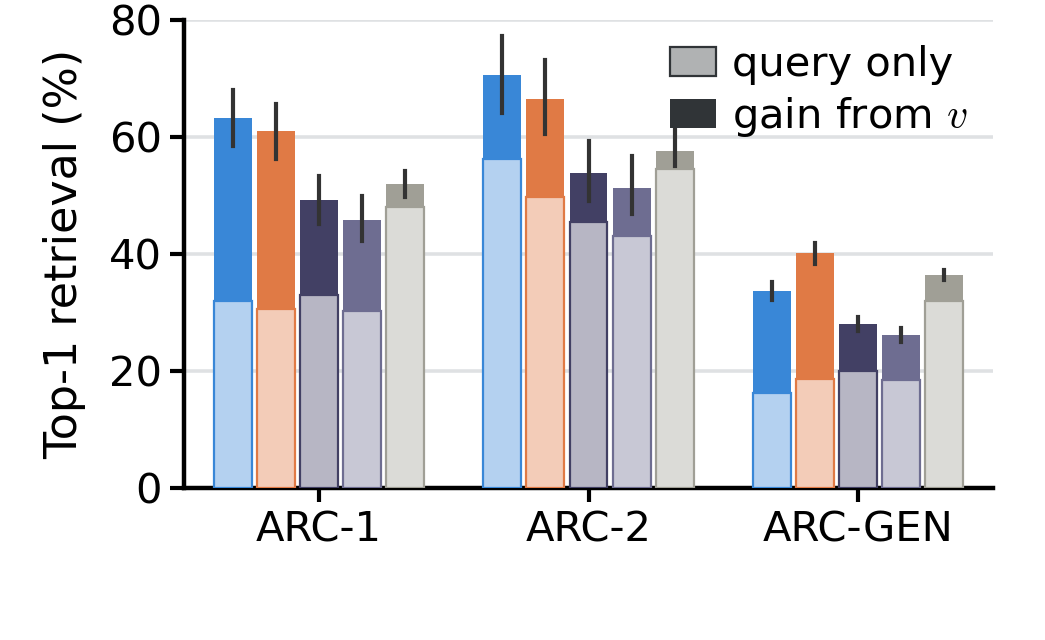}
        \caption{Exact-output retrieval.}
        \label{fig:transport-retrieval}
    \end{subfigure}
    \hfill
    \begin{subfigure}[t]{0.33\textwidth}
        \centering
        \includegraphics[width=\linewidth]{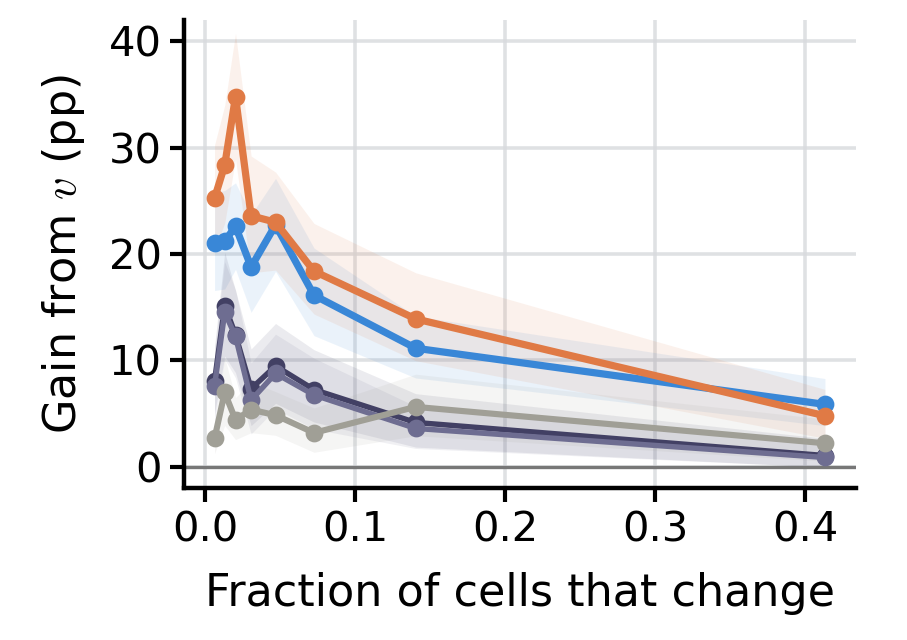}
        \caption{Gain by transformation extent.}
        \label{fig:transport-by-change}
    \end{subfigure}
    \caption{
        Intuition vectors point toward the answer and guide its retrieval (224 px; layers as in
        Fig.~\ref{fig:task-id}).
        \textbf{(a)} Increase in the angle between the intuition vector and the test vector when
        the answer's cells are shuffled, which keeps its size and color counts (evaluation tasks;
        mean over ten shuffles).
        \textbf{(b)} Top-1 retrieval among all outputs of the dataset from the query alone
        (light) and after adding the intuition vector (full bar; dark: gain).
        \textbf{(c)} Gain against the fraction of canvas cells that change (ARC-GEN, eight
        equal-count bins; shading: $95\%$ intervals).
        Whiskers: $95\%$ task-bootstrap intervals of the paired difference.
    }
    \label{fig:transport}
\end{figure*}

\subsection{Intuition vectors capture relational structure and guide retrieval}
\label{sec:alignment}

The preceding analysis asks an across-task question. Here, we instead ask, within individual tasks, whether an intuition vector aligns with a test transformation and whether transporting a query along that vector improves retrieval of its output.

On both ARC-1 and ARC-2, intuition vectors align with the matched test vectors of their own task, whereas cross-task vectors are near-orthogonal. The untrained ViT and patchifier align as closely as the trained encoders, so alignment alone does not require training (App.~\ref{sec:supp-geometric-results}). Crucially, when we break the spatial structure of the test output by shuffling its cells, which preserves grid dimensions and color counts, the trained encoders separate the shuffled from the original test vectors by $15.2^\circ$--$18.9^\circ$, against $4.6^\circ$--$7.1^\circ$ for the untrained controls and $2.5^\circ$--$4.6^\circ$ for pixels (Fig.~\ref{fig:transport-gap}), a sensitivity that grows with depth only in the trained encoders (Supp.~Figs.~\ref{fig:supp-arc-angle-arc1} and~\ref{fig:supp-arc-angle-arc2}).

To test whether this structure can guide output selection, we add the intuition vector to the query embedding and rank every output in the same dataset by distance to the transported query, reporting top-1 accuracy (Fig.~\ref{fig:concept-transport}). On the ARC-1 and ARC-2 evaluation sets, the trained encoders retrieve best ($61.1\%$--$63.2\%$ and $66.5\%$--$70.7\%$ at 224 px; Fig.~\ref{fig:transport-retrieval}). Retrieval from the query alone does not favor the trained encoders; they instead gain most from transport, $30.5$--$31.3$ points on ARC-1 and $14.4$--$16.8$ on ARC-2, against $15.7$--$16.2$ and $8.3$--$8.4$ for the untrained controls and $3.9$ and $3.1$ for pixels (ARC-1: Bonferroni-adjusted $p<.001$; on ARC-2 the advantage over the untrained controls is significant only at 480 px; Supp.~Fig.~\ref{fig:supp-arc-retrieval-480}).

The same holds at scale on ARC-GEN (79,400 generated queries from 794 tasks, 368,442 candidates, layers fixed from the ARC evaluation sets): transport adds $17.4$ and $21.5$ points for DINOv3 and MAE, versus $8.1$, $7.6$, and $4.4$ for the untrained ViT, the untrained patchifier, and pixels (Fig.~\ref{fig:transport-retrieval}). In this denser pool, pixels retrieve best from the query alone and remain comparable to the trained encoders after transport. The transport gain in pretrained encoders is not confined to transformations that leave the output similar to the query (Fig.~\ref{fig:transport-by-change}). In ARC-1/2 near-miss pools, where each answer competes with 99 randomly recolored variants, intermediate DINOv3 layers outperform pixels by 11--22 points at 480 px once at least 5\% of cells are recolored (App.~\ref{app:noisy_arc}; Supp.~Fig.~\ref{fig:supp-arc-knn-noisy}).

Thus, trained intuition vectors capture transformation structure beyond size and color statistics and guide output selection.

Retrieval selects among available candidate grids; generating an answer requires translating the task-level signal into query-specific changes. Exploratory single-cell analyses in DINOv3 suggest a contextual challenge for this step: translation and recoloring compose approximately additively within a shared grid, but their embedding directions change with the surrounding cells (App.~\ref{app:low_level}), motivating learned, context-dependent execution. We next test whether the intuition vector and query together can condition generation of the exact output.

\begin{figure*}[!t]
    \centering
    \captionsetup[subfigure]{skip=1pt}
    \begin{subfigure}[t]{\textwidth}
        \centering
        \includegraphics[width=\linewidth]{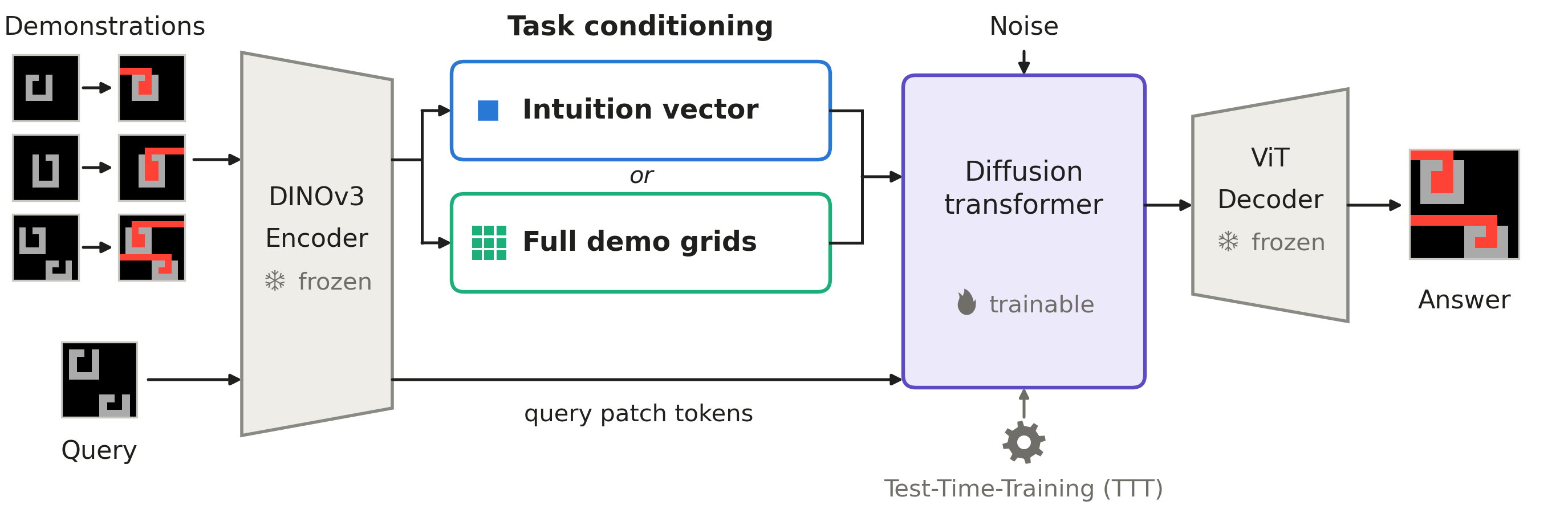}
        \caption{Intuition-vector-conditioned RAE.}
        \label{fig:rae2-method}
    \end{subfigure}

    \vspace{0.2em}
    \includegraphics[width=0.9\textwidth]{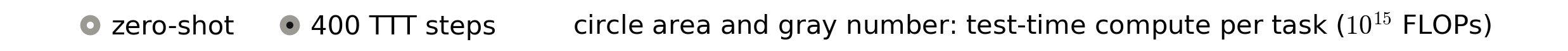}

    \begin{subfigure}[t]{0.325\textwidth}
        \centering
        \includegraphics[width=\linewidth]{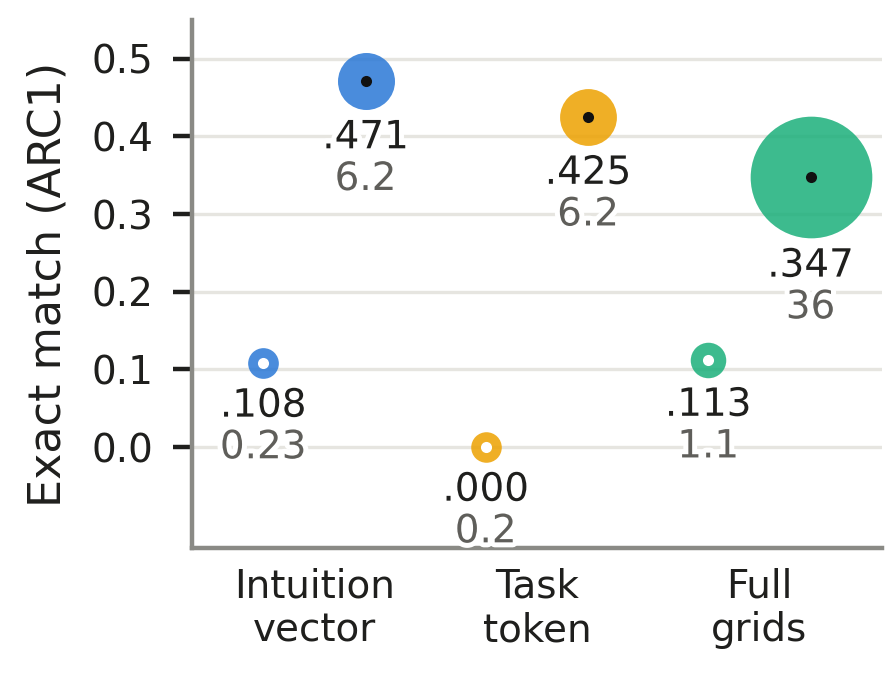}
        \caption{ARC1 held-out, exact match.}
        \label{fig:rae2-arc1}
    \end{subfigure}
    \hfill
    \begin{subfigure}[t]{0.325\textwidth}
        \centering
        \includegraphics[width=\linewidth]{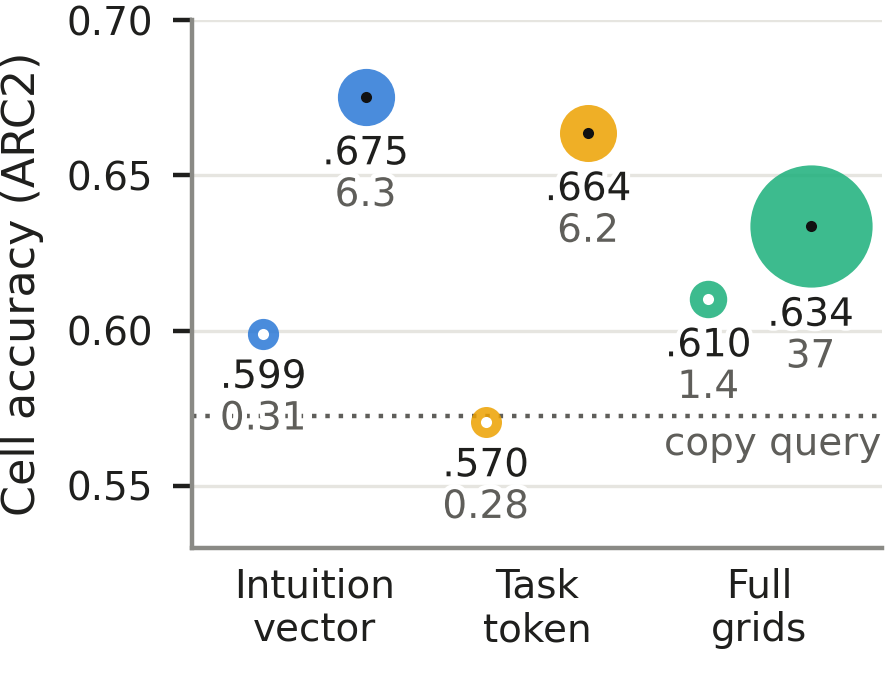}
        \caption{ARC2 evaluation, cell accuracy.}
        \label{fig:rae2-arc2}
    \end{subfigure}
    \hfill
    \begin{subfigure}[t]{0.325\textwidth}
        \centering
        \includegraphics[width=\linewidth]{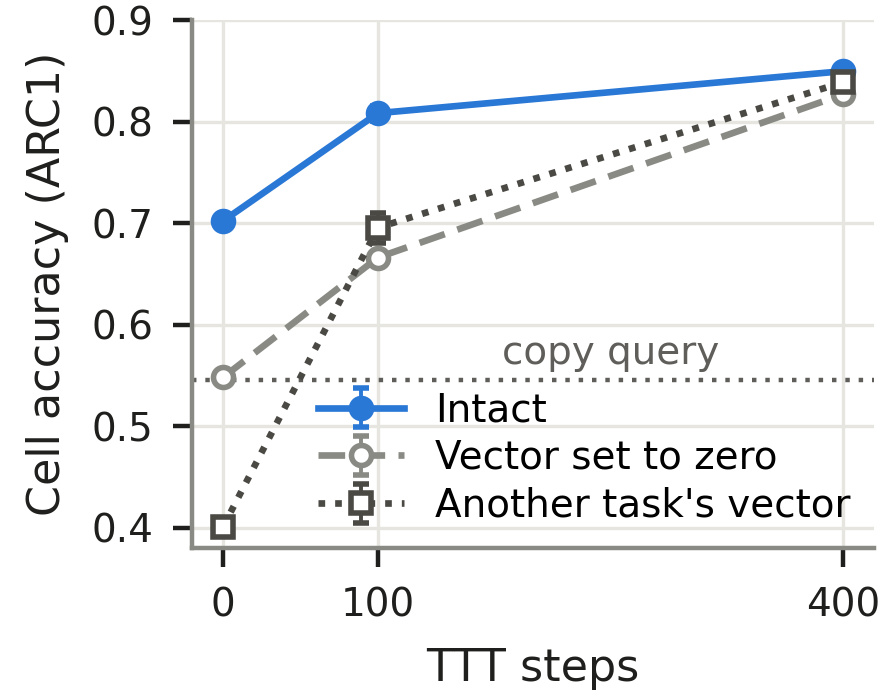}
        \caption{The vector carries the task.}
        \label{fig:rae2-ablation}
    \end{subfigure}
    \caption{
        Intuition-vector-conditioned generation.
        \textbf{(a)} A frozen DINOv3 encoder and ViT decoder surround a trained diffusion
        transformer, conditioned on the query's patch tokens and either the intuition
        vector (one token) or the full demonstration grids (5,406 tokens for three pairs), in the
        colors of (b--d). Gear: test-time training (TTT) of the diffusion transformer on the
        demonstrations; answers are voted over 32 views.
        \textbf{(b,c)} Score per conditioning at approximately matched GPU training time, with a learned task-token
        control. Center dot: zero-shot (white) or 400 TTT steps (black; mean of three TTT seeds);
        circle area and gray numbers: test-time compute per task. Dotted: copying the query.
        \textbf{(d)} The intuition-vector model with its vector intact, set to zero, or taken
        from another task.
    }
    \label{fig:rae2}
\end{figure*}

\subsection{Intuition vectors steer generation}
\label{sec:rae-results}
We next ask whether the intuition vector can be used to steer generation of the query's exact answer. We test this with the proposed Reasoning Representation Autoencoder (RRAE; Fig.~\ref{fig:rae2-method}, Sec.~\ref{sec:methods-rae}), comparing conditioning by the intuition vector (one token), full demonstration grids (5,406 tokens for three pairs), or a learned, and thus domain-specific, task token. We compare checkpoints at approximately matched GPU training time (epochs 192, 48, and 192, respectively); the single-token models use about $24\%$ fewer training FLOPs (App.~\ref{sec:supp-rae}). We evaluate generation zero-shot and after 400 steps of test-time training (TTT).
Zero-shot, the intuition-vector and full-grid RRAEs solve a similar fraction of the ARC1 held-out tasks ($10.8\%$ and $11.3\%$; paired permutation test over tasks, $p=1$), although the full-grid answers are closer cell-wise ($77.9\%$ versus $70.2\%$ cell accuracy), and on ARC2 both slightly exceed copying the query ($59.9\%$ and $61.0\%$ versus $57.2\%$ cell accuracy, with at most $0.8\%$ exact matches; Fig.~\ref{fig:rae2-arc1},~\ref{fig:rae2-arc2}). The task-token RRAE has, by construction, no code for an unseen task and stays at the level of copying the query (task token is zero initialized).

After TTT, the intuition-vector RRAE reaches $47.1\%$ on ARC1, against $42.5\%$ for the task token ($p=.05$) and $34.7\%$ for full grids ($p<.001$), and $4.4\%$, $4.0\%$, and $1.7\%$ exact matches on ARC2 ($67.5\%$, $66.4\%$, and $63.4\%$ cell accuracy), at about one-sixth of the full-grid conditioning model's test-time compute ($6.2$ versus $36\times10^{15}$ FLOPs per task). Longer adaptation does not close the gap to the task token ($50.3\%$ versus $42.1\%$ after 1,600 steps; Supp.~Fig.~\ref{fig:arc-rae-ttt-budget}). The lead over full grids narrows with further training: at twice the compute (epochs 384 and 96; the task-token run ended at epoch 192), the intuition-vector and full-grid RRAEs reach $49.3\%$ and $46.9\%$ on ARC1 after TTT ($p=.52$), $14.2\%$ and $19.2\%$ zero-shot ($p=.17$), and $5.3\%$ and $3.3\%$ on ARC2 after TTT (Supp.~Fig.~\ref{fig:arc-rae}).
The RRAE follows the intuition vector (Fig.~\ref{fig:rae2-ablation}; Supp.~Fig.~\ref{fig:supp-rae-ablation}). With the vector set to zero, the intuition-vector RRAE returns its query unchanged: $88\%$ of its zero-shot ARC1 outputs are exact copies of the query. With another task's vector it does edit the query ($78\%$ of answers differ from it), but wrongly, and falls below copying ($40.1\%$ cell accuracy, versus $54.5\%$ for a copy and $70.2\%$ with the correct vector). These are the generative counterparts of the query-only and cross-task controls of the previous sections: the output depends on the content of a single token, not merely on its presence. TTT on the demonstrations can stand in for a missing or wrong vector, but more slowly: after 100 steps the ablated models solve $17\%$--$19\%$ of the ARC1 tasks against $33.3\%$ intact, and after 400 steps $41\%$ against $47.1\%$, the level of the task-token conditioning approach ($42.5\%$).

Together, a single vector that requires no training replaces thousands of demonstration tokens: it gives comparable exact-match performance to full-grid conditioning at similar GPU training time and a fraction of the test-time compute, and, unlike a learned task token, applies to unseen tasks without adaptation. With TTT, it suffices to solve close to half of the held-out ARC1 tasks exactly. With ARC2 still largely unsolved by all three models, and a single training run per model, we present these results as a proof of concept that the task-level signal of frozen pretrained representations can steer generation on abstract reasoning tasks.

\section{Discussion}
\label{sec:discussion}

These results support a modular approach in which frozen visual representations supply a representation of task-level transformation and a learned executor combines it with the query to generate an answer. The signal is extracted by averaging embedding differences, without learning a specialized task token or adapting the encoder to the task. Its usefulness for both discrimination and generation suggests that pretrained relational structure can contribute directly to inference, beyond serving as a perceptual input or generative latent space. This motivates treating transformation representations as a reusable interface between demonstrations and execution.

\paragraph{Limitations.}
Computational cost restricted each RRAE conditioning variant to one training run leaving variability across independent training runs unmeasured. Generation was tested only with DINOv3 on ARC, with limited ARC2 exact-match performance. Thus, transfer across encoders and domains remains unproven.

Although intuition-vector conditioning reduces compute relative to full demonstrations, the complete RRAE remains large. RAE identifies a width bottleneck in standard DiTs: effective diffusion in pretrained representation spaces requires the model width to match or exceed the encoder dimension \citep{zheng2025rae}. Our implementation retains RAEv2's world-model backbone dimensions \citep{singh2026raev2} and optimization has not been attempted. Smaller, domain-specific models \citep{jolicoeurmartineau2025morerecursivereasoningtiny,hu2026arc} motivate more efficient implementations, although differences in data and evaluation protocols preclude direct performance comparisons.

\paragraph{Future directions.}
Future work could use intuition vectors to condition smaller recursive networks or vision transformers, or learn compact adapters that permit diffusion in lower-dimensional pretrained representations while preserving transformation information and exact reconstruction. These strategies target executor cost and latent dimensionality, respectively.

\section{Conclusion}
\label{sec:conclusion}

Frozen pretrained visual representations contain relational structure that can be extracted from demonstrations through simple vector differences. The extracted intuition vectors identify tasks across instances, guide output retrieval, and condition exact generation. Unlike learned task codes, they are directly available for unseen tasks without task-specific fitting, while their compact form reduces computation relative to full-demonstration conditioning in our generative comparisons. Using DINOv3 and RAEv2, we establish a proof of concept for modular visual reasoning in which pretrained representations supply a transformation signal to a learned executor, motivating more efficient implementations and evaluation across broader visual domains.

{
    \small
    \bibliographystyle{ieeenat_fullname}
    \bibliography{main}
}

\clearpage
\maketitlesupplementary

\appendix
\setcounter{figure}{0}
\renewcommand{\thefigure}{S\arabic{figure}}
\def\theHfigure{S\arabic{figure}} 
\section{Appendix}
\subsection{Representation analyses: implementation and evaluation}
\label{sec:supp-representation-methods}

\paragraph{Rendering and controls.}
The geometric analyses use the CLS representation after each of 24 transformer blocks, with no benchmark-specific parameter fitting. ARC grids are centered on a distinct padding color: the 224-px rendering uses a $32\times32$ canvas with 7-px cells, and the 480-px rendering a $30\times30$ canvas with 16-px cells. For pixel controls, the encoders' standard channel-wise normalization is applied to each RGB canvas before flattening and fixed Gaussian projection to 1024 dimensions, averaged over ten projection seeds. The untrained ViT uses the DINOv3-Large architecture with one random initialization, read out like the trained encoders. The untrained patchifier uses only that architecture's patch-embedding layer, with no transformer blocks; its mean patch token replaces the CLS readout, and results average ten initialization seeds.

\paragraph{Bongard layer selection and tests.}
Layers are selected independently for each encoder, resolution, and benchmark using training problems: all 610 OpenWorld problems, and 600 randomly sampled problems from each of LOGO's three training types (1,800 of 9,300), selecting on their mean. Test LOGO accuracy is the equal-weight mean over its four test subsets. Two-sided sign-flip permutation tests compare problem-level accuracies (100,000 permutations; for LOGO, the equal-weight mean over subsets). Each trained encoder is compared with each control on identical problems and queries, Bonferroni-adjusted over 24 comparisons (two trained encoders, three controls, two benchmarks, two resolutions). Comparisons against $50\%$ chance use an exact sign test, since every nonzero problem-level deviation has the same size.

\paragraph{ARC-GEN color and orientation tests.}
These tests use the 397,000-instance classification reference set at 224 px. Each input--output pair is recolored by a shared permutation of the nine non-black colors and classified against reference sets recolored with 0, 1, 4, or 16 permutations per task, or independently sampled permutations for each reference instance. The classified instance's permutation is always absent from its own task's reference examples. For orientation, each instance is classified under each of the seven non-identity rotations/reflections against reference sets whose instances take one of 1 (the original), 2, 4, or 7 orientations per task, always including the original and excluding the classified instance's orientation, or one of all 8 orientations. Accuracy is averaged across the seven query orientations. Both tests use the same leave-one-out Euclidean readout on unnormalized vectors as the original classification.

\paragraph{Alignment and retrieval.}
Cross-task and cell-shuffled alignment controls each average ten randomizations. Transport gains are compared with two-sided permutation tests that flip the sign of each task's summed paired difference between representations (100,000 permutations), Bonferroni-adjusted over 12 comparisons at each resolution (two trained encoders, three controls, two datasets). For ARC-GEN retrieval, 100 unique generated query-answer pairs per task are sampled after globally excluding pairs matching any original ARC pair. Intuition vectors use the original ARC training pairs, and the candidate pool contains 368,442 unique output grids. The eight equal-count task bins use the fraction of canvas cells that change, computed from the remaining unique generated pairs.

\paragraph{ARC layers and intervals.}
The layer maximizing mean transported top-1 retrieval over the ARC-1 and ARC-2 evaluation sets is fixed for task classification, its color/orientation tests, alignment, and retrieval on ARC and ARC-GEN. The selected zero-indexed layers for DINOv3, MAE, and the untrained ViT are 16, 16, and 22 at 224 px, and 15, 18, and 7 at 480 px. Bongard layers are selected separately on training problems; pixels and the patchifier have no layers. Unless stated otherwise, ARC intervals are $95\%$ percentile intervals from 2,000 task-bootstrap resamples; transport gains and angle differences retain their paired per-task differences.

\subsection{Additional concept-discrimination and retrieval results}
\label{sec:supp-geometric-results}

\paragraph{Bongard resolutions, layers, and controls.}
At 224 px, DINOv3 gives the strongest results: Bongard--LOGO reaches $64.2\%$ accuracy at layer 23 (SEM $=0.8$ percentage points), $3.0$ points below the $67.2\%$ dataset-paper reference \citep{nie2020bongardlogo}, and Bongard-OpenWorld reaches $58.2\%$ at layer 23 (SEM $=2.5$ points), $5.8$ points below its $64.0\%$ reference \citep{wu2024bongardopen}. These are references from the original dataset papers rather than current state-of-the-art comparisons; the LOGO reference is the equal-weight mean of the per-subset maxima over different baselines, which no single baseline reaches. MAE reaches $60.8\%$ on LOGO at layer 19 and $55.8\%$ on OpenWorld at layer 18. Results at 480 px are similar (DINOv3: $63.0\%$ on LOGO and $59.0\%$ on OpenWorld; MAE: $58.0\%$ and $55.8\%$; Supp.~Fig.~\ref{fig:supp-bongard-480}); complete layer profiles appear in Supp.~Figs.~\ref{fig:supp-bongard-layerwise} and~\ref{fig:supp-bongard-layerwise-ow}.

Pixel accuracy remains near chance, ranging from $49.5\%$ to $49.9\%$ on LOGO and from $53.5\%$ to $54.3\%$ on OpenWorld across the two resolutions. The untrained ViT scores $51.0\%$--$51.2\%$ on LOGO and $52.2\%$--$53.2\%$ on OpenWorld, and the untrained patchifier $52.1\%$ and $53.2\%$. On LOGO, every trained encoder exceeds every control by $8.7$ to $14.3$ points at 224 px (paired permutation tests, Bonferroni-adjusted $p<.001$). Subtask breakdowns appear in Supp.~Fig.~\ref{fig:supp-bongard-subtasks}.

\begin{figure*}[p]
    \centering
    \includegraphics[width=0.98\textwidth]{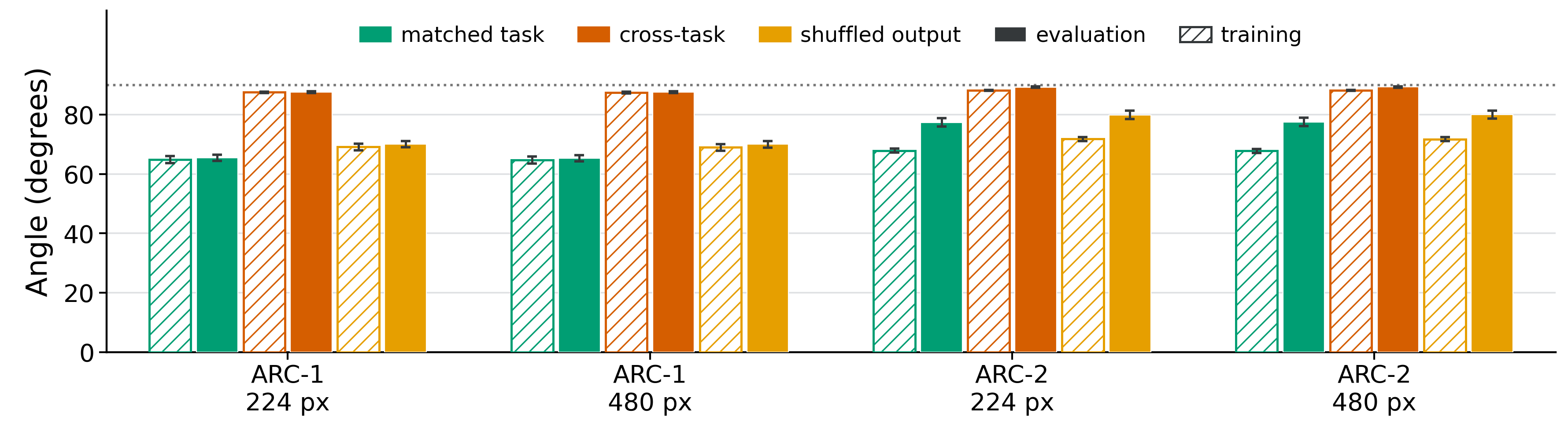}
    \caption{
        Intuition--test-vector angles in random-projection pixel space for ARC-1
        and ARC-2 at 224 and 480 px. Pixel representations have no layer axis and
        are therefore shown categorically. Green bars show matched tasks, red
        bars cross-task controls, and orange bars cell-shuffled answers. Filled
        bars denote evaluation tasks and open hatched bars training tasks. Bars
        average ten projection seeds; error bars combine the mean within-seed
        task SEM with uncertainty across projection seeds. Results at 224 and
        480 px differ by at most $0.2^\circ$.
    }
    \label{fig:supp-arc-angle-pixel}
\end{figure*}

\paragraph{Matched and cross-task alignment.}
On both ARC-1 and ARC-2, in every representation and at both resolutions, intuition vectors align with the matched test vectors of their own task, whereas cross-task vectors are near-orthogonal ($75.8^\circ$--$89.5^\circ$; Supp.~Figs.~\ref{fig:arc1-angle-bars},~\ref{fig:arc2-angle-bars},~\ref{fig:supp-arc-angle-480},~\ref{fig:supp-arc-angle-pixel}). The untrained ViT and patchifier align as closely as the trained encoders (matched angles $32.6^\circ$--$55.6^\circ$ versus $34.3^\circ$--$54.6^\circ$ at 224 px), so alignment alone does not require training.

\paragraph{Retrieval baselines.}
On the ARC-1 and ARC-2 evaluation sets at 224 px, the controls retrieve $45.8\%$--$52.0\%$ and $51.3\%$--$57.7\%$ of outputs after transport, respectively (Fig.~\ref{fig:transport-retrieval}). Retrieval from the query alone, with no added vector, does not favor the trained encoders: on ARC-1, pixels lead ($48.1\%$ versus $30.5\%$--$32.0\%$). All layers and training sets appear in Supp.~Figs.~\ref{fig:arc-knn-new-analysis} and~\ref{fig:supp-arc-retrieval-train}; retrieval examples appear in Supp.~Fig.~\ref{fig:arc-knn-new-examples}.

In ARC-GEN's denser pool of 368,442 candidates, pixels retrieve best from the query alone and remain comparable to the trained encoders after transport. Pretraining nevertheless produces a larger gain from adding the intuition vector (Fig.~\ref{fig:transport-retrieval}). Additional ARC-GEN comparisons and transformation breakdowns appear in Supp.~Figs.~\ref{fig:supp-arcgen-overview},~\ref{fig:arcgen-breakdown}, and~\ref{fig:supp-arcgen-breakdown-480}. Exploratory analyses of context-dependent single-cell transformations are presented separately in App.~\ref{app:low_level}.

\subsection{Additional figures}
\begin{figure*}[tbp]
    \centering

    \begin{minipage}[c]{0.39\textwidth}
        \centering
        \begin{subfigure}[c]{\linewidth}
            \centering
            \includegraphics[width=\linewidth]{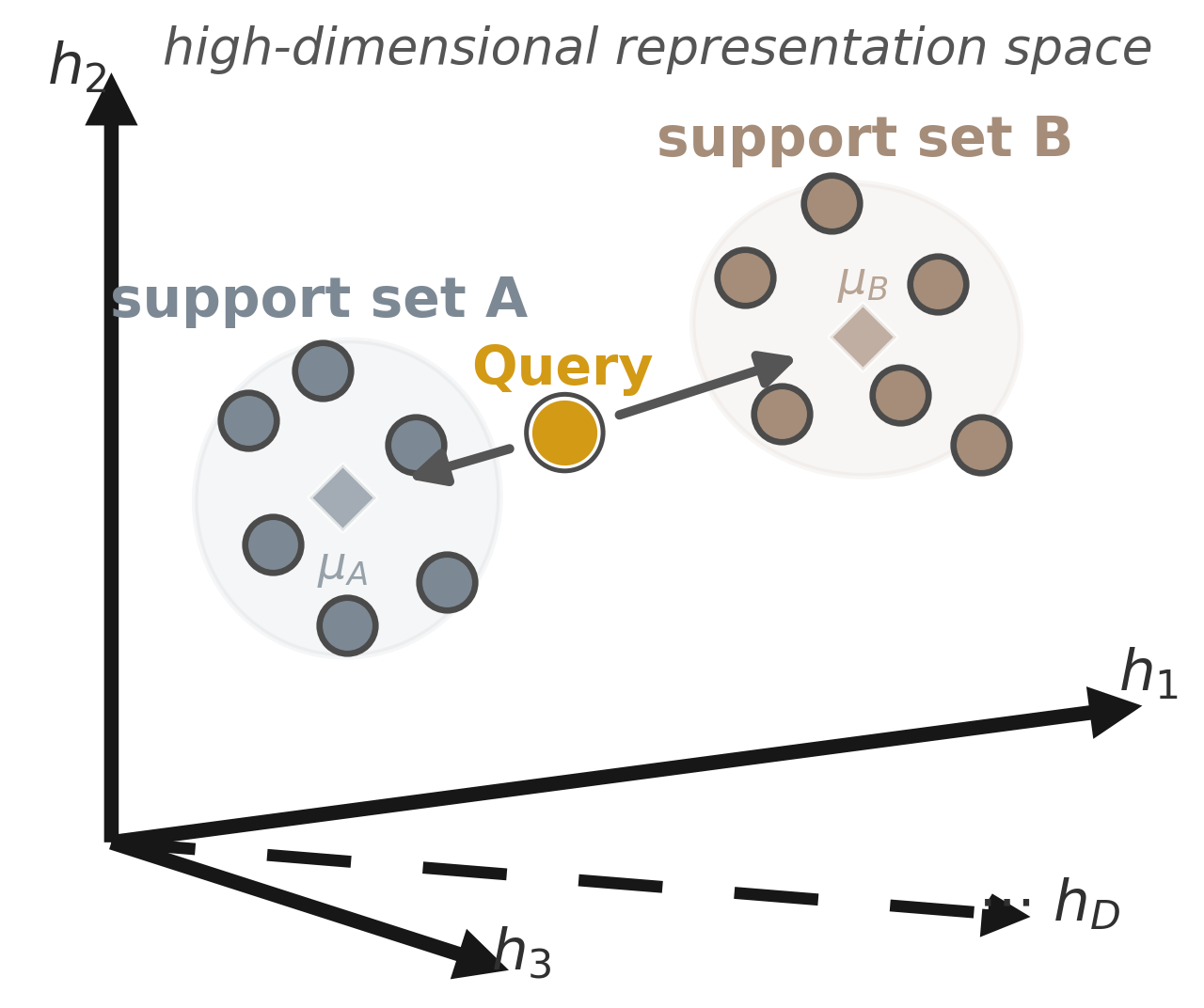}
            \caption{Nearest-centroid readout in representation space.}
            \label{fig:bongard-schematic}
        \end{subfigure}
    \end{minipage}
    \hfill
    \begin{minipage}[c]{0.59\textwidth}
        \centering
        \begin{subfigure}[t]{\linewidth}
            \centering
            \includegraphics[width=\linewidth]{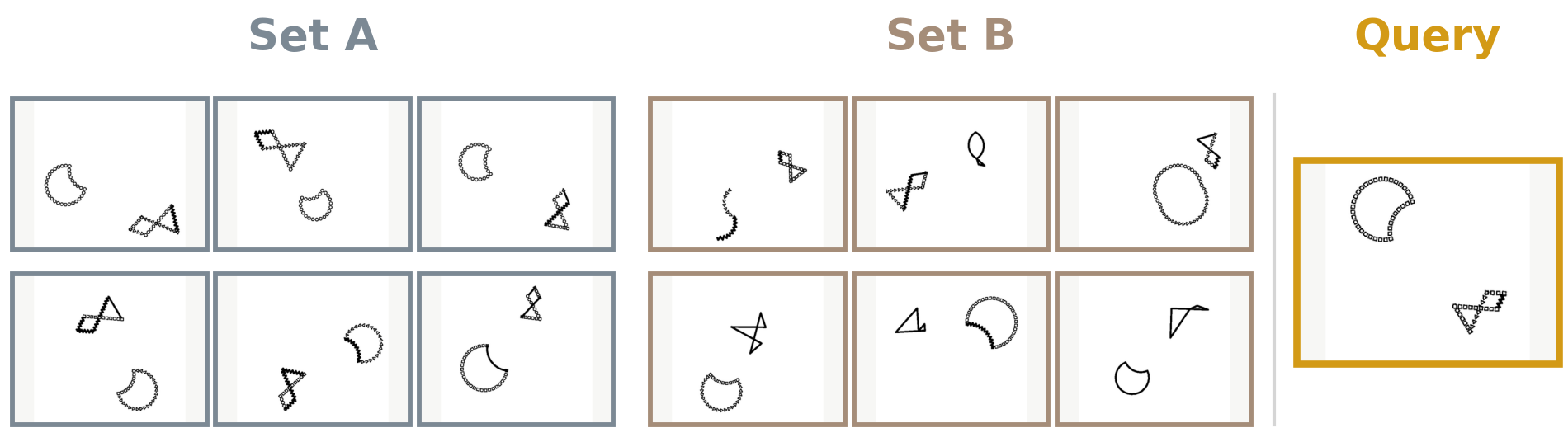}
            \caption{Example Bongard-LOGO problem.}
            \label{fig:bongard-logo-example}
        \end{subfigure}

        \vspace{0.15em}

        \begin{subfigure}[t]{\linewidth}
            \centering
            \includegraphics[width=\linewidth]{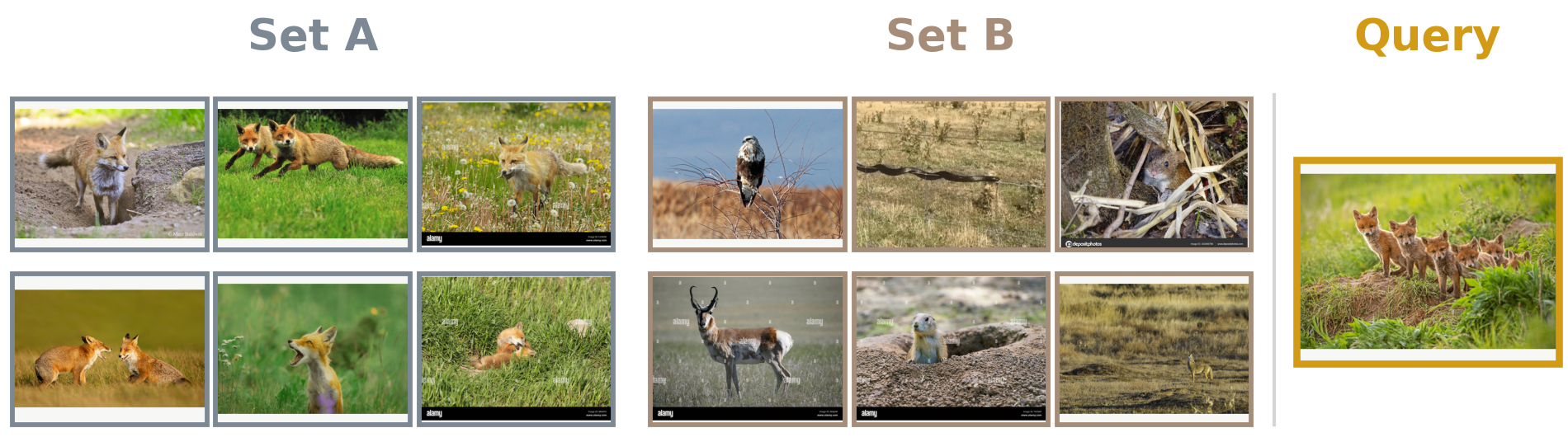}
            \caption{Example Bongard-OpenWorld problem.}
            \label{fig:bongard-ow-example}
        \end{subfigure}
    \end{minipage}

    \vspace{0.65em}

    \begin{subfigure}[t]{0.49\textwidth}
        \centering
        \includegraphics[width=\linewidth]{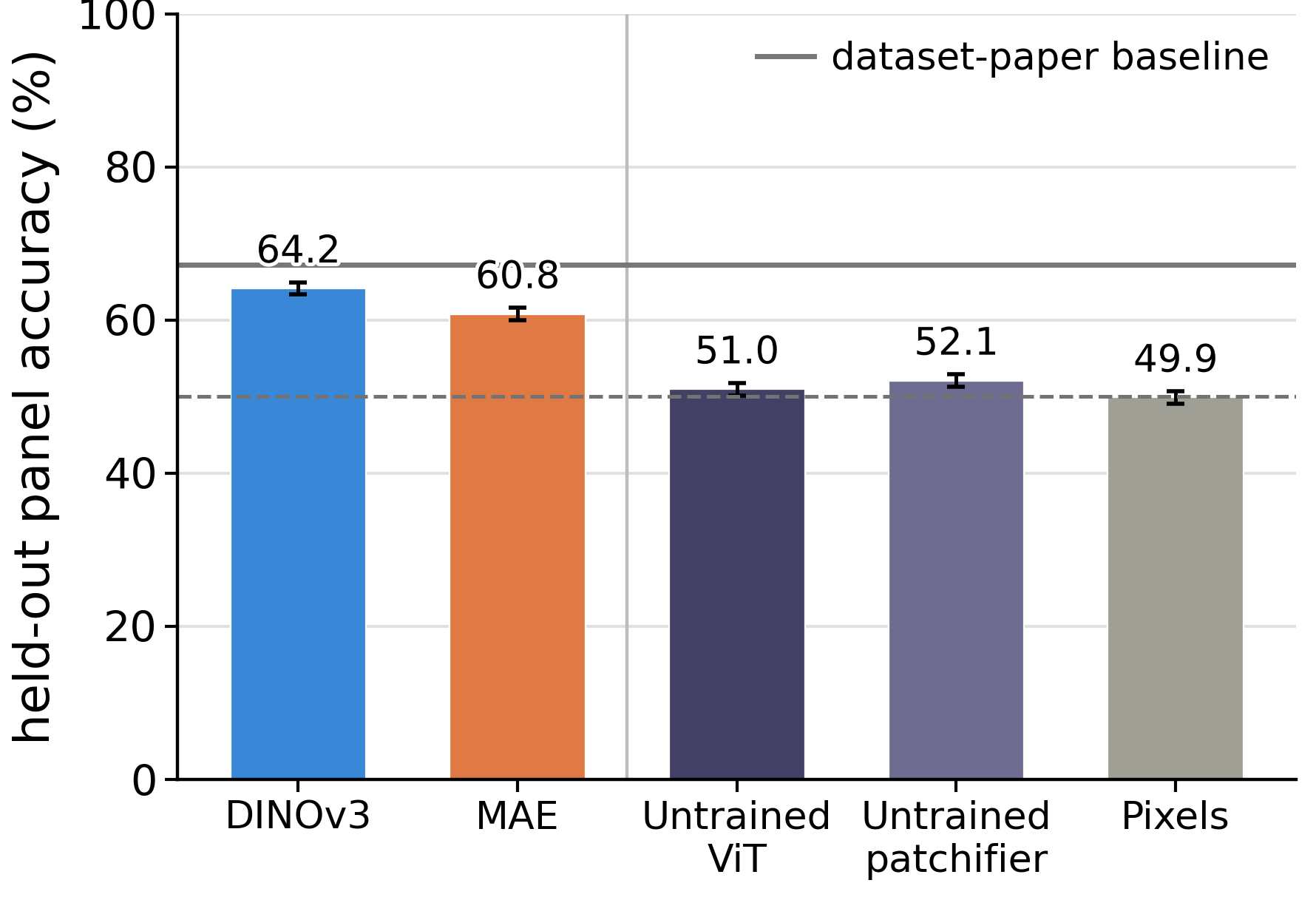}
        \caption{Bongard--LOGO, average over four test subsets.}
        \label{fig:bongard-logo-bars}
    \end{subfigure}
    \hfill
    \begin{subfigure}[t]{0.49\textwidth}
        \centering
        \includegraphics[width=\linewidth]{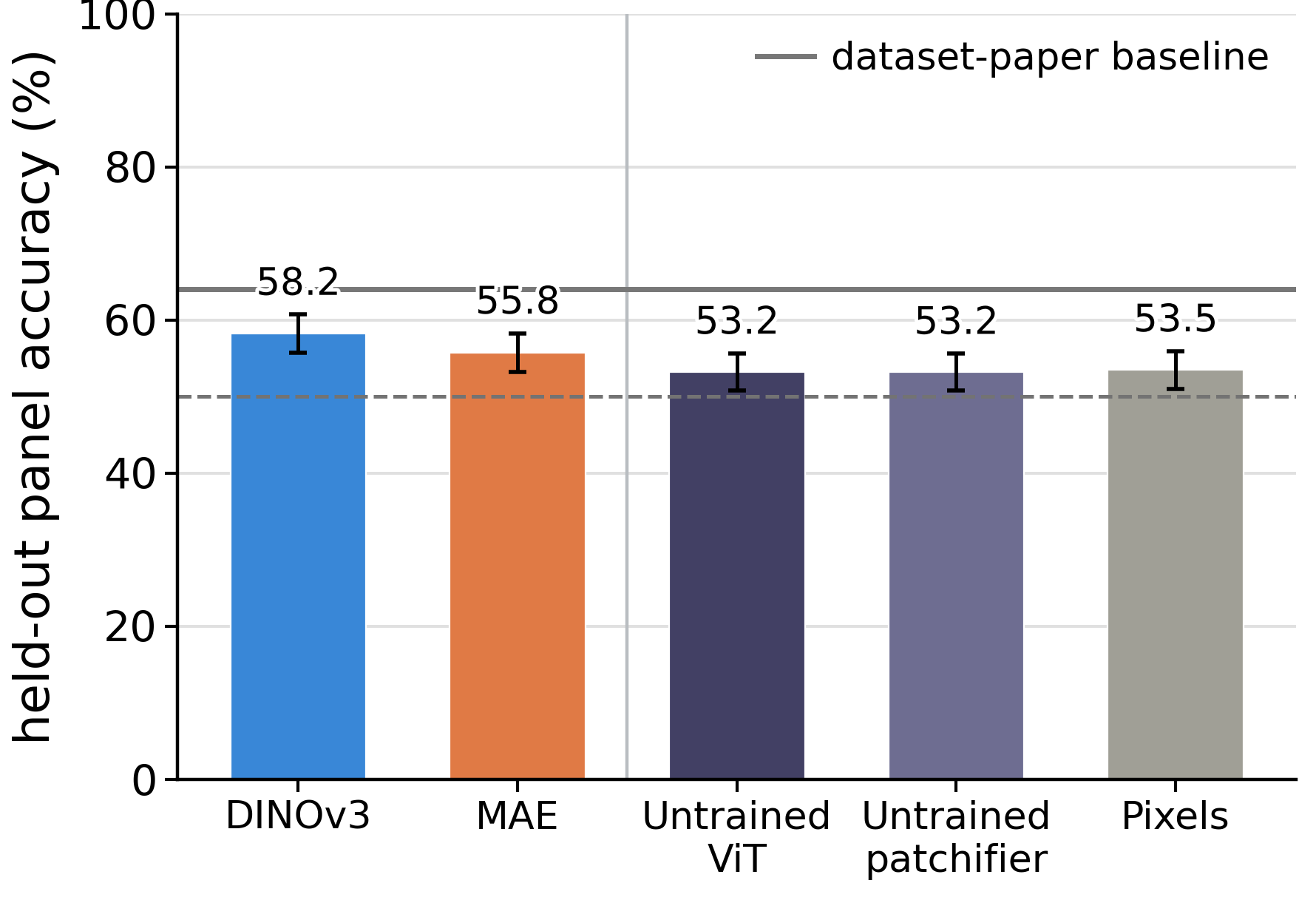}
        \caption{Bongard--OpenWorld, all 200 problems.}
        \label{fig:bongard-ow-bars}
    \end{subfigure}

    \caption{
        Visual concept discrimination in frozen representation spaces on Bongard--LOGO and Bongard-OpenWorld. \textbf{(a)} The two support sets define class centroids, and a held-out query is assigned to its nearer centroid. \textbf{(b, c)} Representative Bongard-LOGO and Bongard-OpenWorld problems. \textbf{(d, e)} Nearest-centroid test accuracy at 224 px for DINOv3, a size-matched MAE, two untrained controls: a randomly initialized ViT with the DINOv3-Large architecture (untrained ViT; seed 0) and that architecture's patch-embedding layer alone, mean-pooled over patches with no transformer blocks (untrained patchifier; averaged over ten initialization seeds), and random-projection pixel controls. For each encoder, including the untrained ViT, and each dataset, the layer is selected on training problems (LOGO: 600 problems from each of its three training types; OpenWorld: its 610 training problems) and test accuracy is reported at that layer.
        LOGO values are equal-weight means across its four disjoint test  subsets; OpenWorld values pool 200 problems. Error bars show SEM across problems for encoders, and the mean problem-level SEM across ten projections for pixels or ten seeds for the untrained patchifier. Gray lines mark the dataset-paper baselines, the best results reported in each dataset's original paper rather than the current state of the art (LOGO: $67.2\%$, the equal-weight mean of the per-subset maxima over its baselines~\citep{nie2020bongardlogo}, which no single baseline reaches; OpenWorld: $64.0\%$, SNAIL with OpenCLIP and a caption auxiliary task~\citep{wu2024bongardopen}); dashed lines mark $50\%$ chance. Results at 480 px appear in Supp.~Fig.~\ref{fig:supp-bongard-480}; per-subtask results and full layer profiles in
        Supp.~Fig.~\ref{fig:supp-bongard-subtasks} and Supp.~Figs.~\ref{fig:supp-bongard-layerwise} and~\ref{fig:supp-bongard-layerwise-ow}, respectively.
    }
    \label{fig:bongard-task-vector-analysis}
\end{figure*}
\begin{figure*}[tbp]
    \centering

    \begin{minipage}[c]{0.38\textwidth}
        \centering
        \begin{subfigure}[c]{\linewidth}
            \centering
            \includegraphics[width=0.9\linewidth]{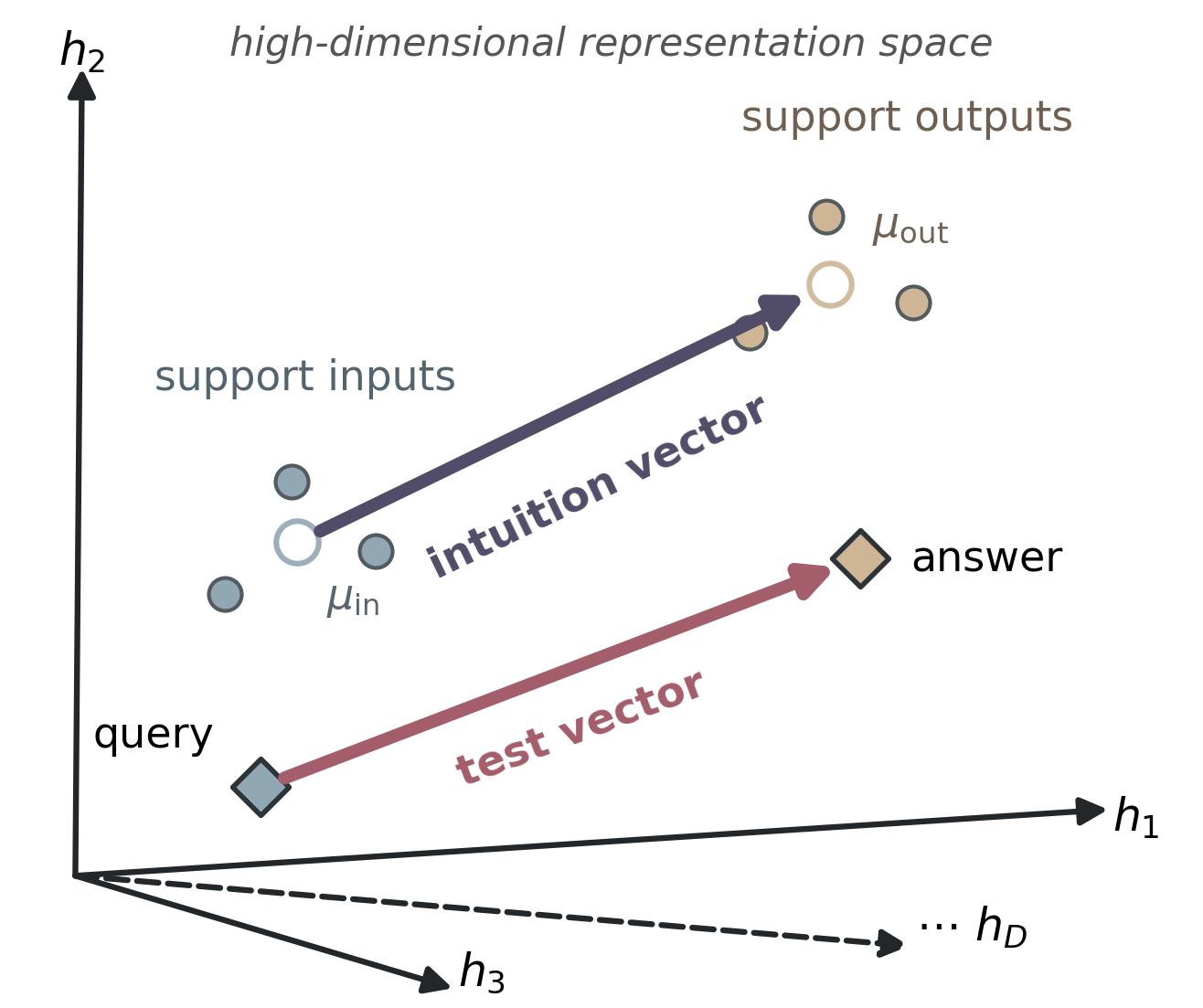}
            \caption{Intuition and test vectors.}
            \label{fig:arc-vector-schematic}
        \end{subfigure}
    \end{minipage}
    \hfill
    \begin{minipage}[c]{0.60\textwidth}
        \centering
        \begin{subfigure}[c]{\linewidth}
            \centering
            \includegraphics[width=0.9\linewidth]{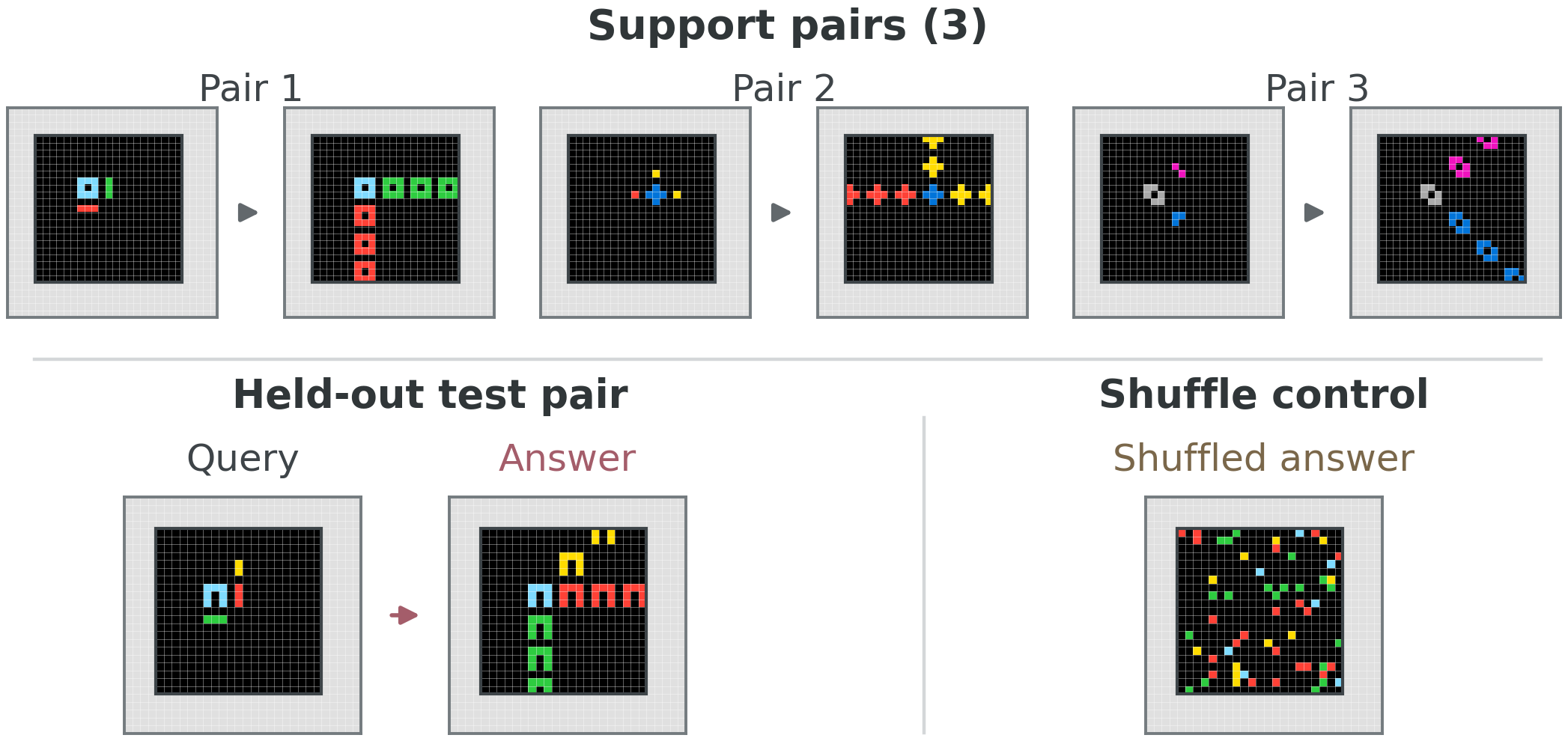}
            \caption{Three ARC support pairs $(K=3)$, held-out pair (bottom left), and shuffled-output control (bottom right).}
            \label{fig:arc-task-example}
        \end{subfigure}
    \end{minipage}

    \vspace{0.35em}

    \begin{subfigure}[t]{\textwidth}
        \centering
        \includegraphics[width=0.9\linewidth]{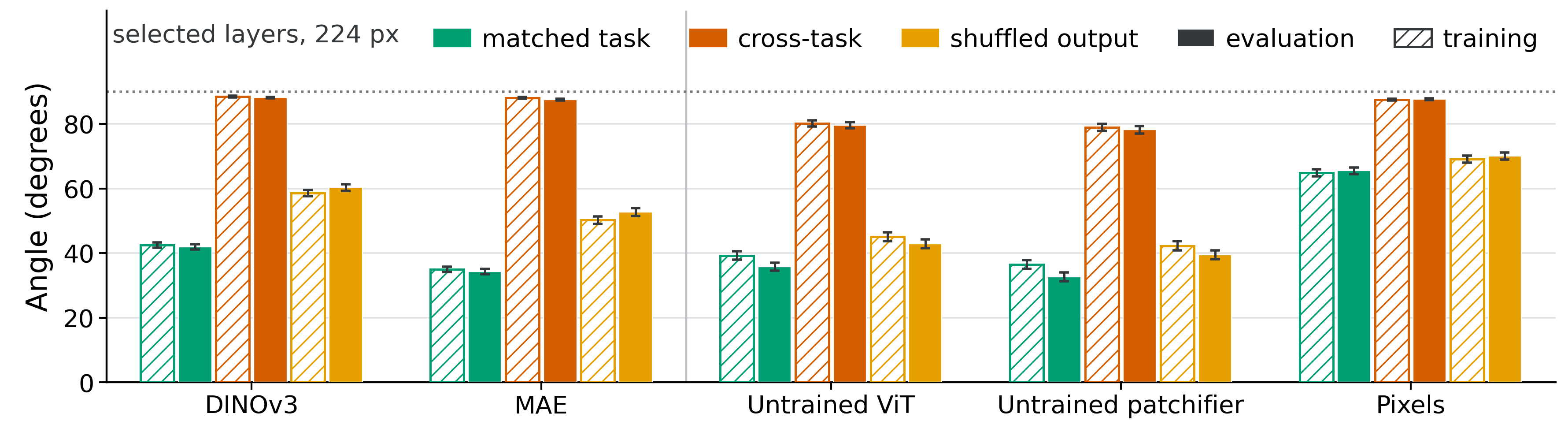}
        \caption{ARC-1: every encoder and control at 224 px.}
        \label{fig:arc1-angle-bars}
    \end{subfigure}

    \vspace{0.2em}

    \begin{subfigure}[t]{\textwidth}
        \centering
        \includegraphics[width=0.9\linewidth]{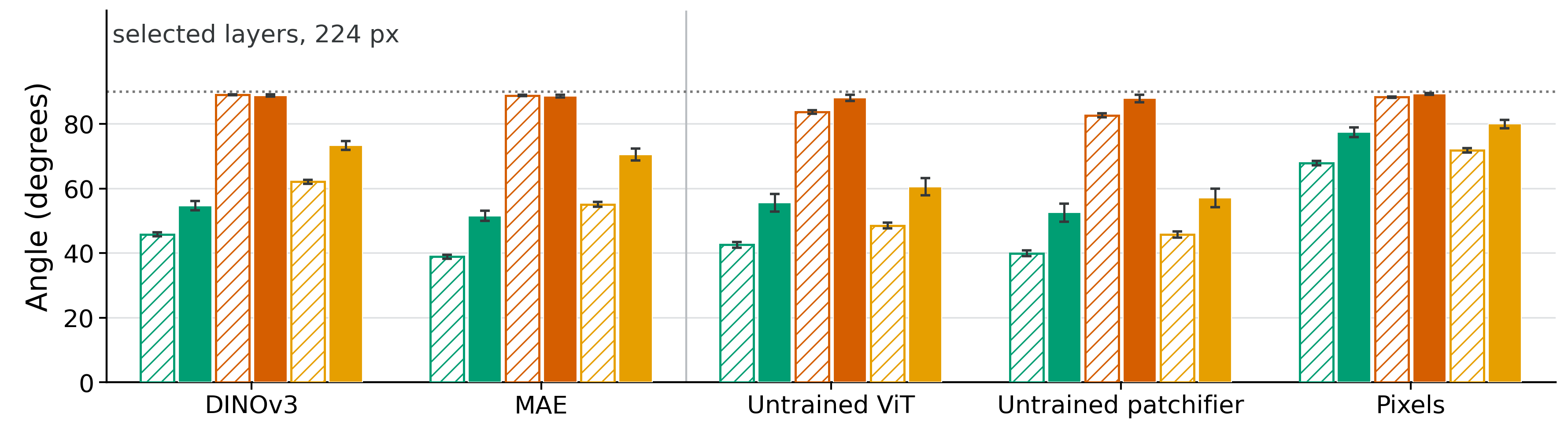}
        \caption{ARC-2: the same comparison.}
        \label{fig:arc2-angle-bars}
    \end{subfigure}

    \caption{
        Intuition vector alignment in frozen DINOv3 and MAE representations of
        ARC-1 and ARC-2. \textbf{(a)} The intuition vector connects the
        support-input and support-output centroids; the test vector connects the
        held-out query and answer. \textbf{(b)} A task with three support pairs,
        its held-out test pair, and a cell-shuffled answer control, which preserves
        grid shape and color counts while disrupting spatial structure.
        \textbf{(c,d)} Intuition to test vector angles at 224 px, at each encoder's selected layer
        (Sec.~\ref{sec:methods-layers}; DINOv3 and MAE 16, untrained ViT 22); right of the divider are the untrained ViT and untrained patchifier of
        Fig.~\ref{fig:bongard-task-vector-analysis} and the random-projection pixel controls.
        Bars compare matched tasks with cross-task and shuffled-output
        controls, each averaged over 10 randomizations. Filled bars denote
        evaluation tasks and open hatched bars training tasks; ARC-1 has $n=400$
        tasks per dataset, and ARC-2 has $n=1{,}000$ training and $n=120$ evaluation
        tasks. Bars show mean $\pm$SEM across tasks; the dotted line marks
        orthogonality. Panel \textbf{(c)} provides the key for both panels.
        Pixel and patchifier controls are
        layer-independent; patchifier bars average ten seeds, with the mean SEM across tasks.
        Results at 480 px appear in Supp.~Fig.~\ref{fig:supp-arc-angle-480}; full layer-wise
        results in Supp.~Fig.~\ref{fig:supp-arc-angle-arc1}, \ref{fig:supp-arc-angle-arc2}.
    }

    \label{fig:arc-task-vector-analysis}
\end{figure*}
\begin{figure*}[tbp]
    \centering
    \captionsetup[subfigure]{skip=1pt}

    \begin{minipage}[c]{0.34\textwidth}
        \centering
        \begin{subfigure}[c]{\linewidth}
            \centering
            \includegraphics[width=0.9\linewidth]{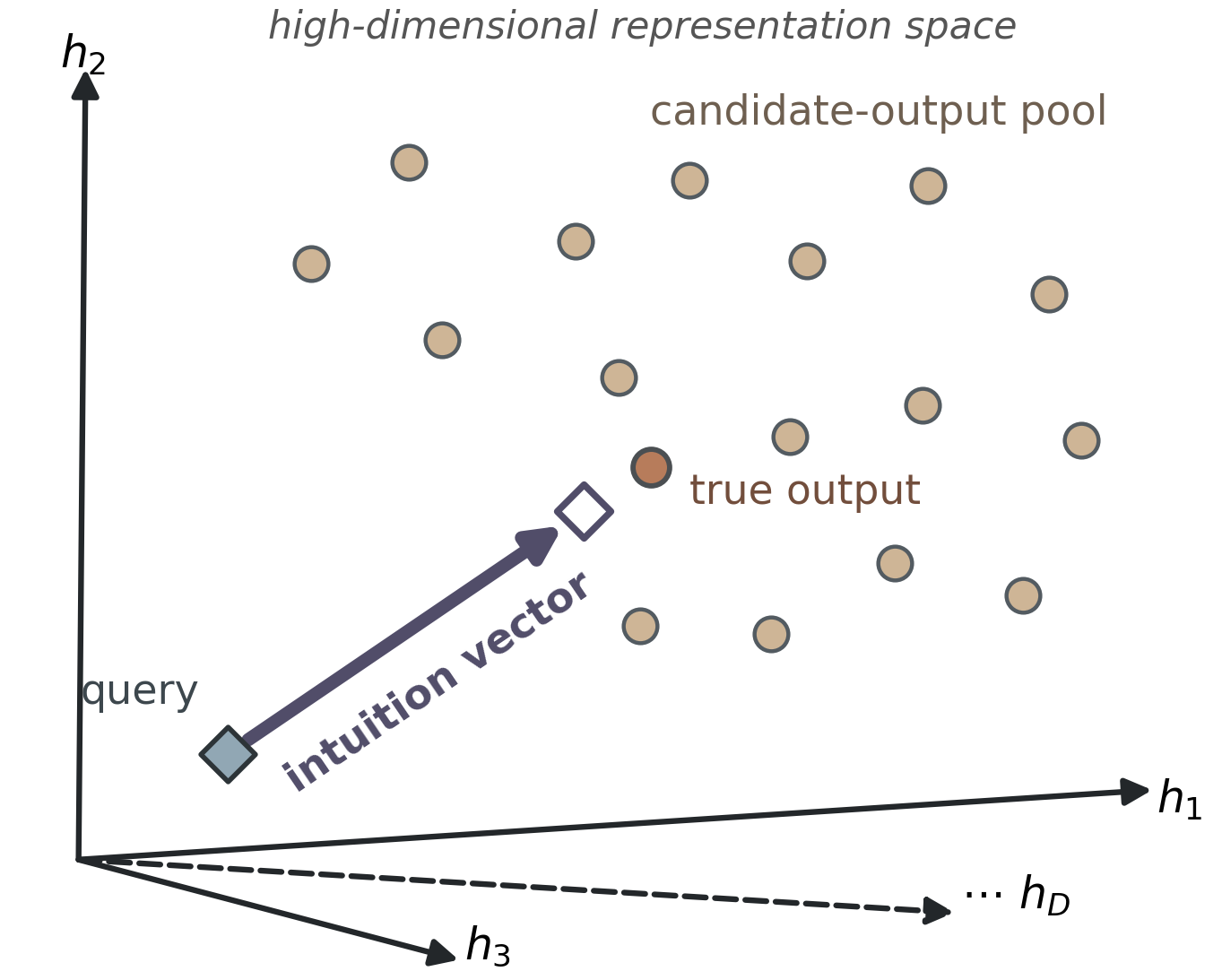}
            \caption{KNN retrieval schematic.}
            \label{fig:arc-knn-new-schematic}
        \end{subfigure}
    \end{minipage}
    \hfill
    \begin{minipage}[c]{0.64\textwidth}
        \centering
        \begin{subfigure}[c]{\linewidth}
            \centering
            \includegraphics[width=0.9\linewidth]{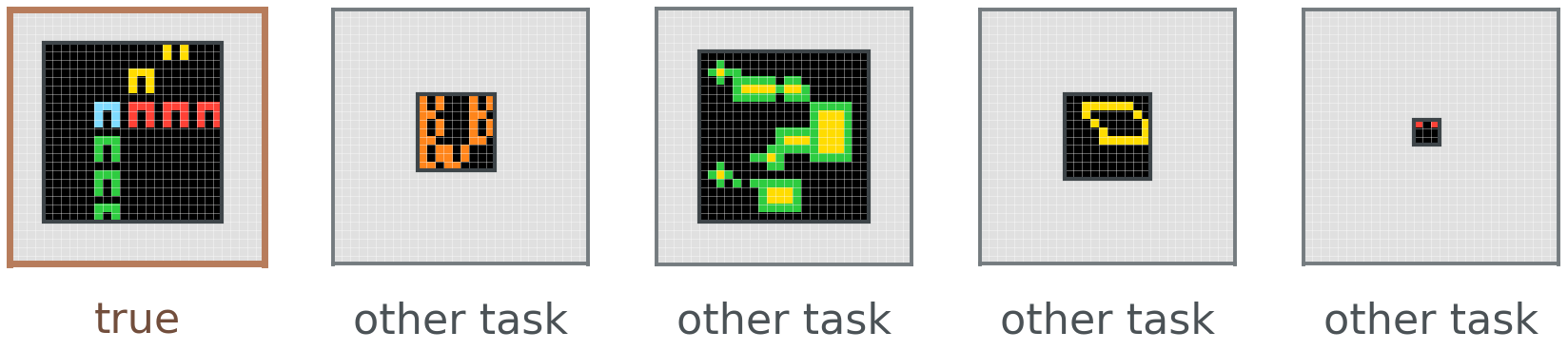}
            \caption{Example candidates.}
            \label{fig:arc-knn-new-examples}
        \end{subfigure}
    \end{minipage}

    \vspace{0.45em}

    \includegraphics[width=0.8\textwidth]{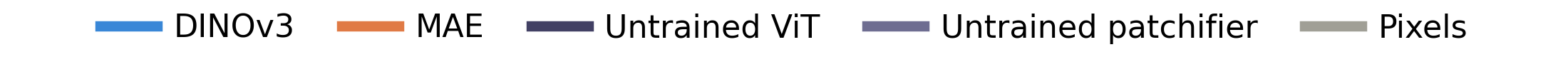}

    \vspace{0.2em}

    \begin{subfigure}[t]{0.49\textwidth}
        \centering
        \includegraphics[width=0.9\linewidth]{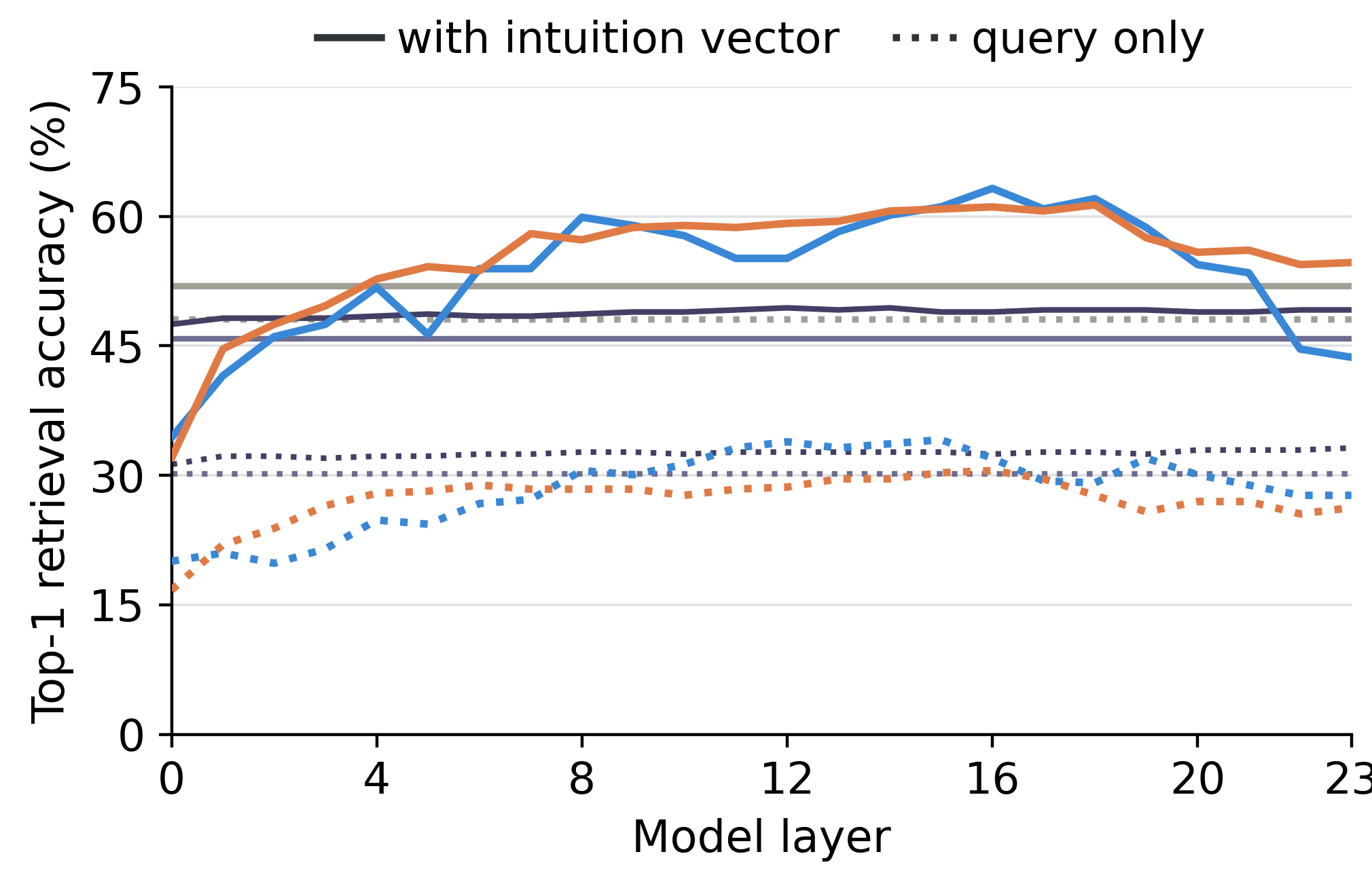}
        \caption{ARC-1, evaluation set.}
        \label{fig:arc1-knn-new-eval}
    \end{subfigure}
    \hfill
    \begin{subfigure}[t]{0.49\textwidth}
        \centering
        \includegraphics[width=0.9\linewidth]{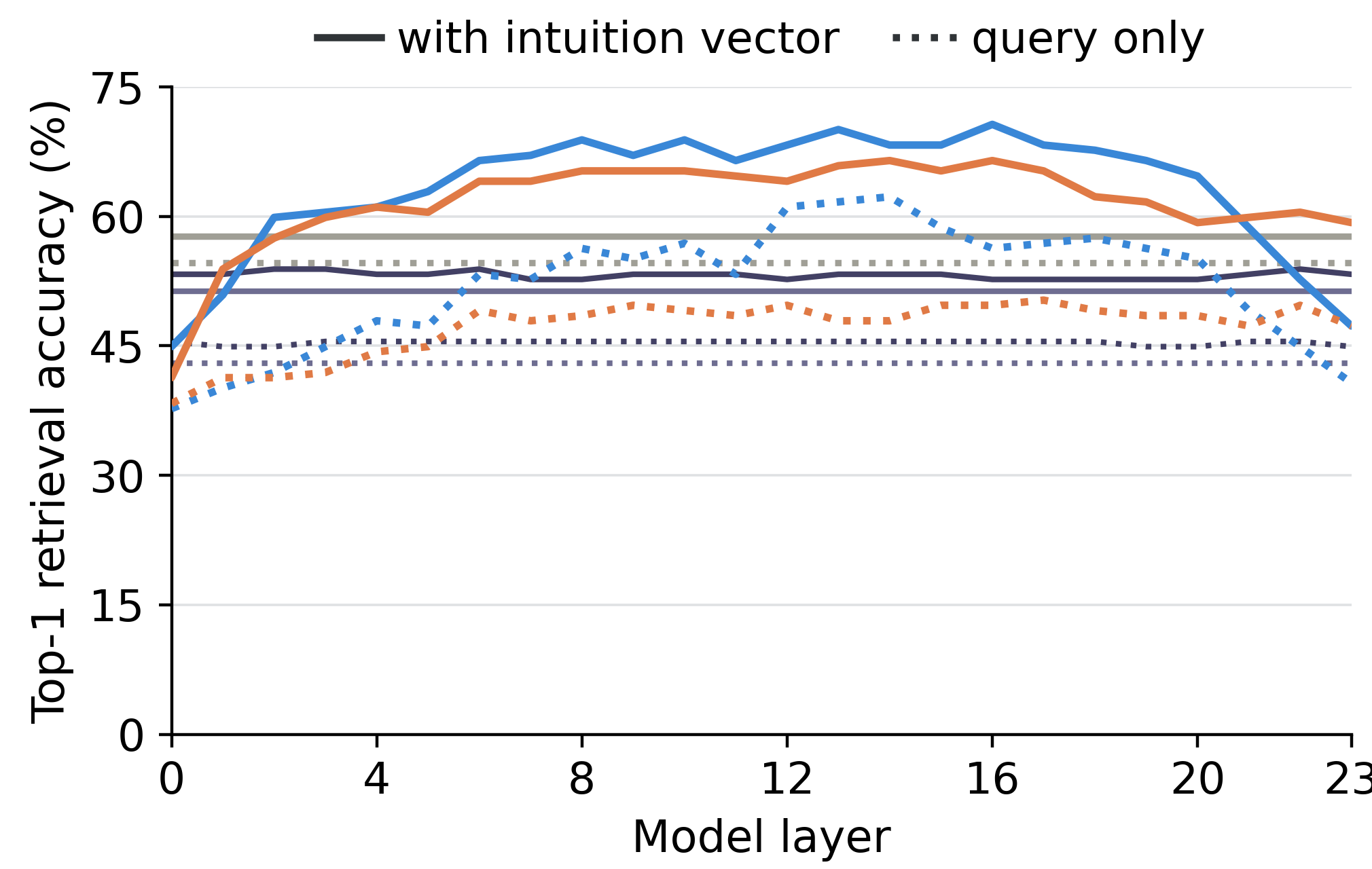}
        \caption{ARC-2, evaluation set.}
        \label{fig:arc2-knn-new-eval}
    \end{subfigure}

    \vspace{0.4em}

    \begin{subfigure}[t]{0.49\textwidth}
        \centering
        \includegraphics[width=0.9\linewidth]{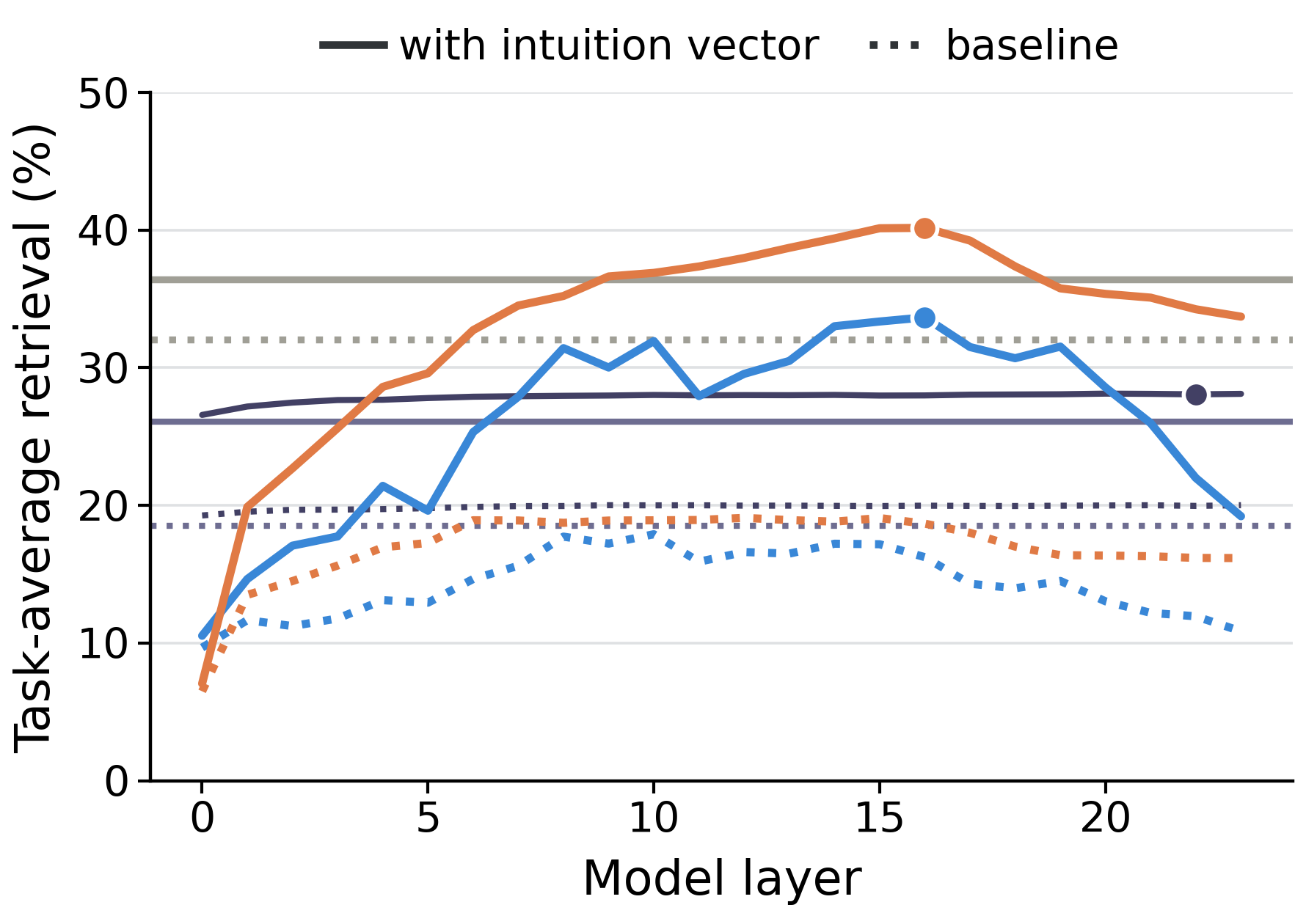}
        \caption{ARC-GEN.}
        \label{fig:arcgen-retrieval-by-layer}
    \end{subfigure}
    \hfill
    \begin{subfigure}[t]{0.49\textwidth}
        \centering
        \includegraphics[width=0.9\linewidth]{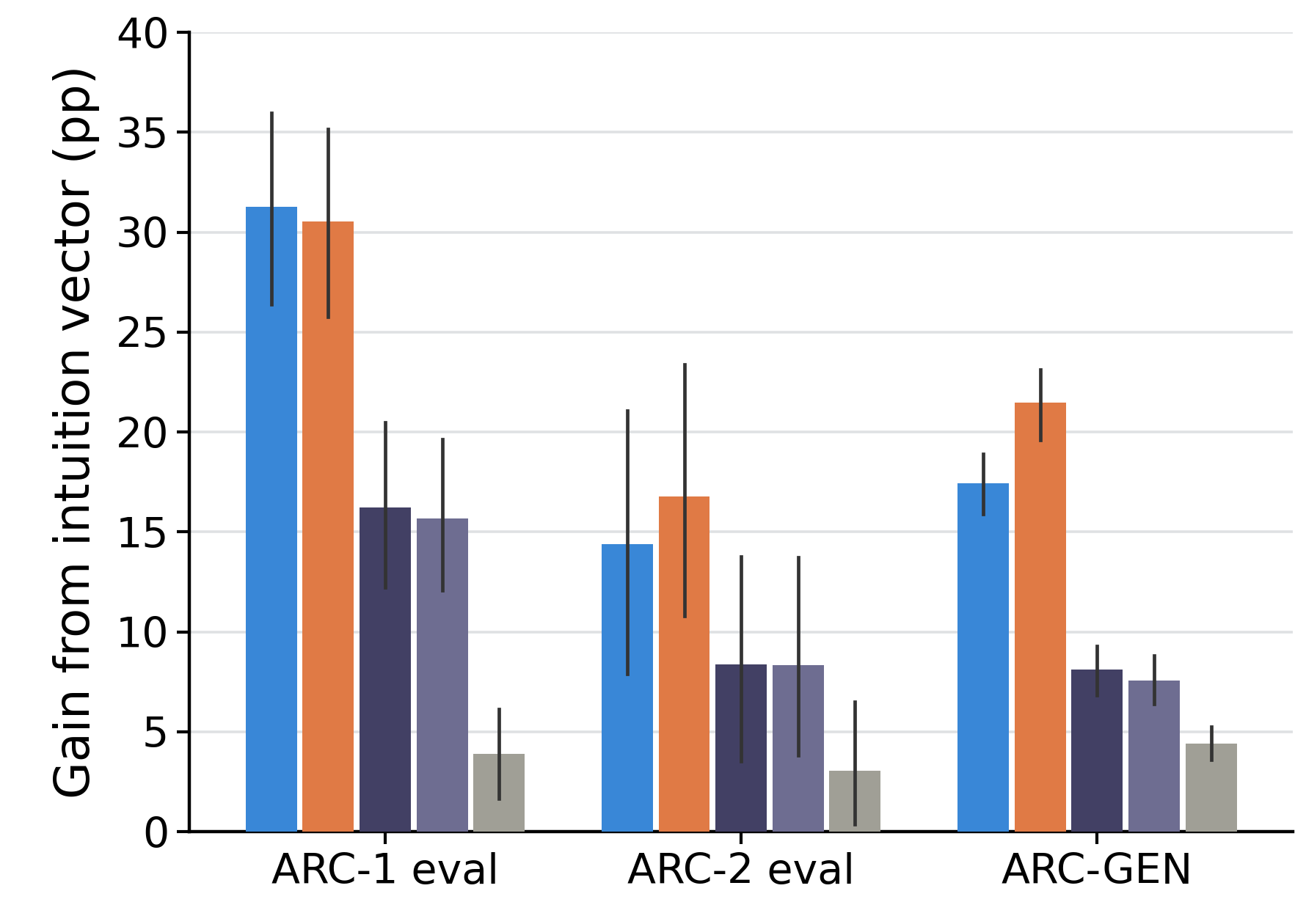}
        \caption{Gain from the intuition vector.}
        \label{fig:retrieval-gain-summary}
    \end{subfigure}

    \caption{
        ARC output retrieval. \textbf{(a)} The inferred intuition vector transports the held-out query representation, and all outputs in the same dataset are ranked by distance to the transported query. \textbf{(b)} Candidate pool is the true output (outlined) competing with every other output in that dataset; four alternatives are shown. \textbf{(c, d)} Top-1 retrieval accuracy at 224 px across the 24 transformer blocks on the ARC-1 and ARC-2 evaluation sets. Curves compare DINOv3, a size-matched MAE, and the untrained ViT; flat lines are the layer-independent random-projection pixel baselines and the untrained patchifier (mean of ten seeds; Fig.~\ref{fig:bongard-task-vector-analysis}). Solid lines use the translated query; dotted lines use the original query, isolating the contribution of the intuition vector. Training set results appear in Supp.~Fig.~\ref{fig:supp-arc-retrieval-train}.  \textbf{(e)} ARC-GEN retrieval at 224 px from a 368{,}442 grid candidate pool, across all 24 layers, with the same models. Markers identify selected encoder layers, chosen for the untrained ViT by the same rule. \textbf{(f)} Gain from the intuition vector (transported minus query-only top-1, in percentage points) at 224 px, at each encoder's selected layer: the layer maximizing mean ARC-1/ARC-2 evaluation transport retrieval, with pixels and the patchifier layer-independent. Whiskers are $95\%$ task-bootstrap intervals; for ARC-1 and ARC-2 they resample tasks over paired per-query outcomes with and without the vector. ARC-1 and ARC-2 layers were selected on these same queries, which makes ARC-GEN the held-out estimate. Results at 480 px appear in Supp.~Fig.~\ref{fig:supp-arc-retrieval-480}.
    }
    \label{fig:arc-knn-new-analysis}
\end{figure*}
\begin{figure*}[tbp]
    \centering
    \captionsetup[subfigure]{skip=2pt}

    \begin{subfigure}[t]{0.49\textwidth}
        \centering
        \includegraphics[width=\linewidth]{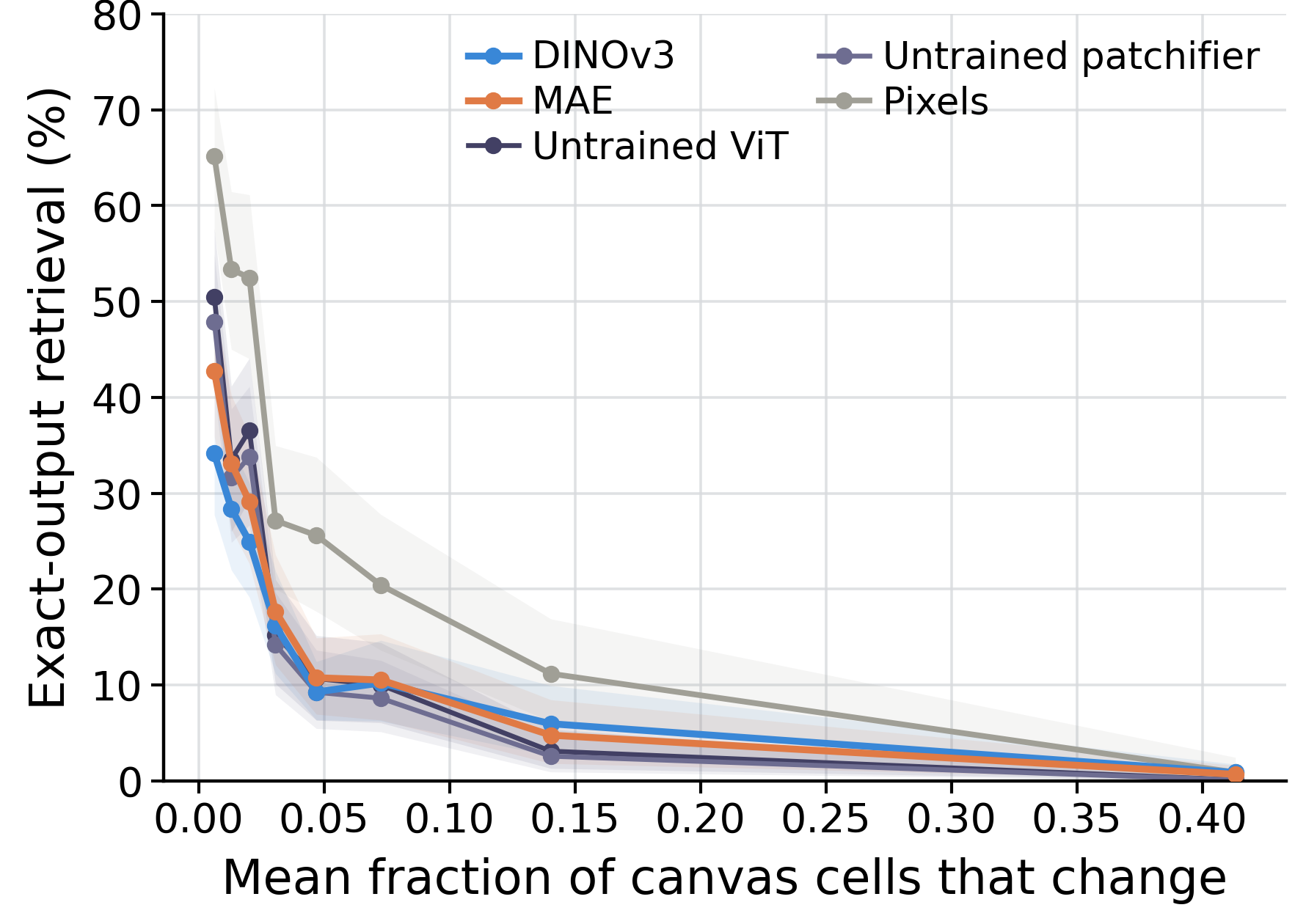}
        \caption{Retrieval without an intuition vector.}
        \label{fig:arcgen-baseline-by-change}
    \end{subfigure}
    \hfill
    \begin{subfigure}[t]{0.49\textwidth}
        \centering
        \includegraphics[width=\linewidth]{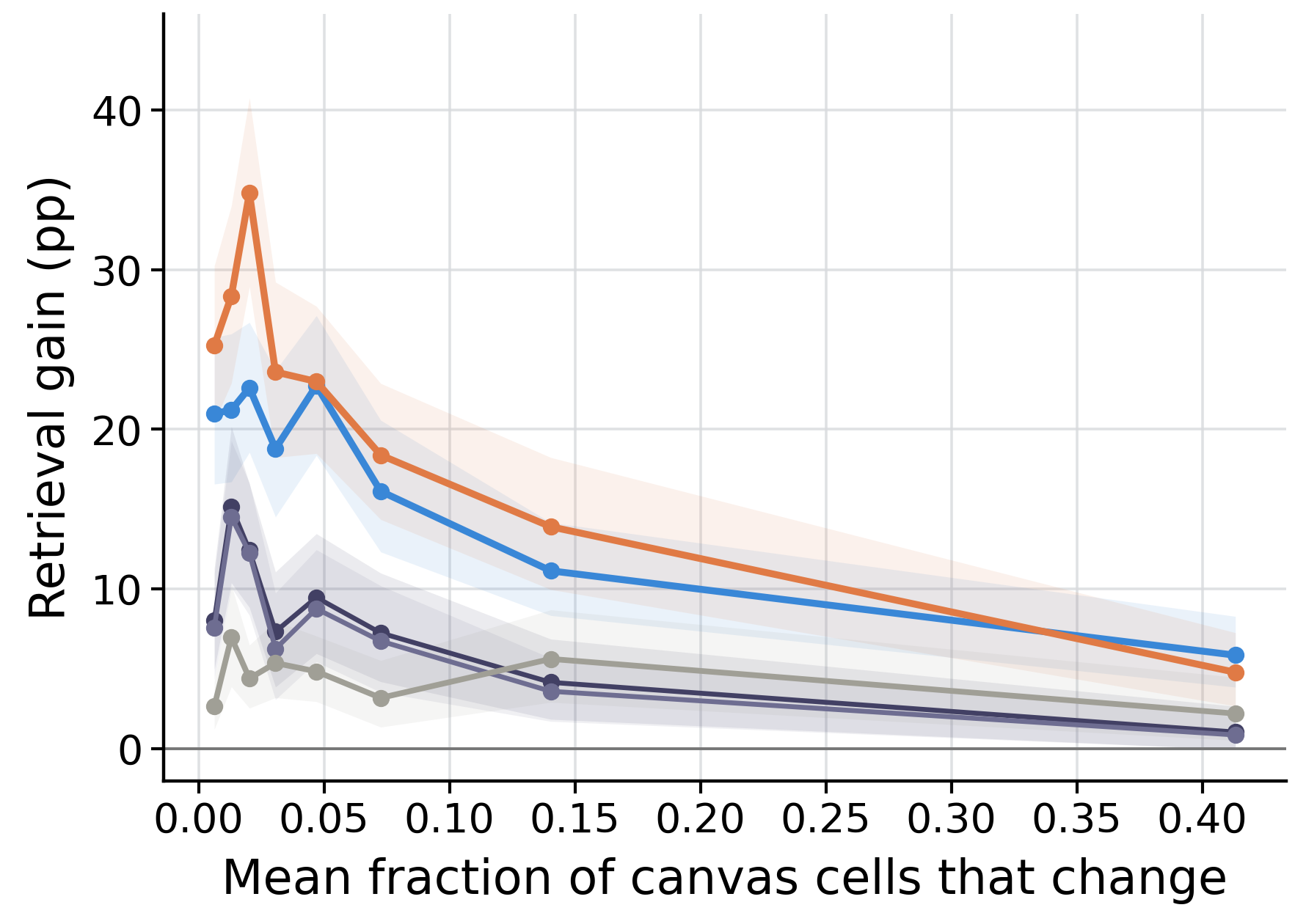}
        \caption{What the intuition vector adds.}
        \label{fig:arcgen-lift-by-change}
    \end{subfigure}

   \caption{
        Baseline retrieval and intuition-vector gain by transformation extent on ARC-GEN at 224 px. Both panels analyze 794 tasks with 100 generated queries each; encoder layers were selected on original ARC queries and held fixed. \textbf{(a,b)} Baseline retrieval (without the intuition vector) and its gain from adding the vector, in percentage points, against the fraction of canvas cells that change. Eight shared, fixed, near-equal-count bins use predictors computed from separate clean instances; shading shows marginal $95\%$ task-bootstrap intervals. The untrained ViT and untrained patchifier (Fig.~\ref{fig:bongard-task-vector-analysis}) use the same bins; patchifier task values average ten seeds. Results at 480 px appear in Supp.~Fig.~\ref{fig:supp-arcgen-breakdown-480}.
    }
    \label{fig:arcgen-breakdown}
\end{figure*}
\begin{figure*}[tbp]
    \centering
    \captionsetup[subfigure]{skip=2pt}

    \begin{subfigure}[t]{0.49\textwidth}
        \centering
        \includegraphics[width=0.9\linewidth]{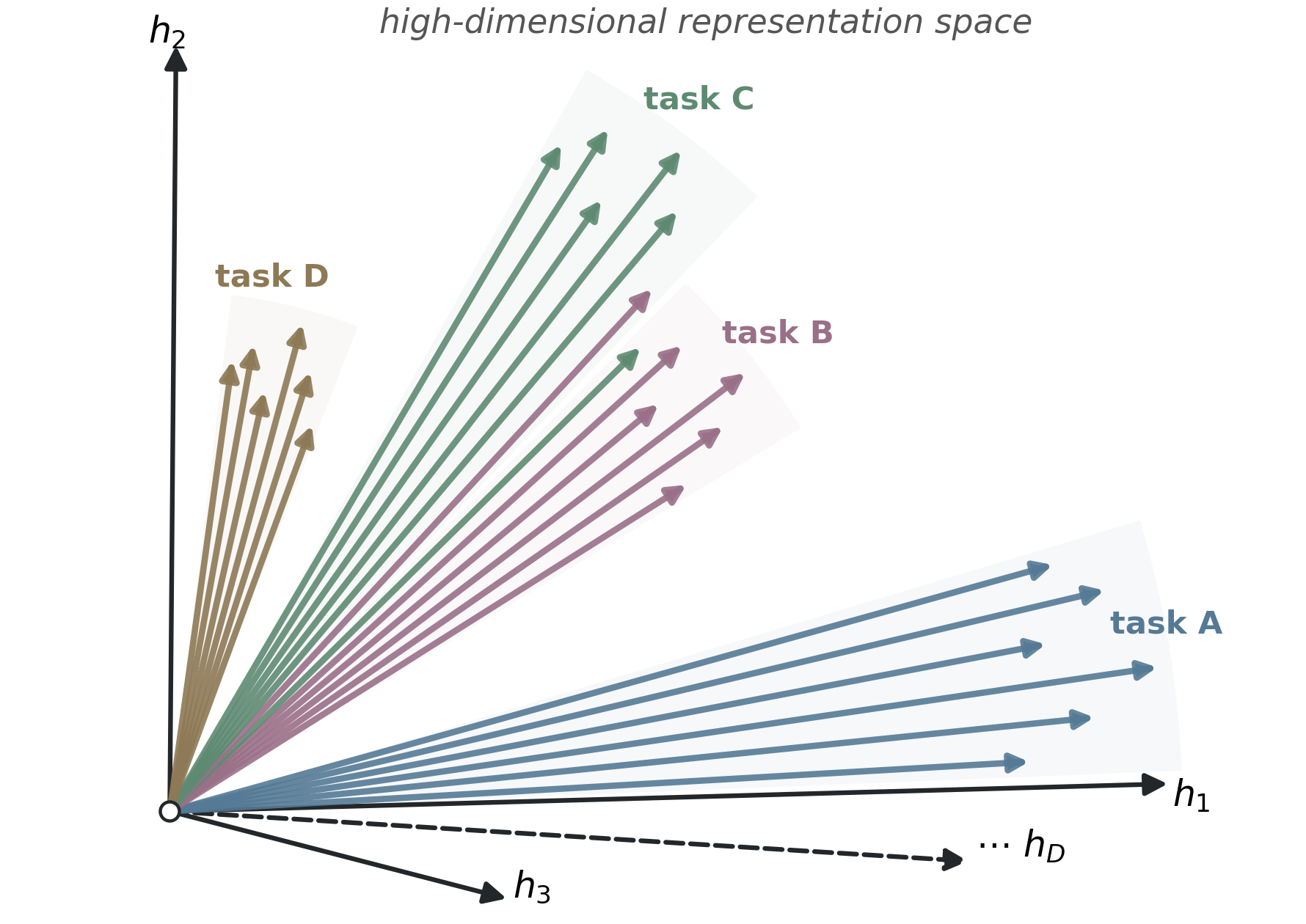}
        \caption{Population of task instance-vectors}
        \label{fig:arcgen-classification-scheme}
    \end{subfigure}
    \hfill
    \begin{subfigure}[t]{0.49\textwidth}
        \centering
        \includegraphics[width=0.9\linewidth]{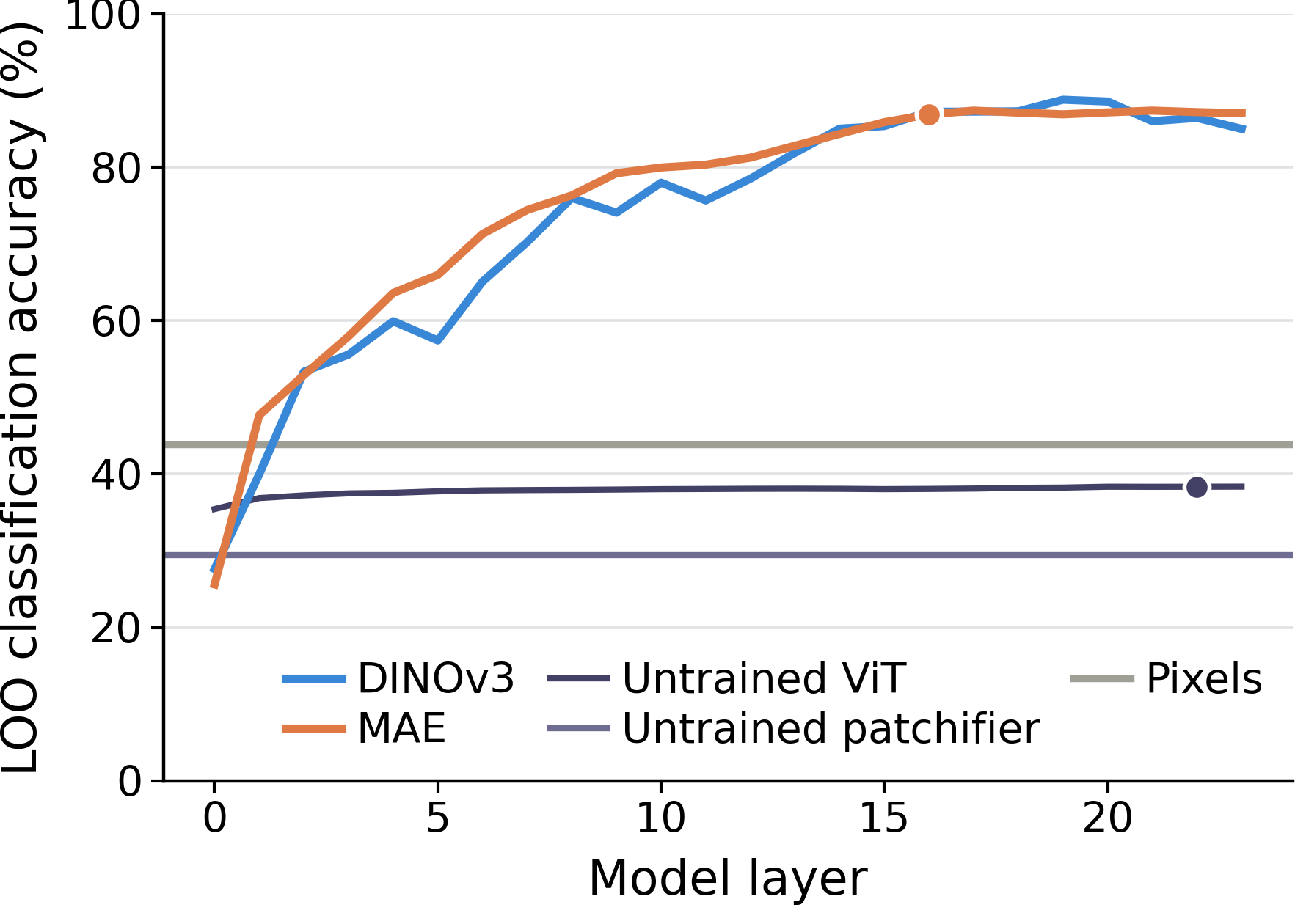}
        \caption{Instance classification across layers.}
        \label{fig:arcgen-classification-by-layer}
    \end{subfigure}
    \vspace{0.4em}

    \begin{subfigure}[t]{0.49\textwidth}
        \centering
        \includegraphics[width=0.9\linewidth]{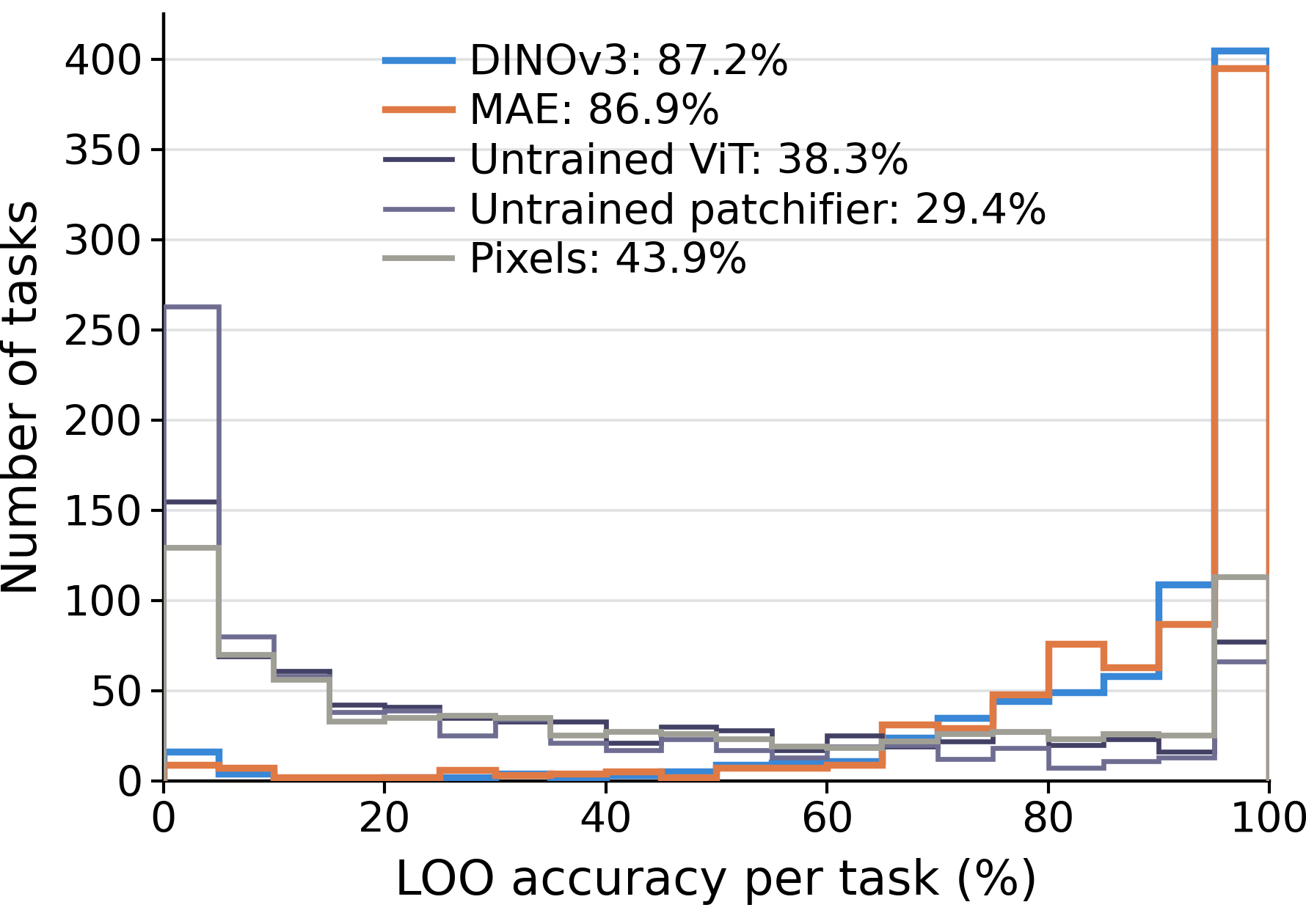}
        \caption{Per-task performance distribution.}
        \label{fig:arcgen-classification-histogram}
    \end{subfigure}

    \caption{
        Population-level organization of single instance ARC-GEN difference vectors (794 tasks; 500 instances each; 397,000 total). DINOv3, MAE, random-projection pixel controls, and the untrained ViT and untrained patchifier of Fig.~\ref{fig:bongard-task-vector-analysis} were evaluated on the same instances. \textbf{(a)} Schematic of task-instance difference vectors.  \textbf{(b)} Leave-one-out nearest-centroid accuracy at 224 px across model layers. Markers denote layers selected independently by retrieval on the original ARC evaluation pairs, by the same rule for the untrained ViT; pixel and patchifier controls are layer-independent, and patchifier values average ten seeds. \textbf{(c)} Per-task accuracy distributions at these layers; legends report means. Results at 480 px appear in Supp.~Fig.~\ref{fig:supp-arcgen-classification-480}.
    }
    \label{fig:arcgen-classification}
\end{figure*}
\begin{figure*}[tbp]
    \centering
    \captionsetup[subfigure]{skip=2pt}

    \begin{subfigure}[t]{0.31\textwidth}
        \centering
        \includegraphics[width=\linewidth]{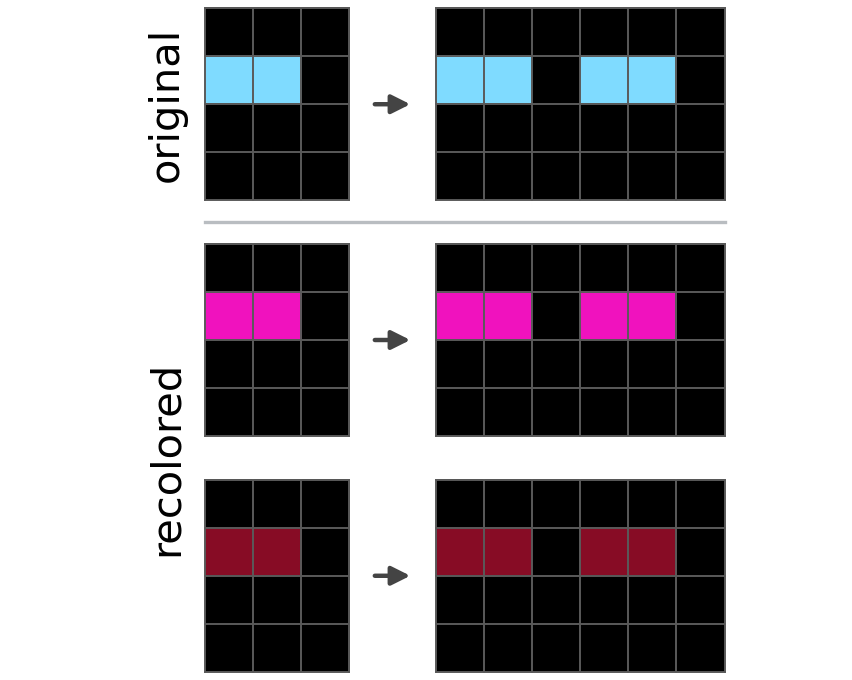}
        \caption{Recolored instances.}
        \label{fig:color-permutation-examples}
    \end{subfigure}
    \hfill
    \begin{subfigure}[t]{0.67\textwidth}
        \centering
        \includegraphics[width=\linewidth]{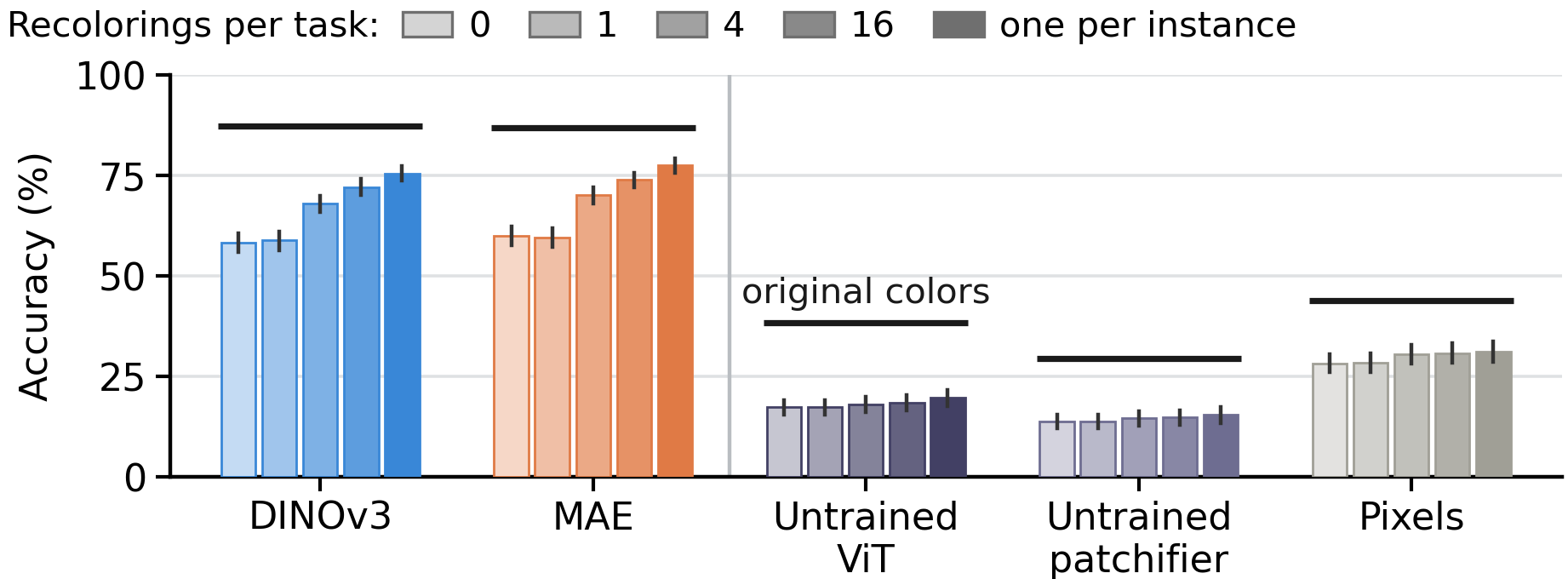}
        \caption{Task classification as the population is recolored.}
        \label{fig:color-permutation-bars}
    \end{subfigure}

    \caption{
        ARC-GEN task classification is not reducible to color identity. Every instance of
        Fig.~\ref{fig:arcgen-classification} (794 tasks $\times$ 500 instances, 224 px) is
        re-rendered with a permutation of the nine non-black colors, applied to input and output
        alike, and classified with the same readout and layers. The classified instance is
        always recolored, with a permutation its task's population never uses. \textbf{(a)} One
        task instance (input and output) in its original colors and in two recolorings.
        \textbf{(b)} The population (the instances that define each task's centroid) is
        recolored with 0, 1, 4 or 16 permutations per task, or with one per instance, in which
        case color carries no task information. Black segments: accuracy in original colors,
        which recoloring a whole task consistently leaves unchanged (DINOv3 87.2\%).
        Mean per-task accuracy with 95\% task-bootstrap
        intervals; chance is 0.13\%. Pixel and patchifier values average ten projections or seeds.
    }
    \label{fig:color-permutation}
\end{figure*}
\begin{figure*}[tbp]
    \centering
    \includegraphics[width=0.67\textwidth]{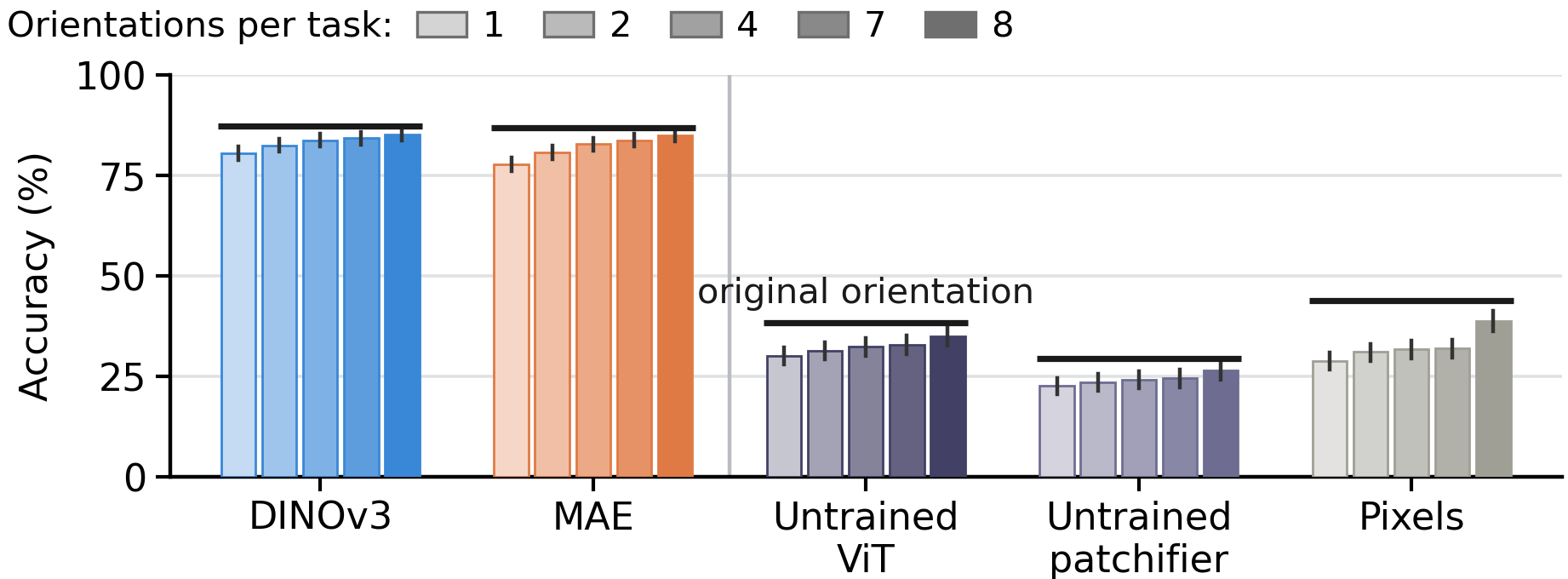}
    \caption{
        ARC-GEN task classification under rotation and reflection, the counterpart of
        Fig.~\ref{fig:color-permutation-bars}. Every instance of Fig.~\ref{fig:task-id}
        (794 tasks $\times$ 500 instances, 224 px) is re-rendered in each of the seven non-identity
        rotations and reflections of the square, applied to input and output alike, and classified
        with the same readout and layers; accuracies average the seven test orientations. The
        population (the instances that define each task's centroid) holds one orientation per
        instance, drawn from 1 (the original only), 2, 4 or 7 orientations per task that always
        include the original and never the classified instance's own, or from all 8, including it.
        Black segments: accuracy in the original orientation, which rotating or reflecting a whole
        task consistently leaves unchanged (DINOv3 87.2\%). Mean per-task accuracy with 95\%
        task-bootstrap intervals; chance is 0.13\%. Pixel and patchifier values average ten
        projections or seeds.
    }
    \label{fig:supp-orientation}
\end{figure*}
\begin{figure*}[tbp]
    \centering
    \captionsetup[subfigure]{skip=2pt}
    \begin{subfigure}[t]{0.49\textwidth}
        \centering
        \includegraphics[width=\linewidth]{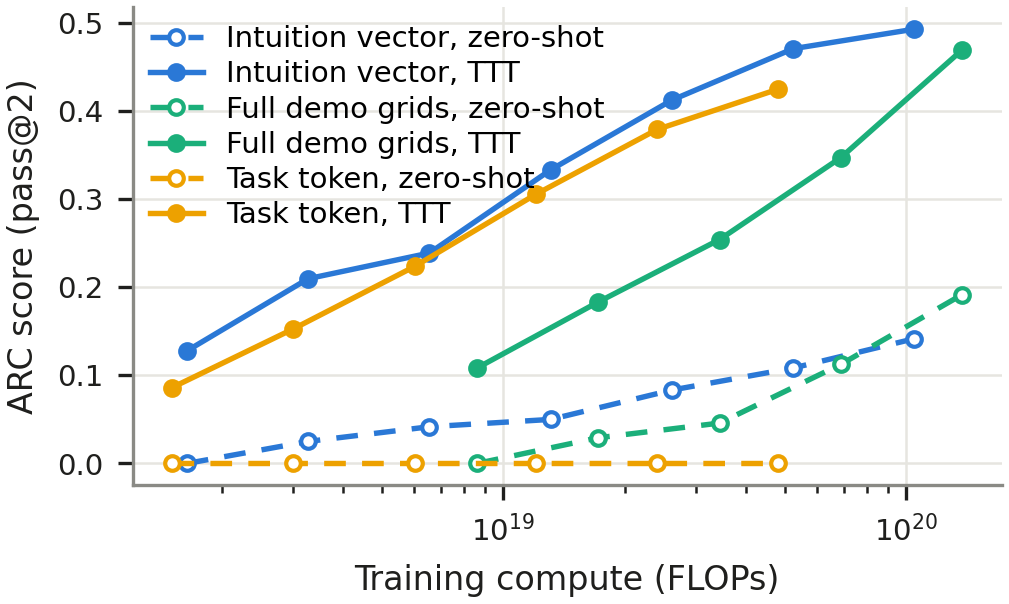}
        \caption{ARC1 performance against training compute.}
        \label{fig:arc-rae-training-compute}
    \end{subfigure}
    \hfill
    \begin{subfigure}[t]{0.49\textwidth}
        \centering
        \includegraphics[width=\linewidth]{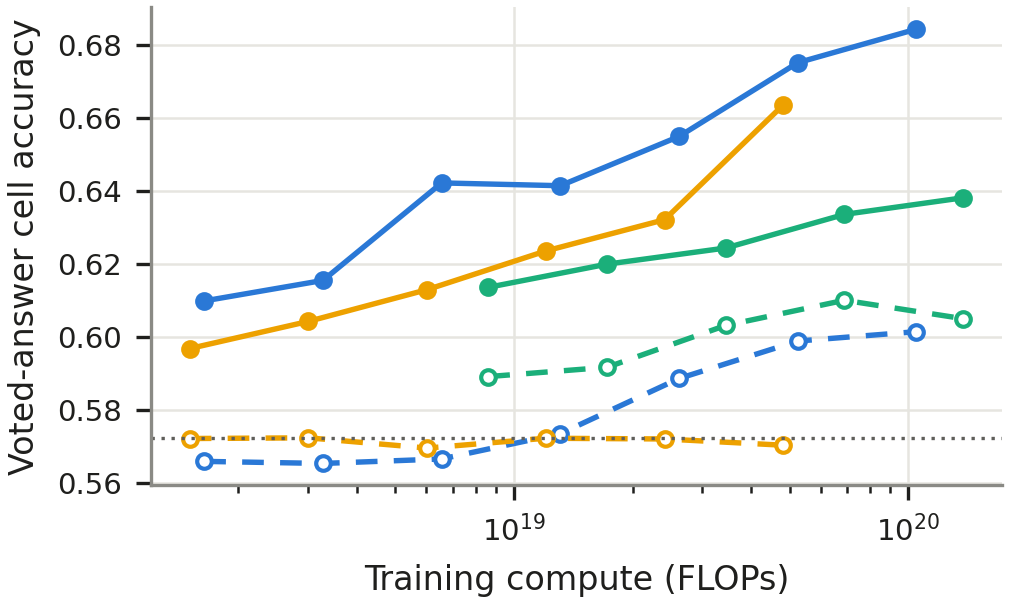}
        \caption{ARC2 performance against training compute.}
        \label{fig:arc-rae-training-compute-arc2}
    \end{subfigure}

    \vspace{0.3em}
    \begin{subfigure}[t]{0.60\textwidth}
        \centering
        \includegraphics[width=\linewidth]{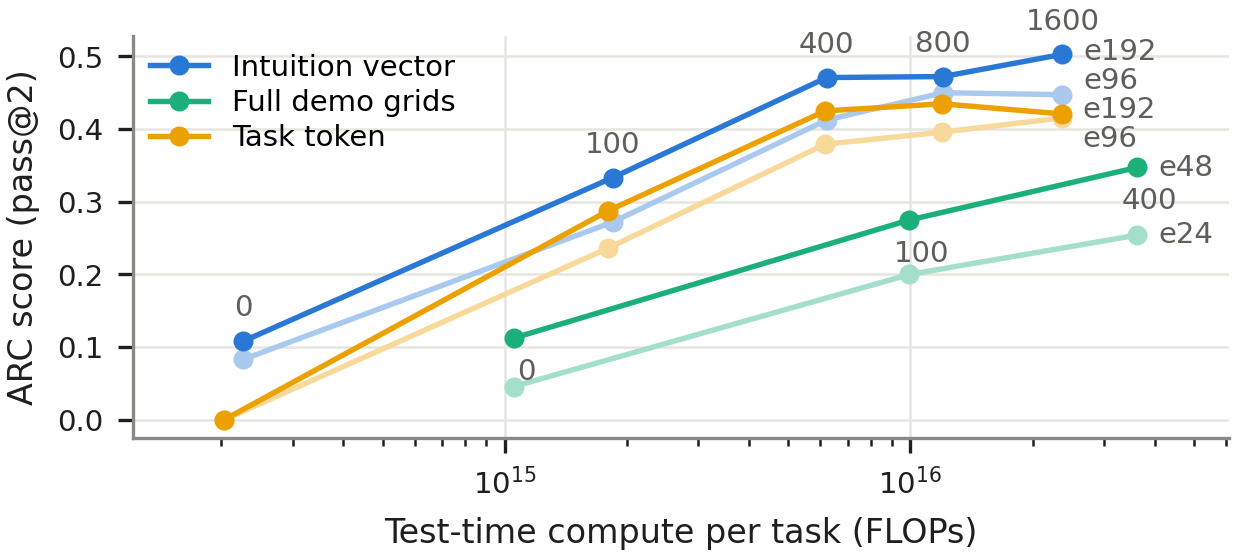}
        \caption{ARC1 performance against test-time compute.}
        \label{fig:arc-rae-ttt-budget}
    \end{subfigure}
    \hfill
    \begin{subfigure}[t]{0.38\textwidth}
        \centering
        \includegraphics[width=\linewidth]{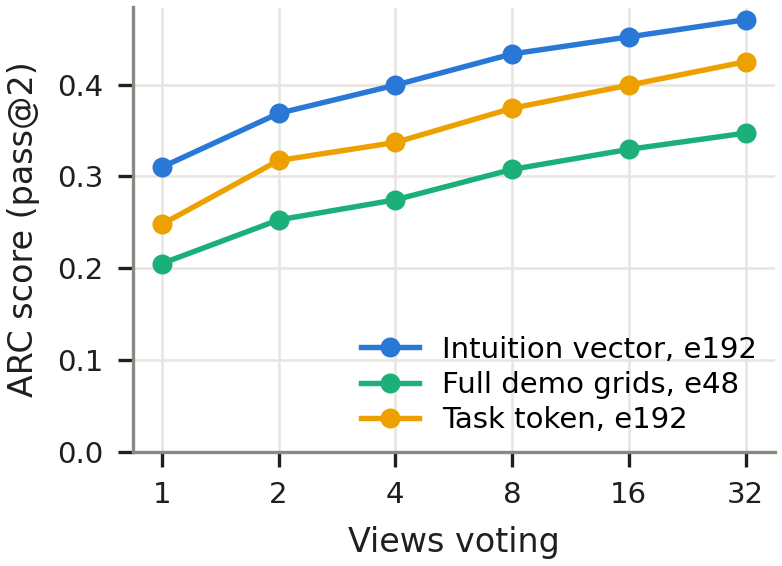}
        \caption{ARC1 performance against views voting.}
        \label{fig:arc-rae-view-voting}
    \end{subfigure}
    \caption{
        \textbf{ARC-RAE: compute--performance tradeoffs.}
        \textbf{(a, b)} Training compute versus ARC1 pass@2 (120 held-out tasks) and
        ARC2 cell accuracy (120 evaluation tasks), with a learned task-token control.
        The legend in (a) applies to (b). Dashed/open: zero-shot;
        solid/filled: three test-time-training (TTT) seeds after 400 whole-network steps.
        The dotted line in (b) copies the query input.
        Training FLOPs count encoder passes and transformer forward/backward passes
        over 29,393 training examples per epoch.
        \textbf{(c)} Numbers denote TTT steps (two fitting examples per step; 0 is zero-shot);
        checkpoint labels give training epochs. Darker curves use later checkpoints,
        paired across arms by training budget (see \emph{Compute}, App.~\ref{sec:supp-rae}).
        Filled: three TTT seeds (zero-shot is deterministic); open: fewer seeds.
        Test-time FLOPs include demonstration encoding, adaptation, and 32-view
        sampling and decoding, averaged over tasks.
        \textbf{(d)} Pass@2 after 400 TTT steps when a random subset of the 32 views
        votes (three TTT seeds), at the later checkpoints of (c).
        One training run per arm; maximum TTT-seed SD in (a, b): 0.034 and 0.011.
    }
    \label{fig:arc-rae}
\end{figure*}

\begin{figure}[tbp]
    \centering
    \includegraphics[width=0.8\linewidth]{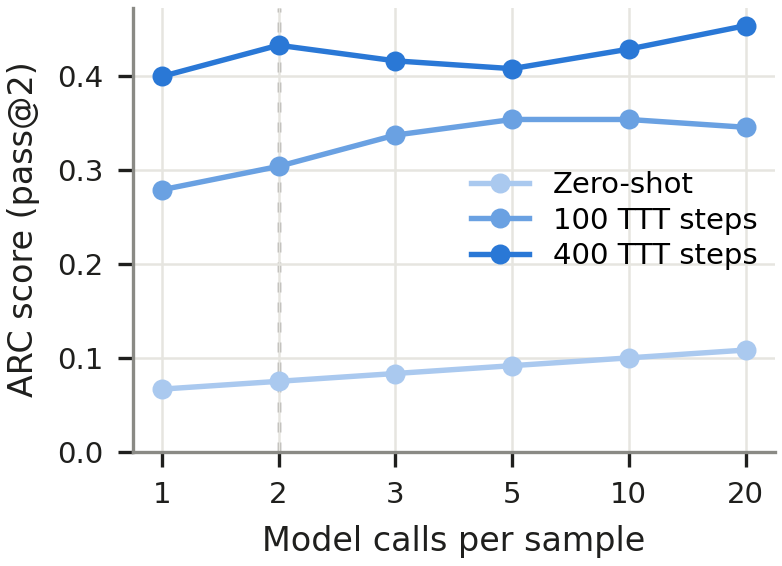}
    \caption{
        ARC1 held-out exact match (pass@2, 120 tasks) of the intuition-vector model (training
        epoch 192) against the number of network evaluations used to generate each answer,
        zero-shot and after 100 and 400 TTT steps. Dashed line: the two evaluations used
        throughout. Answers vote over the 8 rotations and reflections only, with one TTT seed, so
        values lie below those of the 32-view protocol. No count differs significantly from two
        (paired sign-flip tests over tasks, unadjusted, all $p\geq.13$).
    }
    \label{fig:arc-rae-sampler-calls}
\end{figure}

\begin{figure*}[tbp]
    \centering
    \includegraphics[width=\textwidth]{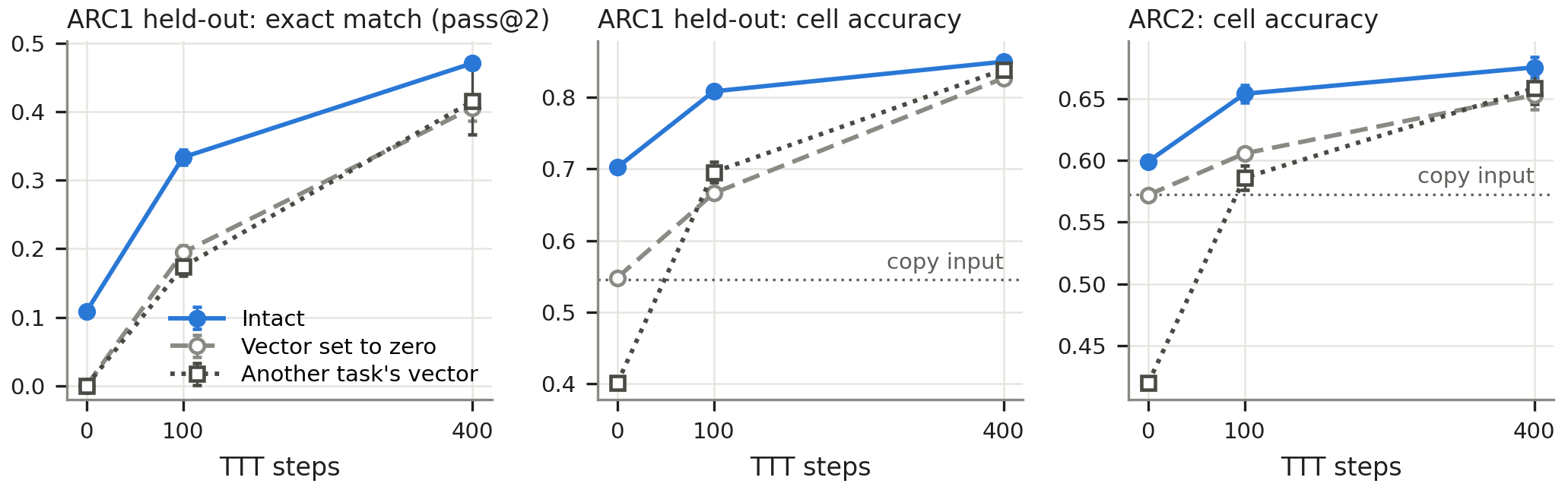}
    \caption{
        What the intuition vector contributes, by test-time-training budget. The intuition-vector
        model (training epoch 192) is evaluated with its vector intact, set to zero, or replaced by
        another task's vector, after 0, 100 and 400 TTT steps: ARC1 held-out exact match (pass@2,
        averaged over tasks), ARC1 cell accuracy and ARC2 cell accuracy (over test queries). Error
        bars: SD over three TTT seeds (zero-shot is deterministic). Dotted: copying the query input.
        Fig.~\ref{fig:rae2-ablation} shows the middle panel.
    }
    \label{fig:supp-rae-ablation}
\end{figure*}

\begin{figure*}[tbp]
    \centering
    \captionsetup[subfigure]{skip=2pt}

    \begin{subfigure}[t]{0.49\textwidth}
        \centering
        \includegraphics[width=\linewidth]{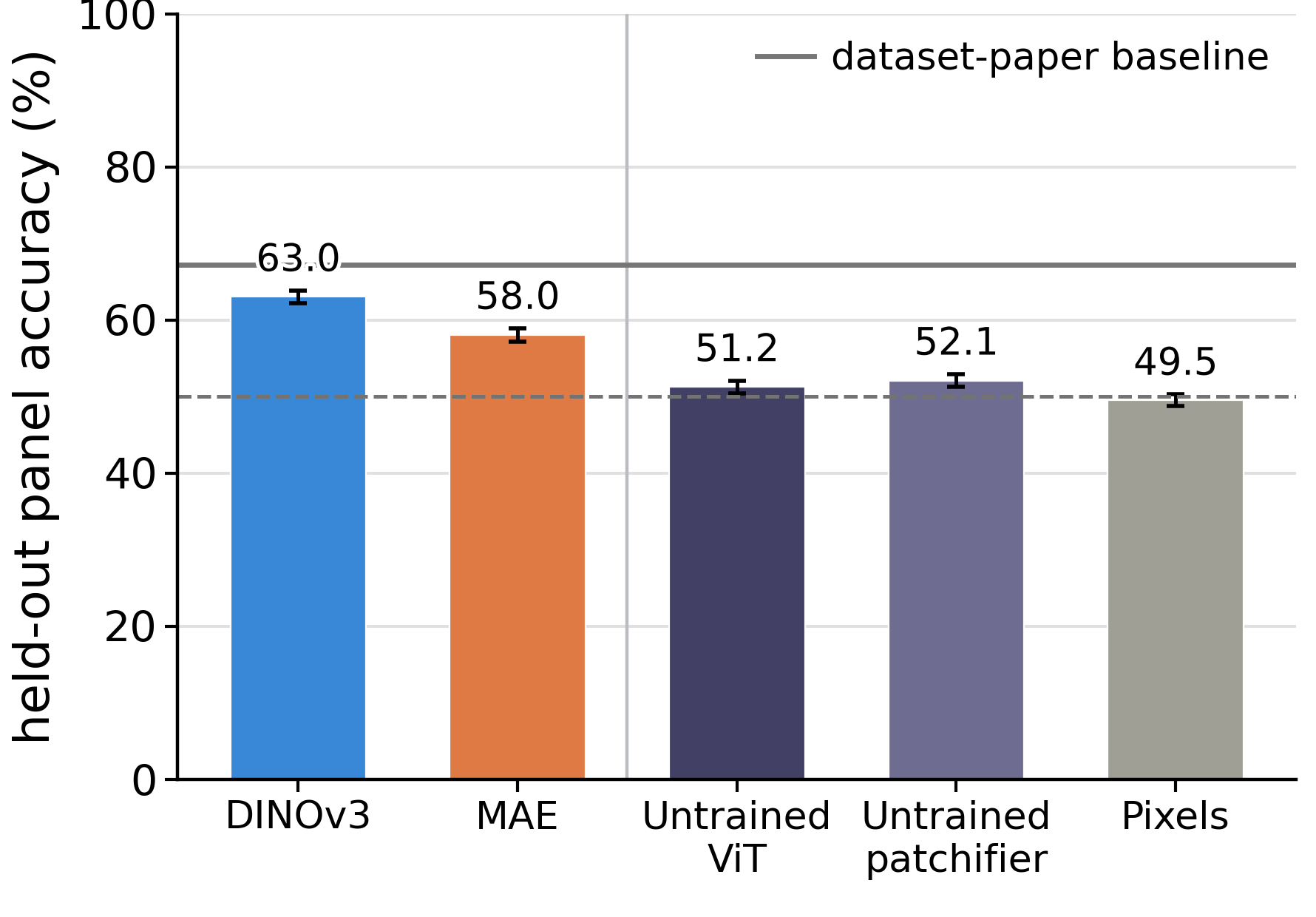}
        \caption{Bongard--LOGO, average over four test subsets.}
        \label{fig:supp-bongard-480-logo}
    \end{subfigure}
    \hfill
    \begin{subfigure}[t]{0.49\textwidth}
        \centering
        \includegraphics[width=\linewidth]{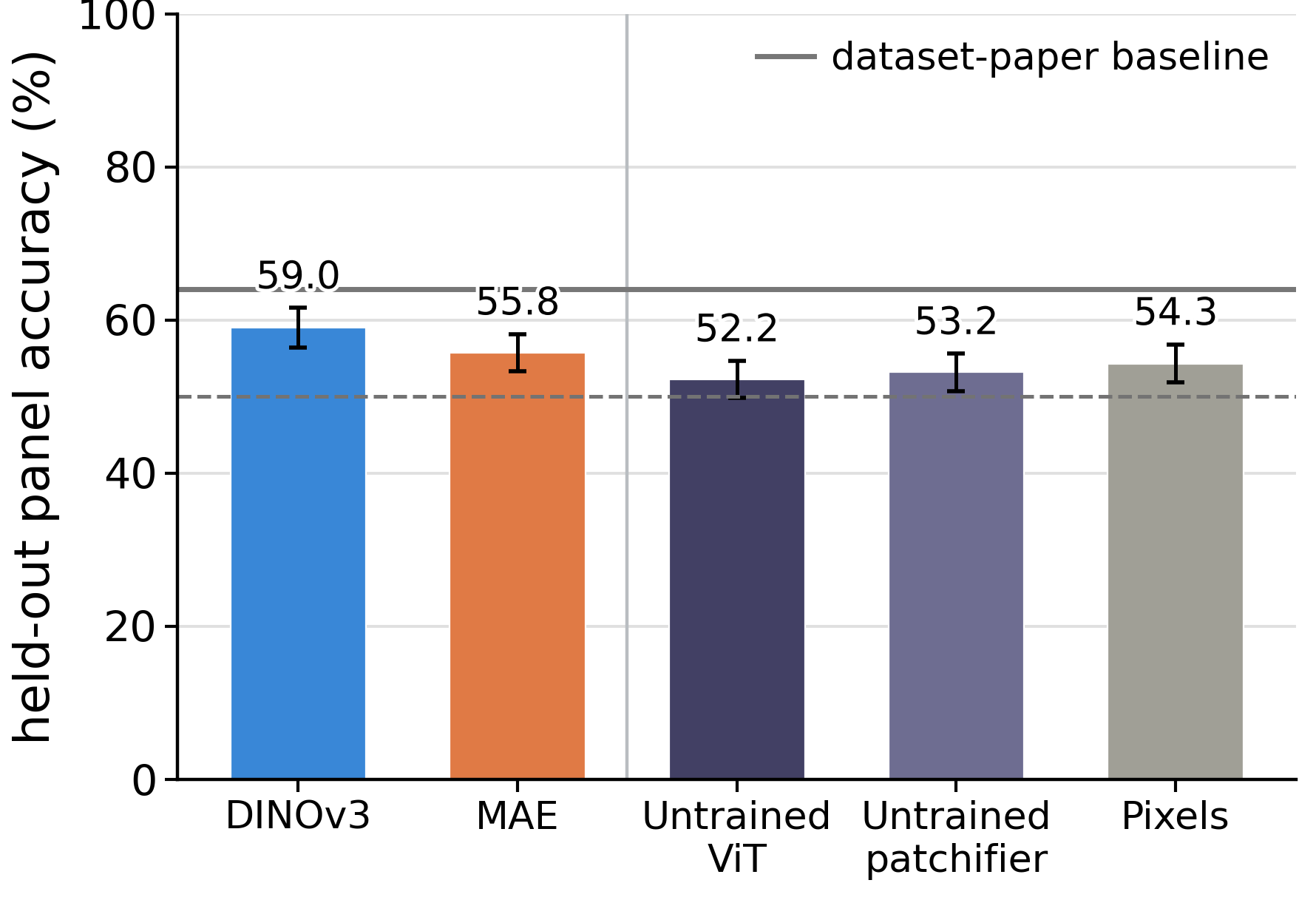}
        \caption{Bongard--OpenWorld, all 200 problems.}
        \label{fig:supp-bongard-480-ow}
    \end{subfigure}

    \caption{
        Bongard nearest-centroid accuracy at 480 px, the counterpart of
        Figure~\ref{fig:bongard-task-vector-analysis}(d,e). Models, layer rule, error bars and
        reference lines are as in that figure.
    }
    \label{fig:supp-bongard-480}
\end{figure*}

\begin{figure*}[tbp]
    \centering
    \captionsetup[subfigure]{skip=2pt}

    \begin{subfigure}[t]{\textwidth}
        \centering
        \includegraphics[width=0.9\linewidth]{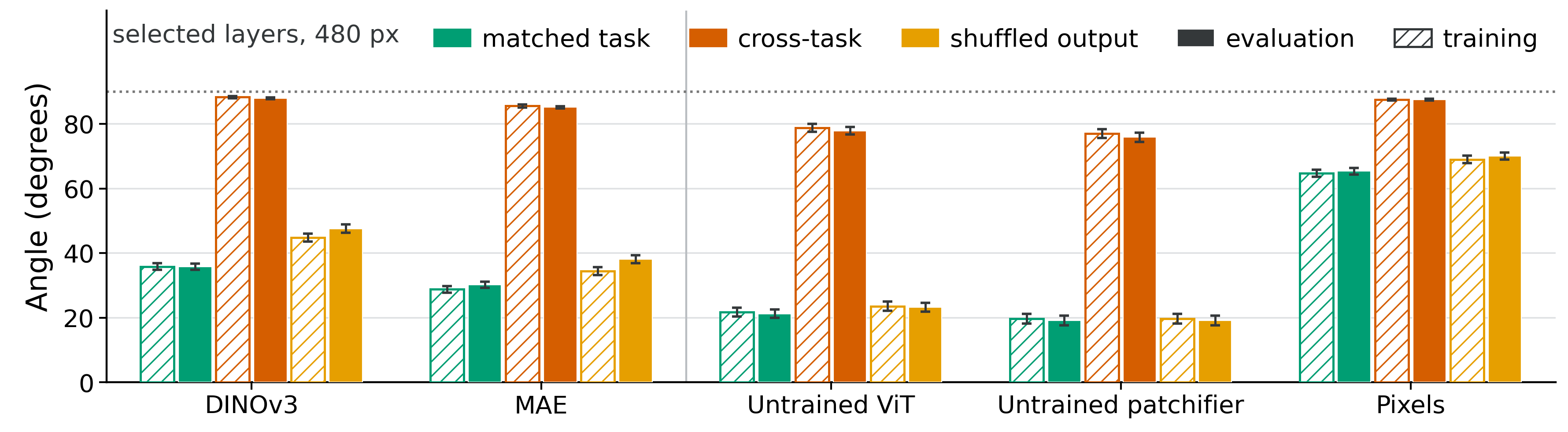}
        \caption{ARC-1 at 480 px.}
        \label{fig:supp-arc1-angle-480}
    \end{subfigure}

    \vspace{0.2em}

    \begin{subfigure}[t]{\textwidth}
        \centering
        \includegraphics[width=0.9\linewidth]{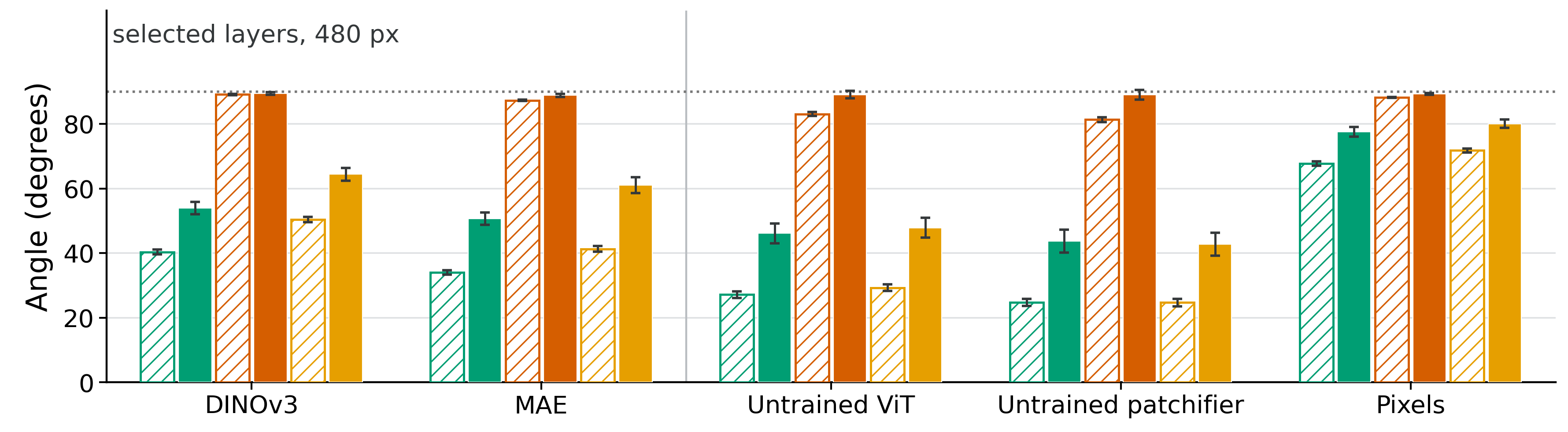}
        \caption{ARC-2 at 480 px.}
        \label{fig:supp-arc2-angle-480}
    \end{subfigure}

    \caption{
        Intuition to test vector angles at 480 px, at each encoder's selected layer (DINOv3 15,
        MAE 18, untrained ViT 7; Sec.~\ref{sec:methods-layers}), the counterpart of
        Figure~\ref{fig:arc-task-vector-analysis}(c,d); conditions, splits, controls and error
        bars are as in that figure. At this resolution each 16-pixel patch covers exactly one
        grid cell, so the untrained patchifier's mean is a linear function of the grid's color
        counts.
    }
    \label{fig:supp-arc-angle-480}
\end{figure*}

\begin{figure*}[tbp]
    \centering
    \captionsetup[subfigure]{skip=2pt}

    \includegraphics[width=0.8\textwidth]{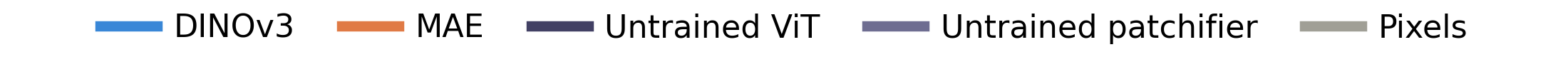}

    \vspace{0.2em}

    \begin{subfigure}[t]{0.49\textwidth}
        \centering
        \includegraphics[width=0.9\linewidth]{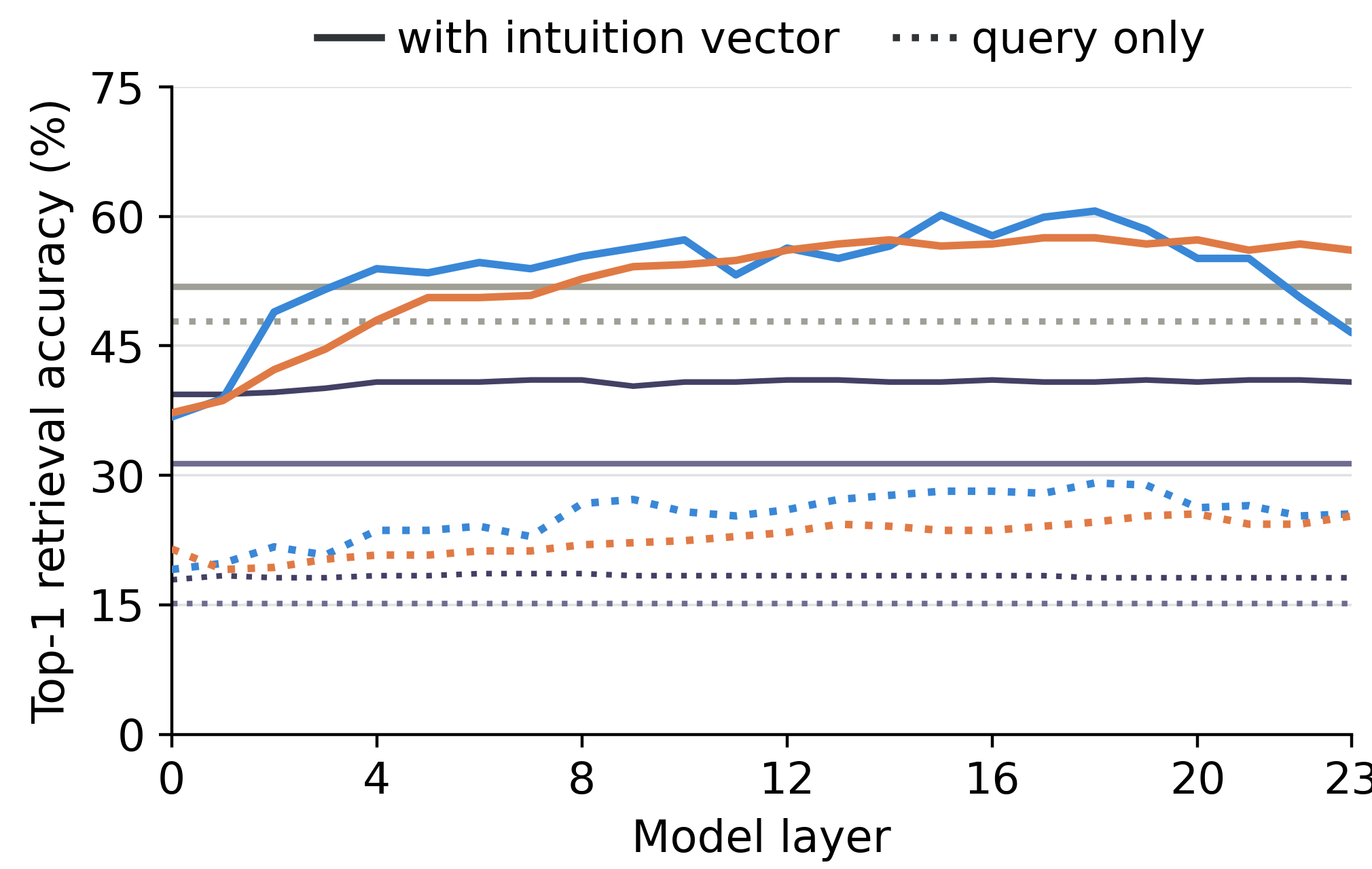}
        \caption{ARC-1, evaluation set.}
        \label{fig:supp-arc1-retrieval-480}
    \end{subfigure}
    \hfill
    \begin{subfigure}[t]{0.49\textwidth}
        \centering
        \includegraphics[width=0.9\linewidth]{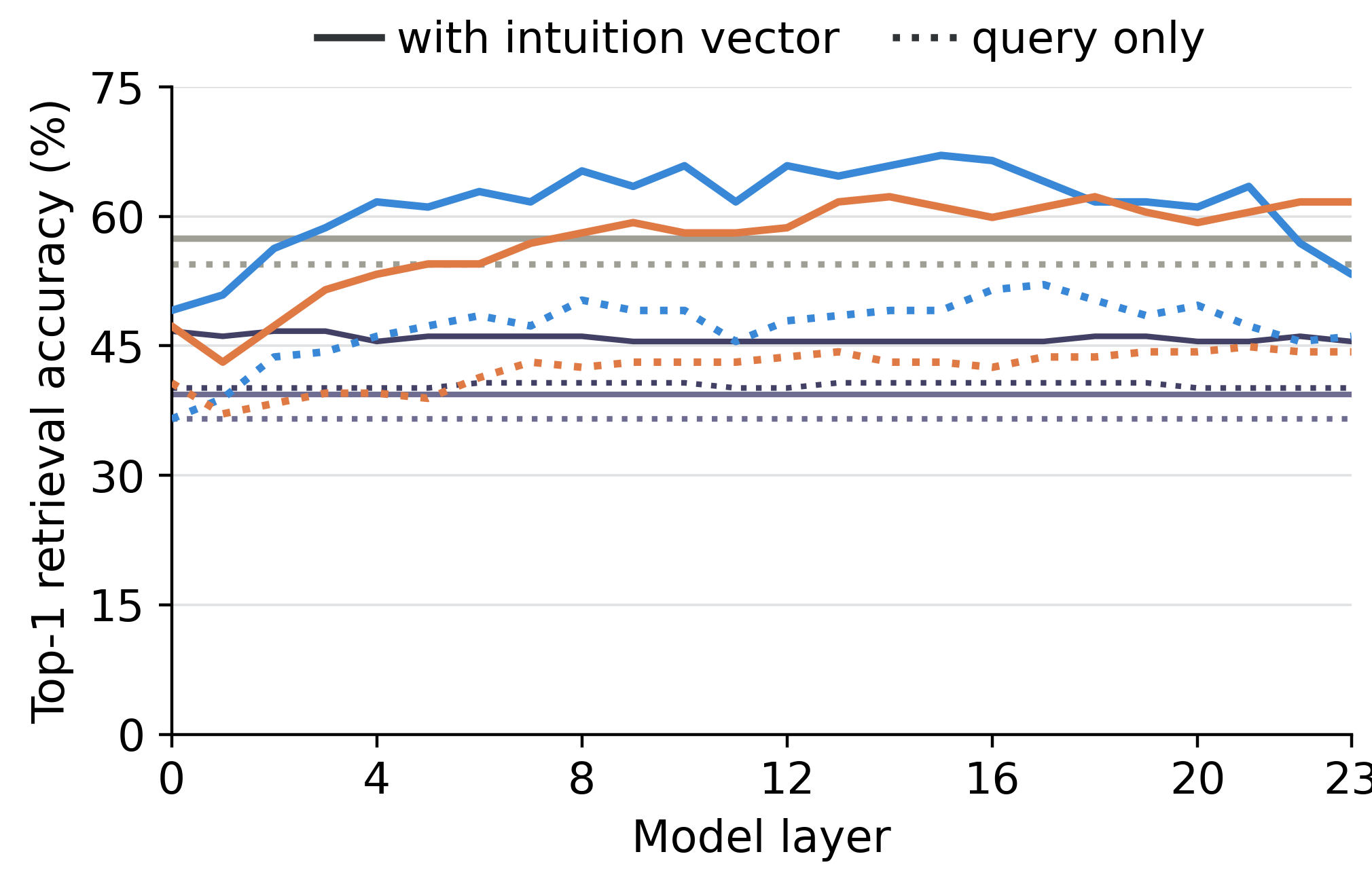}
        \caption{ARC-2, evaluation set.}
        \label{fig:supp-arc2-retrieval-480}
    \end{subfigure}

    \vspace{0.4em}

    \begin{subfigure}[t]{0.49\textwidth}
        \centering
        \includegraphics[width=0.9\linewidth]{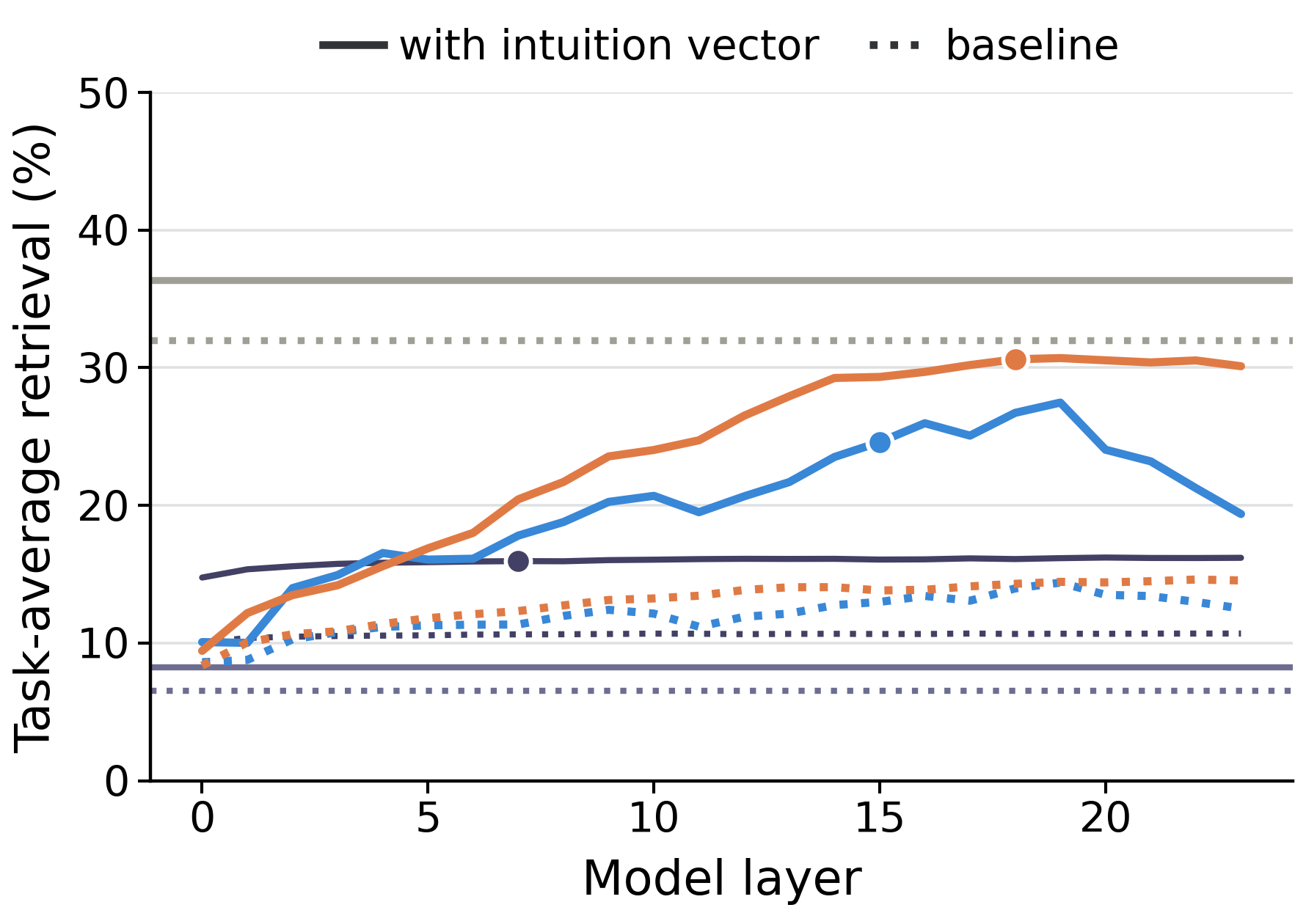}
        \caption{ARC-GEN.}
        \label{fig:supp-arcgen-retrieval-480}
    \end{subfigure}
    \hfill
    \begin{subfigure}[t]{0.49\textwidth}
        \centering
        \includegraphics[width=0.9\linewidth]{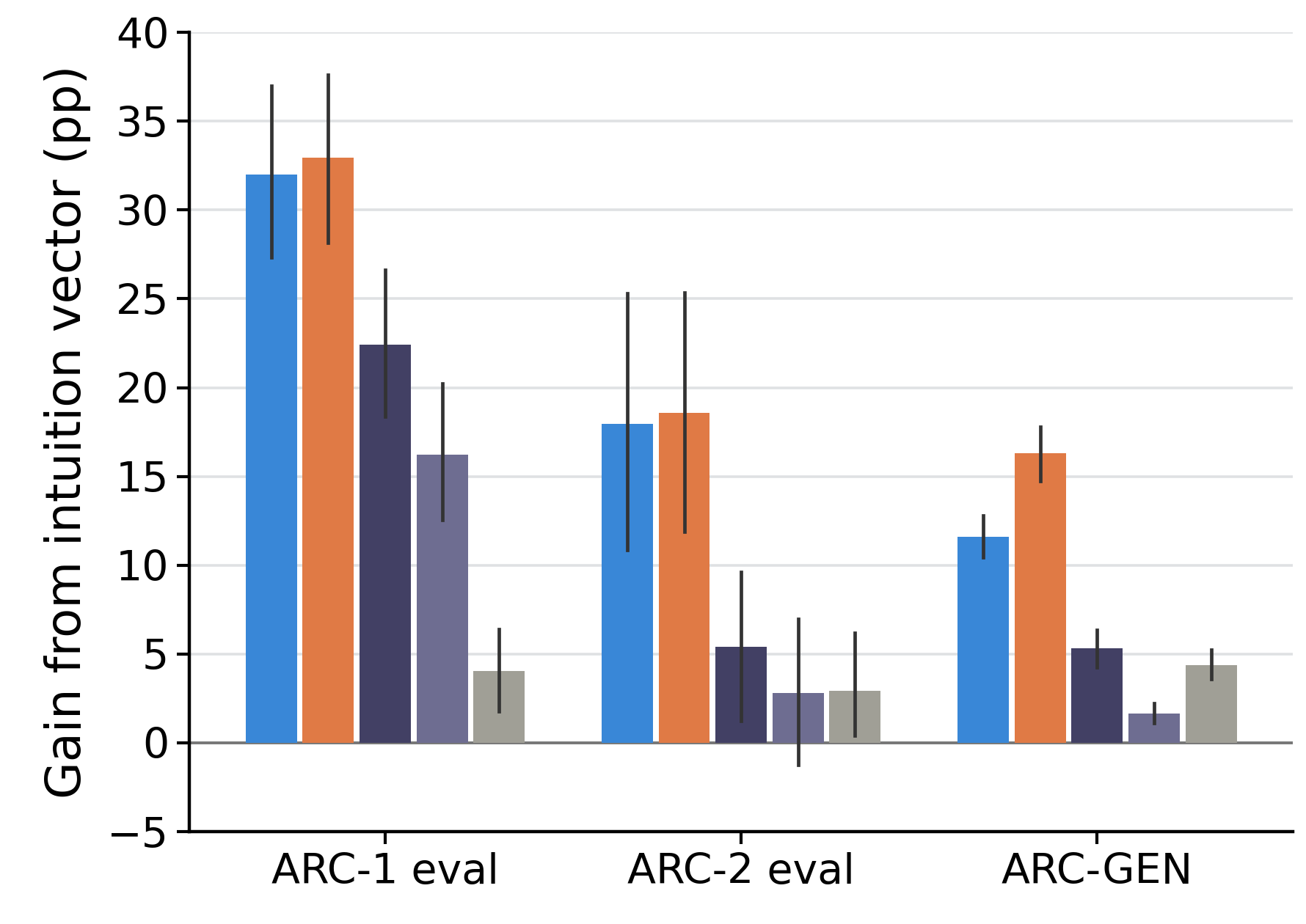}
        \caption{Gain from the intuition vector.}
        \label{fig:supp-retrieval-gain-summary-480}
    \end{subfigure}

    \caption{
        Output retrieval at 480 px, the counterpart of
        Figure~\ref{fig:arc-knn-new-analysis}(c--f); models, protocols, layer markers and
        intervals are as in that figure.
    }
    \label{fig:supp-arc-retrieval-480}
\end{figure*}

\begin{figure*}[tbp]
    \centering
    \captionsetup[subfigure]{skip=2pt}

    \begin{subfigure}[t]{0.49\textwidth}
        \centering
        \includegraphics[width=\linewidth]{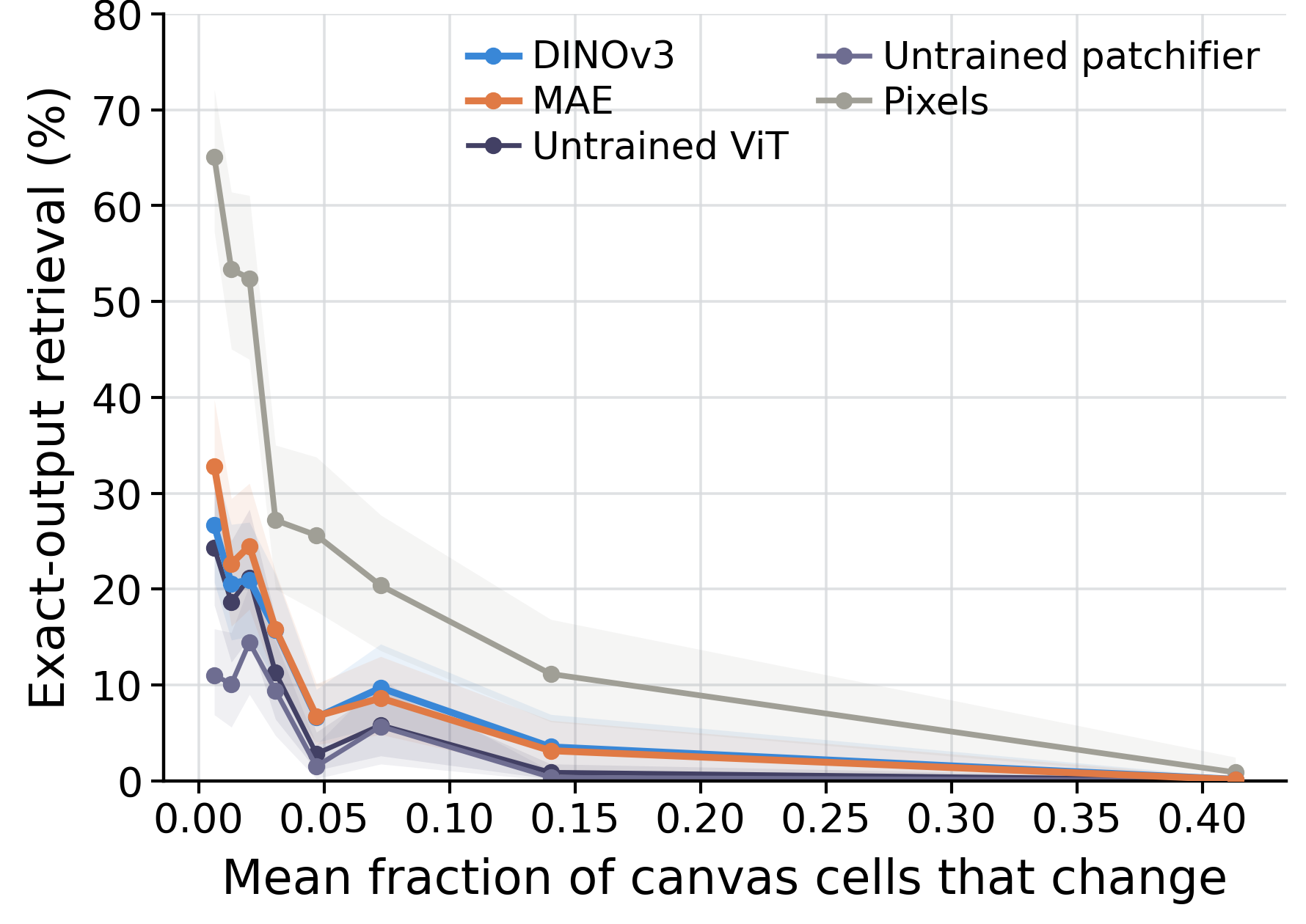}
        \caption{Retrieval without an intuition vector.}
        \label{fig:supp-arcgen-baseline-by-change-480}
    \end{subfigure}
    \hfill
    \begin{subfigure}[t]{0.49\textwidth}
        \centering
        \includegraphics[width=\linewidth]{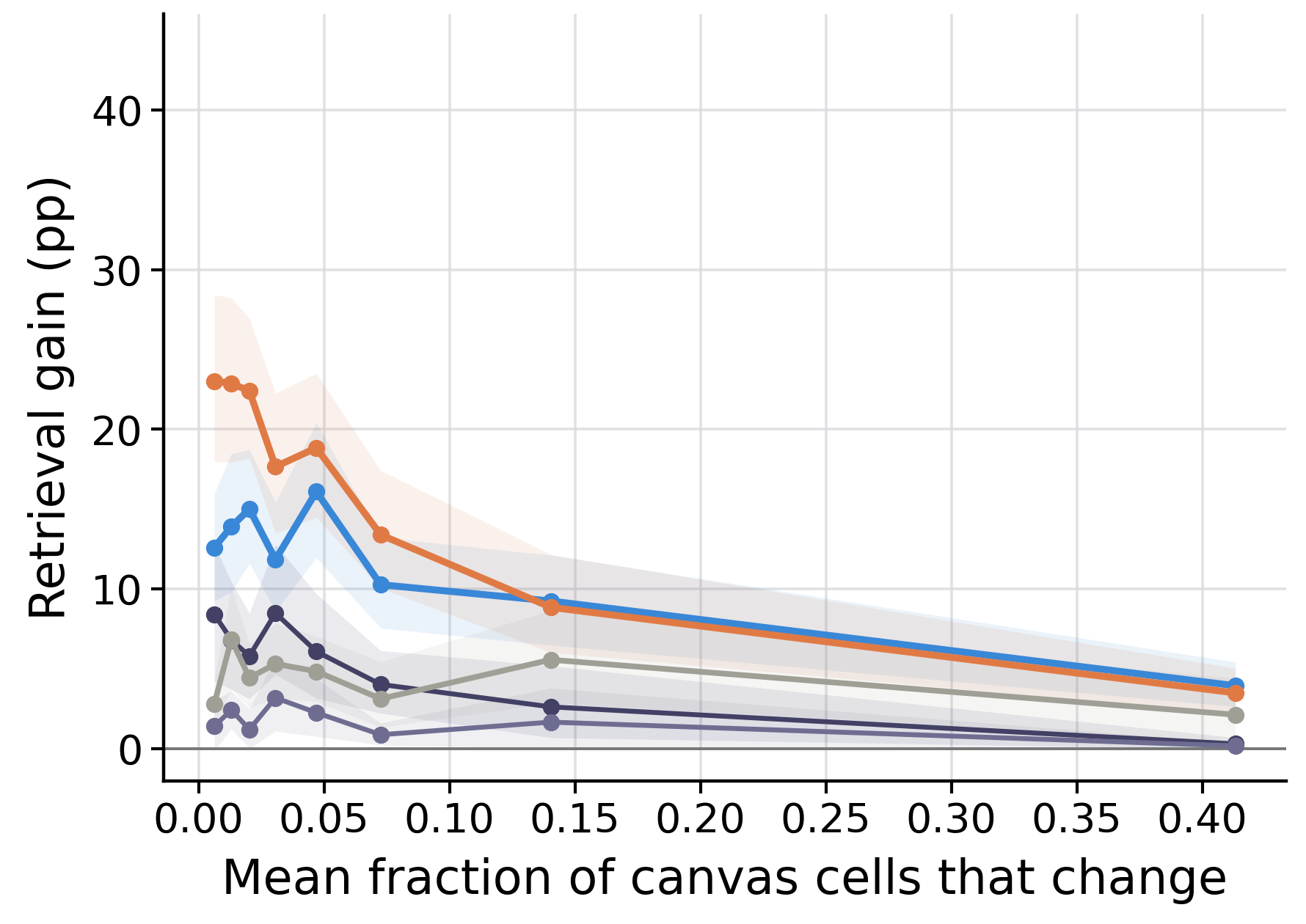}
        \caption{What the intuition vector adds.}
        \label{fig:supp-arcgen-lift-by-change-480}
    \end{subfigure}

    \caption{
        ARC-GEN baseline retrieval and intuition-vector gain by transformation extent at
        480 px, the counterpart of Figure~\ref{fig:arcgen-breakdown}; bins and intervals are
        shared with that figure.
    }
    \label{fig:supp-arcgen-breakdown-480}
\end{figure*}

\begin{figure*}[tbp]
    \centering
    \captionsetup[subfigure]{skip=2pt}

    \begin{subfigure}[t]{0.49\textwidth}
        \centering
        \includegraphics[width=0.9\linewidth]{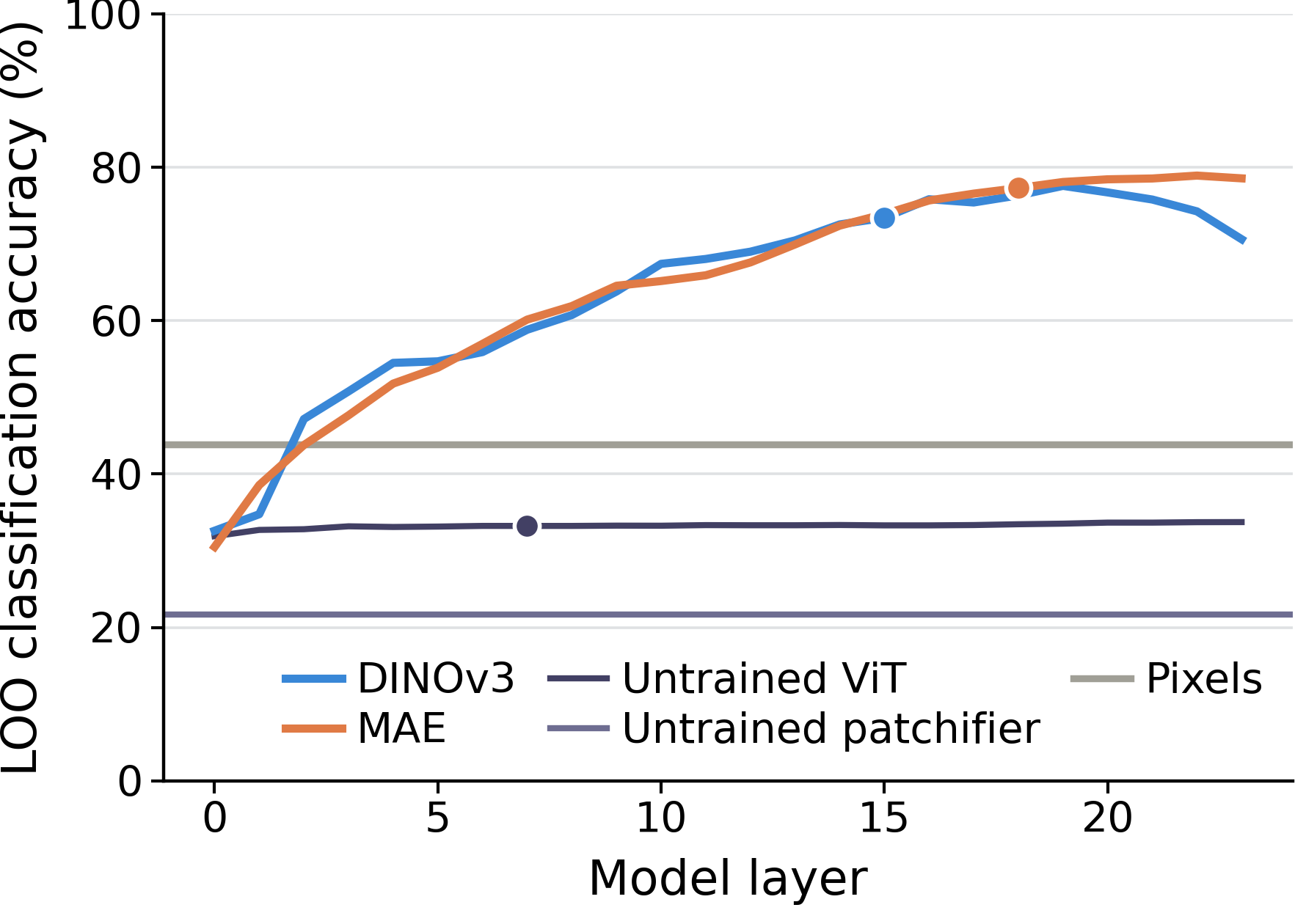}
        \caption{Instance classification across layers.}
        \label{fig:supp-arcgen-classification-by-layer-480}
    \end{subfigure}
    \hfill
    \begin{subfigure}[t]{0.49\textwidth}
        \centering
        \includegraphics[width=0.9\linewidth]{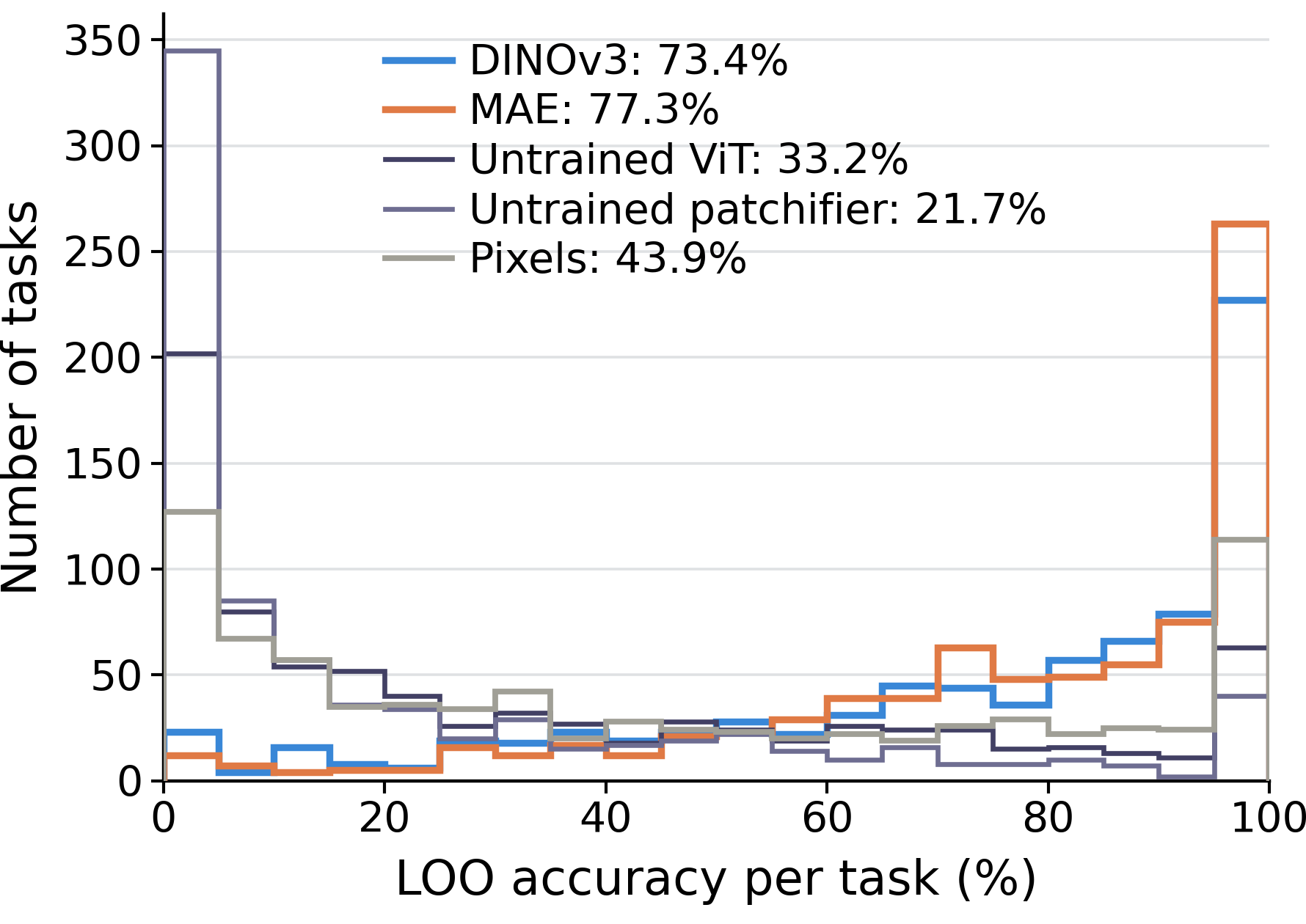}
        \caption{Per-task performance distribution.}
        \label{fig:supp-arcgen-classification-histogram-480}
    \end{subfigure}

    \caption{
        ARC-GEN leave-one-out task classification at 480 px, the counterpart of
        Figure~\ref{fig:arcgen-classification}(b,c); models, instances and layer markers are as
        in that figure.
    }
    \label{fig:supp-arcgen-classification-480}
\end{figure*}

\begin{figure*}[p]
    \centering
    \captionsetup[subfigure]{skip=2pt}

    \begin{subfigure}[t]{0.98\textwidth}
        \centering
        \includegraphics[width=\linewidth]{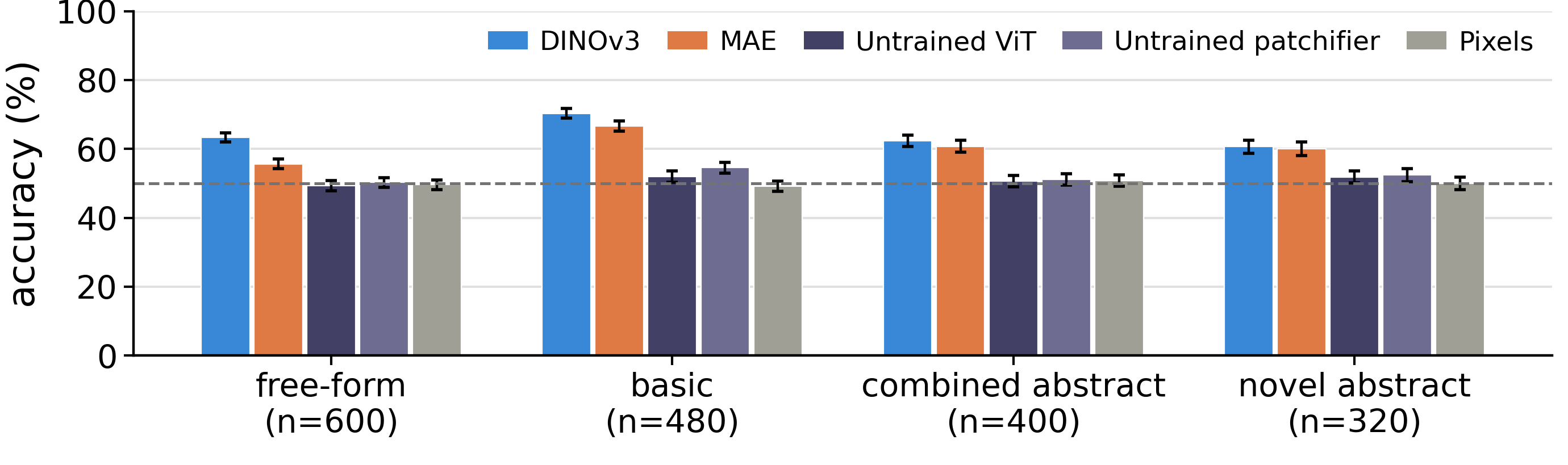}
        \caption{Bongard--LOGO test subsets, 224 px.}
        \label{fig:supp-bongard-subtasks-logo-224}
    \end{subfigure}

    \vspace{0.35em}

    \begin{subfigure}[t]{0.98\textwidth}
        \centering
        \includegraphics[width=\linewidth]{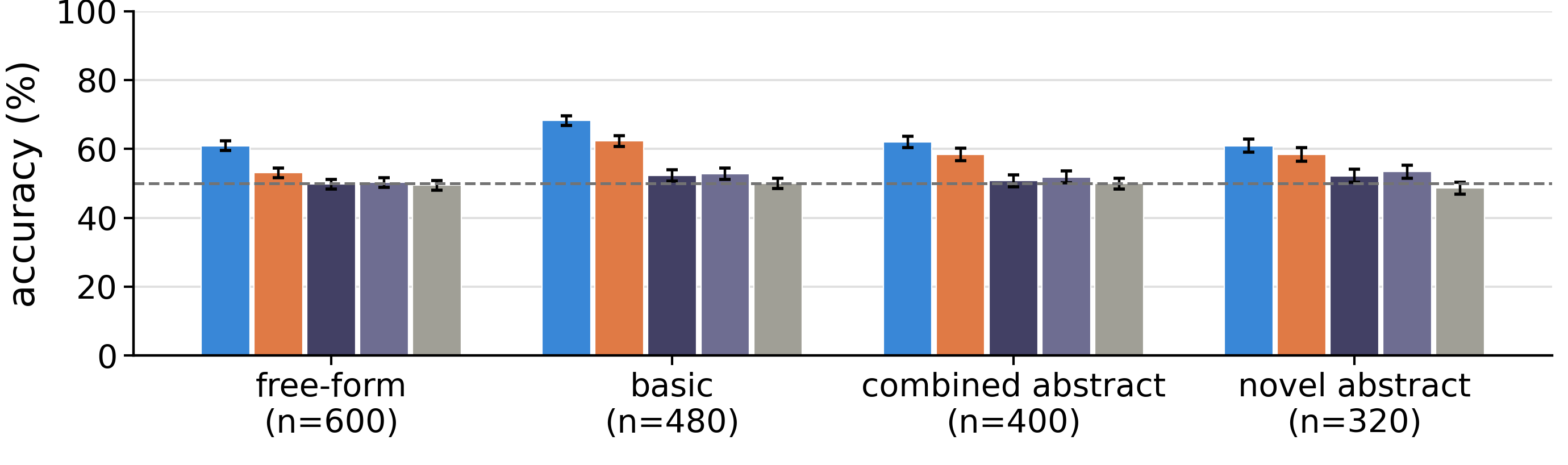}
        \caption{Bongard--LOGO test subsets, 480 px.}
        \label{fig:supp-bongard-subtasks-logo-480}
    \end{subfigure}

    \vspace{0.35em}

    \begin{subfigure}[t]{0.98\textwidth}
        \centering
        \includegraphics[width=\linewidth]{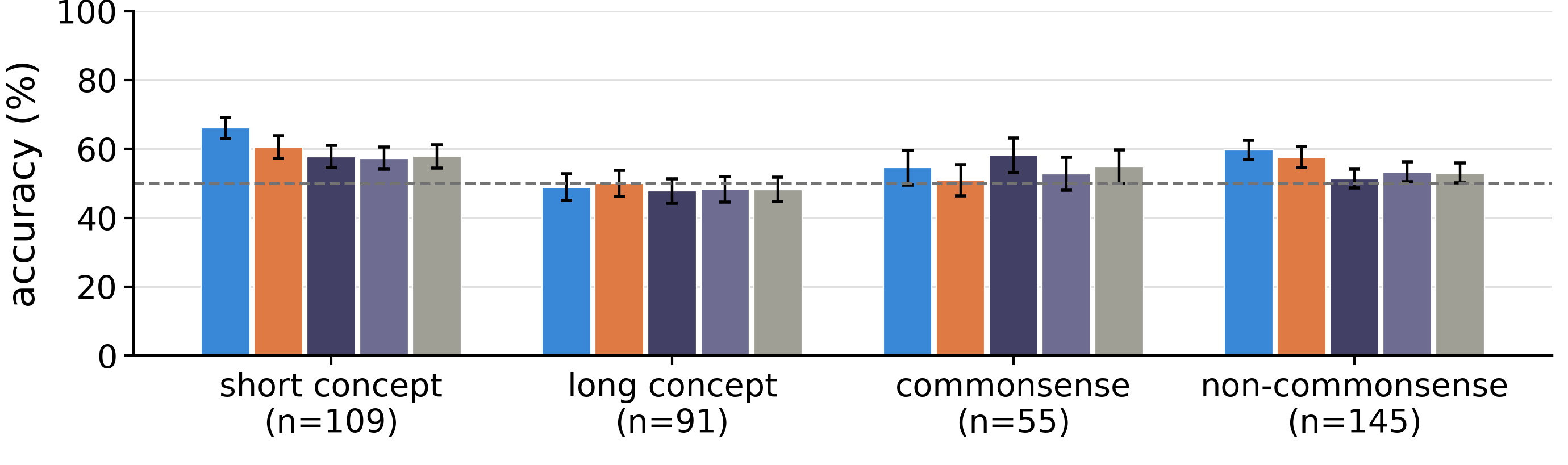}
        \caption{Bongard--OpenWorld subgroups, 224 px.}
        \label{fig:supp-bongard-subtasks-ow-224}
    \end{subfigure}

    \vspace{0.35em}

    \begin{subfigure}[t]{0.98\textwidth}
        \centering
        \includegraphics[width=\linewidth]{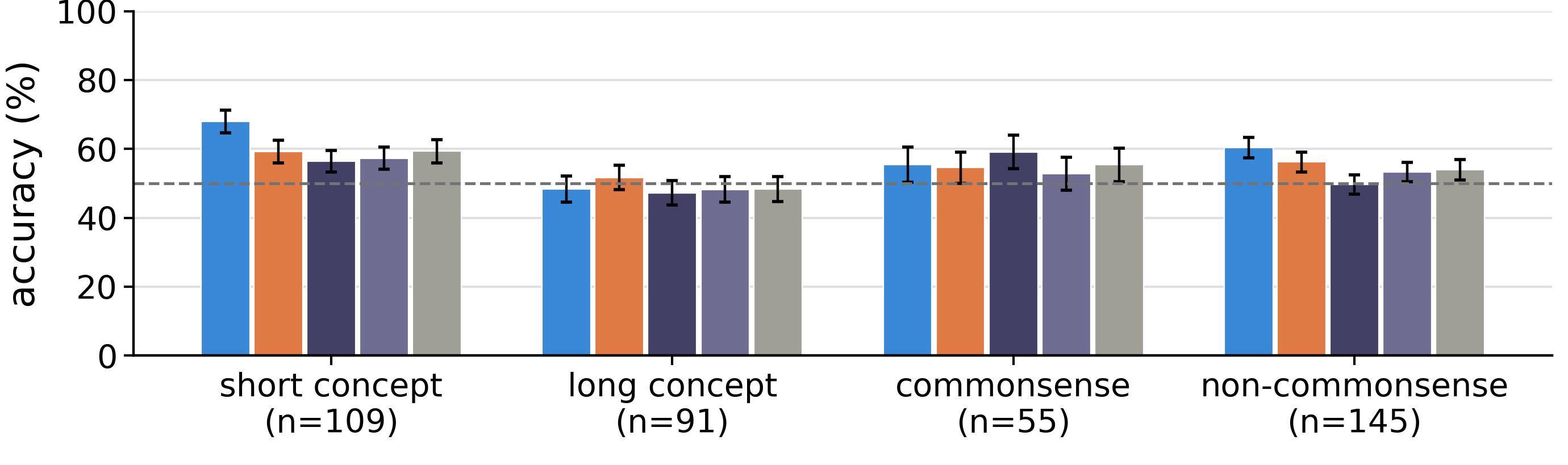}
        \caption{Bongard--OpenWorld subgroups, 480 px.}
        \label{fig:supp-bongard-subtasks-ow-480}
    \end{subfigure}

    \caption{
        Per-subtask nearest-centroid accuracy on the Bongard benchmarks, at the
        settings selected for Figure~\ref{fig:bongard-task-vector-analysis}(d,e).
        \textbf{(a,b)} The four disjoint Bongard--LOGO test subsets at 224 and
        480 pixels. \textbf{(c,d)} The Bongard--OpenWorld subgroups at 224 and
        480 pixels. Bars compare DINOv3, a MAE encoder of matched size, the
        untrained ViT and untrained patchifier of
        Figure~\ref{fig:bongard-task-vector-analysis}, and a random-projection
        pixel control; the legend in \textbf{(a)} applies to every panel. The
        encoder layer is selected on training problems within each dataset and
        resolution, as in Figure~\ref{fig:bongard-task-vector-analysis}, and held
        fixed across all subtasks, so neither the layer nor any subtask is chosen
        on test data. Error bars are the
        SEM across problems for the encoders and, for pixels, a conservative
        upper bound on the SEM of the mean over ten projection seeds; for the
        untrained patchifier, the mean problem-level SEM over its ten seeds. The dashed
        line marks $50\%$ chance. No dataset-paper baseline is drawn, because
        those papers report values per subset rather than for these
        groupings. The OpenWorld subgroups form two overlapping partitions of
        the same 200 problems -- concept length (short/long) and commonsense
        requirement -- and must not be averaged together. The count beneath each
        group is its number of problems; every problem contributes two held-out
        decisions, one per class.
    }
    \label{fig:supp-bongard-subtasks}
\end{figure*}

\begin{figure*}[p]
    \centering
    \captionsetup[subfigure]{skip=2pt}

    \begin{subfigure}[t]{0.49\textwidth}
        \centering
        \includegraphics[width=\linewidth]{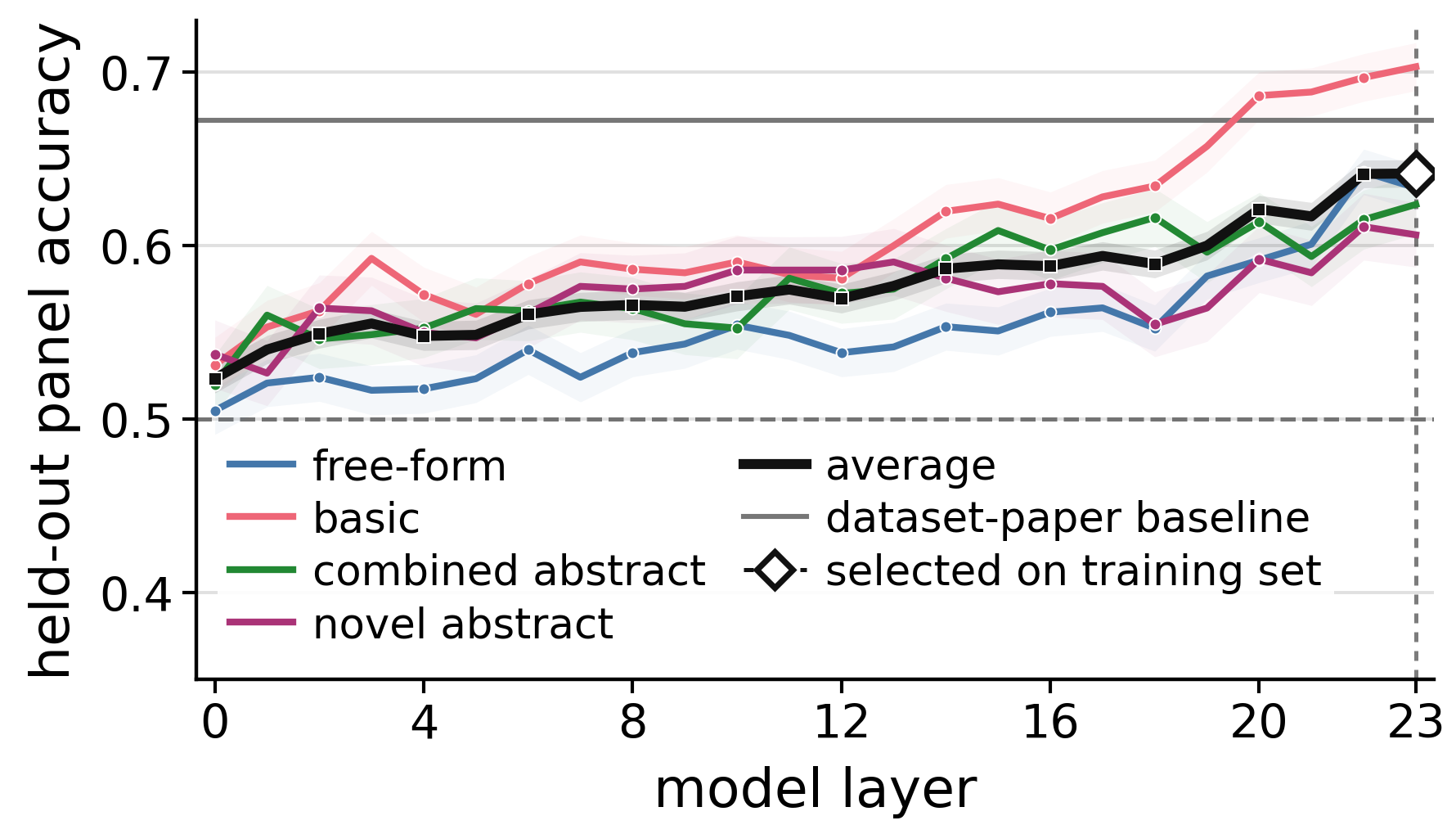}
        \caption{DINOv3, 224 px.}
        \label{fig:supp-bongard-layerwise-logo-dino224}
    \end{subfigure}
    \hfill
    \begin{subfigure}[t]{0.49\textwidth}
        \centering
        \includegraphics[width=\linewidth]{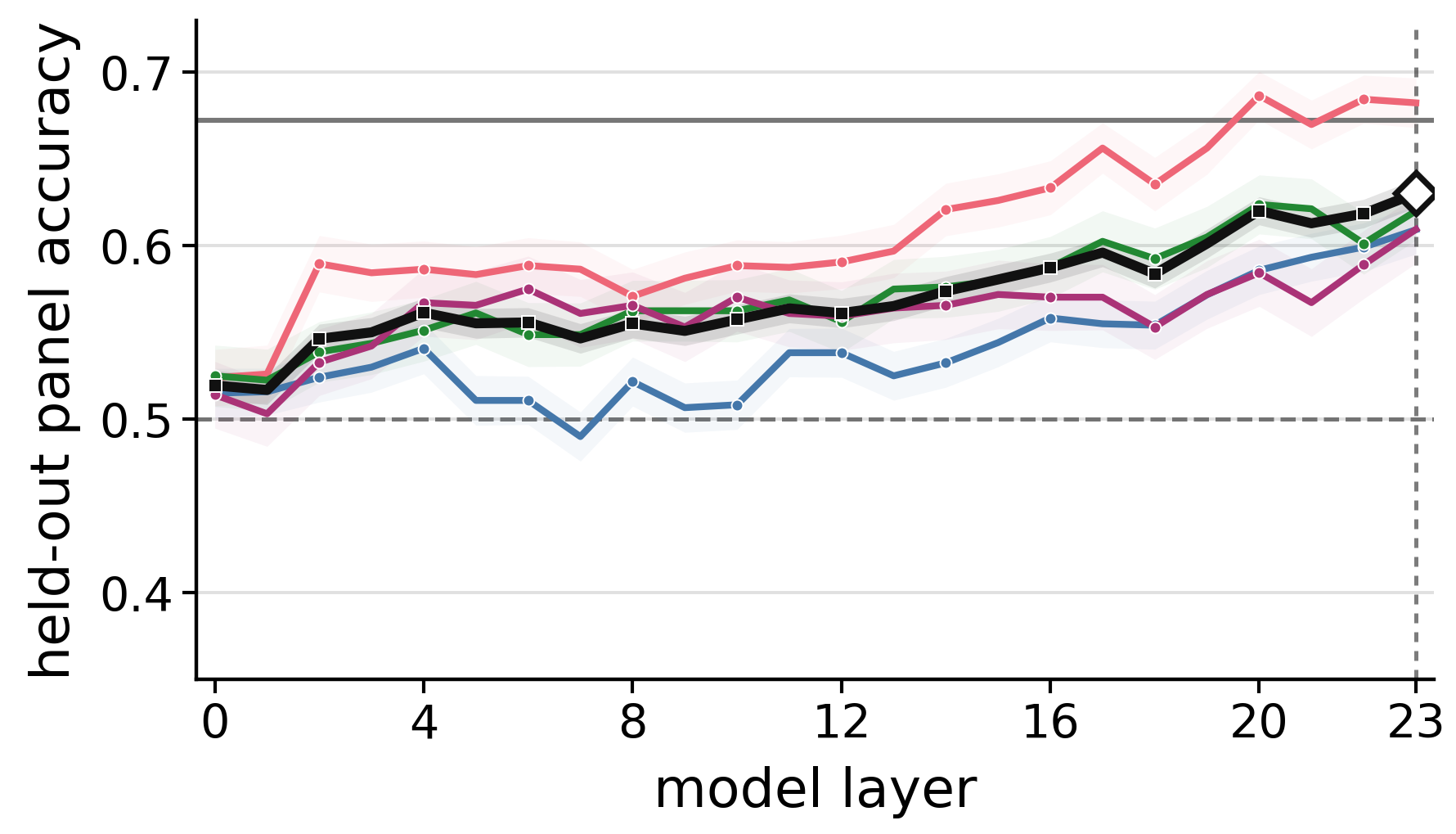}
        \caption{DINOv3, 480 px.}
        \label{fig:supp-bongard-layerwise-logo-dino480}
    \end{subfigure}

    \vspace{0.35em}

    \begin{subfigure}[t]{0.49\textwidth}
        \centering
        \includegraphics[width=\linewidth]{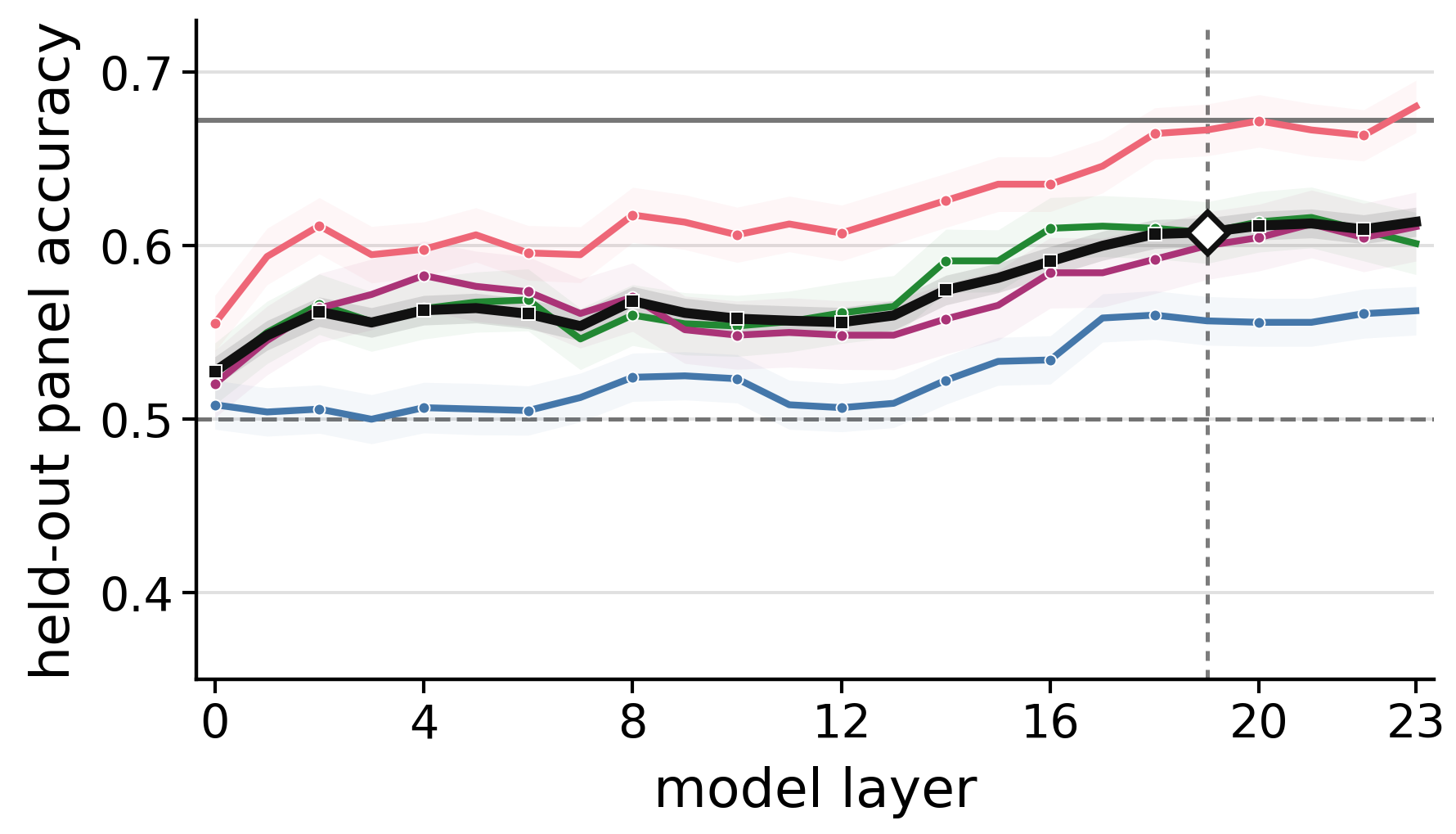}
        \caption{MAE, 224 px.}
        \label{fig:supp-bongard-layerwise-logo-mae224}
    \end{subfigure}
    \hfill
    \begin{subfigure}[t]{0.49\textwidth}
        \centering
        \includegraphics[width=\linewidth]{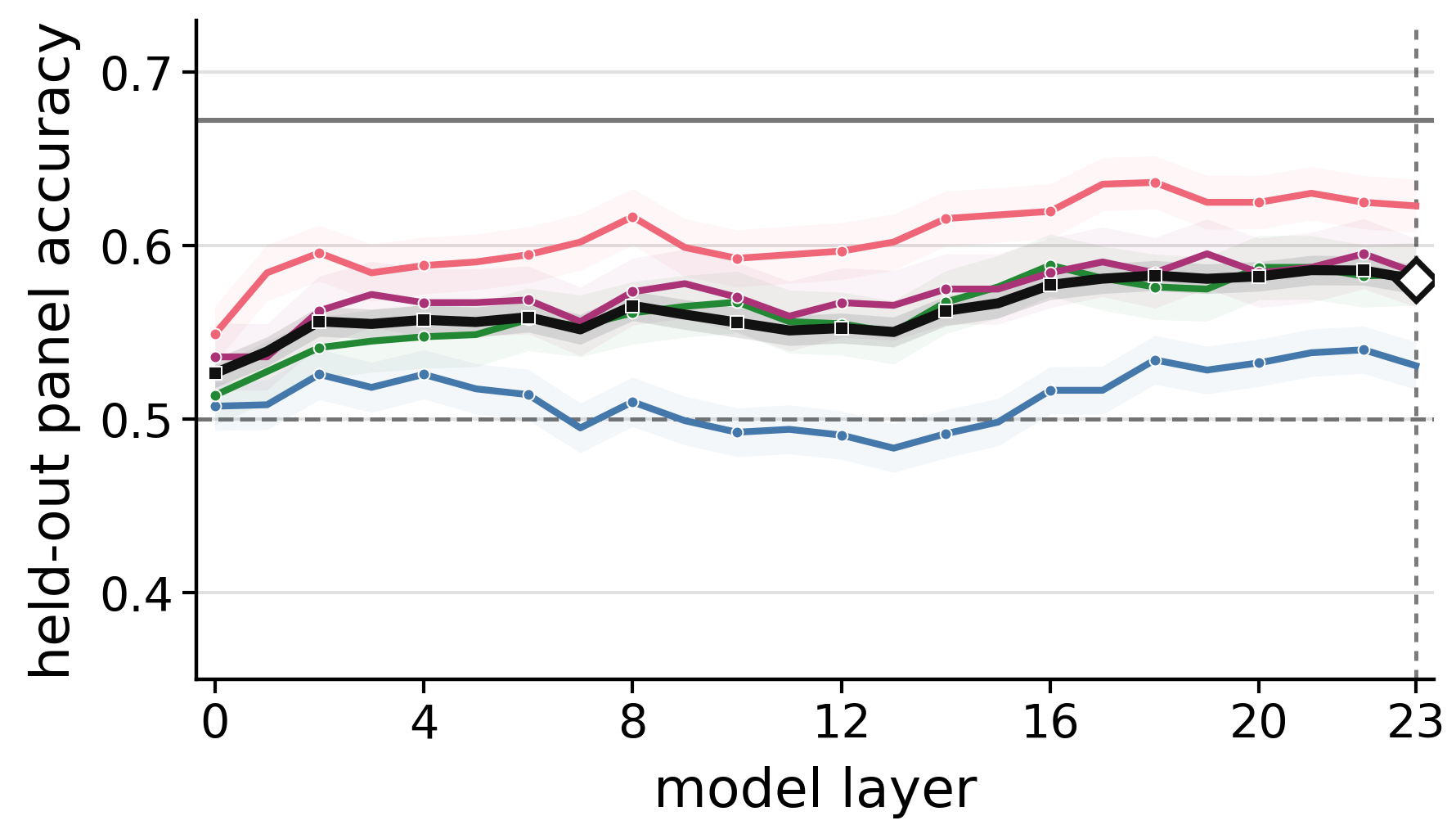}
        \caption{MAE, 480 px.}
        \label{fig:supp-bongard-layerwise-logo-mae480}
    \end{subfigure}

    \vspace{0.35em}

    \begin{subfigure}[t]{0.49\textwidth}
        \centering
        \includegraphics[width=\linewidth]{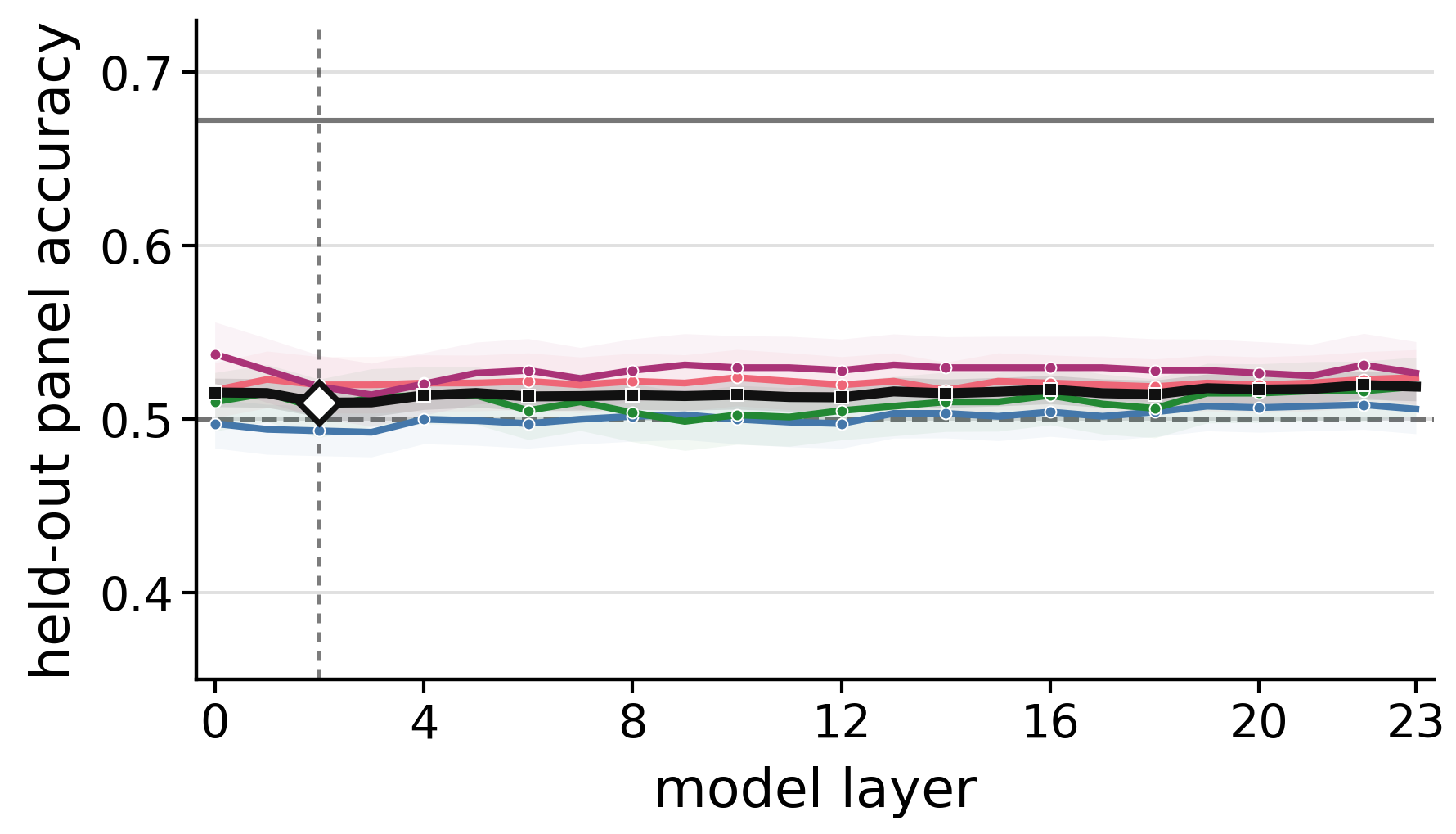}
        \caption{Untrained ViT, 224 px.}
        \label{fig:supp-bongard-layerwise-logo-vit224}
    \end{subfigure}
    \hfill
    \begin{subfigure}[t]{0.49\textwidth}
        \centering
        \includegraphics[width=\linewidth]{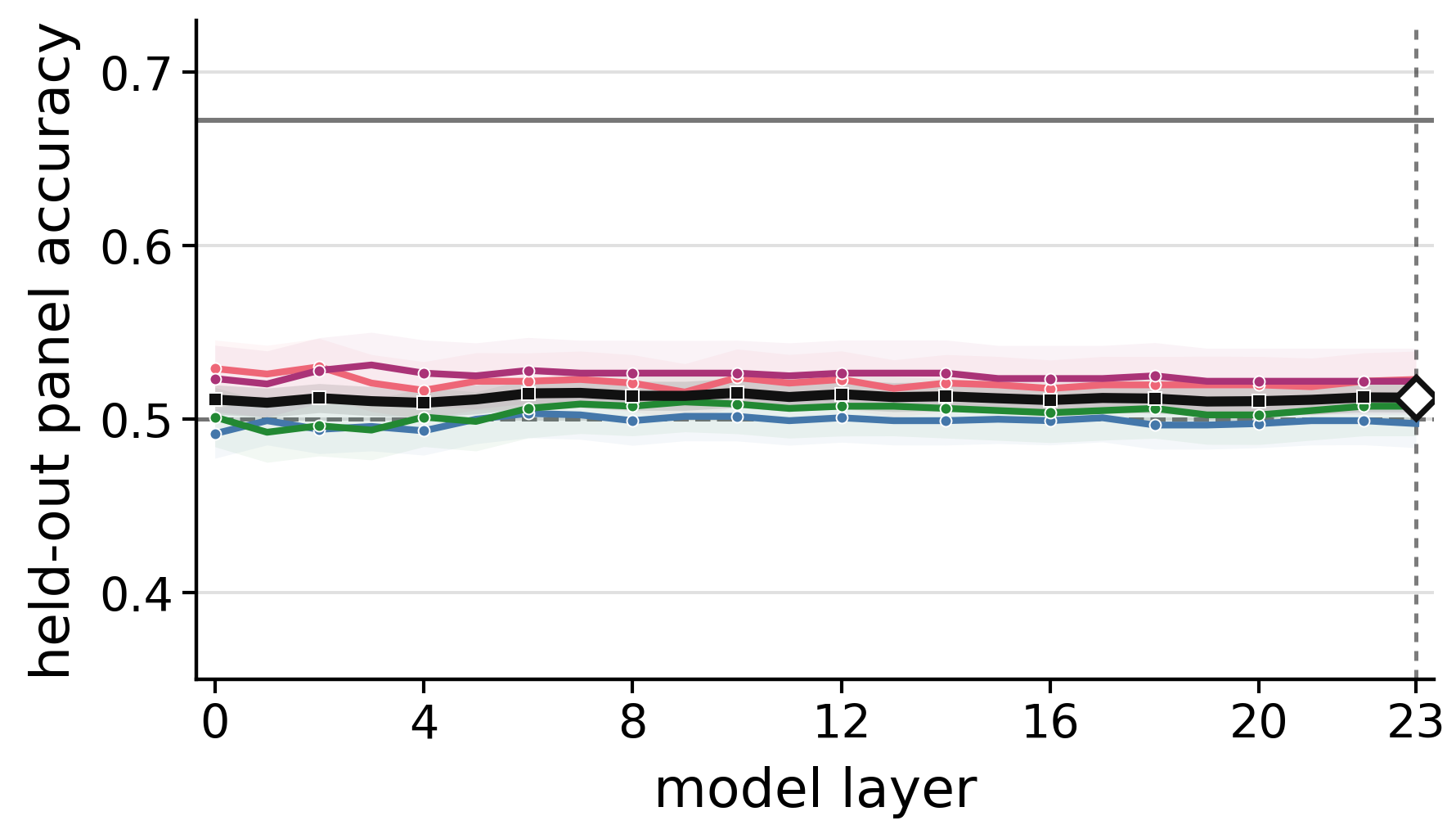}
        \caption{Untrained ViT, 480 px.}
        \label{fig:supp-bongard-layerwise-logo-vit480}
    \end{subfigure}

    \caption{
        Bongard--LOGO nearest-centroid accuracy across all 24 transformer blocks, for
        DINOv3 \textbf{(a,b)}, a MAE encoder of matched size \textbf{(c,d)} and the
        untrained ViT of Figure~\ref{fig:bongard-task-vector-analysis} \textbf{(e,f)}, at
        224 px (left column) and 480 px (right column). Each panel shows one curve per
        disjoint test subset and a black curve for their equal-weight mean; the legend in
        \textbf{(a)} applies to every panel. Shading denotes the SEM across problems. The
        gray line marks the dataset-paper baseline of
        Figure~\ref{fig:bongard-task-vector-analysis} ($67.2\%$, the equal-weight mean of
        the per-subset maxima in \citep{nie2020bongardlogo}) and the dashed line marks
        $50\%$ chance. All panels share the same axes. The open diamond and dotted line
        mark the layer selected on training problems (600 problems from each of LOGO's
        three training types); the bars in Figure~\ref{fig:bongard-task-vector-analysis}(d)
        report test accuracy at that layer, so a curve can peak above the reported value.
        The untrained ViT's residual branches are initialized at $10^{-5}$ (LayerScale), so
        its CLS token changes little with depth and its profile is flat.
    }
    \label{fig:supp-bongard-layerwise}
\end{figure*}

\begin{figure*}[p]
    \centering
    \captionsetup[subfigure]{skip=2pt}

    \begin{subfigure}[t]{0.49\textwidth}
        \centering
        \includegraphics[width=\linewidth]{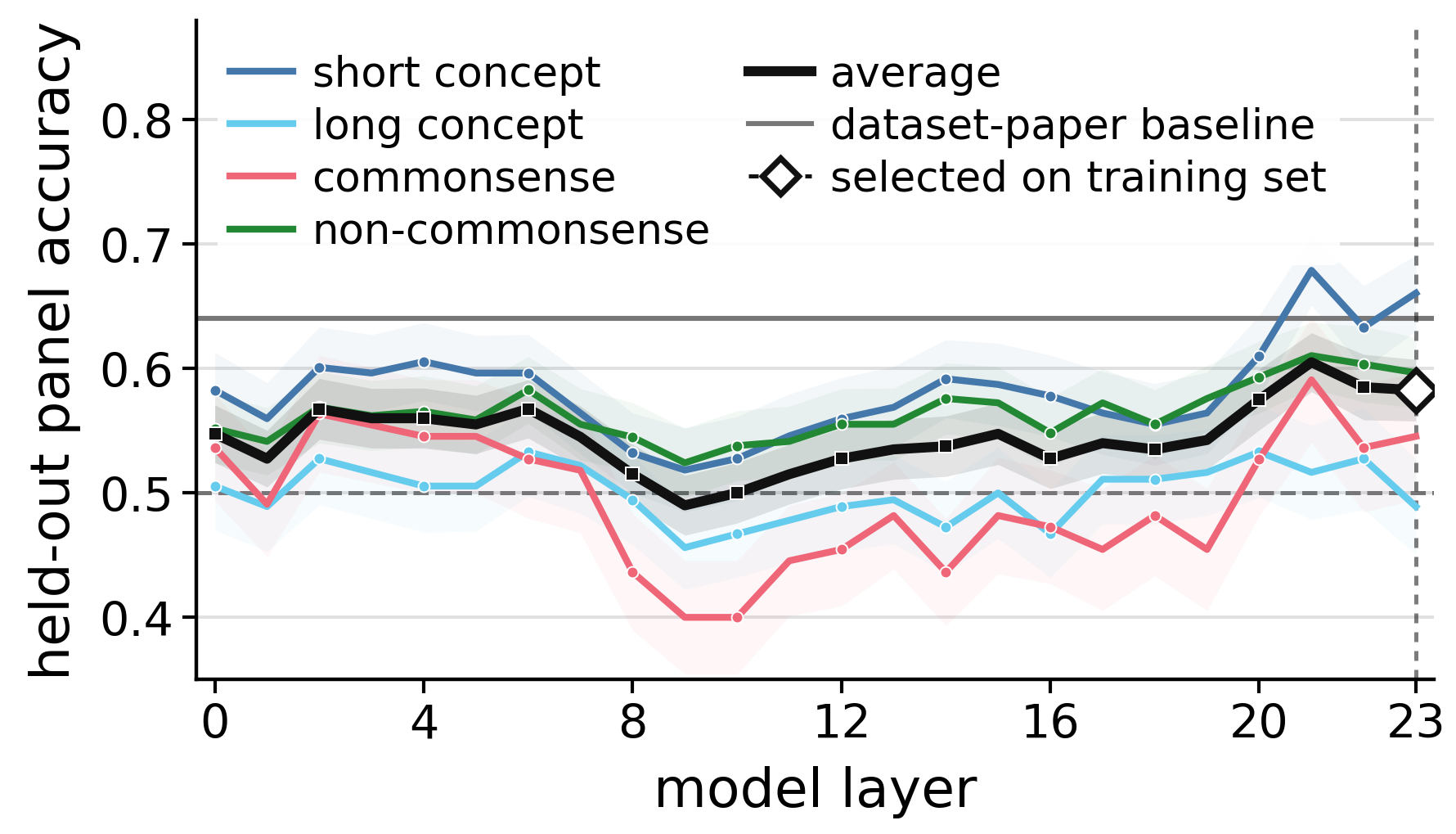}
        \caption{DINOv3, 224 px.}
        \label{fig:supp-bongard-layerwise-ow-dino224}
    \end{subfigure}
    \hfill
    \begin{subfigure}[t]{0.49\textwidth}
        \centering
        \includegraphics[width=\linewidth]{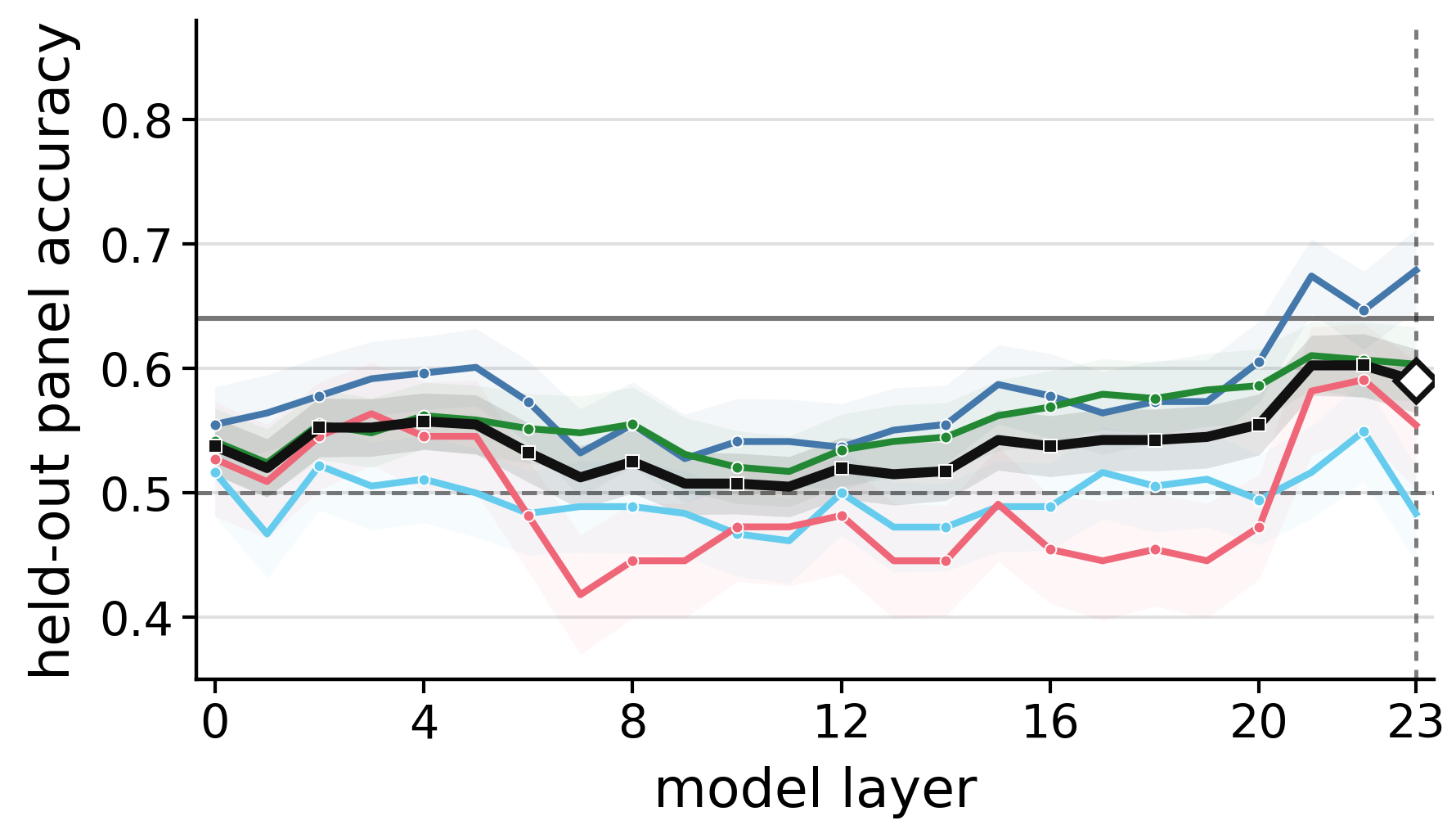}
        \caption{DINOv3, 480 px.}
        \label{fig:supp-bongard-layerwise-ow-dino480}
    \end{subfigure}

    \vspace{0.35em}

    \begin{subfigure}[t]{0.49\textwidth}
        \centering
        \includegraphics[width=\linewidth]{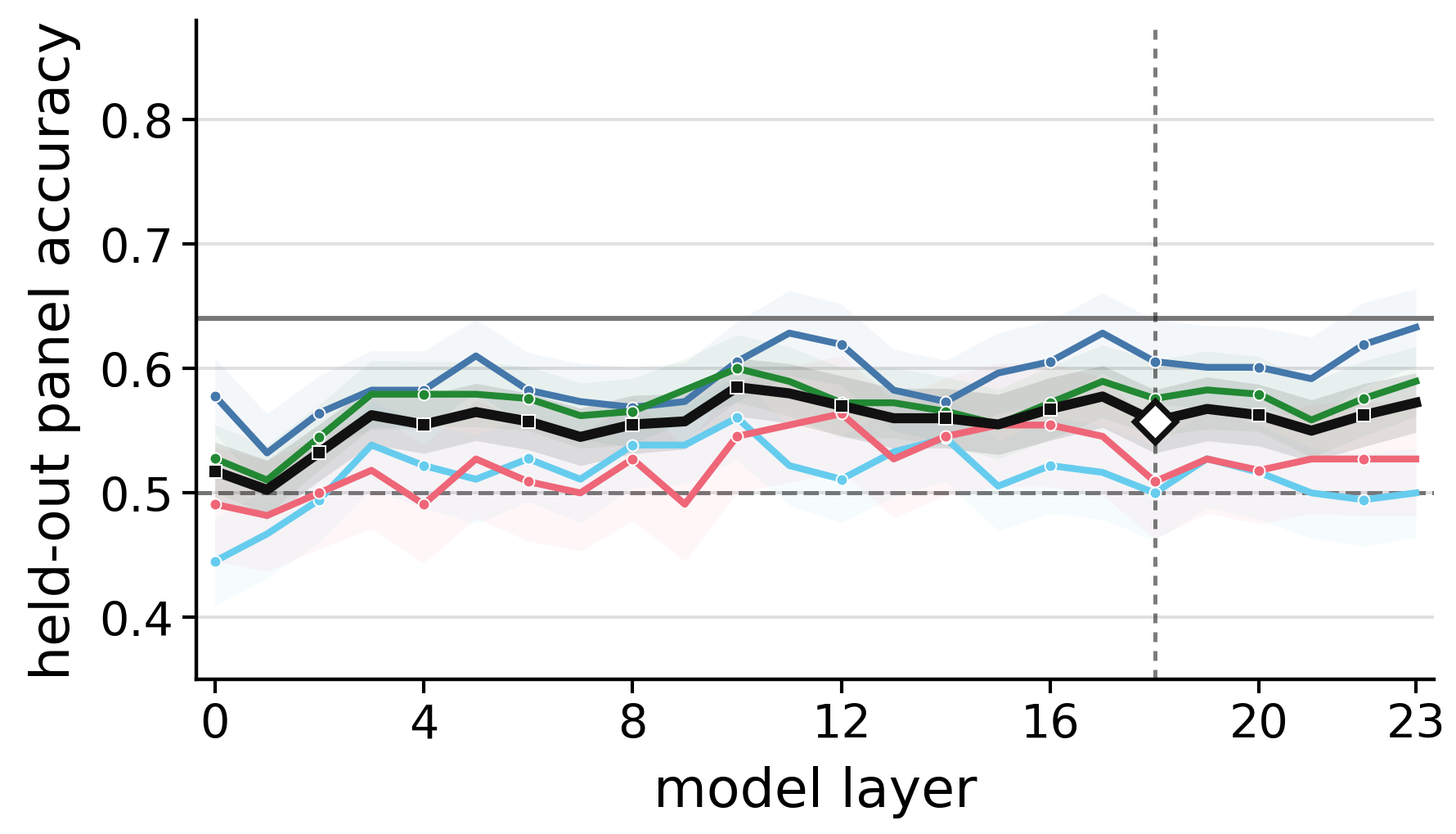}
        \caption{MAE, 224 px.}
        \label{fig:supp-bongard-layerwise-ow-mae224}
    \end{subfigure}
    \hfill
    \begin{subfigure}[t]{0.49\textwidth}
        \centering
        \includegraphics[width=\linewidth]{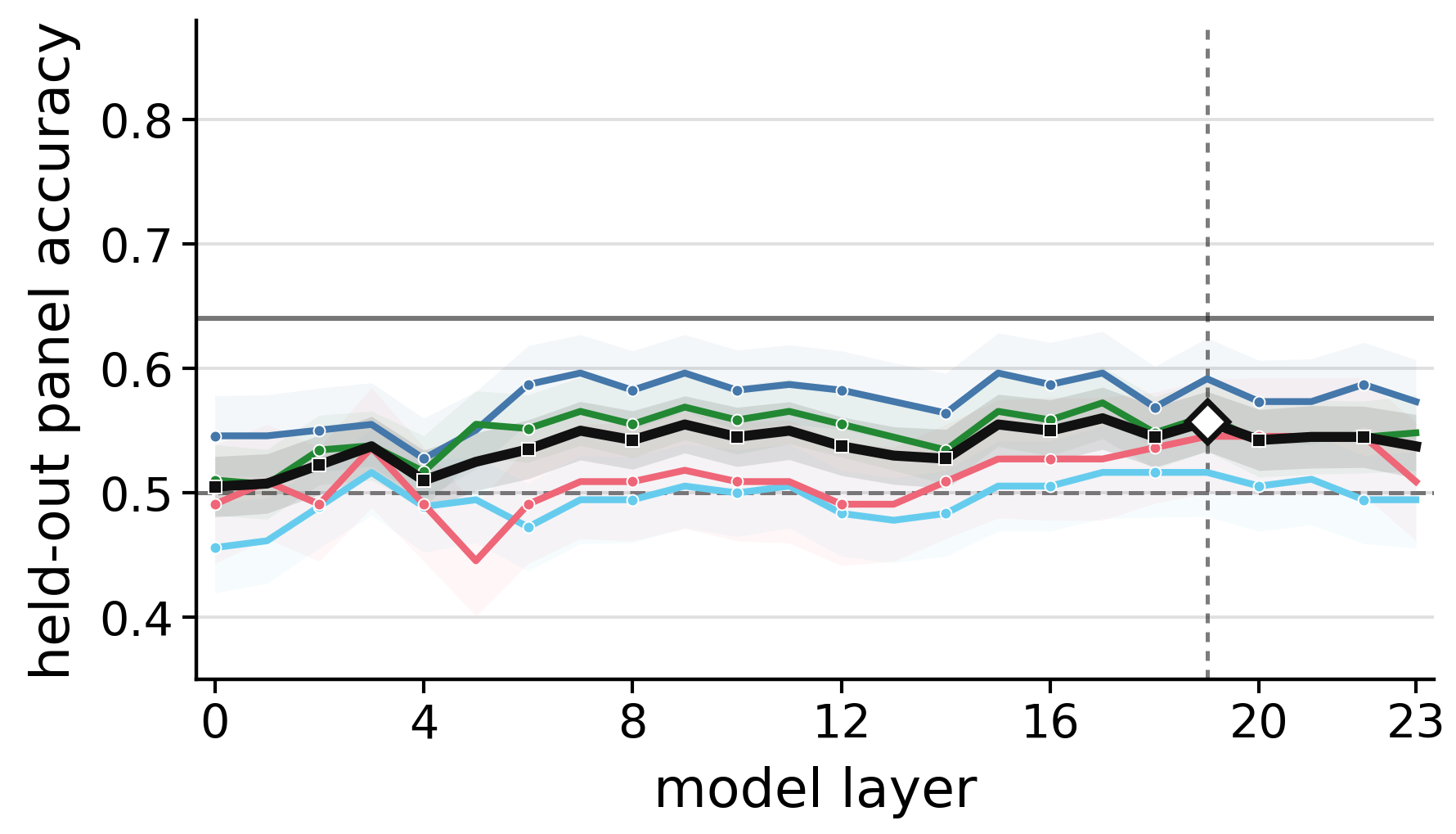}
        \caption{MAE, 480 px.}
        \label{fig:supp-bongard-layerwise-ow-mae480}
    \end{subfigure}

    \vspace{0.35em}

    \begin{subfigure}[t]{0.49\textwidth}
        \centering
        \includegraphics[width=\linewidth]{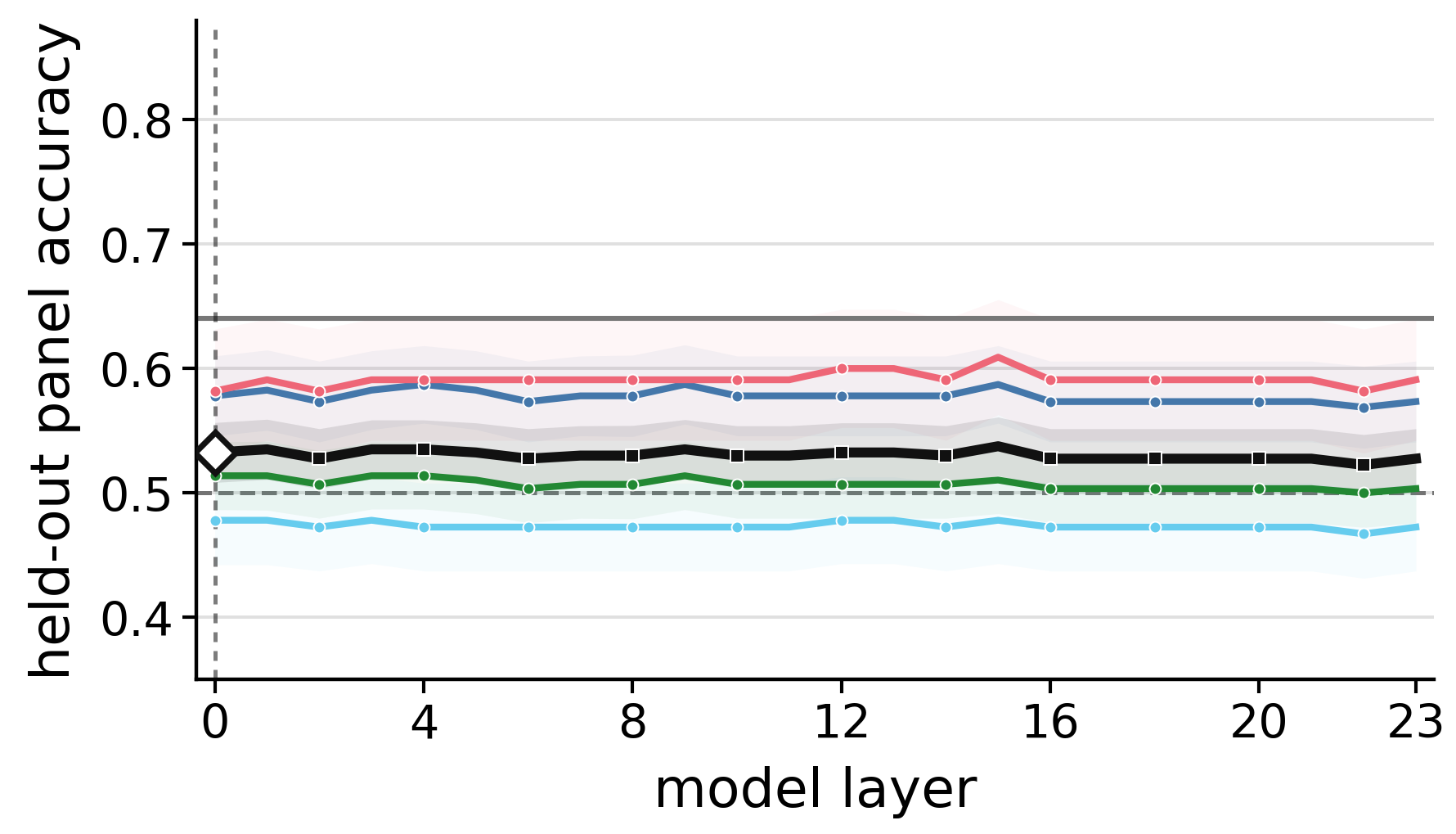}
        \caption{Untrained ViT, 224 px.}
        \label{fig:supp-bongard-layerwise-ow-vit224}
    \end{subfigure}
    \hfill
    \begin{subfigure}[t]{0.49\textwidth}
        \centering
        \includegraphics[width=\linewidth]{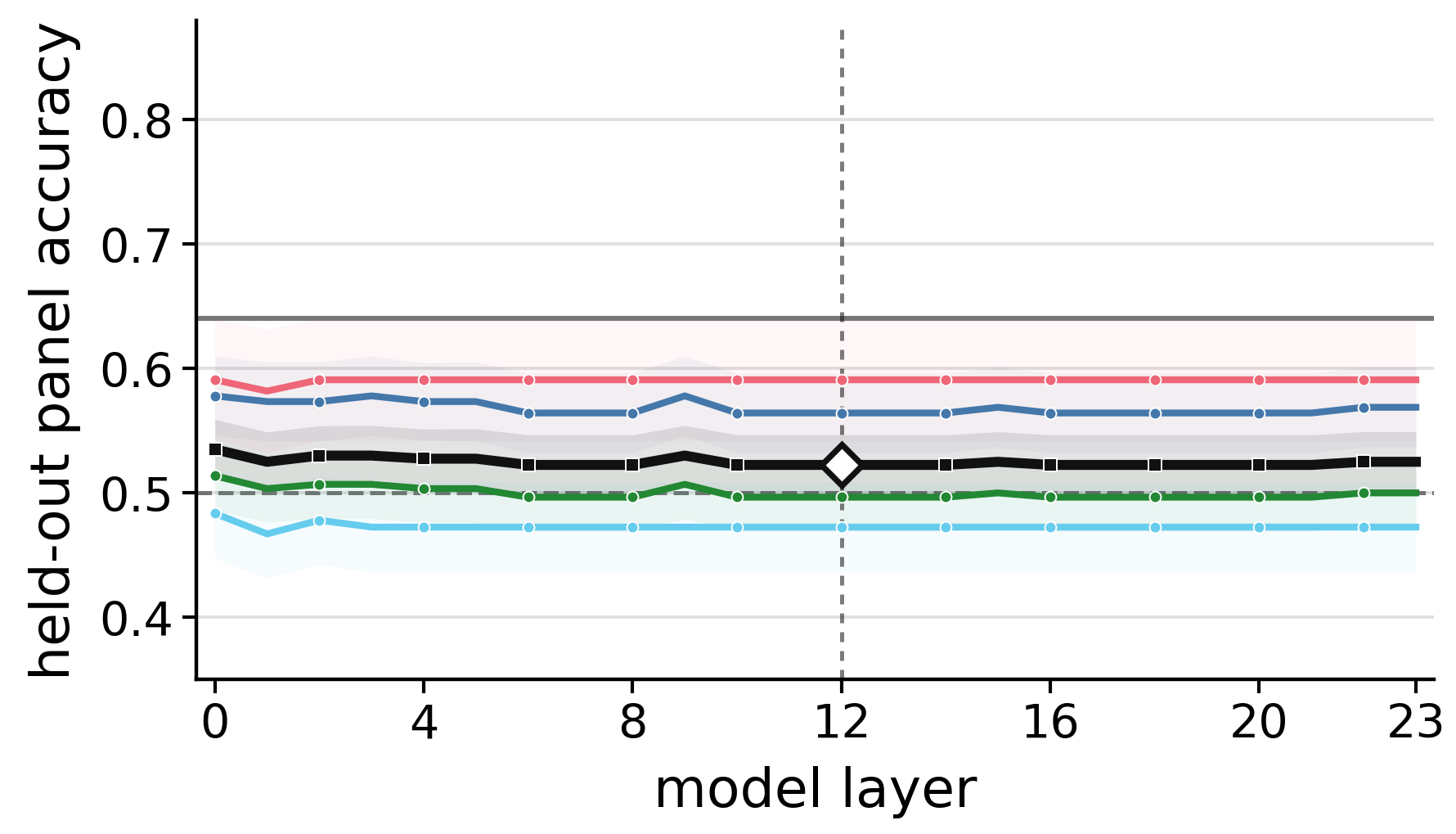}
        \caption{Untrained ViT, 480 px.}
        \label{fig:supp-bongard-layerwise-ow-vit480}
    \end{subfigure}

    \caption{
        Bongard-OpenWorld nearest-centroid accuracy across all 24 transformer blocks,
        arranged as in Supp.~Fig.~\ref{fig:supp-bongard-layerwise}. Each panel shows one
        curve per subgroup and a black curve for all 200 problems; the legend in
        \textbf{(a)} applies to every panel. The subgroups form two overlapping partitions
        of the same problems -- concept length and commonsense requirement -- so their
        curves are not independent strata. The gray line marks the dataset-paper baseline
        ($64.0\%$ \citep{wu2024bongardopen}); shading, chance line and axes are as in
        Supp.~Fig.~\ref{fig:supp-bongard-layerwise}. The open diamond and dotted line mark
        the layer selected on OpenWorld's 610 training problems; the bars in
        Figure~\ref{fig:bongard-task-vector-analysis}(e) report test accuracy at that layer.
    }
    \label{fig:supp-bongard-layerwise-ow}
\end{figure*}

\begin{figure*}[p]
    \centering
    \captionsetup[subfigure]{skip=2pt}

    \begin{subfigure}[t]{0.49\textwidth}
        \centering
        \includegraphics[width=\linewidth]{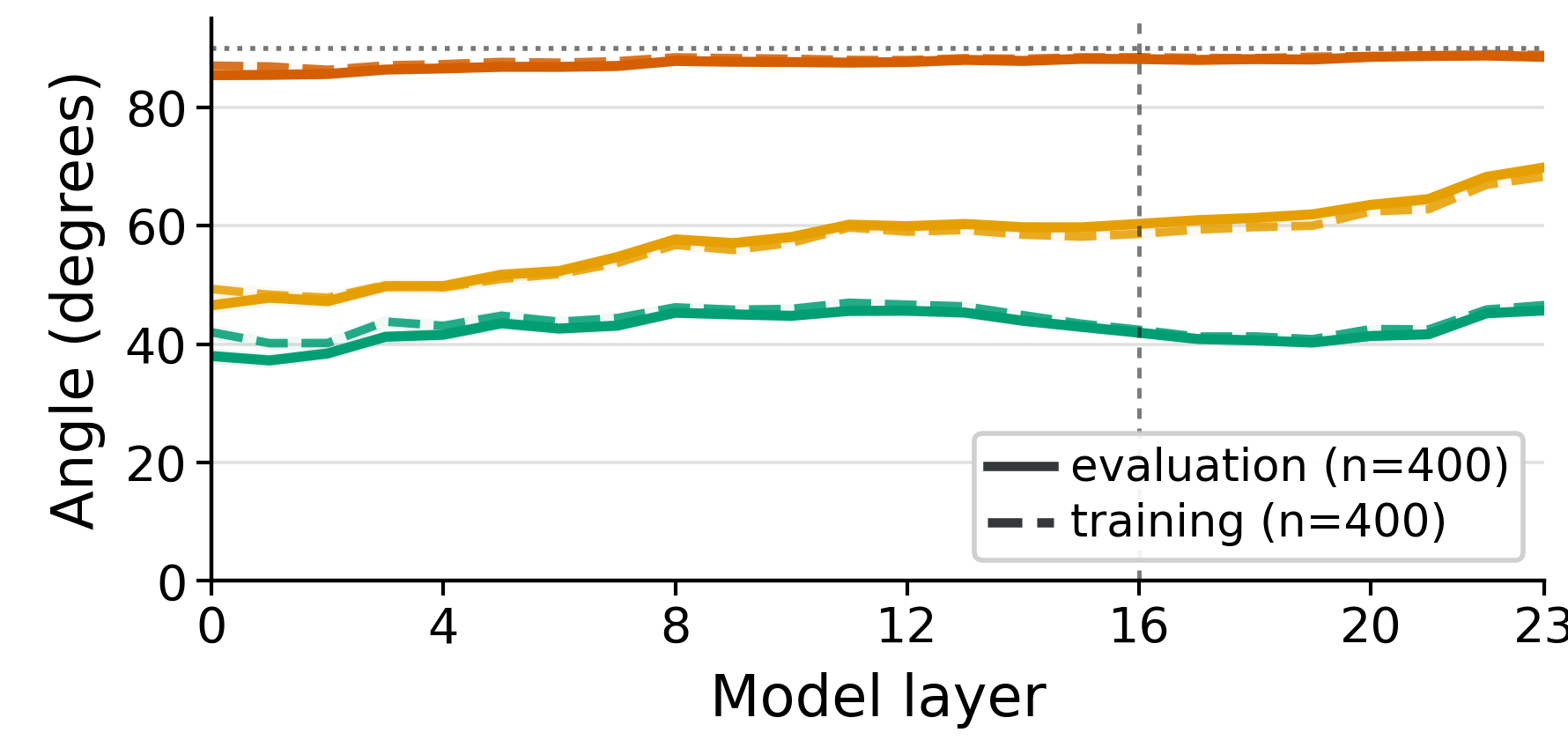}
        \caption{DINOv3, 224 px.}
        \label{fig:supp-arc-angle-arc1-dino224}
    \end{subfigure}
    \hfill
    \begin{subfigure}[t]{0.49\textwidth}
        \centering
        \includegraphics[width=\linewidth]{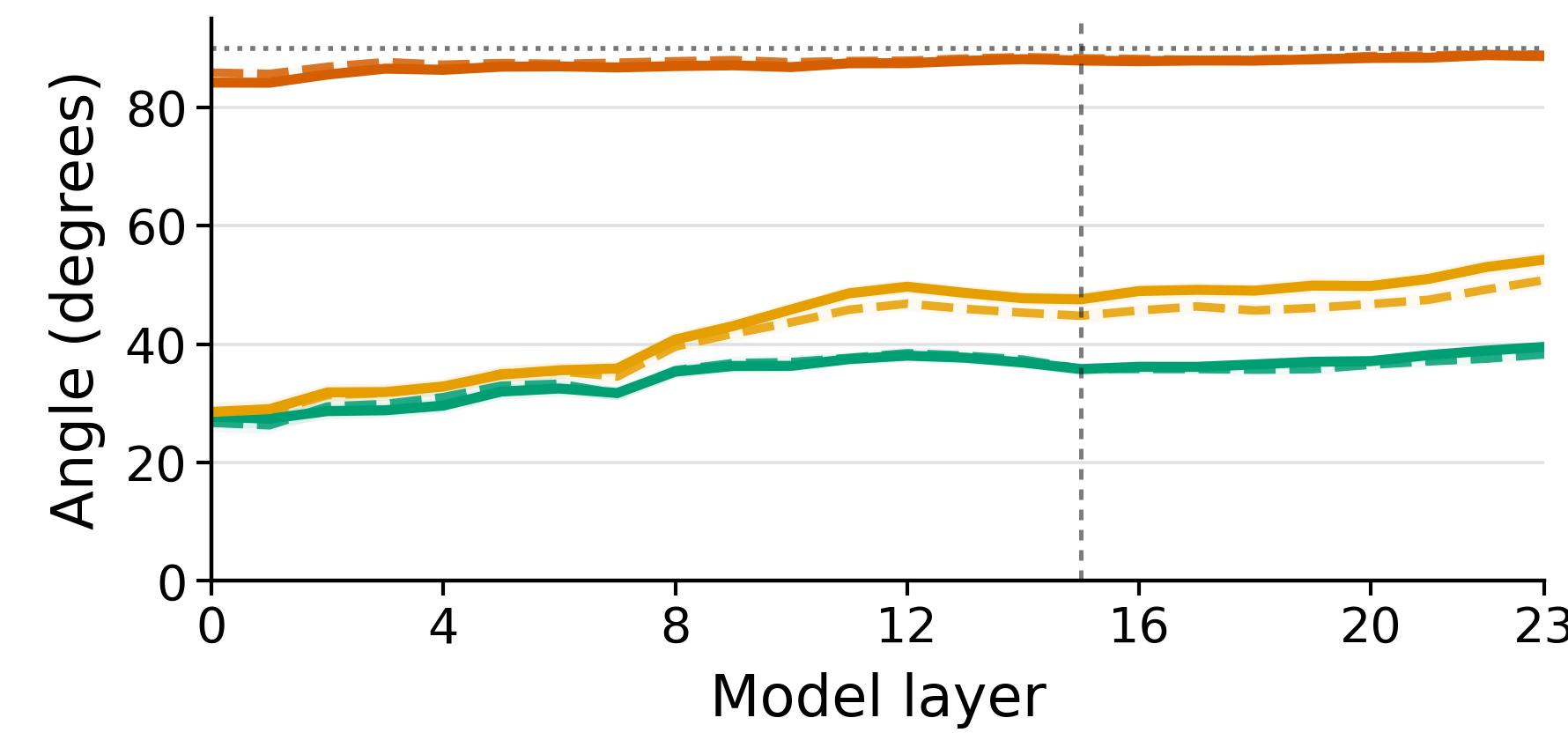}
        \caption{DINOv3, 480 px.}
        \label{fig:supp-arc-angle-arc1-dino480}
    \end{subfigure}

    \vspace{0.3em}

    \begin{subfigure}[t]{0.49\textwidth}
        \centering
        \includegraphics[width=\linewidth]{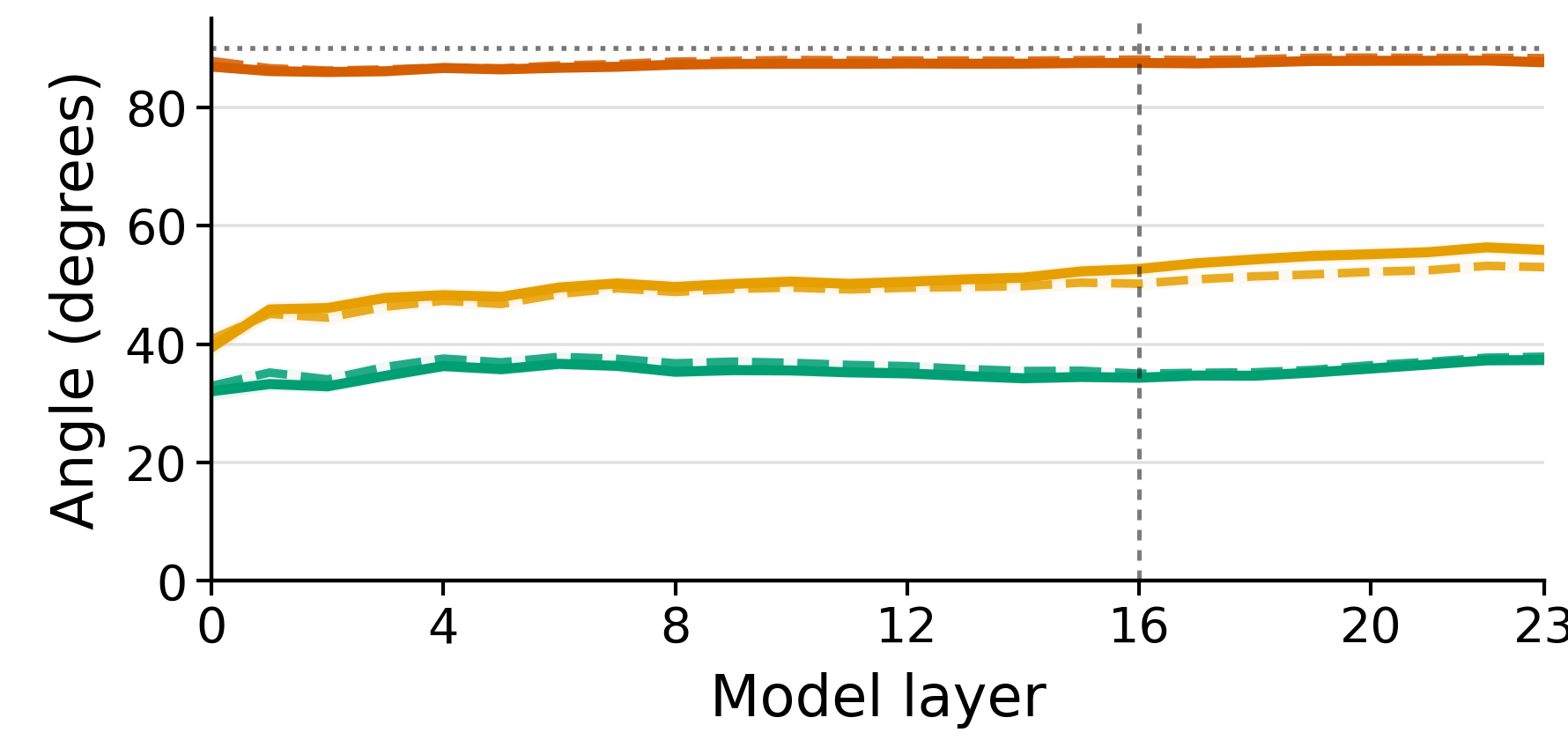}
        \caption{MAE, 224 px.}
        \label{fig:supp-arc-angle-arc1-mae224}
    \end{subfigure}
    \hfill
    \begin{subfigure}[t]{0.49\textwidth}
        \centering
        \includegraphics[width=\linewidth]{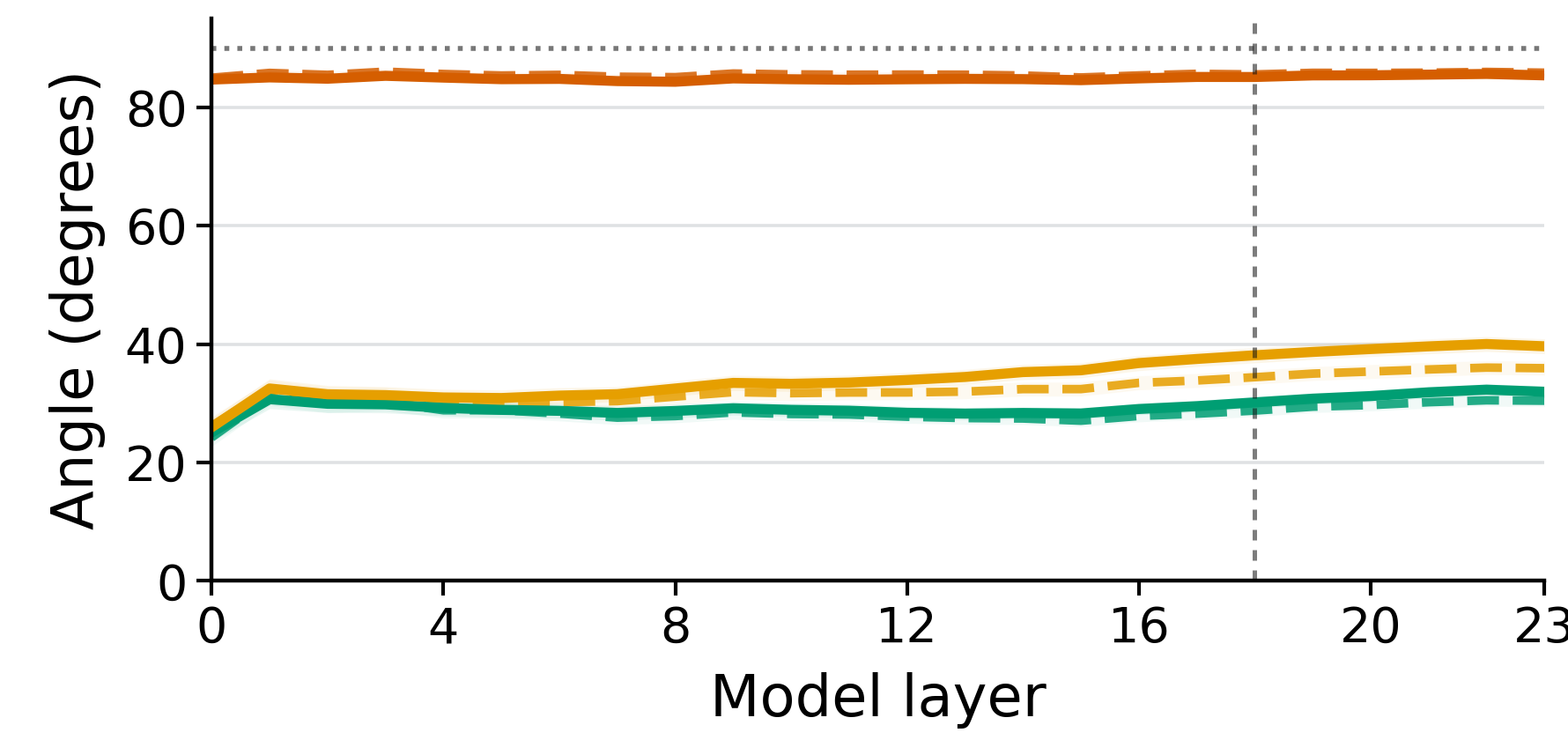}
        \caption{MAE, 480 px.}
        \label{fig:supp-arc-angle-arc1-mae480}
    \end{subfigure}

    \vspace{0.3em}

    \begin{subfigure}[t]{0.49\textwidth}
        \centering
        \includegraphics[width=\linewidth]{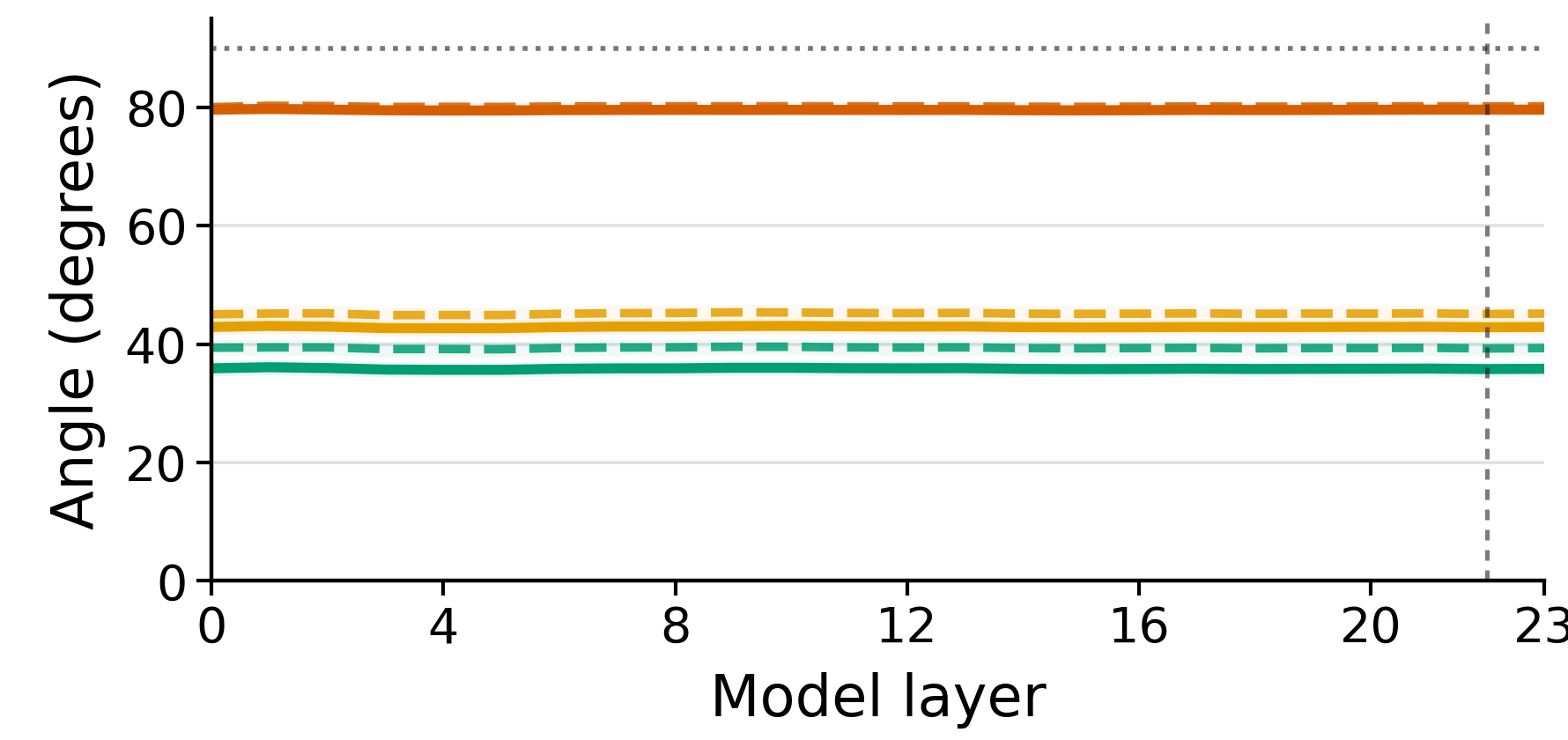}
        \caption{Untrained ViT, 224 px.}
        \label{fig:supp-arc-angle-arc1-vit224}
    \end{subfigure}
    \hfill
    \begin{subfigure}[t]{0.49\textwidth}
        \centering
        \includegraphics[width=\linewidth]{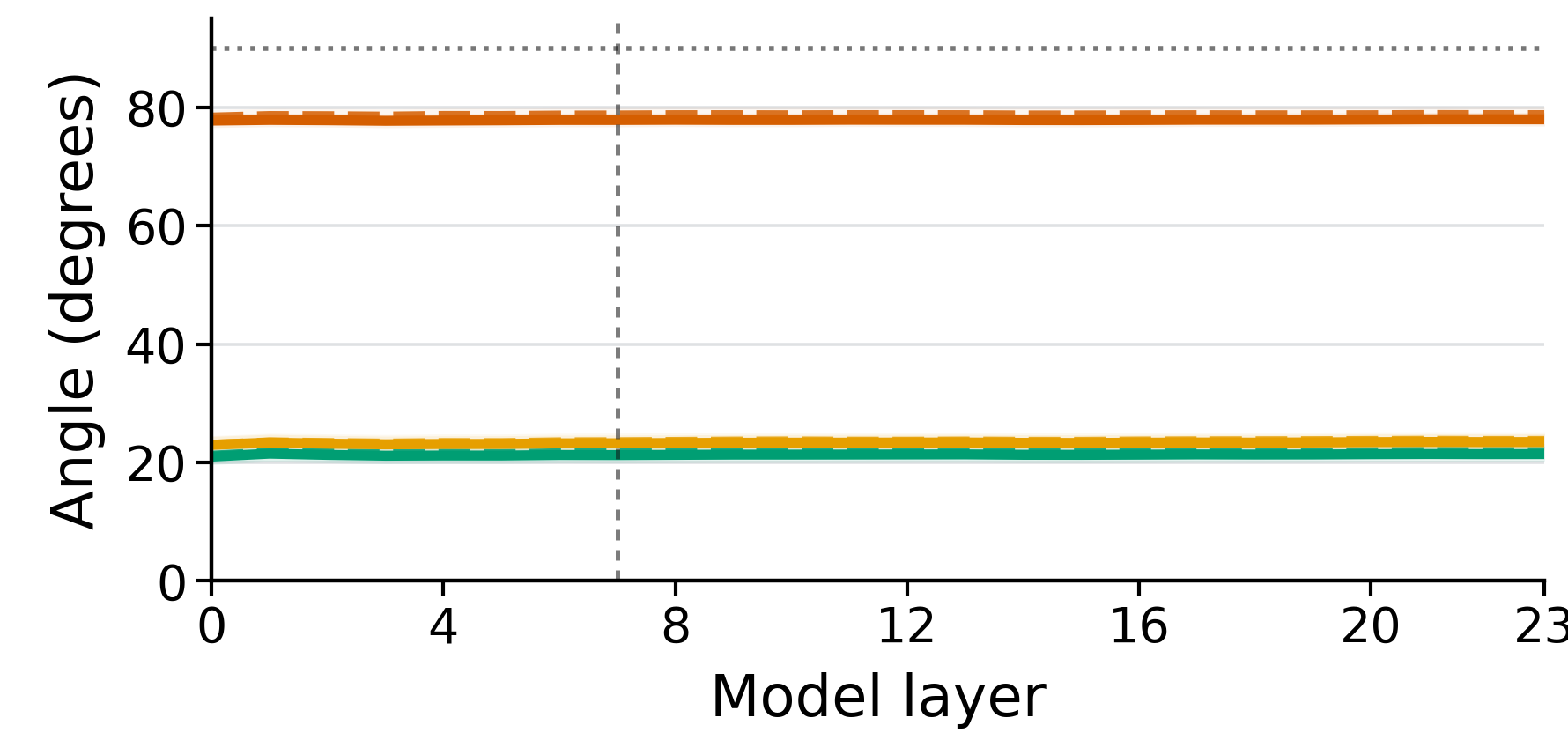}
        \caption{Untrained ViT, 480 px.}
        \label{fig:supp-arc-angle-arc1-vit480}
    \end{subfigure}

    \caption{
        Intuition--test-vector angles across all 24 transformer blocks on ARC-1, for
        DINOv3 \textbf{(a,b)}, MAE \textbf{(c,d)} and the untrained ViT of
        Figure~\ref{fig:bongard-task-vector-analysis} \textbf{(e,f)}, at 224 px (left column)
        and 480 px (right column). Green curves show matched tasks, red curves cross-task
        controls, and orange curves cell-shuffled answers. Solid and dashed curves denote
        the evaluation and training sets, respectively; shading shows the SEM across tasks
        and the dotted line marks $90^\circ$. Both sets contain $n=400$ tasks. Cross-task
        and shuffled-output controls average ten randomizations.
        Figure~\ref{fig:arc-task-vector-analysis}(c) and
        Supp.~Fig.~\ref{fig:supp-arc-angle-480} report the corresponding values at the
        selected layer (dashed vertical line; Sec.~\ref{sec:methods-layers}), together with
        the untrained patchifier. The untrained
        ViT's residual branches are initialized at $10^{-5}$ (LayerScale), so its CLS token
        changes little with depth and its profile is flat.
    }
    \label{fig:supp-arc-angle-arc1}
\end{figure*}

\begin{figure*}[p]
    \centering
    \captionsetup[subfigure]{skip=2pt}

    \begin{subfigure}[t]{0.49\textwidth}
        \centering
        \includegraphics[width=\linewidth]{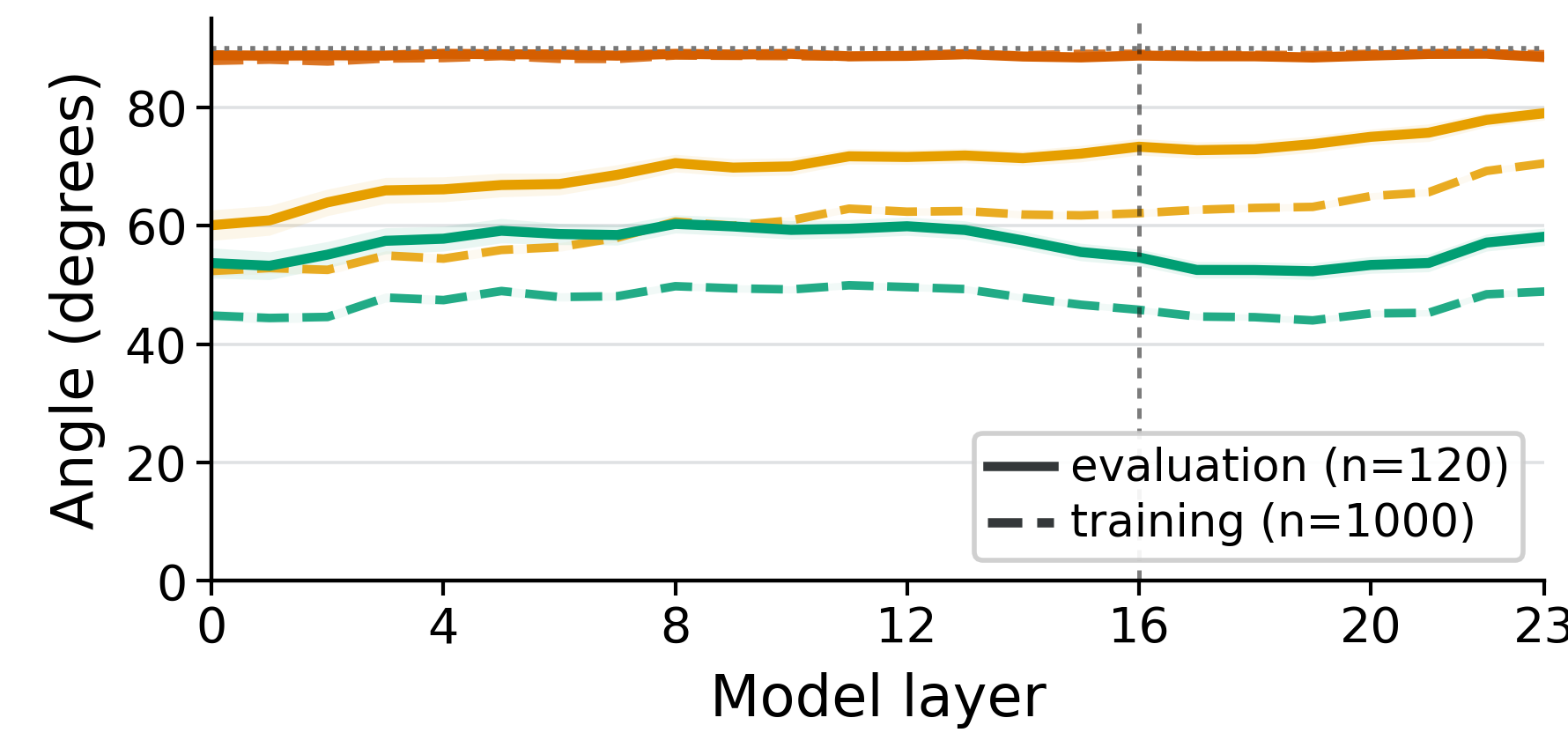}
        \caption{DINOv3, 224 px.}
        \label{fig:supp-arc-angle-arc2-dino224}
    \end{subfigure}
    \hfill
    \begin{subfigure}[t]{0.49\textwidth}
        \centering
        \includegraphics[width=\linewidth]{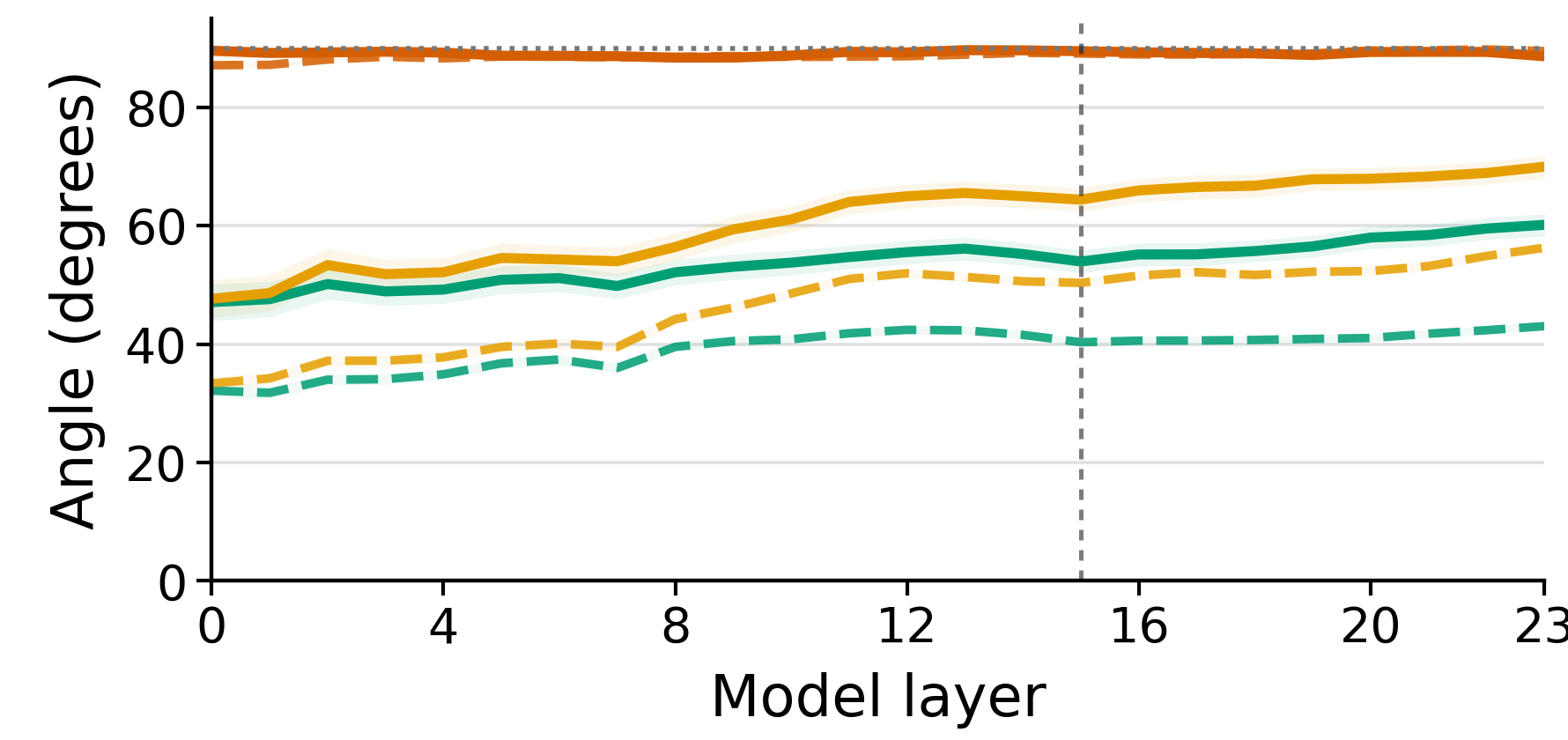}
        \caption{DINOv3, 480 px.}
        \label{fig:supp-arc-angle-arc2-dino480}
    \end{subfigure}

    \vspace{0.3em}

    \begin{subfigure}[t]{0.49\textwidth}
        \centering
        \includegraphics[width=\linewidth]{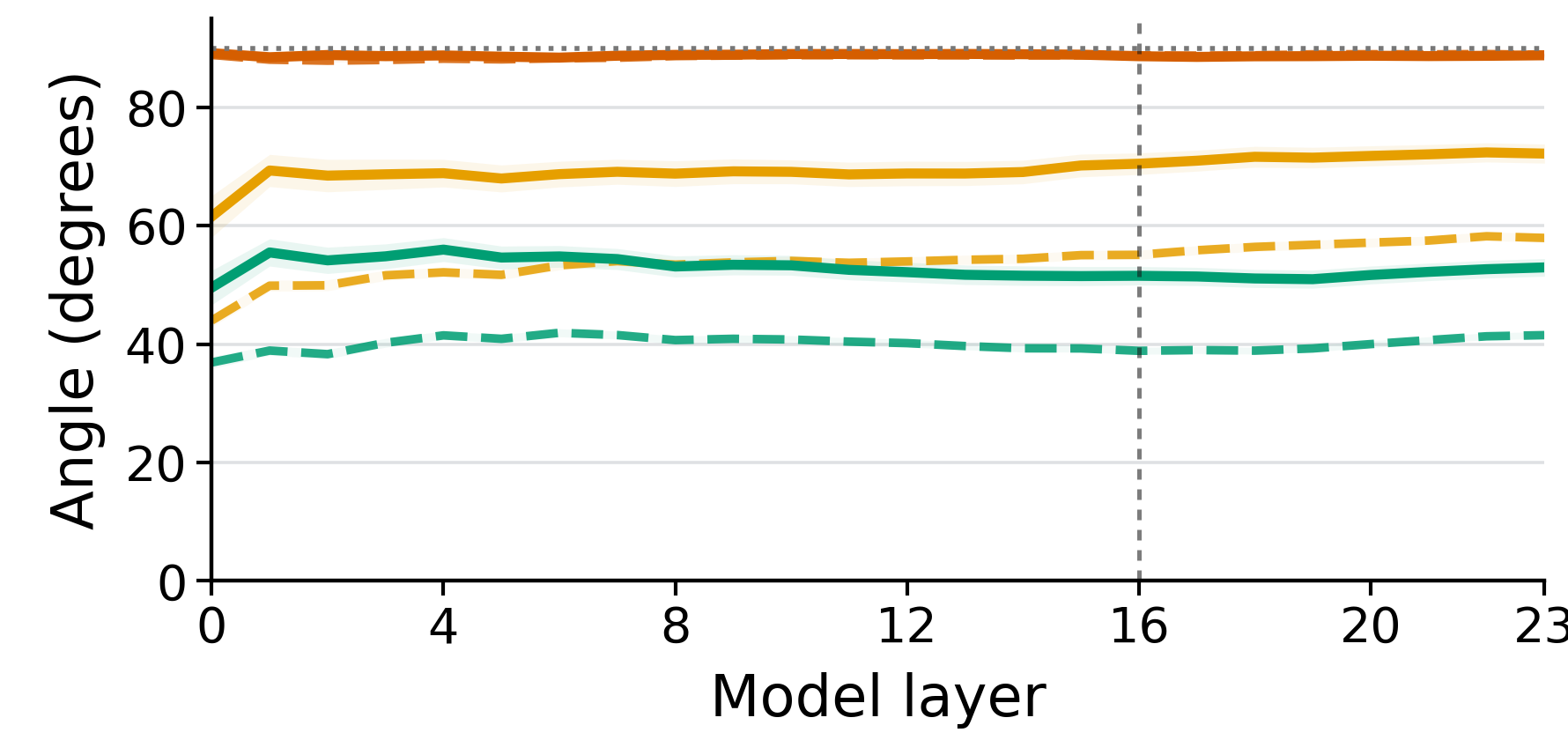}
        \caption{MAE, 224 px.}
        \label{fig:supp-arc-angle-arc2-mae224}
    \end{subfigure}
    \hfill
    \begin{subfigure}[t]{0.49\textwidth}
        \centering
        \includegraphics[width=\linewidth]{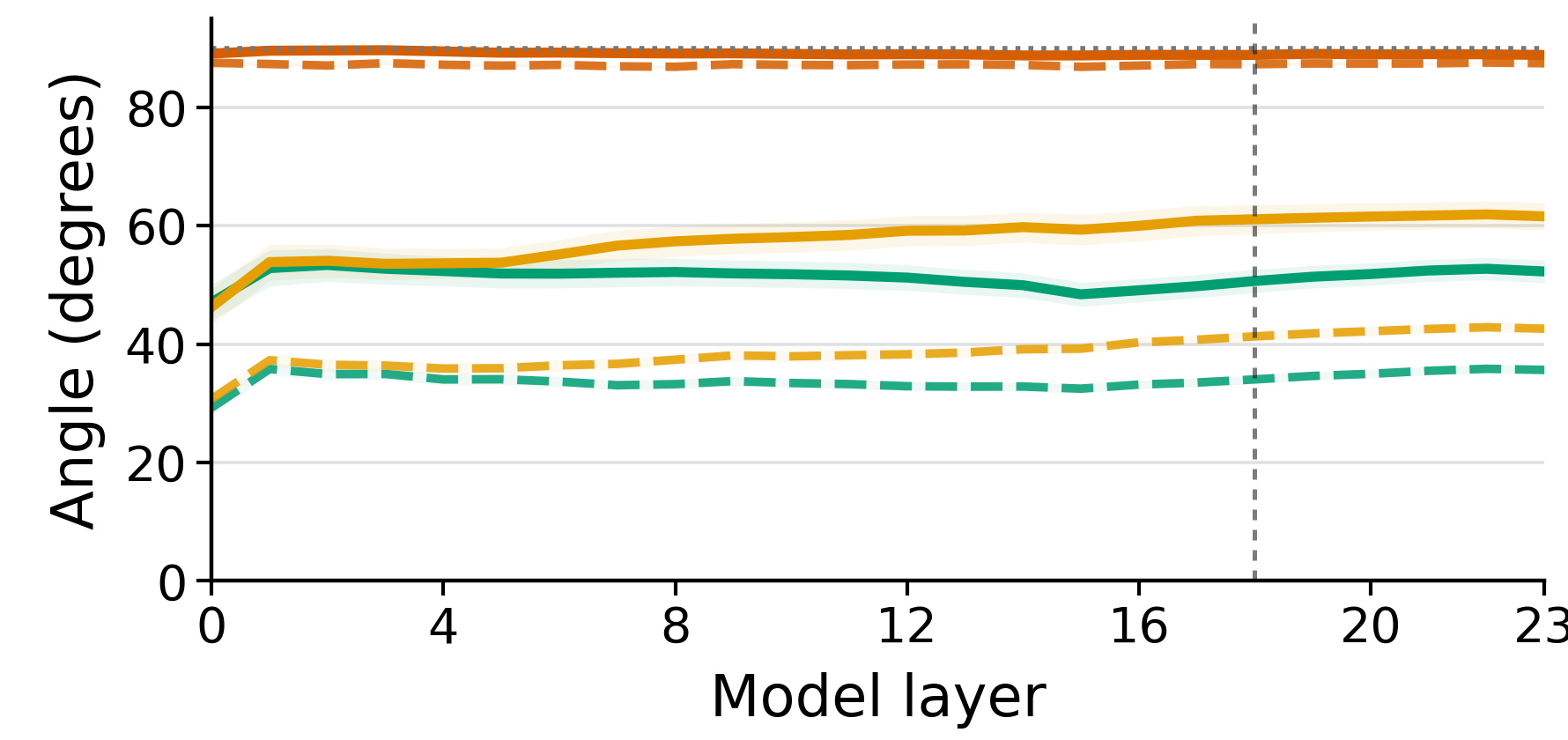}
        \caption{MAE, 480 px.}
        \label{fig:supp-arc-angle-arc2-mae480}
    \end{subfigure}

    \vspace{0.3em}

    \begin{subfigure}[t]{0.49\textwidth}
        \centering
        \includegraphics[width=\linewidth]{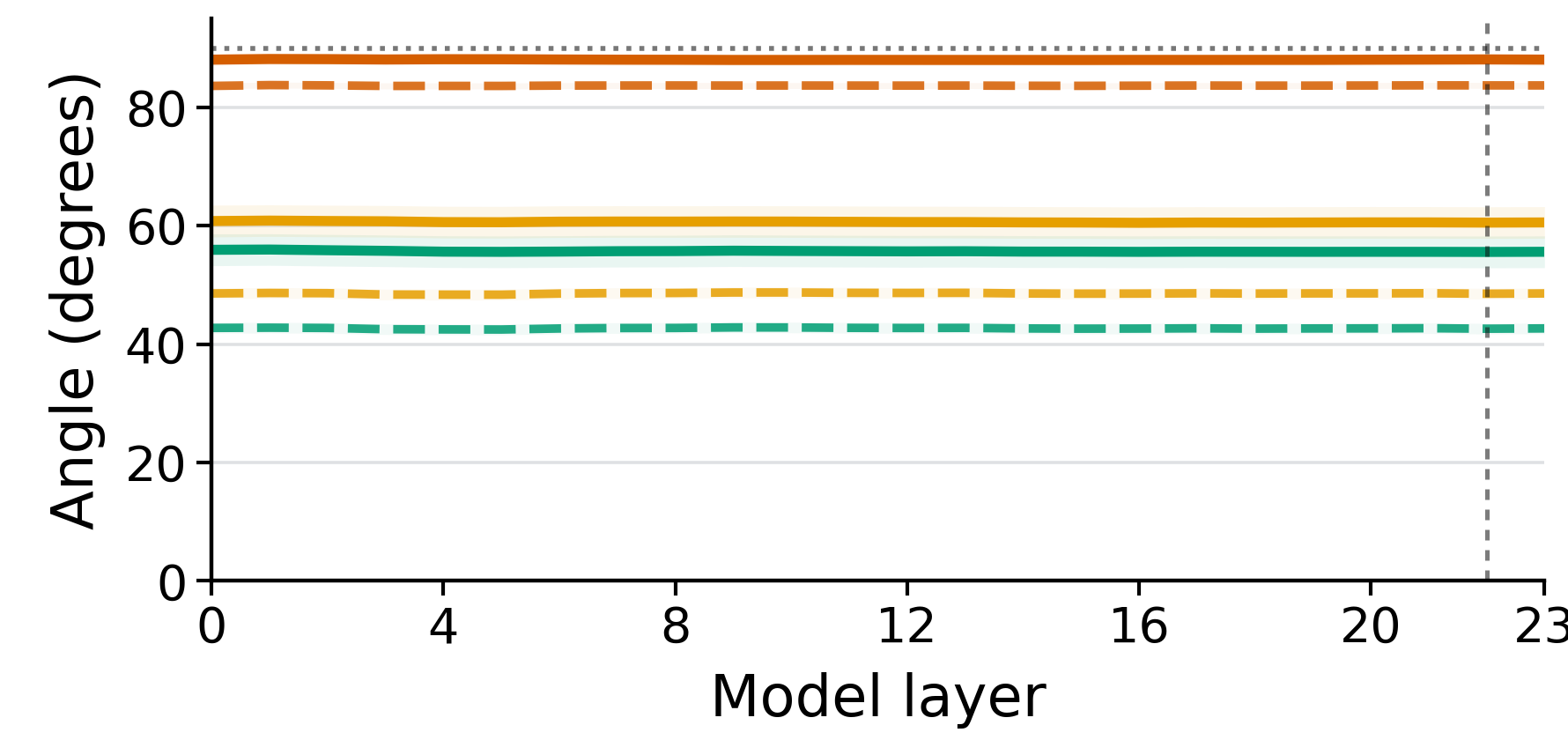}
        \caption{Untrained ViT, 224 px.}
        \label{fig:supp-arc-angle-arc2-vit224}
    \end{subfigure}
    \hfill
    \begin{subfigure}[t]{0.49\textwidth}
        \centering
        \includegraphics[width=\linewidth]{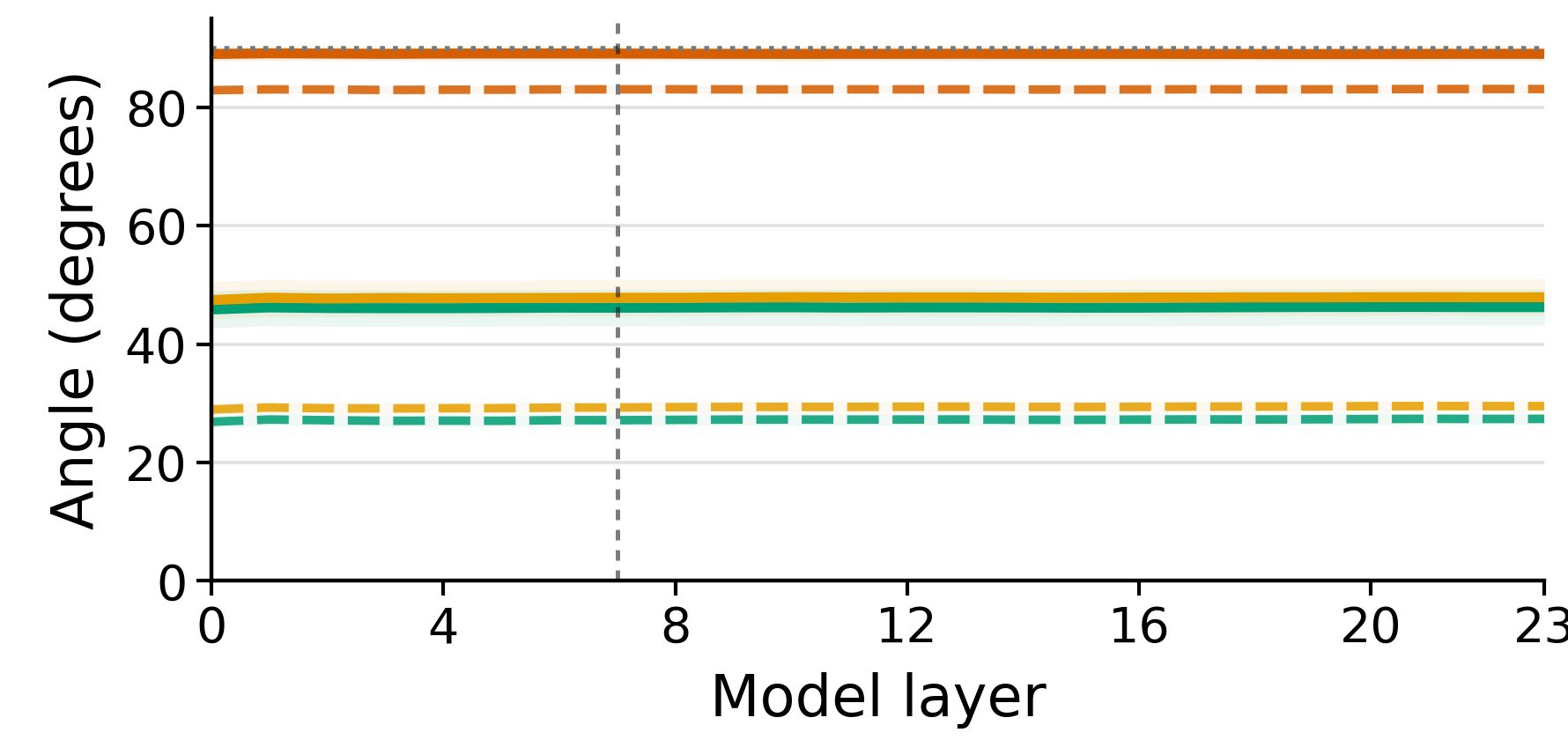}
        \caption{Untrained ViT, 480 px.}
        \label{fig:supp-arc-angle-arc2-vit480}
    \end{subfigure}

    \caption{
        Intuition--test-vector angles across all 24 transformer blocks on ARC-2,
        arranged as in Figure~\ref{fig:supp-arc-angle-arc1} and with the same conventions.
        The training set contains $n=1{,}000$ tasks and the evaluation set
        $n=120$, producing wider evaluation uncertainty. Figure~\ref{fig:arc-task-vector-analysis}(d)
        and Supp.~Fig.~\ref{fig:supp-arc-angle-480} report the corresponding values at the
        selected layer (dashed vertical line).
    }
    \label{fig:supp-arc-angle-arc2}
\end{figure*}

\begin{figure*}[tbp]
    \centering
    \captionsetup[subfigure]{skip=2pt}

    \includegraphics[width=0.8\textwidth]{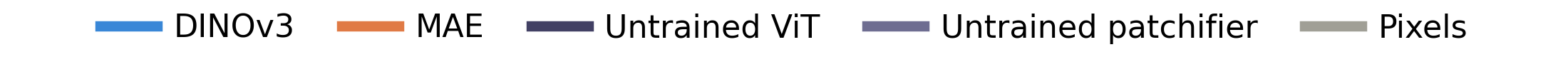}

    \vspace{0.2em}

    \begin{subfigure}[t]{0.49\textwidth}
        \centering
        \includegraphics[width=\linewidth]{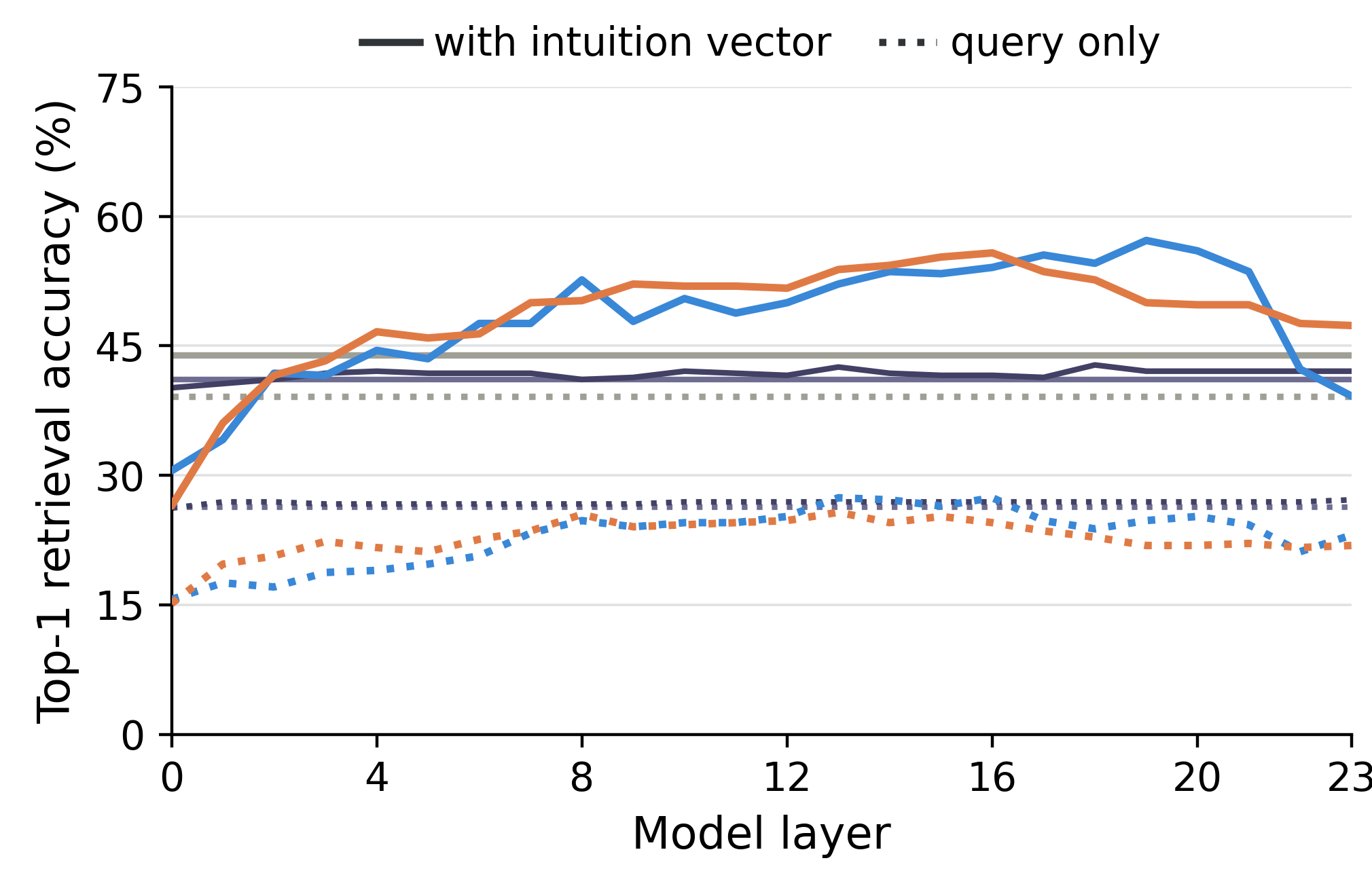}
        \caption{ARC-1, training set, 224 px.}
        \label{fig:supp-arc1-retrieval-train}
    \end{subfigure}
    \hfill
    \begin{subfigure}[t]{0.49\textwidth}
        \centering
        \includegraphics[width=\linewidth]{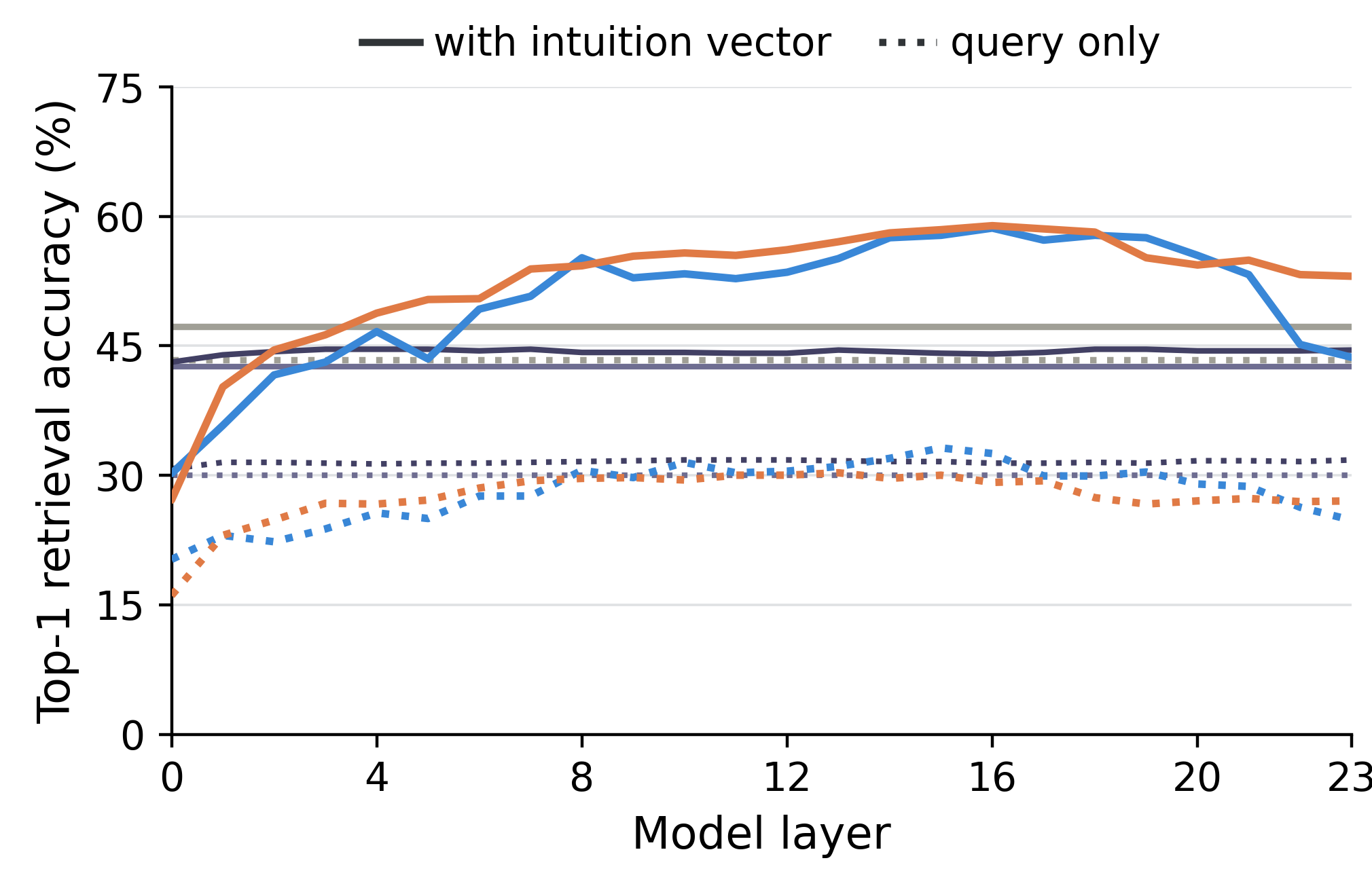}
        \caption{ARC-2, training set, 224 px.}
        \label{fig:supp-arc2-retrieval-train}
    \end{subfigure}

    \vspace{0.4em}

    \begin{subfigure}[t]{0.49\textwidth}
        \centering
        \includegraphics[width=\linewidth]{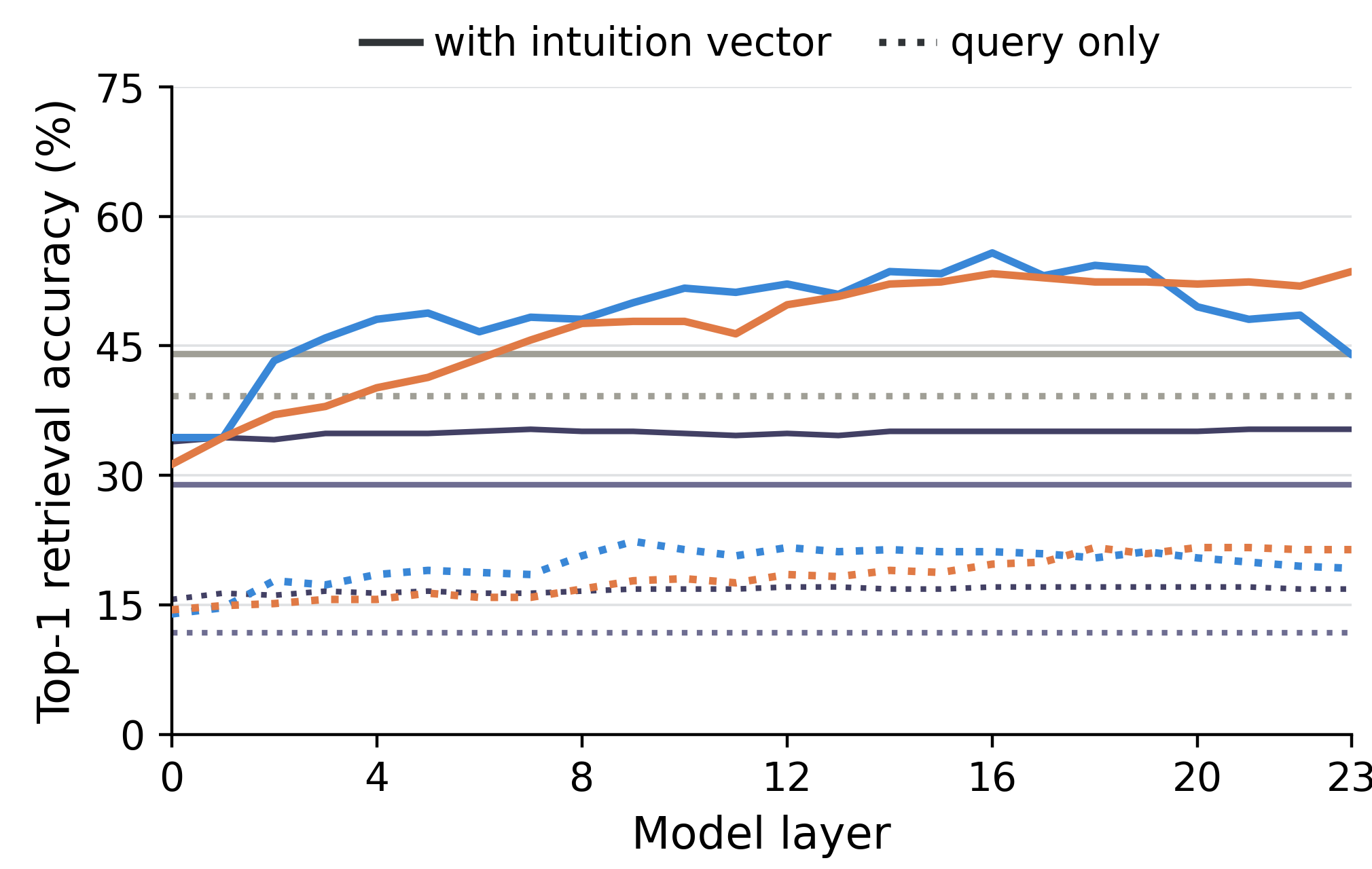}
        \caption{ARC-1, training set, 480 px.}
        \label{fig:supp-arc1-retrieval-train-480}
    \end{subfigure}
    \hfill
    \begin{subfigure}[t]{0.49\textwidth}
        \centering
        \includegraphics[width=\linewidth]{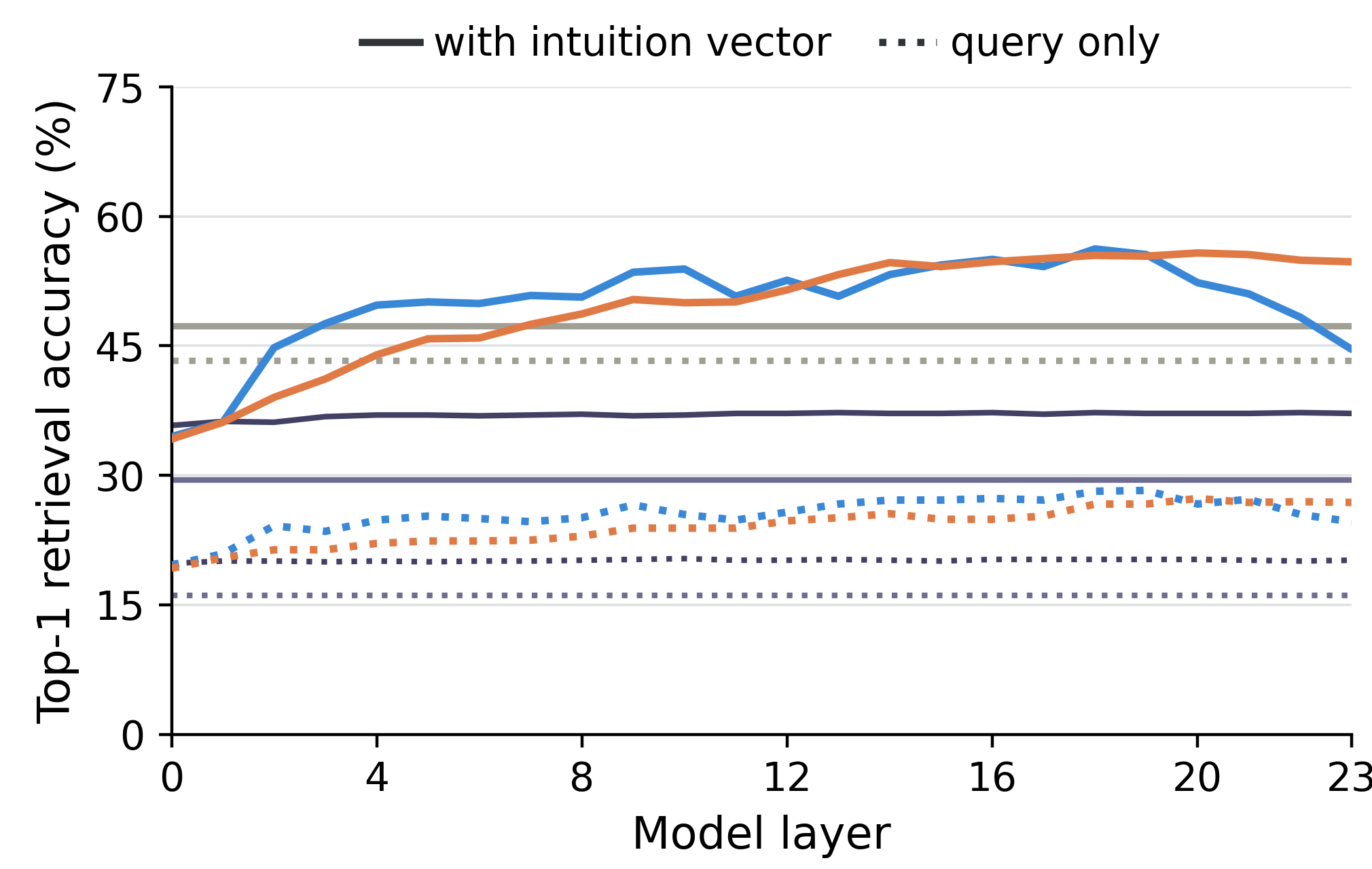}
        \caption{ARC-2, training set, 480 px.}
        \label{fig:supp-arc2-retrieval-train-480}
    \end{subfigure}

    \caption{
        Output retrieval on the ARC training sets, matching
        Figure~\ref{fig:arc-knn-new-analysis}(c,d) and Supp.~Fig.~\ref{fig:supp-arc-retrieval-480}(a,b)
        panel for panel. Top-1 accuracy across all 24 transformer blocks, comparing DINOv3,
        a MAE encoder of matched size, and the untrained ViT, with the layer-independent
        random-projection pixel controls and the untrained patchifier (mean of ten seeds)
        drawn as flat lines. Solid curves add the intuition vector to the held-out query and dotted
        curves rank against the unperturbed query. ARC-1 contains 400 tasks in each
        split; ARC-2 contains 1,000 training and 120 evaluation tasks. These
        training-split curves provide an additional descriptive comparison.
        Encoder layers were selected by mean transported retrieval accuracy on the
        ARC-1 and ARC-2 evaluation sets (Sec.~\ref{sec:methods-layers}).
    }
    \label{fig:supp-arc-retrieval-train}
\end{figure*}

\begin{figure*}[tbp]
    \centering
    \captionsetup[subfigure]{skip=2pt}

    \begin{subfigure}[t]{0.49\textwidth}
        \centering
        \includegraphics[width=\linewidth]{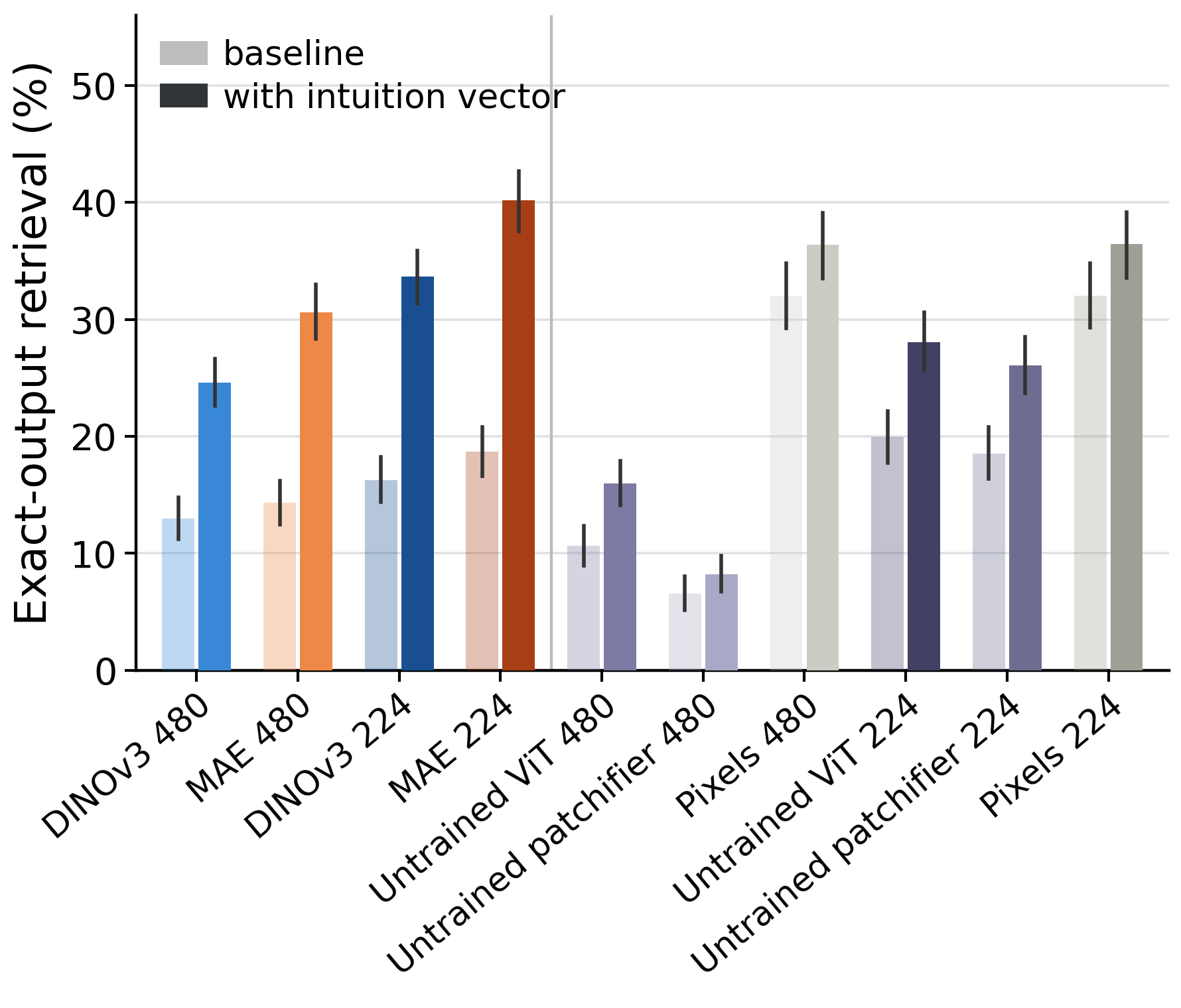}
        \caption{Retrieval at the selected layer.}
        \label{fig:arcgen-baseline-transport}
    \end{subfigure}
    \hfill
    \begin{subfigure}[t]{0.49\textwidth}
        \centering
        \includegraphics[width=\linewidth]{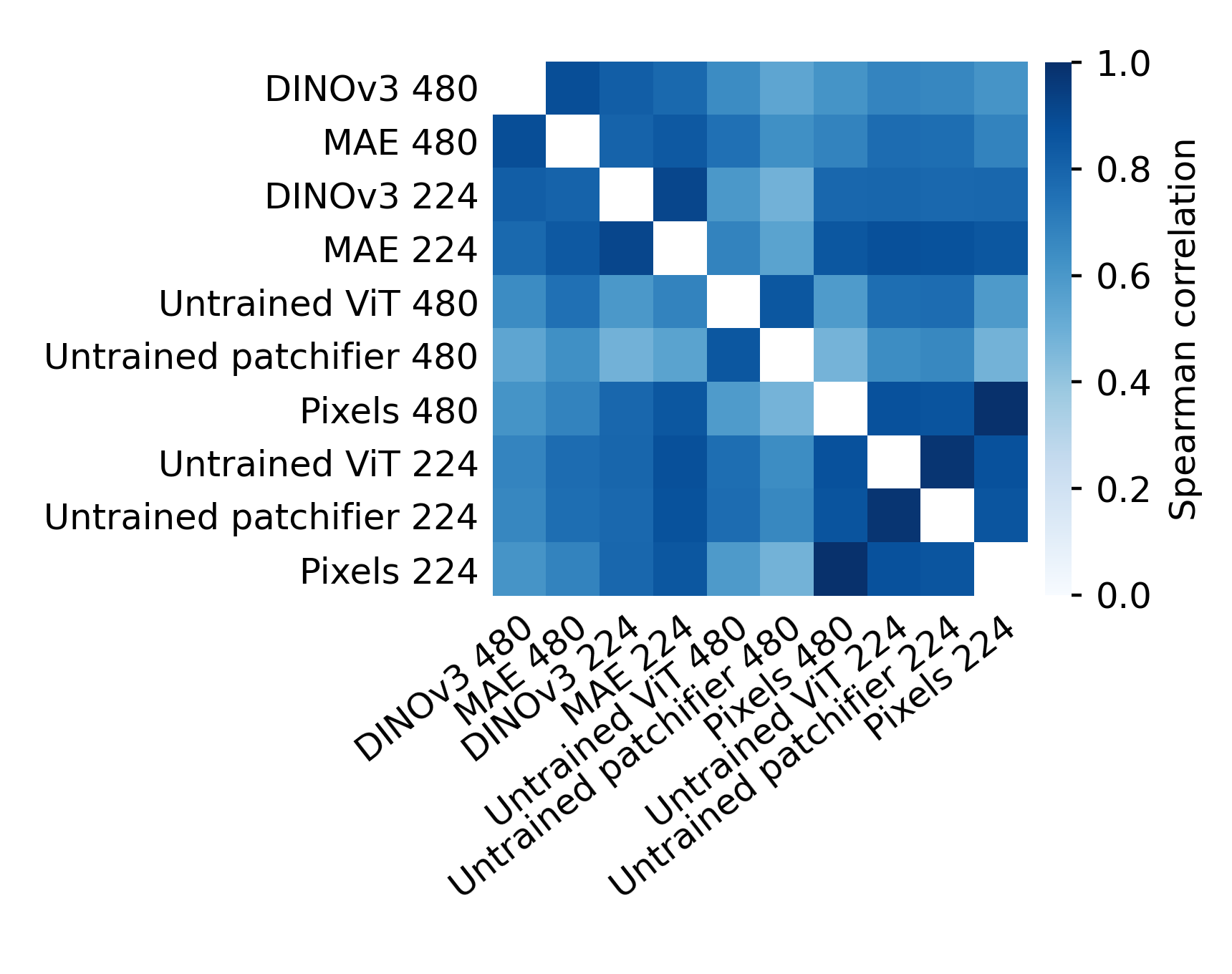}
        \caption{Agreement between representations across tasks.}
        \label{fig:arcgen-retrieval-matrix}
    \end{subfigure}

    \caption{
        Additional ARC-GEN retrieval diagnostics. DINOv3, MAE, and random-projection pixel controls are compared at 224 and 480 px, with the untrained ViT and untrained patchifier (Figure~\ref{fig:bongard-task-vector-analysis}) added in (a,b). Retrieval uses 100 generated queries from each of 794 tasks and a 368{,}442-grid candidate pool; encoder layers were selected on original ARC queries and then held fixed. \textbf{(a)} Exact-output retrieval without the intuition vector (pale) and with it (solid) at each encoder's selected layer, alongside the pixel controls; pixels and the untrained controls are right of the divider, and light and dark shades denote 480 and 224 px. Whiskers are marginal $95\%$ task-bootstrap intervals. \textbf{(b)} Pairwise Spearman correlations between representations' task-wise transported retrieval rates (color scale at right).
    }
    \label{fig:supp-arcgen-overview}
\end{figure*}

\subsection{Near-miss ARC 1\&2 retrieval}
\label{app:noisy_arc}
Near-miss pools probe finer discrimination. Intuition vector calculation and application is identical to ARC1\&2 procedure but here, the candidate pool contains the true output and 99 independently recolored variants at each corruption level. The analysis uses corruption fractions $\{0.01,0.05,0.10,0.30,0.60\}$; $\max(1,\operatorname{round}(\rho HW))$ cells are sampled without replacement and each is replaced uniformly by one of the other nine ARC colors. Corruption occurs before rendering and never changes padding. All test queries are included, regardless of whether whole-pool retrieval succeeds. We evaluated DINOv3 at 480 px (layers 6, 12, 17, and 22) and the pixel control on the identical corrupted grids, averaging the latter over the ten projection seeds; MAE and 224-px renderings were not evaluated in this setting. Retrieval accuracy uses the binomial standard error $\sqrt{\hat p(1-\hat p)/N}$; for pixels, this is combined with the SEM across projection seeds. Consequently, error bars describe query-level sampling variation and do not cluster multiple queries from the same task.

\begin{figure*}[p]
    \centering
    \captionsetup[subfigure]{skip=2pt}

    \begin{subfigure}[t]{0.49\textwidth}
        \centering
        \includegraphics[width=\linewidth]{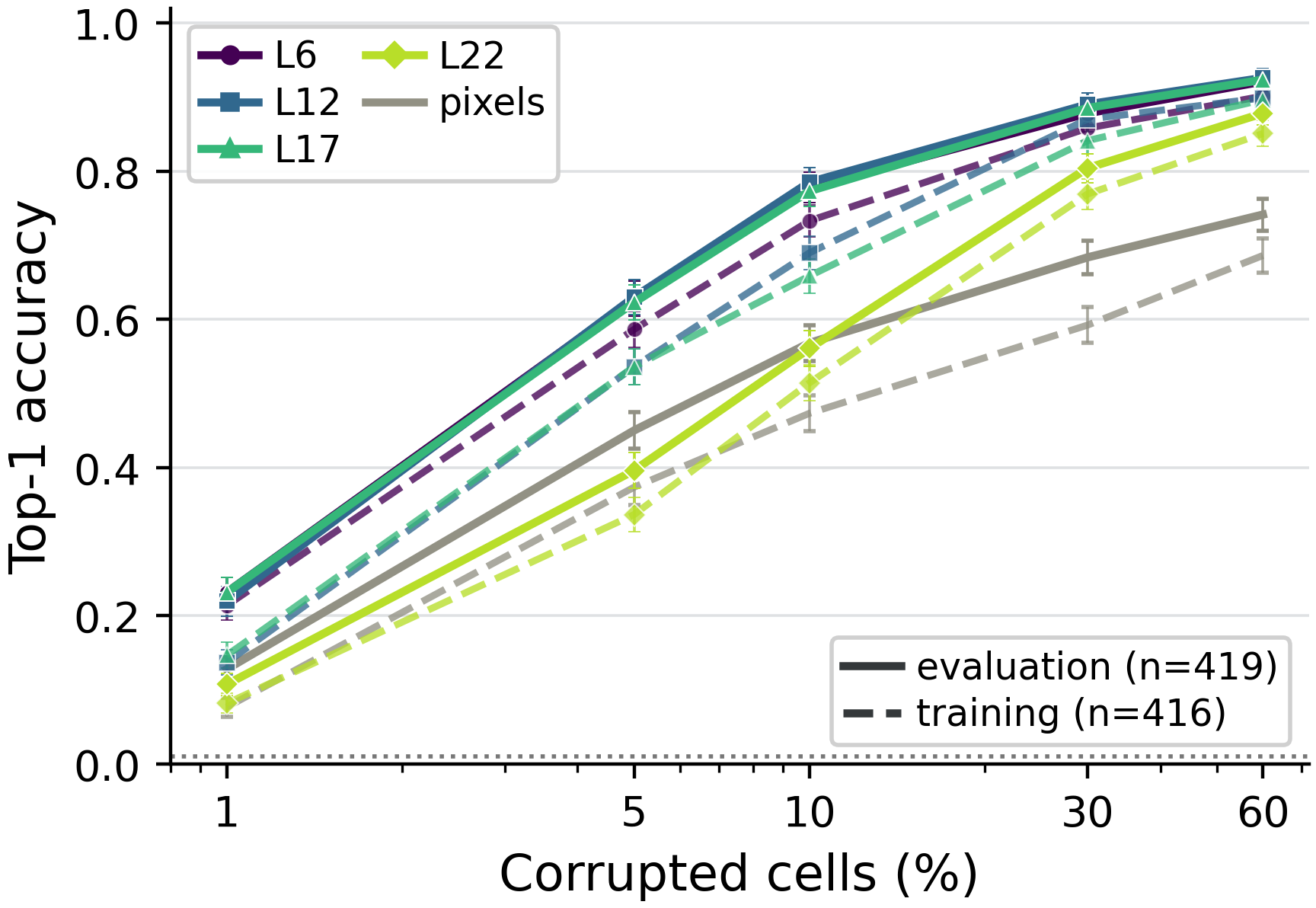}
        \caption{Noisy-grid discrimination on ARC-1.}
        \label{fig:arc1-knn-noisy}
    \end{subfigure}
    \hfill
    \begin{subfigure}[t]{0.49\textwidth}
        \centering
        \includegraphics[width=\linewidth]{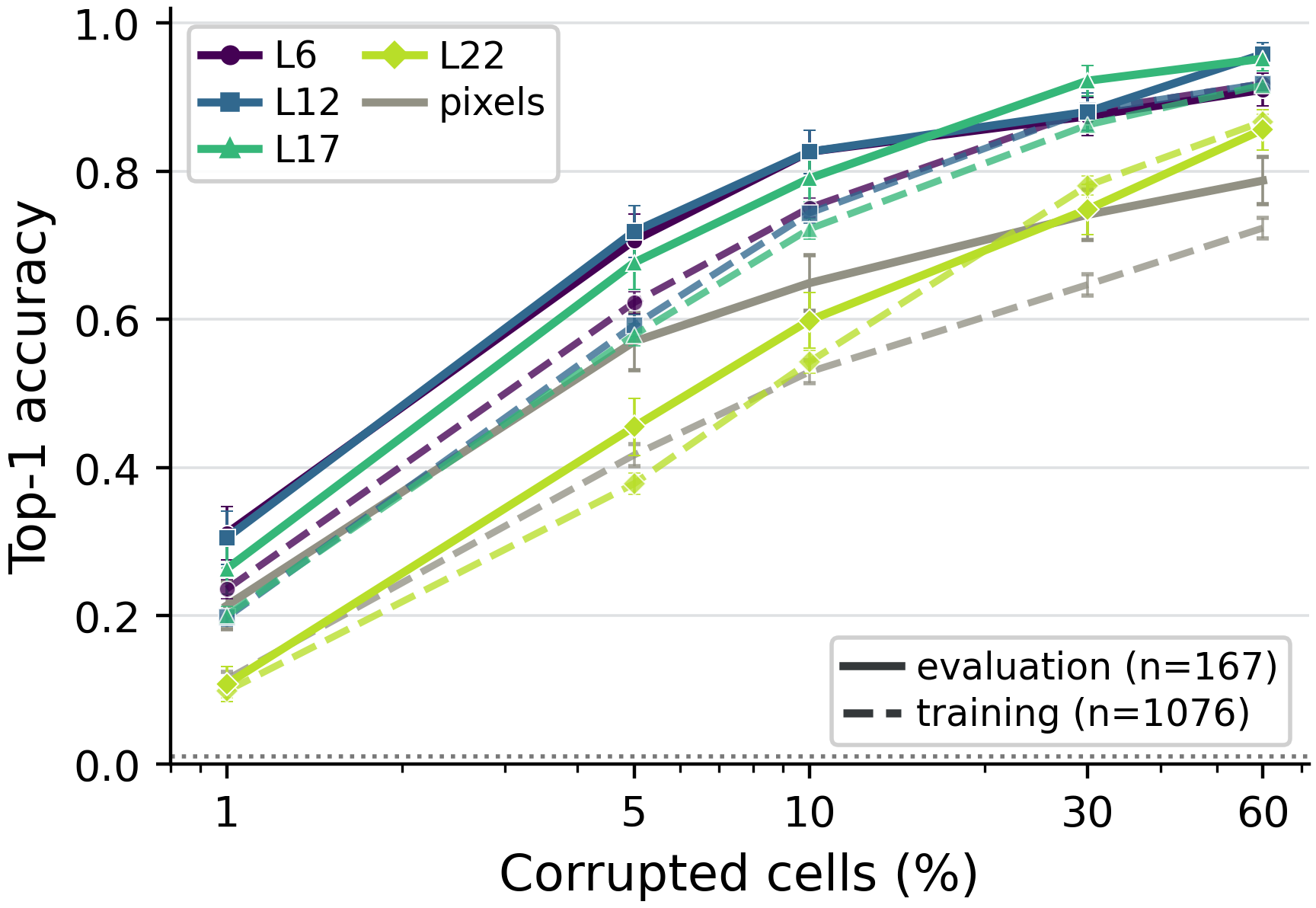}
        \caption{Noisy-grid discrimination on ARC-2.}
        \label{fig:arc2-knn-noisy}
    \end{subfigure}

    \caption{
        Nearest-neighbor discrimination after intuition-vector translation on noisy
        ARC-1 and ARC-2 for DINOv3 (480 px) and 10 random projections of pixel-space. \textbf{(a,b)} Top-1
        retrieval of the true answer from a 100-grid pool containing the answer
        and corrupted variants, shown as a function of the fraction of corrupted
        cells. Colors identify sampled model layers; error bars are binomial SEM
        estimates. For pixels, error bars combine the mean binomial SEM with the SEM across seeds.
    }
    \label{fig:supp-arc-knn-noisy}
\end{figure*}

To test whether the pixel lead reflects low-level overlap between query and answer rather than a finer representation of the outputs, these pools make each answer compete only with corrupted copies of itself. Accuracy rose with the recolored fraction in both representations, but once at least 5\% of cells were recolored, pixels remained 11--22 points below intermediate DINOv3 layers at 480 px (Fig.~\ref{fig:supp-arc-knn-noisy}).

\subsection{The Reasoning Representation Autoencoder (RRAE): implementation details}
\label{sec:supp-rae}

\paragraph{Overview.}
The RRAE has three components (Fig.~\ref{fig:rae2-method}). A frozen DINOv3-L/16 \emph{encoder} turns a rendered grid into a $30\times30$ array of latent tokens, one per grid cell, and a CLS token. A \emph{pixel decoder} maps latent tokens back to a rendered grid. A \emph{diffusion transformer} (DiT) generates the latent tokens of the answer grid from Gaussian noise, given the latent tokens of the query input and a task signal: the intuition vector, the latent tokens of the demonstration grids, or a learned task token. Training has two phases: the decoder is trained to reconstruct grids from their latents and frozen (phase 1), and the DiT is then trained with flow matching on few-shot training examples (phase 2). At inference the generated latents are decoded to a grid under 32 augmented views that vote on the answer, optionally after test-time training on the task's demonstrations. The paragraphs below follow this order: rendering and latents, the decoder, the DiT and its three conditioning variants, the data, the training examples, the objective, compute, and finally inference and test-time training.

\paragraph{Rendering, latents and CLS tokens.}
Every grid is rendered at 480 px as a $30\times30$ canvas of 16-px cells, so that each cell coincides with one DINOv3 patch. Unlike the centered renders of the representation analyses, the grid is placed at the top-left corner of the canvas and the remaining cells take a padding color, so that cell $(r,c)$ always occupies the same latent token and the output size is read off as the extent of non-pad cells. Latents are the DINOv3-L/16 patch tokens averaged across the seven blocks $L=\{11,13,\dots,23\}$ at each spatial position (zero-indexed, final layer norm applied). The final block's spatial mean patch token is then added to every position, followed by per-channel normalization with dataset statistics, giving a $30\times30\times1024$ latent per grid. Every CLS token the model uses (query, demonstrations, and the auxiliary target) is the average of the same seven layers' CLS tokens after the encoder's final layer norm. The intuition vector $\bar v$ uses the same layer average, computed from 224-px renders ($32\times32$ canvas, 7-px cells, grid at the top-left), as these identify tasks better than the 480-px renders (Sec.~\ref{sec:task-identity}; Supp.~Fig.~\ref{fig:supp-arcgen-classification-480}). It thus differs from the single-layer $v_l$ of the representation analyses in the layer average, in the placement of the grid, in applying the encoder's final layer norm to each layer's CLS token, and in using at most three demonstrations.

\paragraph{Pixel decoder (phase 1).}
We use the RAEv2 ViT-XL decoder config (28 blocks, width 1152, 16 heads). It is trained with L1 pixel loss plus a palette (cell-color) loss, AdamW lr 2e-4, batch size 1, 16 epochs, on 7,000 real ARC grids plus 7,000 synthetic random grids, without noise augmentation, and the checkpoint with the best validation accuracy is kept (epoch 8); it reconstructs 99.6\% of validation grids exactly (99.998\% of cells). So that perception and readout are not a bottleneck for generation, the decoder is trained on a broad set of grids: the real grids are drawn from the whole ARC1 and ARC2 training sets, which include grids of 115 of the 120 held-out tasks. The decoder only learns to reconstruct a grid from its own latents, so this exposes neither a task's rule nor its query-answer mapping to the DiT, whose training data exclude these tasks. Decoded images are read out as the nearest palette color, or pad, per cell, and the output grid is the top-left bounding box of non-pad cells.

\paragraph{Diffusion transformer (phase 2).}
The DiT is the RAEv2 diffusion transformer with a wide DDT head (DiT$^{\mathrm{DH}}$-XL in RAE~\citep{zheng2025rae}), at the size RAEv2 uses for its world model (its image models use a wider trunk): a 28-block transformer trunk (width 1152, 16 heads) processes the token sequence, and a 2-block head (width 2048, 16 heads) predicts the clean target latent, modulated (adaLN) by the trunk's output tokens and the flow time. As in the RAE, the trunk is wider than the encoder's 1024-d tokens, so every input is mapped to the trunk width by a learned linear map: latent tokens by a $1\times1$ patch embedding, and the task code by a linear projection.

The trunk sequence contains the 900 noisy target tokens (one per cell), four time tokens (Gaussian-Fourier embeddings of the flow time), a learned output-summary token from which the target CLS is predicted (an auxiliary loss, below), and the conditioning tokens, which differ between the models. For the intuition-vector and task-token models these are the 900 query-latent tokens, the query CLS token, the task code and the orientation token (903 conditioning tokens, 1,808 in total). The task code is $\bar v$ or $\tau_k$, a row of a learned table with one 1024-d row per training task; both enter as one token through the same learned projection, so the two models differ only in where the code comes from. The orientation token is a learned embedding of the rotation or reflection (one of the eight dihedral transforms of the square) applied to the example's grids. For the full-grid model the conditioning tokens are the latent grids and CLS tokens of up to three demonstration pairs, with learned role (input/output) and pair embeddings, followed by the query (6,307 conditioning tokens, 7,212 in total); missing demonstrations are masked out of attention. Grid-shaped token sets share 2D rotary position embeddings, so a cell has the same position in the query, demonstration and target grids. For computational efficiency the full-grid model sees at most three demonstrations; for comparability the single-token models use the same limit, although they could use any number at only the cost of encoding them. Parameter counts are given in Table~\ref{tab:rae-params}.

\begin{table}[h]
\centering
\small
\setlength{\tabcolsep}{4pt}
\begin{tabular}{@{}lrrr@{}}
\toprule
 & $\bar v$ & Full grid & $\tau_k$\\
\midrule
Trunk ($28\times1152$) & 446.3M & 446.3M & 446.3M\\
DDT head ($2\times2048$) & 151.1M & 151.1M & 151.1M\\
Embeddings, heads & 26.6M & 25.4M & 26.6M\\
Task table ($166{,}503\times1024$) & -- & -- & 170.5M\\
\midrule
Trainable total & 624.0M & 622.8M & 794.5M\\
\midrule
Encoder (frozen) & \multicolumn{3}{c}{303.2M}\\
Decoder (frozen) & \multicolumn{3}{c}{416.4M}\\
\bottomrule
\end{tabular}
\caption{Parameter counts of the RRAE variants (trunk and head: blocks $\times$ width). Encoder: DINOv3-L/16; decoder: phase-1 ViT-XL. The intuition-vector count leaves out a 1.9M per-task table that its checkpoint carries but never uses.}
\label{tab:rae-params}
\end{table}

\paragraph{Held-out tasks and training data.}
The holdout is 120 tasks sampled at random from the 393 eligible ARC1 evaluation tasks, excluding the six that are also ARC2 evaluation tasks and one that shares an exact pair with an ARC1 training task. It is removed from every training source by task id, ARC-GEN generator id, and content fingerprint, and contains 123 test queries; the ARC2 evaluation set has 120 tasks and 167 test queries. Phase 2 trains on five sources: ARC1 training (400 tasks); ARC2 training, which overlaps ARC1 (493 tasks after removing the holdout and exact duplicates of ARC1 training tasks, 261 of them ARC1 evaluation tasks; the remaining 19 non-held-out ARC1 evaluation tasks are not trained on); ARC-GEN, which generates further pairs for existing tasks, with up to 1,000 generated pairs per task for 400 ARC1-training and 373 ARC2-training tasks ($\sim$700k pairs); BARC, synthetic tasks that recombine the rules of ARC1 training tasks (164,678 rules, $\sim$1.6M pairs, after holding out 10\% of the rules for validation); and ConceptARC (159 tasks). In total, 166,503 tasks and 2.3M pairs.

\paragraph{Training examples.}
A training example is a query pair with up to three demonstration pairs of the same task (Sec.~\ref{sec:methods-rae}). The query pair is drawn uniformly from all pairs of the task, including the original test pairs of training tasks, and up to three of the remaining pairs form the demonstrations. One random dihedral transform and one uniformly random permutation of the 10 colors (pad fixed) are applied to every grid of the example. Because the sources differ in size by orders of magnitude, an epoch is not a pass over all pairs but a fixed expected number of examples per task: 4 for ARC1, ARC2 and ConceptARC, 14 for ARC-GEN, and 0.0863 for BARC, so that BARC makes up $\sim$48\% of the 29,393 examples per epoch.

\paragraph{Objective and optimization (phase 2).}
The target latent $z$ is mixed with Gaussian noise $\epsilon$ as $z_t=(1-t)\,z+t\,\epsilon$, where $t\in[0,1]$ is the flow time ($t=1$ is pure noise), and the network predicts the clean latent from $z_t$, the query and the task signal. The loss is computed in velocity space, \ie on the prediction error divided by $t$ (clamped below at $t_{\min}=0.05$). Flow time is sampled from a logit-normal distribution with RAE's dimension-dependent shift $s=\sqrt{30\cdot30\cdot1024/4096}=15$. The flow loss weights cells that differ from the target's dominant color by 2. Following RAEv2~\citep{singh2026raev2}, we also use representation alignment through an intermediate clean patch-latent prediction from the internal-guidance base head, read out after block 8 of 28. This prediction receives the same weighted velocity-space flow loss as the final prediction, with coefficient 1; the base head is not used for guidance at inference. Separately, we add a cosine loss predicting the target's layer-averaged DINO CLS from the output-summary token, with coefficient 0.1. The total objective is
\[
\begin{aligned}
\mathcal{L}&=\mathcal{L}_{\mathrm{flow}}+\mathcal{L}_{\mathrm{base}}+0.1\,\mathcal{L}_{\mathrm{CLS}},\\
\mathcal{L}_{\mathrm{CLS}}&=1-\cos(\hat c,c),
\end{aligned}
\]
where $\hat c$ is the predicted CLS and $c$ is the frozen encoder's target CLS; the cosine loss is averaged over the batch.

We use AdamW ($\beta_1=0.9$, $\beta_2=0.95$) at a constant learning rate of $7\times10^{-5}$ without weight decay, gradient-norm clipping at 1.0, bf16 autocast, and an effective batch of 8 examples on 4 H100 GPUs (3,675 updates per epoch). Task-token rows are trained with Adam at $3\times10^{-3}$ (weight decay 0.1). We keep an exponential moving average of the weights (decay 0.9999) and evaluate the EMA weights. Each model is a single run at constant learning rate, evaluated at EMA snapshots along it: epochs 6, 12, 24, 48 and 96 for all models, 192 for the intuition-vector and task-token models, and 60, 144, 240, 300 and 384 for the intuition-vector model. The runs end at epoch 384 (intuition vector), 192 (task token) and 96 (full grids).

\paragraph{Compute.}
One training example costs 9.3 TFLOP for the single-token models and 48.8 TFLOP for the full-grid model (forward and backward, including encoding), i.e.\ $5.2\times$ more. Comparisons at approximately matched GPU training time pair full-grid epoch $e$ with single-token epoch $4e$ (12/48, 24/96, 48/192), which gives the single-token models about 24\% fewer training FLOPs (\eg $5.2$ vs.\ $6.9\times10^{19}$ FLOP at 192/48). In wall-clock time on 4 H100 GPUs, including per-epoch monitoring, an epoch took 0.42 h for the single-token models and 1.58 h for the full-grid model, \ie 1.7 and 6.3 H100-hours. This amounts to about 320 H100-hours for the single-token models at epoch 192 and 300 for the full-grid model at epoch 48 (640 and 610 at epochs 384 and 96), giving similar GPU training time for the paired checkpoints.

\paragraph{Sampling and voting.}
Two sampling steps already give most of the performance of longer schedules ($43.3\%$ versus $45.4\%$ with 20 steps after 400 TTT steps, and $7.5\%$ versus $10.8\%$ zero-shot, neither difference significant; Supp.~Fig.~\ref{fig:arc-rae-sampler-calls}), and additional augmentations keep improving accuracy (Supp.~Fig.~\ref{fig:arc-rae-view-voting}); we use two steps and cap the augmentations at 32.
Each answer uses two network evaluations: one at $t=1$ on pure noise, an Euler step to $t=0.4$, and a second evaluation whose clean-latent prediction is decoded as the answer. The 32 inference views combine the 8 orientation transforms with 4 color permutations (the identity and three fixed random permutations); each view uses its own noise sample, receives its orientation token (single-token models), and its answer is mapped back to the original frame. The cell-wise vote takes the most common output size and then each cell's most common color among the answers of that size, so it need not equal any single answer.

\paragraph{Test-time training.}
All DiT parameters are fine-tuned, starting from the EMA weights and reset for every task, with Adam ($\beta_1=0.9$, $\beta_2=0.95$, no weight decay) at a learning rate of $3\times10^{-5}$, a 10-step linear warm-up and gradient-norm clipping at 1.0, using the training loss and its weights. For the task-token model, $\tau_k$ is fitted at a learning rate of $3\times10^{-3}$. The demonstrations of each fitting example are a random subset of up to three of the demonstrations other than the pseudo-query, so tasks with more than four demonstrations use all of them over a run. Its augmentation is drawn uniformly from the same 32 views as at inference (three random color permutations, fixed across tasks, plus the identity). Examples are drawn independently with replacement, so over a run each (demonstration, augmentation) combination is visited several times with different noise and flow times.

\paragraph{Intuition-vector ablations.}
The epoch-192 intuition-vector model is evaluated with its vector intact, set to zero, or replaced by another task's vector, at 0, 100, and 400 TTT steps. Donor tasks are assigned by a fixed seeded derangement within each evaluation set, so no task receives its own vector. The donor's demonstrations undergo the same rotation/reflection and color permutation as the recipient's grids. The vector replacement is applied during both TTT and inference; the query and training targets remain those of the recipient task. Thus, TTT still fits the recipient's demonstrations, with the missing or incorrect task signal held in place. Adapted results average three TTT seeds (Supp.~Fig.~\ref{fig:supp-rae-ablation}).

\subsection{Low-level compositionality analyses}
\label{app:low_level}

\subsubsection{Scope, construction, and controls}
\label{app:low-level-methods}
These exploratory analyses use the frozen DINOv3-Large/16 encoder at 480 px; MAE and the 224-px rendering are not evaluated. Grids use the same $30\times30$ canvas as the main analyses. Unless otherwise stated, scalar results are reported at layer 17. Translation, recoloring, and addition vectors are differences between the relevant endpoint embeddings, and composition is evaluated using cosine similarity and the relative residual $\lVert M-\widehat M\rVert/\lVert M\rVert$.

For the context-free analysis, the local cosine matrix uses the 784 source positions at which all four unit translations are valid, four translations of a blue cell, and its eight directed recolorings. For the layerwise composition test, translation primitives are averaged over valid positions and all nine foreground colors for each displacement $d$, and recoloring primitives are averaged over all positions for each ordered color pair. We test all 796 nonzero displacements satisfying $\lVert d\rVert_2\leq16$ and all 72 directed recolorings, sampling eight valid source positions without replacement for each combination. This yields $458{,}496$ joint transformations per layer. The shuffled control independently permutes displacement and recoloring assignments while preserving both marginals and is averaged over ten permutations.

Contextual analyses use nine interior anchor positions. For each anchor and occupancy, four independently generated placement sets contribute eight layouts each, giving 32 layouts per anchor and $9\times32=288$ spatial units. Positions are sampled uniformly without replacement from unoccupied cells, and conditions cross all nine non-black ARC colors. In the addition-transfer analysis, contexts contain one to five cells and all 81 ordered added-cell/context-color pairs are tested, producing 23,328 transitions per occupancy and 116,640 overall. In the shared-context composition analysis, the anchor is the one-cell base $B$ and two to five sampled cells form the jointly added set $R$. Wrong-location controls average ten permutations of the added sites while retaining the base, occupancy, and colors. Layerwise analyses use a balanced subset of 16 layouts per anchor (144 spatial units).

Color conditions sharing a spatial layout are not treated as independent replicates. Values are first averaged over colors within each anchor-layout unit and then across spatial units; 95\% confidence intervals are percentile intervals from 2,000 bootstrap resamples of spatial units. The exhaustive context-free analyses are treated as fixed-design descriptive summaries, and their ten shuffles estimate the control mean rather than a sampling interval. For position decoding, a standardized ridge classifier ($\alpha=100$) predicts the nine-way anchor position from $\Delta_\ell(a;C)$ at layer 17 with five context cells. Four-fold grouped cross-validation holds out one independently generated placement set per fold and keeps all color variants of a layout in the same fold. Confidence intervals use the same spatial-unit bootstrap.

\subsubsection{Context-dependent compositionality of single-cell transformations}
We begin with the occupancy-preserving operations, translation and recoloring, on otherwise empty synthetic $30\times30$ ARC-format grids. Let $S(p,c)=E_\ell(\mathrm{grid}(p,c))-E_\ell(\varnothing)$ represent one cell of color $c$ at position $p$ (Fig.~\ref{fig:single-cell-design}). Differences between these states define individual low-level transformations, while averages over controlled positions or colors define canonical family directions. Unlike intuition vectors, these averages estimate a predefined operation across known realizations rather than infer a latent task from its demonstrations. At layer 17, opposite unit translations are strongly opposed (cosines $-0.90$ for up--down and $-0.85$ for left--right), whereas orthogonal translations and translation--recoloring pairs are near orthogonal. Recoloring directions are mutually aligned to varying degrees (Fig.~\ref{fig:single-cell-primitive-cosine}), separating the two transformation families in local geometry.

We then predict individual joint translation--recoloring changes by summing family-average primitives: the translation for the required displacement averaged over valid positions and colors, and the directed recoloring averaged over positions. Across $458{,}496$ balanced transformations per layer, the matched prediction has mean cosine near $0.999$ and relative residual $0.03$--$0.05$ in early layers. It remains strong at layer 17 (cosine $0.948$, residual $0.298$), whereas shuffled displacement--recoloring assignments have near-zero cosine and residuals of approximately $1.4$--$2.6$ (Fig.~\ref{fig:single-cell-primitive-composition}). Thus, context-free translation and recoloring admit stable, approximately additive primitives.

Having established that canonical family directions predict individual transformations on empty grids, we next examine the individual transformation vectors directly. We compare each transformation measured on an empty grid with the corresponding transformation in a grid containing one to five static cells (Fig.~\ref{fig:single-cell-contextual-design}). Context reduces transfer in both families, especially when context colors match the moved cell or a recoloring endpoint. At layer 17 with five context cells, matched-color translation has mean cosine $0.239$ with its isolated counterpart, and matched-endpoint recoloring has mean cosine $0.528$ (Fig.~\ref{fig:single-cell-context-transfer}). The five-cell layerwise analysis shows increasing context sensitivity after approximately layer 12 (Fig.~\ref{fig:supp-layerwise-context-transfer}).

Despite this change in the primitive vectors, transformations measured from the same contextual source remain nearly additive. At layer 17, the cosine between a joint transformation and the sum of its contextual translation and recoloring decreases only from $0.9984$ with one context cell to $0.9977$ with five; the relative residual remains $0.049$--$0.053$. Empty-grid primitives degrade to cosine $0.646$ and residual $1.166$ at five cells, and an incorrect contextual recoloring also performs worse (Fig.~\ref{fig:single-cell-contextual-composition}). Across layers, contextual primitives remain essential through the middle and deep blocks (Fig.~\ref{fig:supp-layerwise-context-composition}). Translation and recoloring are therefore locally compositional, but their vectors depend on the state in which they are measured.

\newlength{\figthreepanelheight}
\begin{figure*}[tbp]
\centering
    \captionsetup[subfigure]{skip=1pt}

    \setlength{\figthreepanelheight}{0.235\textwidth}

    \begin{subfigure}[t]{0.335\textwidth}
        \centering
        \vspace{0pt}
        \includegraphics[height=\figthreepanelheight]{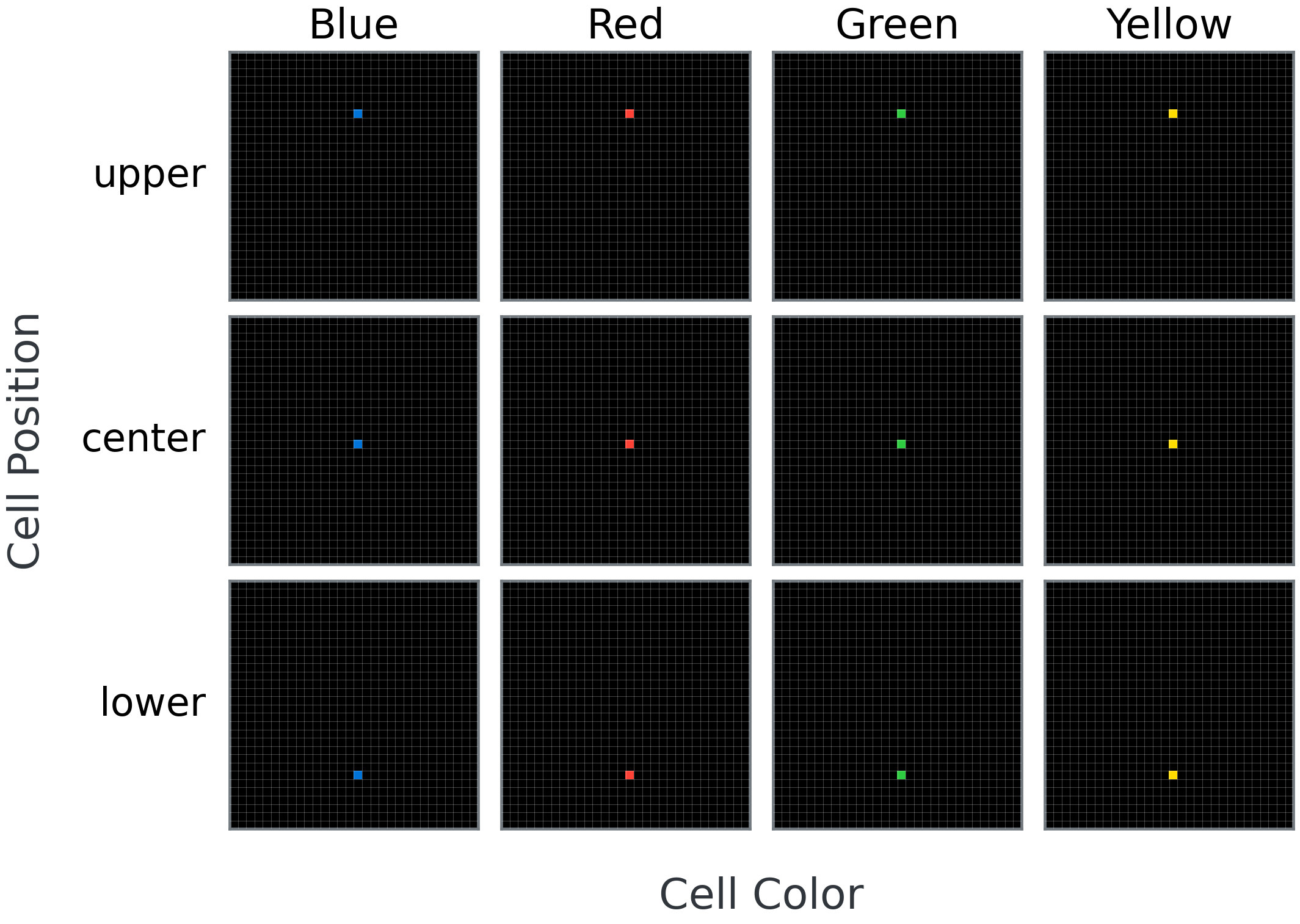}
        \caption{Context-free single-cell design.}
        \label{fig:single-cell-design}
    \end{subfigure}
    \hfill
    \begin{subfigure}[t]{0.305\textwidth}
        \centering
        \vspace{0pt}
        \includegraphics[height=\figthreepanelheight]{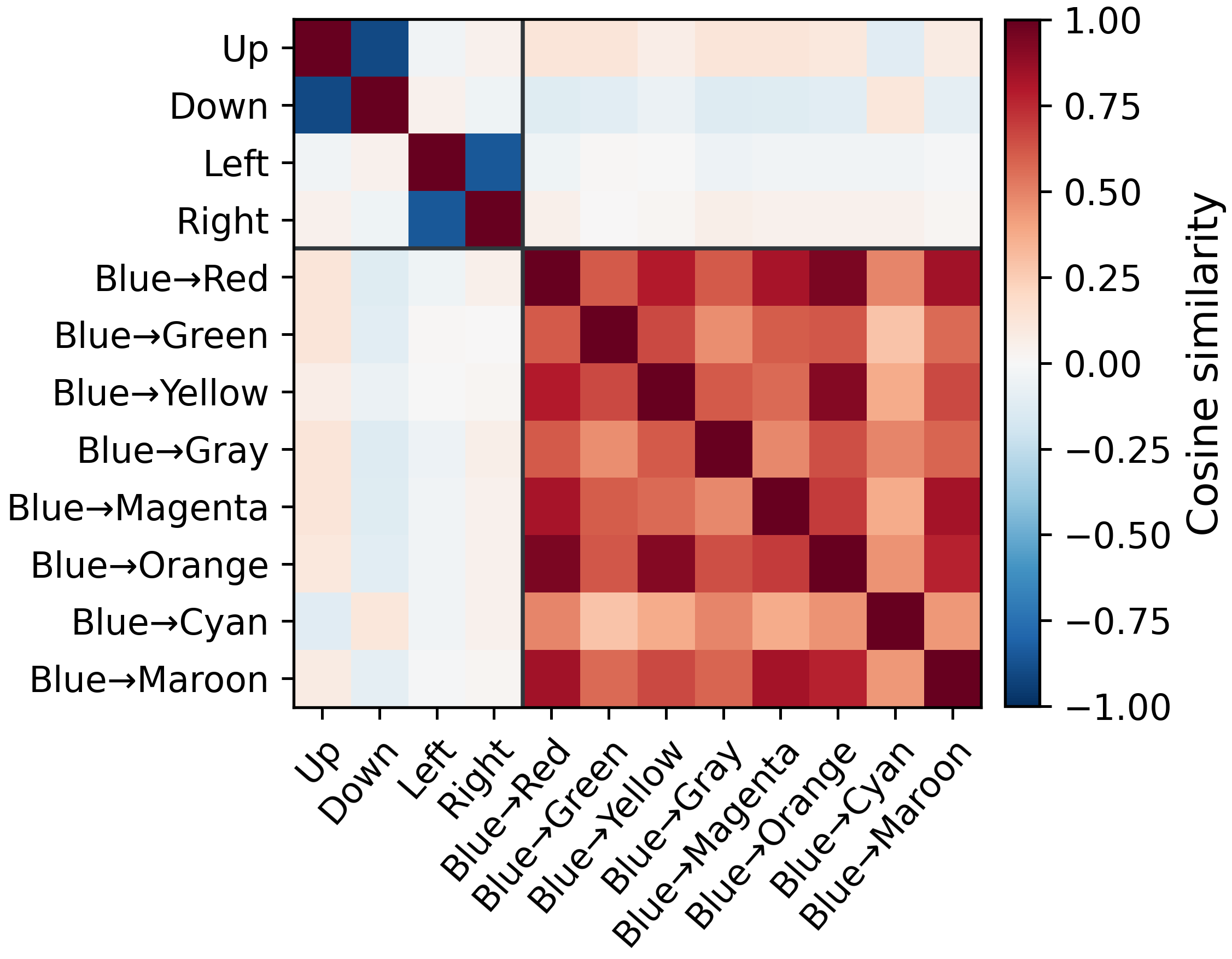}
        \caption{Context-free primitive geometry.}
        \label{fig:single-cell-primitive-cosine}
    \end{subfigure}
    \hfill
    \begin{subfigure}[t]{0.315\textwidth}
        \centering
        \vspace{0pt}
        \includegraphics[height=\figthreepanelheight]{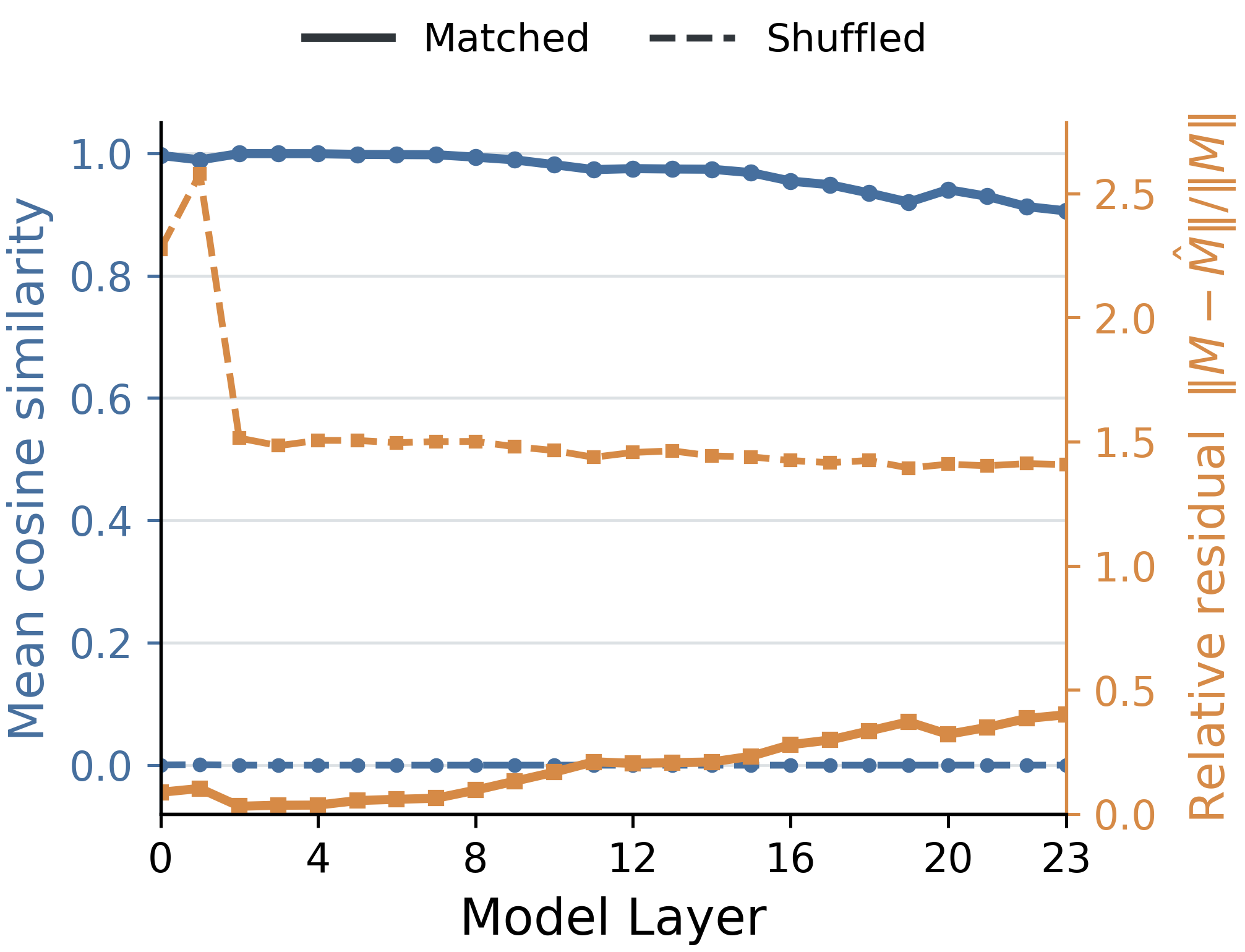}
        \caption{Context-free primitive composition.}
        \label{fig:single-cell-primitive-composition}
    \end{subfigure}

    \vspace{4pt}

    \begin{subfigure}[t]{0.335\textwidth}
        \centering
        \vspace{0pt}
        \includegraphics[height=\figthreepanelheight]{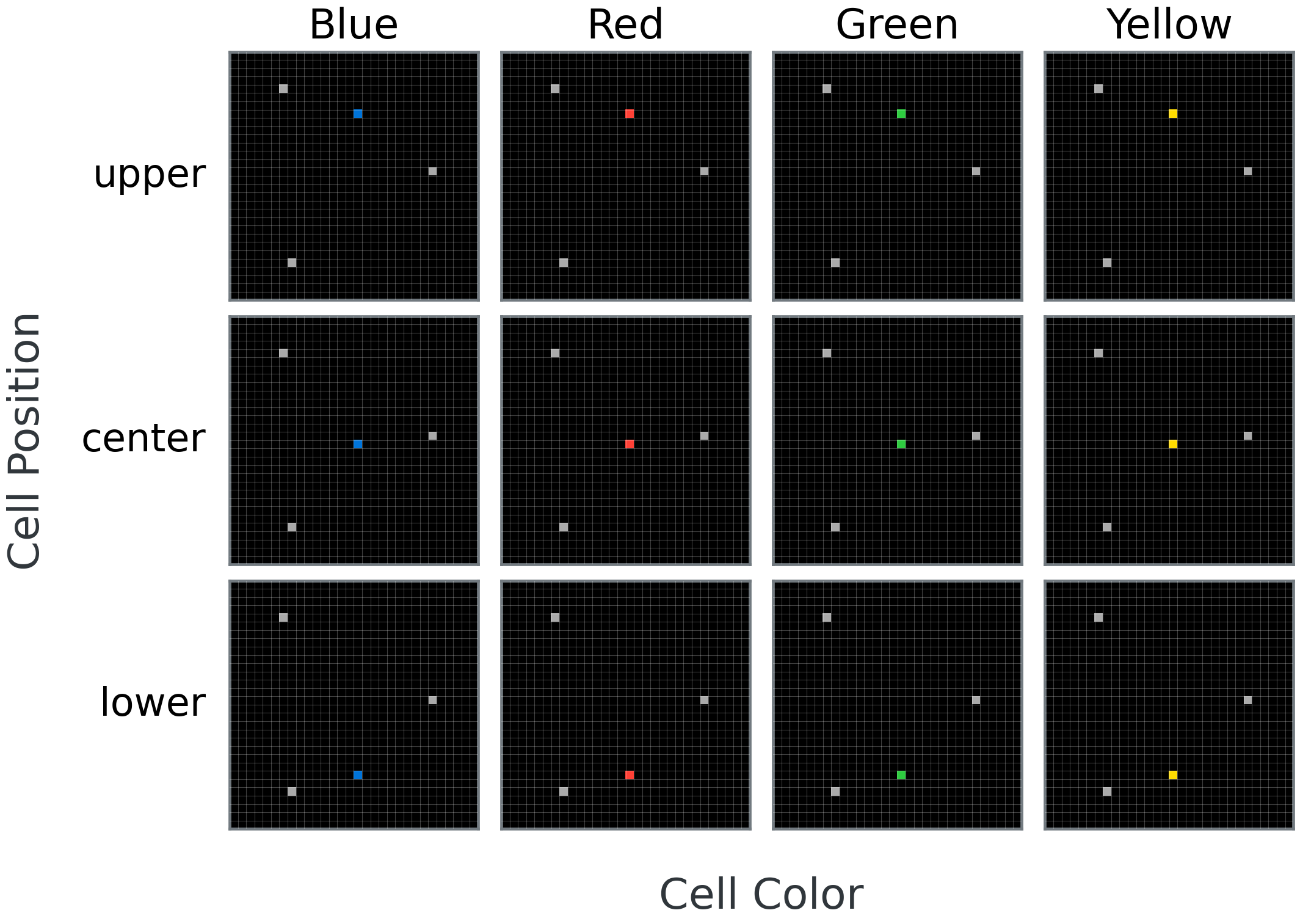}
        \caption{Single-cell design in a shared context.}
        \label{fig:single-cell-contextual-design}
    \end{subfigure}
    \hfill
    \begin{subfigure}[t]{0.315\textwidth}
        \centering
        \vspace{0pt}
        \includegraphics[height=\figthreepanelheight]{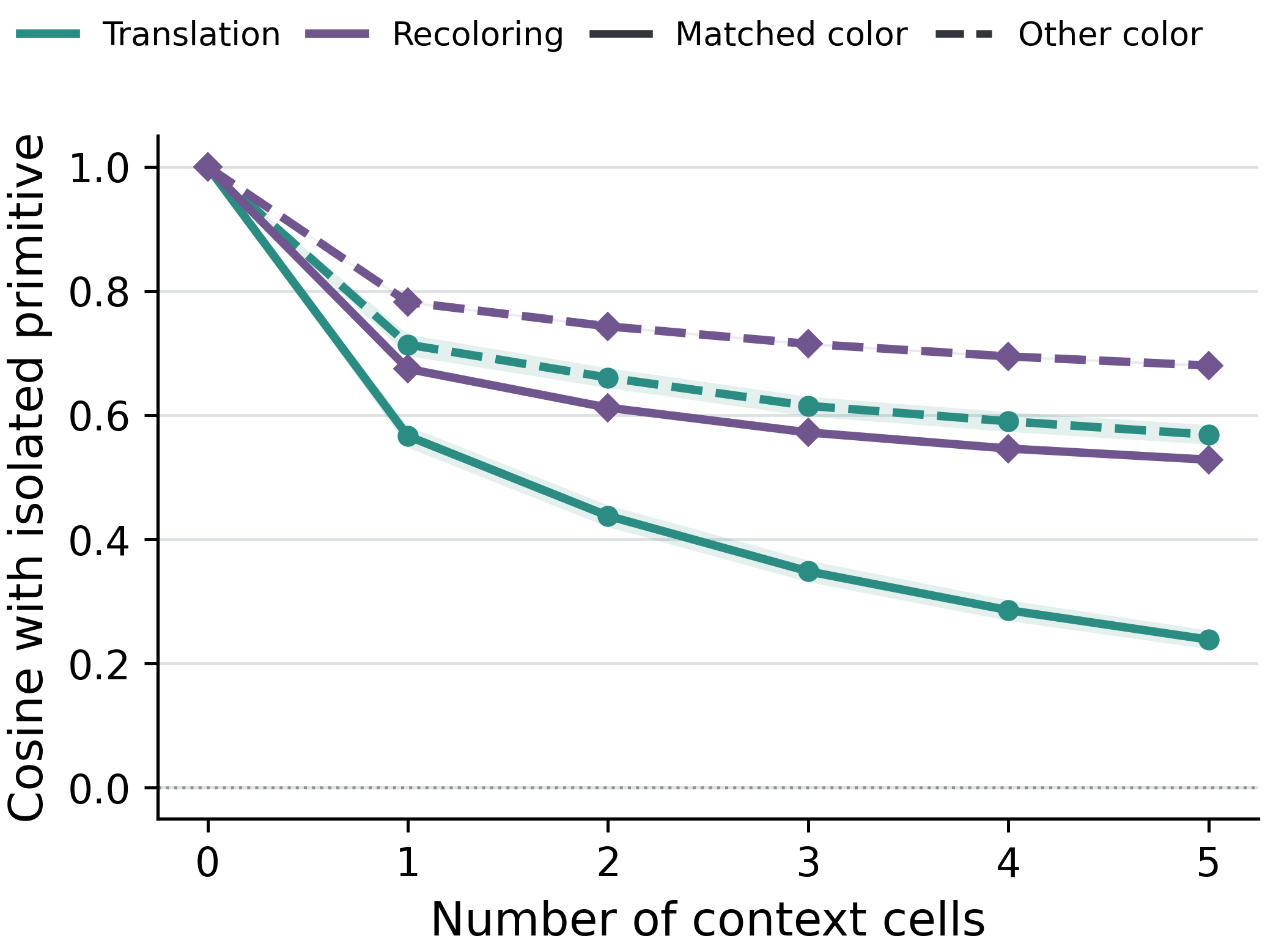}
        \caption{Transfer of isolated primitives into context.}
        \label{fig:single-cell-context-transfer}
    \end{subfigure}
    \hfill
    \begin{subfigure}[t]{0.315\textwidth}
        \centering
        \vspace{0pt}
        \includegraphics[height=\figthreepanelheight]{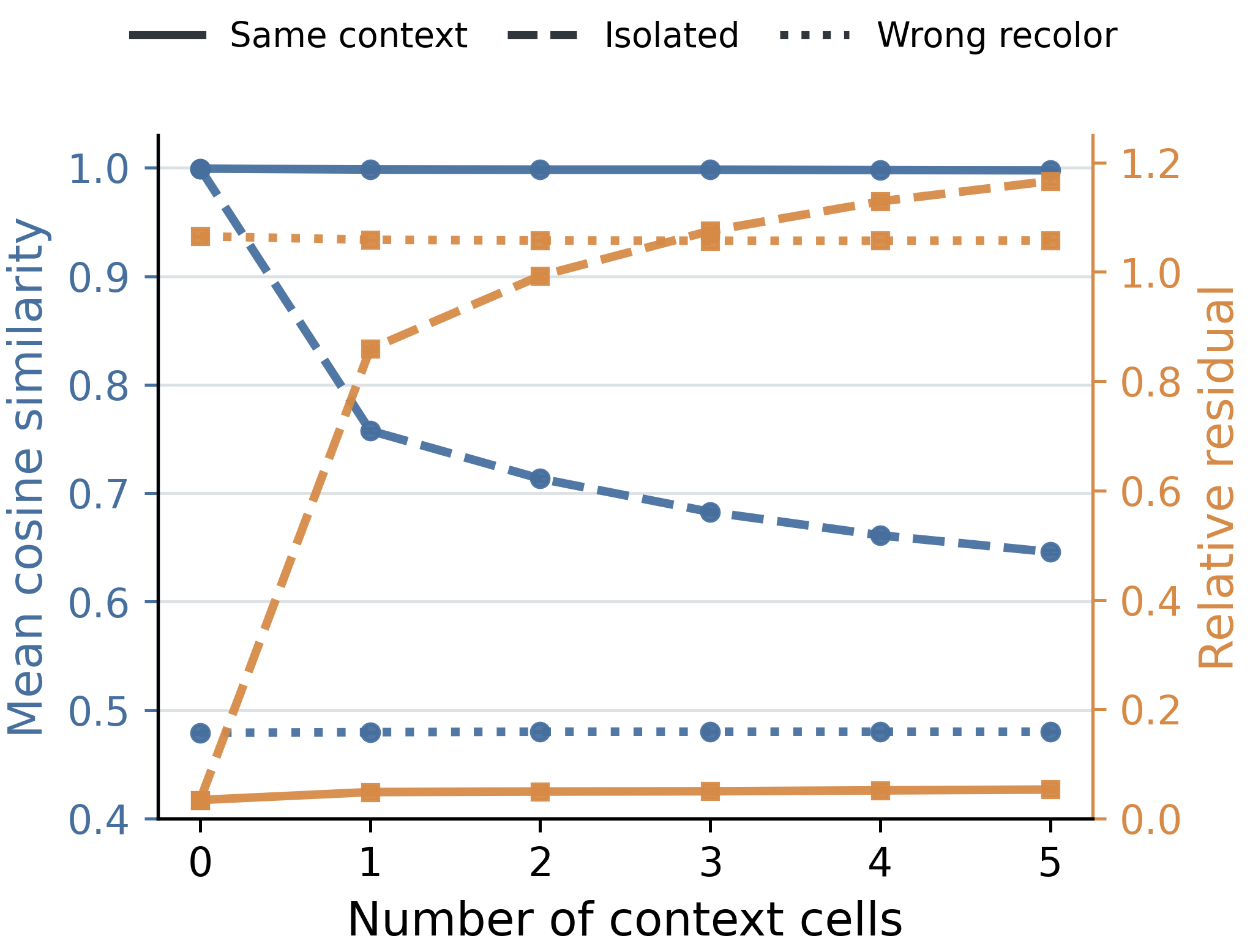}
        \caption{Composition around the same context.}
        \label{fig:single-cell-contextual-composition}
    \end{subfigure}

    \caption{
    Primitive transformations in DINOv3-Large at 480 px compose locally but depend on their source configuration. Layer numbers are zero-indexed.
    \textbf{(a,d)} Synthetic single-cell example designs without and with shared context.
    \textbf{(b)} Layer-17 geometry of four unit translations and eight
    recolorings from blue.
    \textbf{(c)} Layer-wise prediction of individual joint transformations
    from family-average translation and recoloring vectors, with shuffled
    controls.
    \textbf{(e)} Similarity between contextual primitives and their
    empty-grid counterparts by transformation family and color context.
    \textbf{(f)} Joint contextual transformations predicted using primitives
    measured in the same context, on an empty grid, or with an incorrect
    contextual recoloring. Blue and orange denote cosine similarity and
    relative residual, respectively; bands and error bars denote 95\%
    confidence intervals.
    }
    \label{fig:single-cell-primitives}
\end{figure*}

\subsubsection{Context dependence of cell addition}
\begin{figure*}[tbp]
    \centering
    \captionsetup[subfigure]{skip=1pt}

    \begin{subfigure}[t]{0.43\textwidth}
        \centering
        \vspace{0pt}
        \includegraphics[width=\linewidth]{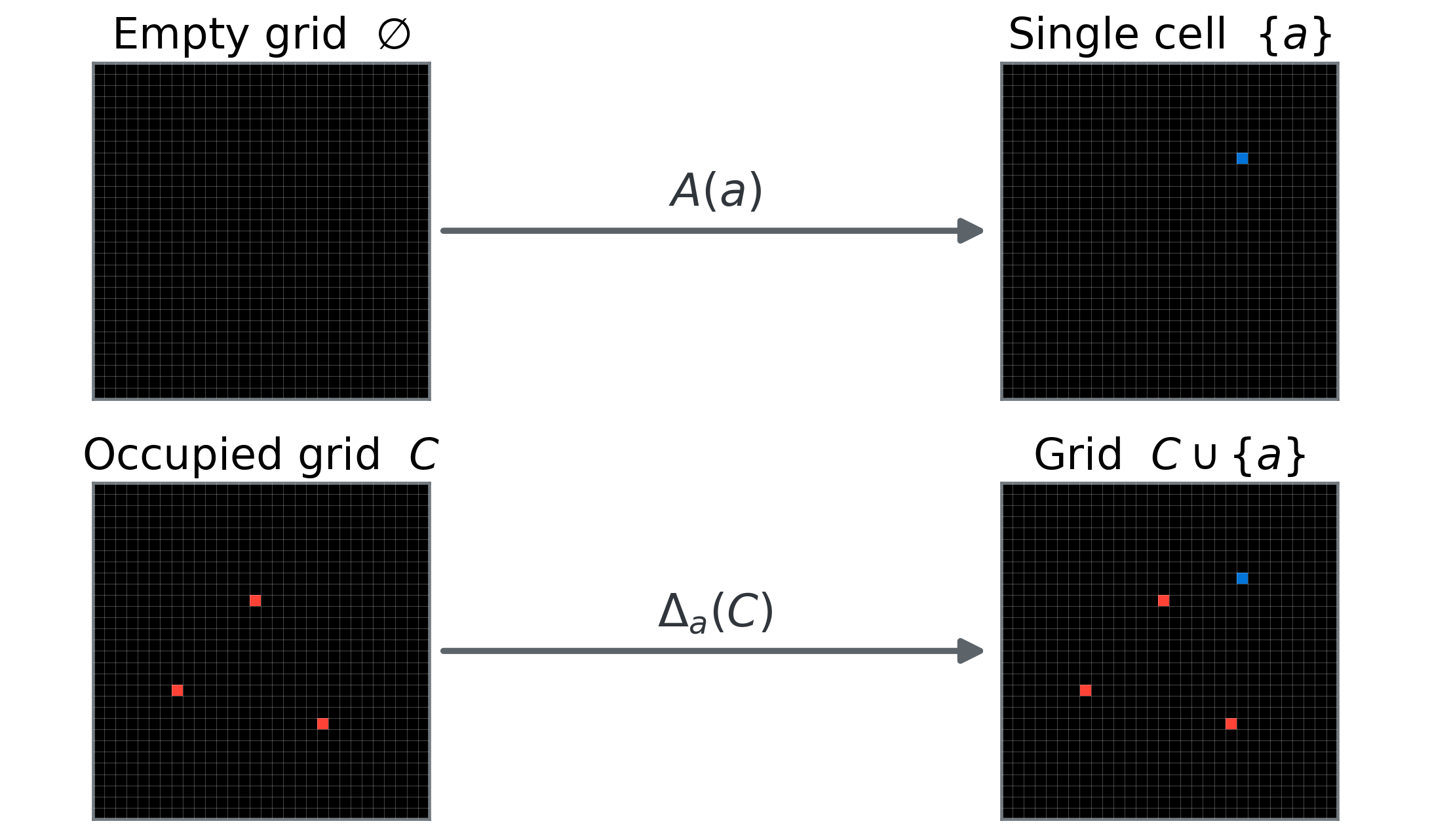}
        \caption{Context-free and contextual cell addition.}
        \label{fig:context-addition-examples}
    \end{subfigure}
    \hspace{0.035\textwidth}
    \begin{subfigure}[t]{0.43\textwidth}
        \centering
        \vspace{0pt}
        \includegraphics[width=\linewidth]{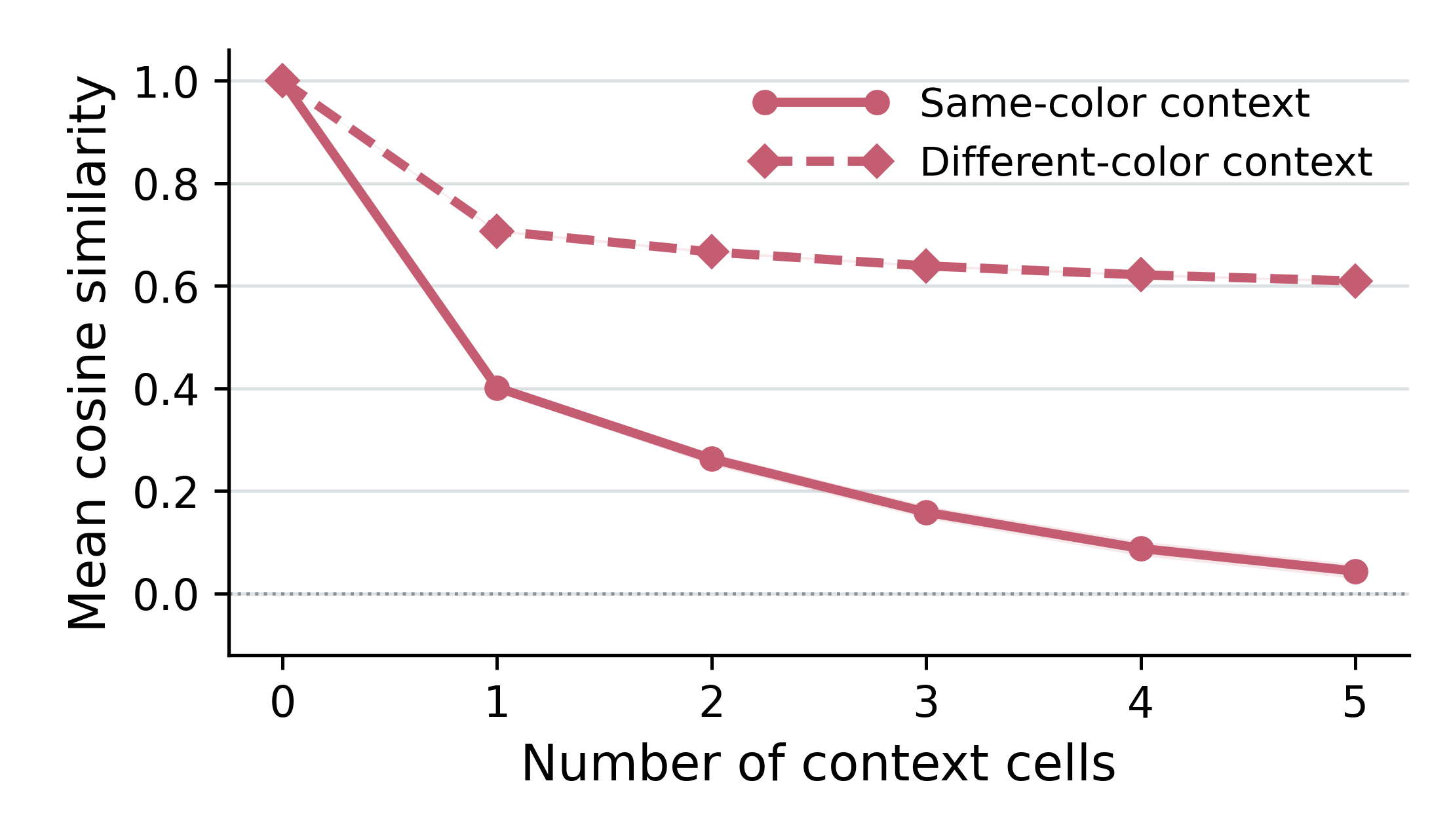}
        \caption{Loss of context-free transfer with grid occupancy.}
        \label{fig:context-addition-mass}
    \end{subfigure}

    \vspace{0.15em}

    \begin{subfigure}[t]{0.43\textwidth}
        \centering
        \vspace{0pt}
        \includegraphics[width=\linewidth]{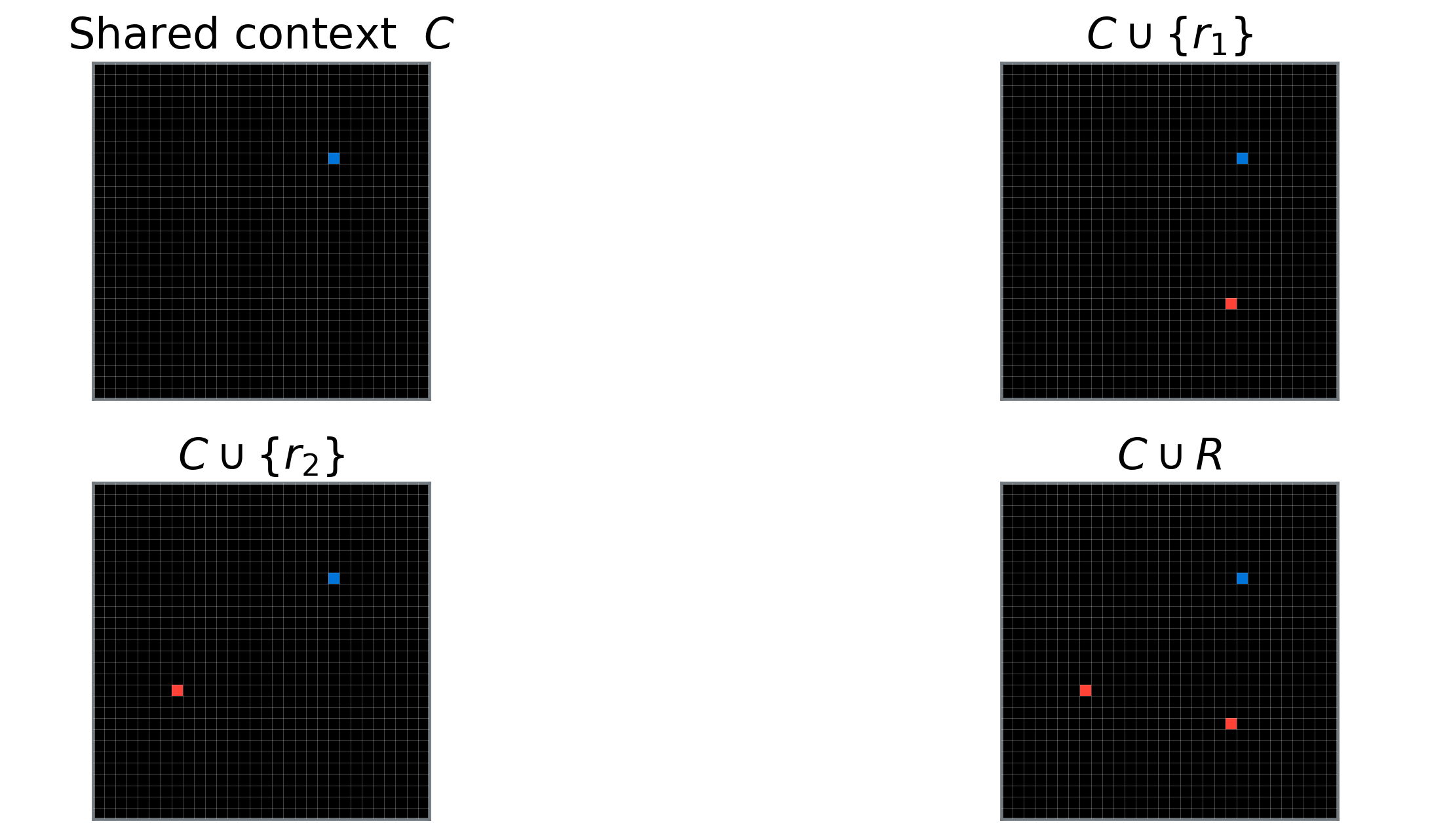}
        \caption{Single-cell additions measured from a shared context.}
        \label{fig:conditioned-addition-examples}
    \end{subfigure}
    \hspace{0.035\textwidth}
    \begin{subfigure}[t]{0.43\textwidth}
        \centering
        \vspace{0pt}
        \includegraphics[width=\linewidth]{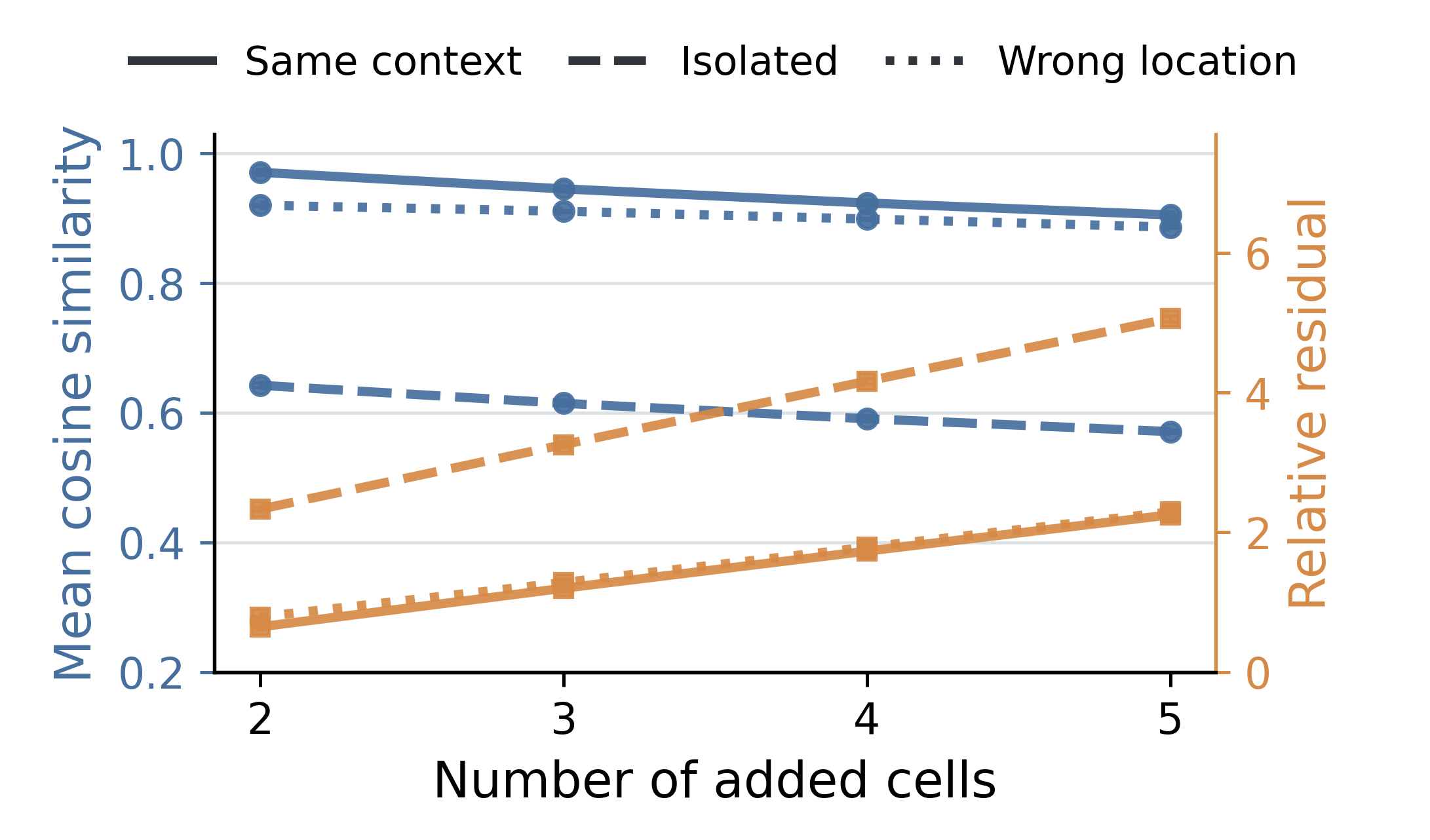}
        \caption{Context-conditioned composition.}
        \label{fig:conditioned-addition-composition}
    \end{subfigure}

    \caption{
    Cell-addition vectors in DINOv3-Large at 480 px are context-dependent and directionally aligned, but not magnitude-additive. Layer numbers are zero-indexed.
    \textbf{(a)} The same cell is added to an empty or occupied grid.
    \textbf{(b)} Similarity between isolated and contextual addition vectors
    as occupancy increases, separated by color context.
    \textbf{(c)} Multiple additions measured separately and jointly from a
    shared source context.
    \textbf{(d)} The joint addition is compared with sums of
    context-conditioned vectors, isolated additions, and a wrong-location
    control. Blue and orange denote cosine similarity and relative residual,
    respectively; bands and error bars denote 95\% confidence intervals.
    }
    \label{fig:context-conditioned-addition}
\end{figure*}

We finally examine single-cell addition on occupied synthetic grids; removal reverses the same pair of endpoints and therefore negates the difference vector. For an added cell $a$ and context $C$, define
\[
\Delta_\ell(a;C)=E_\ell(C\cup\{a\})-E_\ell(C),
\qquad
A_\ell(a)=E_\ell(\{a\})-E_\ell(\varnothing).
\]
The design crosses nine anchor positions, full-canvas random contexts containing one to five cells, and all ordered foreground-color pairs (Fig.~\ref{fig:context-addition-examples}; Appendix~\ref{app:low-level-methods}).

\begin{figure*}[tbp]
    \centering
    \captionsetup[subfigure]{skip=2pt}

    \begin{subfigure}[t]{0.98\textwidth}
        \centering
        \includegraphics[width=\linewidth]{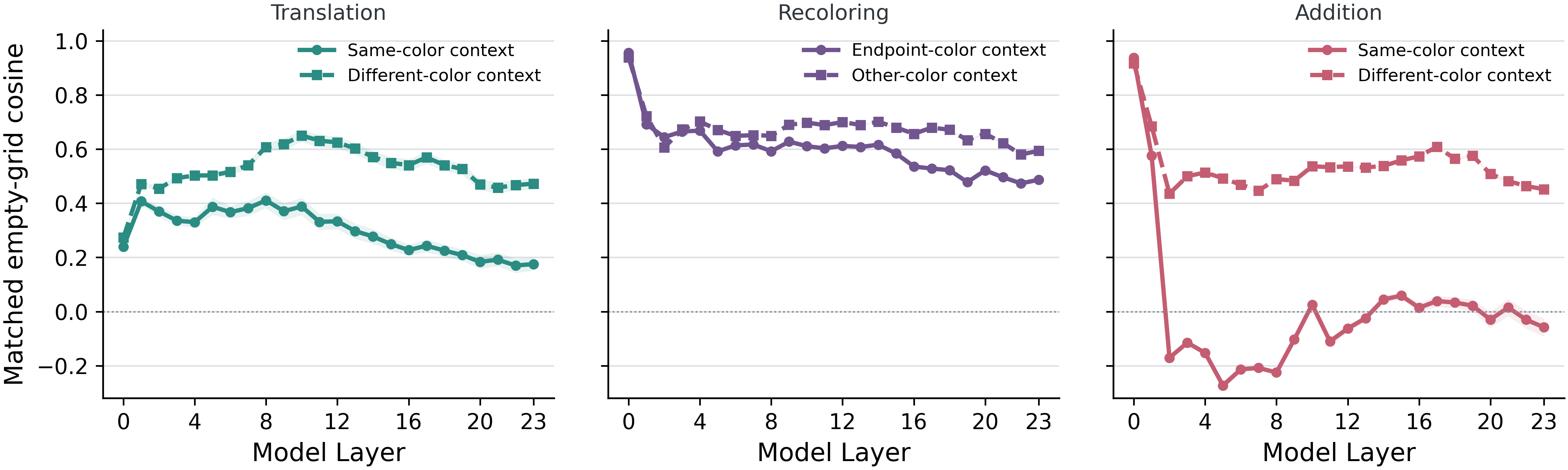}
        \caption{Layerwise context sensitivity with five context cells.}
        \label{fig:supp-layerwise-context-transfer}
    \end{subfigure}

    \vspace{3pt}

    \begin{subfigure}[t]{0.33\textwidth}
        \centering
        \includegraphics[width=\linewidth]{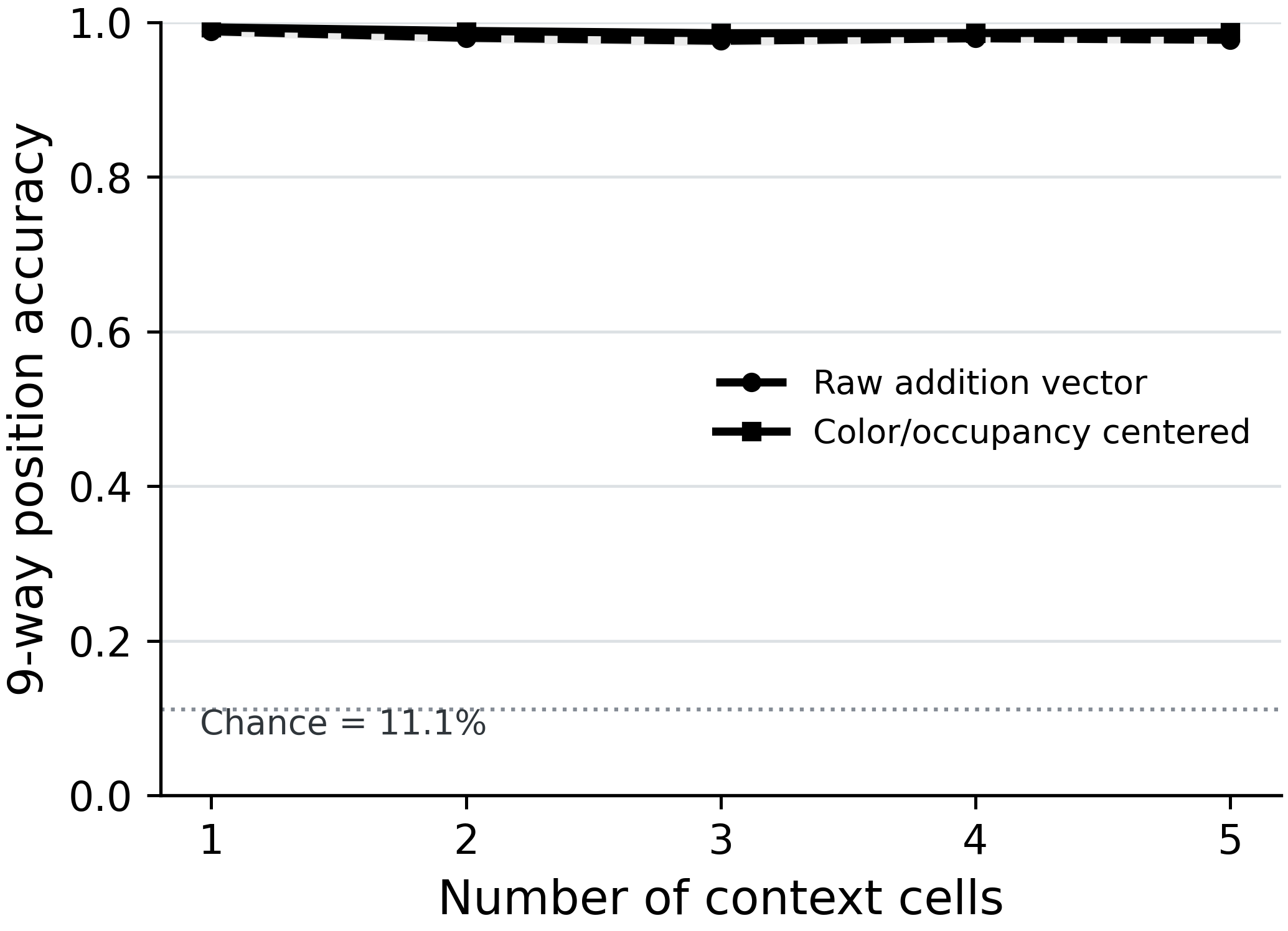}
        \caption{Cross-validated decoding of the added cell's position.}
        \label{fig:supp-addition-location-decoding}
    \end{subfigure}
    \hfill
    \begin{subfigure}[t]{0.65\textwidth}
        \centering
        \includegraphics[width=\linewidth]{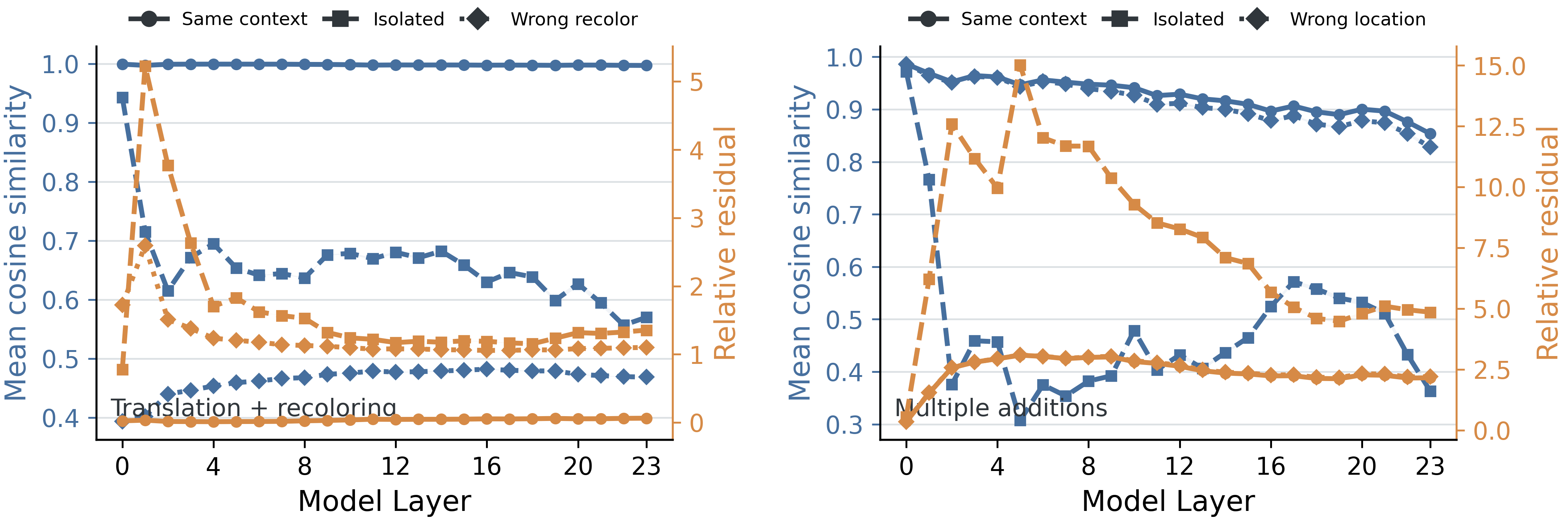}
        \caption{Layerwise contextual composition at the largest tested set size.}
        \label{fig:supp-layerwise-context-composition}
    \end{subfigure}

    \caption{
        Additional analyses of context-dependent single-cell geometry in DINOv3-Large at 480 px. Position decoding uses four-fold grouped cross-validation over independently sampled spatial layouts, with all color variants of a layout confined to the same fold. Layerwise analyses use a balanced subset containing every unit direction in each of four placement sets. The transfer analyses use five context cells; the multi-addition analysis uses five additions from a one-cell base. Bands and error bars denote 95\% spatial-bootstrap confidence intervals. Layer numbers are zero-indexed.
    }
    \label{fig:supp-context-geometry}
\end{figure*}

Addition is strongly context-dependent. At layer 17 with five context cells, the mean cosine between $A_\ell(a)$ and $\Delta_\ell(a;C)$ is $0.044$ (95\% CI: $0.030$--$0.058$) when context and added cells share a color, versus $0.610$ (95\% CI: $0.607$--$0.612$) when they differ (Fig.~\ref{fig:context-addition-mass}). A five-cell analysis across depth shows that this context effect appears after the first block and persists thereafter (Fig.~\ref{fig:supp-layerwise-context-transfer}), consistent with the preceding translation and recoloring results.

To test composition around a shared source state, an anchor cell forms a one-cell base $B=\{b\}$ and $R=\{r_1,\ldots,r_k\}$ contains $k\in\{2,\ldots,5\}$ randomly positioned additions. We compare the joint transition $B\rightarrow B\cup R$ with the sum of the $k$ changes obtained by adding each $r_i$ separately to the same $B$ (Fig.~\ref{fig:conditioned-addition-examples}). At layer 17 and $k=5$, the sum remains directionally aligned with the joint transition (cosine $0.905$, 95\% CI: $0.904$--$0.906$), but its relative residual reaches $2.249$ (95\% CI: $2.237$--$2.262$). Across layers, cosine declines to approximately $0.85$ by the final block while the residual remains near $2.5$ (Fig.~\ref{fig:supp-layerwise-context-composition}). Contextual additions therefore remain in broad directional superposition without being magnitude-additive.

Summing isolated-cell additions performs worse at $k=5$ (cosine $0.571$, residual $5.060$). A wrong-location control, which permutes spatial layouts while preserving the base, colors, and occupancy, remains comparatively close (cosine $0.887$, residual $2.291$; Fig.~\ref{fig:conditioned-addition-composition}). These aggregate measures are therefore dominated by occupancy- and color-related components. Nevertheless, a cross-validated linear decoder recovers the added cell's nine-way location from $\Delta_\ell(a;C)$ with $97.8\%$ accuracy, versus $11.1\%$ chance (Fig.~\ref{fig:supp-addition-location-decoding}; Appendix~\ref{app:low-level-methods}). Spatial information remains linearly accessible even when it contributes little to aggregate cosine geometry.

Together, the single-cell analyses identify context as a boundary of primitive transfer. Translation and recoloring compose accurately when measured from the same state; addition vectors remain aligned but are not magnitude-additive. At a higher level, each ARC intuition vector averages instance transformations measured from different support grids. These observations motivate distinguishing a task-level transformation signal from its query-dependent execution. Establishing how the controlled single-cell geometry relates to task-level intuition vectors will require experiments linking these transformations to individual ARC tasks.

\end{document}